\documentclass[]{fairmeta}

\usepackage[colorinlistoftodos]{todonotes}

\usepackage{graphicx}
\usepackage{url}
\usepackage{graphicx}
\usepackage{tikz}
\usepackage{hyperref}
\usepackage{multirow}
\usepackage{subcaption}
\usepackage{amsmath}
\usepackage{xspace}
\usepackage{calc}
\usepackage{float}
\usepackage{booktabs}
\usepackage{algorithm}
\usepackage{algpseudocode}
\usepackage{multirow}
\usepackage{makecell}
\usepackage{listings}
\usepackage[table]{xcolor}
\usepackage{tcolorbox}
\usepackage{pdflscape}
\usepackage{geometry}
\usepackage{cleveref}
\AddToHook{cmd/appendix/before}{\crefalias{section}{appendix}} 
\usepackage{tabularx}
\usepackage{algorithm}
\usepackage{algorithmicx}
\usepackage{algpseudocode}
\usepackage{soul}
\usepackage{changepage}
\usepackage{pifont}
\usepackage{fontspec}
\usepackage{amsfonts}
\usetikzlibrary{positioning}

\newcommand{\ccmark}{\textcolor{blue}{\ding{51}}}
\newcommand{\xxmark}{\textcolor{red}{\ding{55}}}

\definecolor{codegreen}{rgb}{0,0.6,0}
\definecolor{codegray}{rgb}{0.5,0.5,0.5}
\definecolor{codepurple}{rgb}{0.58,0,0.82}
\definecolor{backcolour}{rgb}{0.95,0.95,0.92}
\sethlcolor{green!15} %

\lstdefinestyle{mystyle}{
    commentstyle=\color{codegreen},
    keywordstyle=\color{magenta},
    numberstyle=\tiny\color{codegray},
    stringstyle=\color{codepurple},
    basicstyle=\ttfamily\footnotesize,
    breakatwhitespace=false,         
    breaklines=true,
    captionpos=b,                    
    keepspaces=true,                 
    numbers=left,                    
    numbersep=5pt,                  
    showspaces=false,                
    showstringspaces=false,
    showtabs=false,                  
    tabsize=2
}

\lstnewenvironment{promptlisting}{
\lstset{
    basicstyle=\ttfamily,
    numbers=none,
    tabsize=1,
    breaklines=true,
    breakautoindent=false,
    breakindent=0pt,
    moredelim=[is][\color{red}]{@}{@},
    moredelim=[is][\color{cyan}]{\#}{\#},
    moredelim=[is][\color{orange}]{£}{£},
    moredelim=[is][\color{brown}]{!}{!},%
}
}{}

\title{\huge{Omnilingual SONAR:} \\\LARGE{Cross-Lingual and Cross-Modal Sentence Embeddings Bridging Massively Multilingual Text and Speech}}

\author[]{Omnilingual SONAR Team}
\author[\dagger]{João Maria Janeiro}
\author[\dagger,\ddagger]{Pere-Lluís Huguet Cabot}
\author[\dagger]{Ioannis Tsiamas}
\author[\dagger]{Yen Meng}
\author[\ddagger]{Vivek Iyer}
\author[\S]{Guillem Ramírez}
\author[\ddagger]{Loic Barrault}
\author[]{Belen Alastruey}
\author[]{Xiang “Tony” Cao}
\author[]{Yu-An Chung}
\author[]{Marta R. Costa-Jussa}
\author[]{David Dale}
\author[]{Kevin Heffernan}
\author[]{Jaehyeong Jo}
\author[]{Artyom Kozhevnikov}
\author[]{Alexandre Mourachko}
\author[]{Christophe Ropers}
\author[]{Holger Schwenk}
\author[\dagger,\ddagger]{Paul-Ambroise Duquenne}

\affiliation[]{FAIR at Meta}
\contribution[\dagger]{OmniSONAR core contributors}
\contribution[\ddagger]{Spectrum core contributors}
\contribution[\S]{OmniSONAR-Token core contributor}

\abstract{Cross-lingual sentence encoders have traditionally been limited to a few hundred languages, and have sacrificed downstream performance to achieve better alignment across languages, limiting their adoption.
In this work, we introduce \sonar{} a novel family of omnilingual, cross-lingual and cross-modal sentence embedding models that breaks this barrier.
We establish a unified semantic space, natively encompassing text, speech, code and mathematical expressions, while achieving state-of-the-art downstream performance for an unprecedented scale of thousands of languages, from high-resource languages to extremely low-resource varieties.

To achieve this scale without representation collapse and while maintaining top-tier performance in the high-resource languages, we employ a progressive training strategy. We first build a state-of-the-art foundational embedding space for 200 languages using an LLM-initialized Encoder-Decoder, combining token-level decoding with a novel split-softmax contrastive loss and synthetic hard negatives. Leveraging this strong foundational space, we expand to several thousands of language varieties via a specialized two-stage teacher-student encoder distillation framework.
Further modeling extensions derived from \sonar{} address long context inputs and token-centric representations. %
Finally, we demonstrate the cross-modal extensibility of this space by seamlessly mapping 177 spoken languages into it.

\sonar{} redefines the state of the art for multilingual representation learning. It halves the cross-lingual similarity search error rate of the previous best models on the 200 languages of FLORES, while also achieving a staggering 15-fold error rate reduction across 1,560 languages in the BIBLE benchmark. Furthermore, our embedding model enables unprecedented translation capabilities, outperforming NLLB-3B on several multilingual benchmarks, and surpassing all previous models, including multi-billion-parameter LLMs, by 15 chrF++ points in 1,560$\rightarrow$English translation in the BIBLE benchmark. Beyond alignment and translation, \sonar{} demonstrates strong general-purpose capabilities across downstream embedding tasks on MTEB and programming languages on XLCoST. For the speech modality, our massively multilingual extension exhibits a 43\% lower error rate in cross-lingual and cross-modal similarity search, while achieving 97\% of SeamlessM4T performance in speech-to-text translation, despite being a zero-shot translation model trained only with ASR data. 
Finally, by training an encoder-decoder language model, Spectrum, exclusively on English text that processes \sonar{} sequences, we unlock immediate high-performance transfer to thousands of languages and the speech modality for complex downstream tasks.
These outstanding results position \sonar{} as a robust, language- and modality-agnostic foundation for any downstream usage.

}

\date{\today}
\correspondence{Paul-Ambroise Duquenne at \email{padqn@meta.com}}

\newcommand{\sonarname}{OmniSONAR}  %

\newcommand{\sonarlogo}{%
  \raisebox{-0.10em}{%
    \includegraphics[height=0.8em]{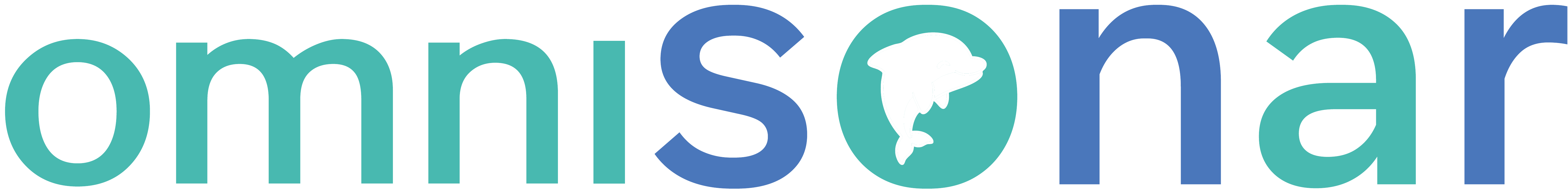}%
  }%
}
\DeclareRobustCommand{\sonarlogoinline}{%
  \texorpdfstring{\sonarlogo\hspace{0.20em}}{OmniSONAR}\xspace%
}
\newcommand{\sonar}{\sonarname{}\xspace}
\newcommand{\dev}{\texttt{dev}\xspace}
\newcommand{\test}{\texttt{test}\xspace}
\newcommand{\train}{\texttt{train}\xspace}
\newcommand{\devtest}{\texttt{devtest}\xspace}
\newcommand{\chrf}{chrF++\xspace}

\newcommand{\sonarhundred}{\sonarname-200\xspace}
\newcommand{\sonarlanguages}{1.5k}
\newcommand{\meFive}{mE5\textsubscript{large} \xspace}
\newcommand{\embGemma}{EmbeddingGemma \xspace}
\newcommand{\embQwen}{Qwen3-Embedding-0.6B \xspace}

\newcommand{\mdocu}{Spectrum}
\newcommand{\sonartower}{SONAR tower}
\newcommand{\tokentower}{token tower}

\newcommand{\tokensonar}{\sonarname{}-Token}

\newcommand{\longcontextsonar}{\sonarname{}-LC}

\DeclareMathOperator*{\argmax}{argmax}

\begin{document}

\maketitle

\newpage
\tableofcontents
\newpage

\begin{figure}
    \centering
    \includegraphics[width=\linewidth]{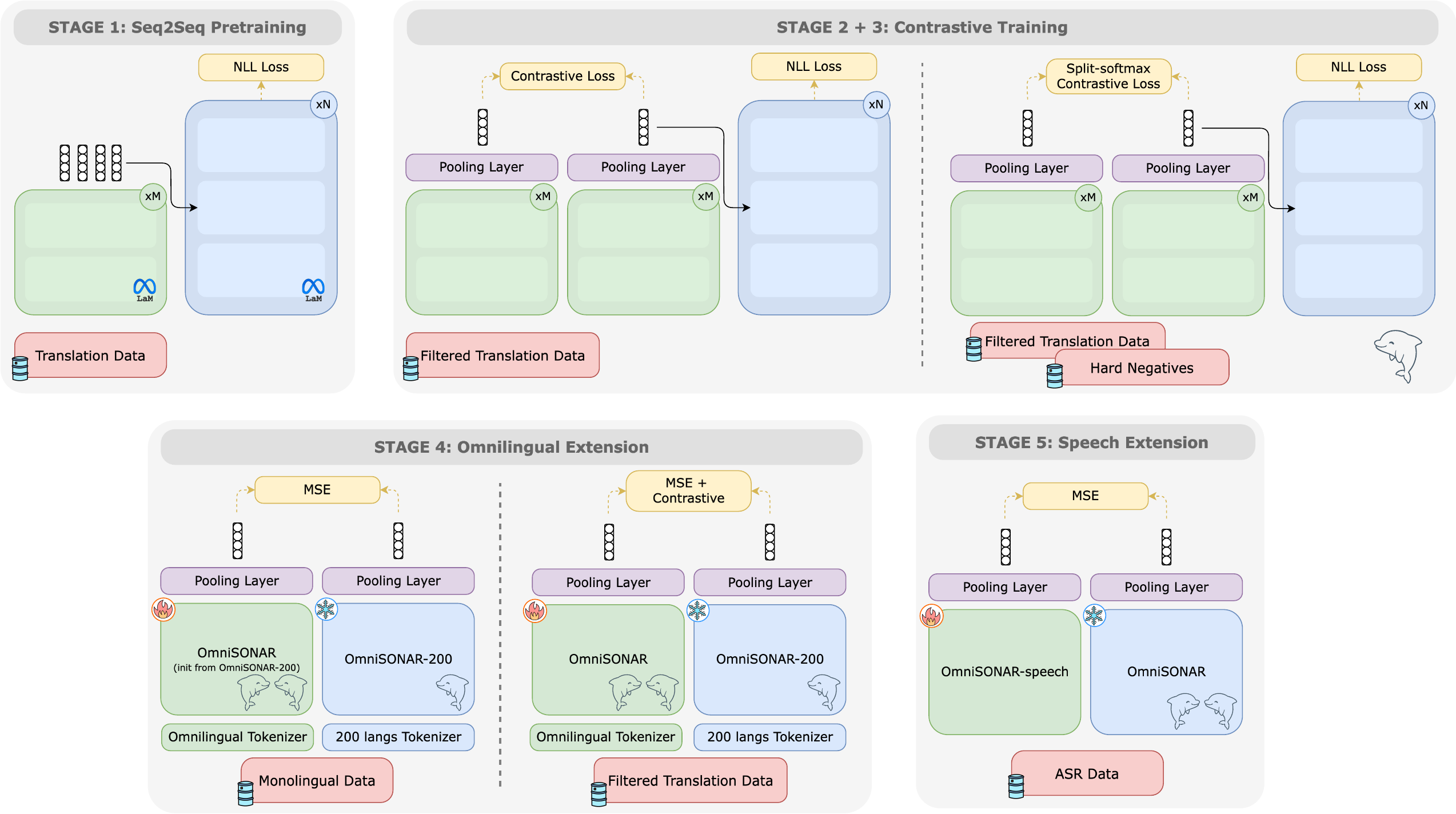}
    \caption{The \sonarlogoinline{} training stages. In \textbf{Stage 1}, we train our LLM-initialized encoder-decoder on translation data with a decoding loss. In \textbf{Stage 2}, we introduce an encoder bottleneck via pooling and train with a combination of contrastive and decoding objectives. In \textbf{Stage 3}, we introduce hard negatives and continue training with a split-softmax contrastive objective and the decoding loss.
    In \textbf{Stage 4}, we extend the space to omnilingual-level language coverage by training with teacher-student distillation on 4,200 language varieties with a combination of MSE and Contrastive objectives, while first warming-up the omnilingual tokenization with MSE-based distillation.
    Lastly, in \textbf{Stage 5}, we extend the omnilingual space to the speech modality with teacher-student distillation using ASR data.
    }
    \label{fig:sonar2_method}
\end{figure}

\section{Introduction}
\label{section:intro}

Multilingual representation learning has long been a central focus in Natural Language Processing, spanning from traditional Machine Translation \citep{nllb,kocmi-etal-2025-findings} to the recent surge in multilingual large language models \citep{workshop2022bloom,aya,gemma3}. Furthermore, there has been growing interest in the speech modality, with advances in both representation learning \citep{w2v_bert,chen2022wavlm} and language modeling \citep{zhang2023speechgpt,moshi,roy2026personaplex}.
However, a persistent challenge remains: the extreme scarcity of training data for the vast majority of the world's languages for both text and speech.
This scarcity has motivated the development of cross-lingual \citep{sonar,mexma,labse} and cross-modal \citep{duquenne2021multimodal,khurana2022samu,clip} sentence encoders, models that establish a shared semantic space where sentences with similar meanings are embedded closely together regardless of their language or modality.
These aligned embeddings act as the vital engine for critical applications, including large-scale parallel data mining for text and speech \citep{schwenk-etal-2021-ccmatrix,duquenne2023speechmatrix}, zero-shot classification \citep{costa-jussa-etal-2024-mutox}, translation quality estimation for text and speech \citep{chen-etal-2023-blaser}, and expanding multilingual and multimodal coverage of language modeling, even while training on monolingual data, as shown in the Large Concept Model \citep{barrault2024large}.
In general, since their representations are aligned across languages (and potentially modalities), they unlock multilingual zero-shot downstream performance for tasks without the need of data in all languages.

Despite their utility, existing encoders face two critical limitations that restrict their widespread adoption. First, a fundamental performance trade-off exists: achieving good cross-lingual alignment often degrades individual representation quality, leaving these models trailing behind general-purpose embeddings \citep{wang2024multilingual, qwen3embedding, embedding_gemma_2025} that do not exhibit language-agnostic alignment, but perform well in downstream evaluations.
Second, coverage is typically restricted to roughly 100 to 200 languages because the field has lacked a methodology that can effectively scale coverage in data-scarce regimes.
Scaling beyond this barrier is often additionally hindered by the well-documented `\emph{curse of multilinguality}' \citep{massively_multilingual_nmt,lifting_the_curse_of_multilinguality,alastruey2025interferencematrixquantifyingcrosslingual}, where adding more languages to a fixed-capacity model degrades performance due to parameter competition.

In this work, we introduce \sonar{}, a novel family of omnilingual, cross-lingual, and cross-modal sentence embedding models designed to break these barriers. \sonar{} establishes a unified semantic space spanning an unprecedented 4,200 language varieties, supporting speech, code, and mathematical expressions. To achieve this scale without sacrificing representation quality, we employ a three-stage progressive training strategy (\Cref{fig:sonar2_method}):

\begin{itemize}
    \item \textbf{Step 1: Establishing a state-of-the-art foundation.} We first build a foundational embedding space for 200 languages using an LLM-initialized Encoder-Decoder architecture. By combining token-level decoding \citep{mexma, sonar} with a novel split-softmax contrastive loss and synthetic hard negatives, we capture deep semantic nuances often lost in standard alignment techniques.
    \item \textbf{Step 2: Omnilingual expansion.} Leveraging this strong foundation, we expand to thousands of language varieties through a teacher-student distillation framework. We project new languages into the space using a hybrid Mean Squared Error (MSE) and contrastive loss objective.
    \item \textbf{Step 3: Speech expansion.} This space is then expanded into speech through distillation, aligning spoken sentences and their transcriptions through an MSE objective.
\end{itemize}

\sonar{} redefines the state of the art for multilingual and cross-lingual representation learning.
\sonar{} halves the cross-lingual similarity search error rate of previous best models on the 200 languages of FLORES while achieving a staggering 15-fold error rate reduction across 1,560 languages in the BIBLE benchmark.
\sonar{}-speech also achieves a 43\% error rate reduction compared to the previous state of the art.
Furthermore, these representations are powerful enough to enable unprecedented translation capabilities, surpassing multi-billion-parameter LLMs by 15 chrF++ points in 1,560$\rightarrow$English translation.

The ultimate validation of \sonar{}'s representational strength is showcased through Spectrum, our encoder-decoder language model that operates on \sonar{}'s embeddings. By training Spectrum exclusively on English text to process \sonar{} sequences, we unlock high-performance, zero-shot transfer to thousands of languages and the speech modality for complex reasoning tasks.
Spectrum achieves a 16\% improvement over LLaMA3.2 3B in XBelebele, due its better multilingual representations, powered by \sonar{} and seamlessly transfer this high performance to Speech-XBelebele.
These results demonstrate that \sonar{} is more than a retrieval tool, it is a robust, language- and modality-agnostic foundation for a wide range of multilingual speech/text tasks.%

Our main contributions are as follows:
\begin{itemize}
\item \textbf{A Novel LLM-based Embedding Framework:} We introduce an Encoder-Decoder architecture initialized from an English-centric pretrained LLM that establishes a state-of-the-art foundational space for 200 languages, natively encompassing code and math. In this framework, we introduce a sequence-to-sequence pre-training stage to provide the multilingual and translation capabilities the base LLM lacks. Then, we couple a translation reconstruction objective with a novel \textit{split-softmax contrastive loss}, forcing the model to capture nuanced semantic information. This space double the performance of the current state of the art in multilingual alignment in FLORES, and closes the gap in downstream performance to general purpose models.

\item \textbf{A Lossless Omnilingual Extension Framework:} We provide a novel method for language expansion combining contrastive and MSE objectives. This method enables new languages to be natively integrated into the representation space, while also ensuring that the performance of existing languages is preserved.
With this expansion we boost the coverage to 4,200+ language varieties.
It achieves a 15-fold error rate reduction across 1,560 evaluated languages in the BIBLE framework.

\item \textbf{Massively Multilingual Speech Integration:} We map the speech modality into this shared space, creating a unified speech encoder covering 177 languages that achieves a 43\% reduction in cross-lingual cross-modal similarity search error rates.

\item \textbf{The First Omnilingual Space:} We present the most massive sentence embedding space to date, natively encompassing code, math, speech and trained on 4,200+ language varieties.
Models were trained at various scales, ranging from 1.5B to 39M parameters, in order to accommodate a wide range of compute budget constraints.

\item \textbf{Unprecedented Omnilingual Decoding:} We demonstrate that our omnilingual representations preserve enough fine-grained semantic information to drastically outperform multi-billion-parameter LLMs in translation benchmarks when evaluated with the paired model decoder.

\item \textbf{General-Purpose Capabilities \& Omnilingual Analysis:} Beyond alignment, \sonar{} demonstrates strong general-purpose performance on MTEB and XLCoST. 
We provide analysis showing how our methodology transforms the multilinguality curse into a blessing for zero-shot generalization, and several ablations for model components.

\item \textbf{Zero-shot Omnilingual Speech \& Text Language Modeling with \sonar{}:} We show how training an encoder-decoder language model (Spectrum) on \sonar{} representations of English text alone,  can unlock zero-shot massively multilingual and speech understanding. Achieving 61\% on XBelebele and 89\% on SpeechSIB zero-shot, Spectrum outperforms bespoke fine-tuned models, demonstrating how \sonar{} can pave the way for radically simple multilingual and multimodal transfer in LLMs. 
\end{itemize}

\section{Related Work}
\label{section:related_work}

The field of multilingual sentence embeddings has grown rapidly, driven by benchmarks like MTEB \citep{muennighoff-etal-2023-mteb}, xsim/xsim++ \citep{laser,chen-etal-2023-xsim}, and MIRACL \citep{zhang-etal-2023-miracl}. In our work, we differentiate between \textit{multilingual} and \textit{cross-lingual} sentence embeddings. The former provides multilingual coverage to general-purpose embeddings, where alignment across languages is only one sub-task among many others. On the other hand, cross-lingual sentence embeddings build semantic representations by focusing explicitly on cross-lingual alignment between translations.

\paragraph{Cross-lingual Alignment.} Cross-lingual embedding models map vector representations across languages into a shared space. Training on translation data typically enables semantic alignment via contrastive objectives using encoders only \citep{yang2019improving, feng-etal-2022-language, miao-etal-2024-enhancing} or non-contrastive objectives with decoder signals \citep{sonar, janeiro-etal-2025-mexma}. In \sonar{}, we combine both decoder and contrastive losses to build a foundational embedding space for 200 languages.

\paragraph{Contrastive Learning.} While contrastive learning dominates sentence embedding training \citep{gao-etal-2021-simcse}, hard negatives remain underexplored in cross-lingual alignment, with LaBSE \citep{feng-etal-2022-language} reporting negative results. General-purpose models \citep{wang2024multilingual, sturua2024jinaembeddingsv3multilingualembeddingstask} have successfully used mined and synthetic negatives. In \sonar{}, we unlock contrastive objectives with synthetic hard negatives for better cross-lingual alignment.

\paragraph{Teacher-Student Distillation.} Teacher-student distillation is commonly used to extend existing embedding spaces to new languages or new modalities like speech. This was introduced by \citet{reimers-gurevych-2020-making} for text and extended to more languages with LASER3 \citep{heffernan2022bitext}. \citet{duquenne2021multimodal} introduced teacher-student training to extend text-only embedding spaces to the speech modality, extracting a fixed-size semantic representation from speech utterances. \citet{khurana2022samu} and \citet{sonar} followed a similar approach for the LaBSE and SONAR embedding spaces, respectively. \citet{charsonar} employed teacher-student distillation to adapt the SONAR encoder to a character-level tokenization, addressing tokenization bottlenecks in unseen scripts. Although Mean Squared Error (MSE) is the gold standard for distilling representations, \citet{mult_representation_distill} demonstrated that contrastive learning objectives can yield sharper decision boundaries and superior retrieval performance. Our approach for the omnilingual extension synthesizes these insights: we employ a student-teacher framework similar to \citet{reimers-gurevych-2020-making}, but scale it to thousands of languages by combining the stability of MSE with the discriminative power of contrastive losses \citep{mult_representation_distill}, while explicitly adapting the vocabulary to handle the immense linguistic diversity of the 4,200 language varieties we use for training.

\paragraph{Massively Multilingual Models.} XLM-R \citep{xlmr} was one of the earliest highly multilingual MLM encoders, while more recently Glot500 \citep{glot500} scaled the coverage to 500 languages. Several works have proposed massively multilingual encoder-decoders for translation-oriented tasks, with NLLB \citep{nllb} and SeamlessM4T \citep{seamless} covering 100 languages for speech and 200 for text, respectively, while Madlad \citep{madlad} offers support for 400 languages. Recent efforts in speech models have scaled coverage for ASR to an omnilingual level with MMS \citep{mms} and Omni-ASR \citep{omni_asr}.

\paragraph{Code and Math.} Recent general-purpose models \citep{wang2024multilingual, nussbaum2025trainingsparsemixtureexperts} and code-specific embeddings \citep{zhang2024code, liu2025codexembed, sureshcornstack} incorporate code and math data. Most code embedding systems use docstring-implementation pairs \citep{Husain2019CodeSearchNetCE, zhang2024code, sureshcornstack}, focusing on function-level rather than sentence-level representations. 

\paragraph{Speech Encoders and Embeddings.} Extending text-centric semantic spaces to the speech modality enables powerful cross-modal applications, such as zero-shot speech translation \citep{duquenne2022t, duquenne2023modular, duquenne:hal-04629427, zeroswot} and cross-lingual speech mining \citep{duquenne2021multimodal,barrault2023seamlessm4t}. Significant gains were observed when training speech-to-text and speech-to-speech translation models with such mined data \citep{lee2022textless,chen2023speech}. SONAR \citep{sonar} utilized the self-supervised representations of w2v-BERT encoders \citep{w2v_bert} to map the speech modality to the embedding space, while charSONAR \citep{charsonar} utilized the highly multilingual CTC-based MMS encoder \citep{mms}. Here we build upon these foundations by initializing our student speech encoder with the massively multilingual wav2vec 2.0 model from Omni-ASR \citep{omni_asr} and projecting speech into the \sonar{} space. %

\paragraph{Multilingual and Multimodal Language Modeling.} A plethora of multilingual decoder-only LLMs have been proposed recently, including Llama \citep{llama3}, Qwen \citep{yang2025qwen3}, Gemma \citep{gemma3} and Aya \citep{salamanca2025tinyaya}. Despite strong progress, recent works on multilingual \citep{bandarkar-etal-2024-belebele, singh2024globalmmluunderstandingaddressing} and cross-lingual \citep{marchisio-etal-2024-understanding, iyer-etal-2025-xl} benchmarking have shown that even frontier LLMs continue to underperform for low-resource languages,  underscoring the need for representations that better bridge the multilingual gap.
A similar challenge emerges in the context of multimodal transfer. Speech/Text Language Models \citep{mitsui2024pslm,moshi} are actively working to bridge the gap between modalities with respect to downstream performance. Various works have introduced novel methods to improve cross-modal transfer, such as interleaving techniques \citep{nguyen2025spirit} and chain-of-modality \citep{zhang2023speechgpt} approaches. Finally, several projects have addressed cross-modal transfer by leveraging shared cross-modal embedding spaces \citep{agostinelli2023musiclm,wang2025mats}.

\section{Data} \label{section:data}

In this section, we introduce the datasets used to train the \sonar{} text and speech models, encompassing monolingual text, parallel translation pairs, code and mathematical expressions, and ASR audio data. We define the specific data regimes employed across our training stages, outline our sources, and detail our filtering, synthetic generation, and upsampling strategies. Lastly, we discuss the evaluation datasets used to measure \sonar{}'s performance.

\subsection{Language Sets and Training Stages}
\label{subsec:data_stages}
Throughout our training pipeline, we distinguish between a \emph{foundational} set of base languages and an extended \emph{omnilingual} set of thousands of language varieties. The foundational set includes 200 languages \cite{nllb,sonar}, which overall benefit from extensive data sources and well-established evaluation benchmarks (i.e., FLORES \cite{nllb}). The base set additionally includes code and math data. We structure our data usage across the training stages as follows:
\begin{itemize}
    \item \textbf{Stage 1 (Sequence-to-Sequence Pre-training):} We utilize parallel translation data spanning 200 $\leftrightarrow$ 200 directions among the foundational set of languages.
    \item \textbf{Stages 2 \& 3 (Contrastive Fine-tuning):} To effectively leverage contrastive signals and hard negatives, we restrict the parallel data to 200 $\rightarrow$ English translation pairs.
    \item \textbf{Stage 4 (Omnilingual Expansion):} We use parallel translation data covering all 4,200+ language varieties. The strict requirement for this stage is that at least one language in the translation pair must be part of the 200+ foundational languages, enabling the frozen teacher model to encode it, while the student model learns the new language representation. 
    \item \textbf{Stage 5 (Cross-Modal Speech Extension):} We use audio-transcription pairs (ASR data) for 177 languages.
\end{itemize}

\subsection{Natural Language Text Data} \label{subsec:text_data}
\paragraph{Training Datasets.}

Translation data aligned at the sentence level has become the standard source of supervised data for learning multilingual sentence embeddings \citep{sonar,wang2024multilingual,janeiro-etal-2025-mexma}. Prior massive multilingual efforts, such as NLLB \citep{nllb}, relied on three primary data streams: human-annotated translations, mined parallel data, and back-translated segments.

We adopt a similar, but modernized, protocol to construct our training corpus. First, to establish our foundational data for the 200 foundational languages, we utilize a mixture of human-translated and mined datasets roughly reproducing the original data composition used to train the NLLB models. Then we generate massive amounts of synthetic translation data sourced from recent, large-scale monolingual document-based web corpora covering 200 languages and segment these raw documents into sentences using a custom SaT model \citep{minixhofer-etal-2023-wheres} that has been fine-tuned for extensive language coverage (see \Cref{app:SaT} for further details). Leveraging the NLLB-3.3B model\footnote{\url{https://huggingface.co/facebook/nllb-200-3.3B}}, we translate English sentences from these document sources into the 200 NLLB-supported languages and non-English sentences into English. Such synthetic data can either be used as back-translated data (source text is synthetic) or forward translated data (target text is synthetic).

To successfully scale our coverage to an omnilingual level, we aggregate a diverse set of high-quality, massively multilingual human-annotated translation datasets, including Bible texts, PanLex \citep{panlex} and Tatoeba \citep{tatoeba}. Extensive details for the massively multilingual datasets used in our omnilingual training pipeline are provided in Appendix \ref{app:translation_data}.

\paragraph{Evaluation Datasets.}
We evaluate our models on a series of highly multilingual translation benchmarks:
\begin{itemize}
    \item \textbf{FLORES} \citep{nllb}: An n-way parallel benchmark for 202 languages, utilizing English hard negatives for challenging similarity search evaluations \citep{xsimplusplus}. This covers our foundational set of languages.
    \item \textbf{FLORES+} \citep{wmt_oldi_24,wmt_oldi_25}: An extension of FLORES with 212 \test languages, to measure performance on new languages within the FLORES domain.
    \item \textbf{BOUQuET} \citep{bouquet}: A multi-centric, multi-domain benchmark. We use the X$\rightarrow$English directions of version \texttt{v2025.11.13} \citep{omtbigpaper}, covering 177 languages, $\sim$40\% of which are outside our foundational language set.
    \item \textbf{AfroLingu-MT} \citep{afrolingumt}: A benchmark dedicated to 38 low-resource African languages.
    \item \textbf{BIBLE}: Our primary omnilingual benchmark, covering 1,560 languages (1,420 added during Stage 5). We use John's Gospel (chapters 1-10 for \dev, 11-22 for \test).
\end{itemize}

Additionally, we evaluate general-purpose downstream capabilities using the sentence-level MTEB benchmark suite, which includes the following tasks and benchmarks:

\begin{itemize}
    \item \textbf{Classification:} MassiveIntent \& MassiveScenario \citep{fitzgerald2022massive}, MTOPDomain \& MTOPIntent \citep{li-etal-2021-mtop}, AmazonCounterfactual \citep{oneill-etal-2021-wish} and SIB200 \citep{adelanietal2024sib}.
    \item \textbf{Pair Classification:} XNLI \citep{conneau2018xnli} and its extension, XNLIV2 \citep{upadhyay2023xnli}.
    \item \textbf{Semantic Textual Similarity (STS):} STS17 \citep{cer-etal-2017-semeval}.
\end{itemize}

\subsection{Code and Math data}
\paragraph{Training datasets.}
Although our primary focus is on sentence-level, modality-agnostic representations, we treat code and mathematical expressions as semantic units that can be mapped into this shared embedding space. In this framework, programming languages like JavaScript or Go are considered alongside natural languages such as Catalan or Portuguese.
To create translation data that encompasses both programming and natural languages, we have developed a comprehensive pipeline that overcomes the limitations of traditional docstring-based methods.
We focus on sentence-level code snippets and mathematical expressions whose semantics can be described in a single natural language sentence.
Our approach involves the following steps:
\begin{enumerate}
    \item[(1)] syntax-aware segmentation of code from 7 programming languages using Abstract Syntax Trees
    \item[(2)] extraction of LaTeX mathematical expressions from scientific corpora
    \item[(3)] generation of natural language descriptions using LLaMA3.3 70B Instruct,
    \item[(4)] creation of multilingual versions through back-translation. Quality is ensured through consistency filtering of the synthetic data.
\end{enumerate}

Some examples of code and math data are presented in \Cref{tab:code_math_examples}.

\begin{table}[ht]
\centering
\small
\begin{tabularx}{\textwidth}{lX}
\toprule
\multicolumn{2}{l}{\textbf{Example 1: Python}} \\
\midrule
Source & \texttt{if input\_event["enable\_points"] or event\_info.get("bonus"):} \\
\addlinespace[0.2em]
Target & The script determines if the \texttt{"enable\_points"} in the \texttt{input\_event} dictionary is set to True, or if the \texttt{event\_info} dictionary includes a key called \texttt{"bonus"} with any associated value. \\
\midrule
\multicolumn{2}{l}{\textbf{Example 2: JavaScript}} \\
\midrule
Source & \texttt{const DataHandler = require('./lib/dataHandler.js'); let windowRef; const dataHandler = new DataHandler(\{\});} \\
\addlinespace[0.2em]
Target & A constant named \texttt{DataHandler} is initialized by importing a module from a file called \texttt{dataHandler.js} located in a folder named \texttt{lib}, then a variable \texttt{windowRef} is declared, and a constant \texttt{dataHandler} is created as a new instance of \texttt{DataHandler}, passing an empty object to its constructor. \\
\midrule
\multicolumn{2}{l}{\textbf{Example 3: Math}} \\
\midrule
Source & $\Psi \in \mathop{\rm L}\nolimits^{2}(W)$ \\
\addlinespace[0.2em]
Target & The function $\Psi$ is a square-integrable function defined on the set W, meaning its square has a finite integral over W. \\
\bottomrule
\end{tabularx}
\caption{Examples of paired data for code and mathematical expressions.}
\label{tab:code_math_examples}
\end{table}

For complete technical details, implementation procedures, and filtering methods, please refer to \Cref{app:code_math_data_generation}.

\paragraph{Evaluation datasets.}
To further assess \sonar{} performance on domains including mathematical expressions and programming languages, we evaluate on GMMLU \citep{singh2024globalmmluunderstandingaddressing}, MMLU translated to 41 languages, by pairing questions in any language to their English equivalent and XLCoST \citep{xlcost}, to our knowledge the only snippet-level Code2Code benchmark. It was built for C++, Java, Python, C\#, Javascript, PHP, C and natural language. It contains parallel programs in all 6 programming languages that were split into parallel code snippets and natural text comments paired to them. Here we focus solely on the Code2Code snippet retrieval benchmark in a zero-shot fashion, as we never train \sonar{} on Code2Code pairs.

\subsection{Speech Data}
\paragraph{Training Datasets.}
For training speech encoders, we use a portion of the Omnilingual ASR Corpus~\citep{omni_asr}.  The total volume of the data portion we use is approximately~121k hours covering a total of~177 languages.  The selection of these~177 languages is based on the overlap between the 200 NLLB languages and those covered by the entire Omnilingual ASR Corpus.  The data is composed of publicly available data and internal data.  The publicly available data include ALFFA \citep{abate2005alffa,gelas2012alffa,gauthier2016alffa}, LibriSpeech ASR \citep{panayotov2015librispeech}, the South African language data of \citet{vanniekerk2017rapid}, ASR and TTS data by \citet{kjartansson2018crowd}, \citet{kjartansson2018tts} and \citet{he2020open}, CSS10 \citep{park2019css10}, FOSD \citep{fosd}, Zeroth Korean dataset,\footnote{\url{https://github.com/goodatlas/zeroth}} Burmese Speech Corpus \citep{oo2020burmese}, Common Voice v22 \citep{ardila2020common}, VoxPopuli \citep{wang2021voxpopuli}, VoxLingua-107 \citep{valk2021slt}, RuLS,\footnote{\url{https://www.openslr.org/96/}} the Kokoro Speech Dataset,\footnote{\url{https://github.com/kaiidams/Kokoro-Speech-Dataset}} MLS \citep{pratap2020mls}, {S}amr{\'o}mur \citep{mollberg2020samromur}, the Kazakh Speech Corpus \citep{khassanov2021crowdsourced}, iMaSC \citep{gopinath2022imascic}, ParlaSpeech-HR \citep{ljubesic2022parlaspeech}, NPSC \citep{solberg2022norwegian}, FLEURS \citep{conneau2023fleurs} and NaijaVoices \citep{emezue2025naijavoices}.

\paragraph{Evaluation dataset.} We evaluated \sonar{} speech encoders on the massively multilingual FLEURS test set \citep{conneau2023fleurs}, which extends FLORES-101 \citep{goyal2022flores} to the speech modality. It can be used as a Speech Translation evaluation set, as it provides speech recordings in 101 languages paired with their English transcriptions.

\subsection{Data Filtering and Upsampling Strategies}
\label{section:data_processing/filtering}

\paragraph{Filtering.}
Given the vast amount of data available, and that the data regimes required across our experimental setup varies, we will use a different set of filtering strategies across our work:
\begin{itemize}
    \item We estimate direction-specific thresholds by applying BLASER2~\citep{blaser2} to the high-quality data of FLORES~\citep{nllb} \dev set. We then take the mean, $\mu(\text{scores}_{xy})$, and standard deviation, $\sigma(\text{scores}_{xy})$, where $\text{scores}_{xy}$ is the BLASER2 scores for the pair of languages x-y for the 997 examples in the set. We score our paired translation data for the languages covered by BLASER2 to be used later as filtering criterion. The filtering criteria applied is $\mu(\text{scores}_{xy}) - k \cdot \sigma(\text{scores}_{xy})$, where k depends on the training stage.
    \item The vast majority of the languages covered in our data are not supported by BLASER2. To filter this data, we use an early version of our omnilingual encoder. Similar to our approach with Blaser-based filtering, we calibrate the language-specific similarity thresholds in BIBLE \dev. For languages that are not included in the BIBLE development set, we apply a relaxed similarity threshold of 0.25. This helps us filter out pairs that are clearly noisy or incorrect. We also remove pairs that have extreme source-to-target length ratios, after accounting for the expected length of each language.
    \item Given the origin of our data, with some sources being n-way parallel, there are numerous duplicates in either source or target sides. To address this, we will apply exact deduplication to both sides of the translation data.
    \item The provenance of our data is diverse, with many of our sources originating from synthetic generation. As a result, we will differentiate between `primary' translation data and `synthetic' data.
\end{itemize}

Data statistics are reported in \Cref{tab:appendix/data/seq2seq_contrastive}.
Finally, we did not apply any specific data filtering on ASR training data.

\paragraph{Upsampling.}
For text modality, in stages 1-3 of training, we sample according to the natural frequencies of the data in our data mix. For the omnilingual extension, we apply temperature-based sampling with a temperature of 0.6.
For ASR data, we follow an upsampling strategy to balance training data across domains and languages. To this end, we employ a two-step sampling procedure. First, for each data source, we sample the data for the $L$ different languages from a distribution
\begin{equation}
  \label{eq:betas}
  p_l \sim \left(\frac{n_l}{N}\right)^{\beta_{L}},
\end{equation}
where $l = 1, . . . , L$,~$n_l$ is the amount of unlabeled audio for each language in the current data source,~$N$ is the total amount of unlabeled audio in the current data source, and~$\beta_L$ is the upsampling factor which controls the trade-off between high- and low-resource languages during pre-training.  Second, we balanced the different data sources by treating each source as a language and applying the same sampling scheme with a sampling parameter~$\beta_D$.  In practice, we set both~$\beta_L$ and~$\beta_D$ to~0.5.  

\subsection{Hard Negatives Generation}
\label{section:data_processing/hard_negatives_generation}
We leverage both in-batch and hard negatives for contrastive training. Based on the intuition behind \citet{chen-etal-2023-xsim}, the optimal hard-negative for a translation pair is an approximate paraphrase of the original translation that incorporates a subtle or traditionally challenging semantic modification.
We synthetically generate these hard negatives using LLaMA3.3 70B Instruct. 
An example of a generated hard negative is provided in \Cref{tab:data_processing/hard_negatives_examples}.
For more details see \Cref{app:data_processing/hard_negatives}. 

\begin{table}[h]
    \centering
    \begin{tabular}{l|l}
        \toprule
        Original Sentence & Generated hard negatives \\
        \midrule
        \multirow{5}{*}{It was carrying a special mission unit} 
          & It was \colorbox{green!15}{not} carrying a special mission unit \\
          & It \colorbox{green!15}{is} carrying a \colorbox{green!15}{normal} mission unit \\
          & \colorbox{green!15}{She} was carrying a special mission unit \\
          & It \colorbox{green!15}{will be} carrying a special mission \colorbox{green!15}{team} \\
          & It was carrying \colorbox{green!15}{12} special mission \colorbox{green!15}{units} \\
        \bottomrule
    \end{tabular}
    \caption{Examples of natural language hard negatives.}
    \label{tab:data_processing/hard_negatives_examples}
\end{table}

\section{Model}
\label{section:model}

In this section, we present the \sonar{} model, detailing its methodology and training recipe. Our primary objective is to create a unified semantic space that achieves robust cross-lingual alignment and strong downstream performance across the entire linguistic spectrum. To solve the language scaling limitations inherent in massively multilingual models, as well as modality competition between text and speech, we employ a deliberate, progressive training strategy, as depicted in \Cref{fig:sonar2_method}.

This methodology is structured around three core milestones:
\begin{enumerate}
\item \textbf{The Foundational Space} (\sonarhundred): We begin by focusing on the 200 languages included in NLLB, which benefit from extensive data sources and well-established benchmarks. By leveraging an initialization from LLaMA3 (\Cref{subsection:model/Llama_init}), sequence-to-sequence pretraining (\Cref{subsection:model/seq2seq_pretraining}) and a novel combination of token-level translation objectives and split-softmax contrastive loss (\Cref{subsection:model/contrastive_training}), we establish a state-of-the-art foundational embedding space. This highly optimized model, referred to as \sonarhundred, serves as the robust anchor for our subsequent expansions.
\item \textbf{The Omnilingual Space} (\sonar): Building upon this foundational embedding space, we expand it to achieve true omnilinguality. We utilize a specialized two-stage teacher-student distillation framework using MSE and contrastive objectives to project thousands of additional language varieties into the \sonarhundred semantic space (\Cref{subsec:methodology_omni_expansion}). This process yields our unified omnilingual encoder, \sonar.
\item \textbf{The Cross-Modal Extension} (\sonar-Speech): Finally, we trained a massively multilingual speech encoder to project the speech modality into our omnilingual space (\Cref{subsection:model/speech_extension}).
\end{enumerate}

This section is organized into several subsections, each describing the key steps involved at every stage of development. We systematically cover the model architecture and initialization, the sequence-to-sequence pretraining phase, the contrastive embedding learning stages, and the subsequent omnilingual and speech extensions.

\subsection{Tokenizers}
\label{section:tokenizers}

Our model is based on LLaMA3, but since it is officially designed to support only eight languages, its default tokenizer is inherently limited in its multilingual capacity. To achieve our goal of building an omnilingual sentence encoder, we cannot rely solely on this English-centric vocabulary. Therefore, we developed two distinct tokenizers for the \sonar{} family, both employing a vocabulary size of 256k (effectively doubling the original capacity of the LLaMA3 tokenizer): a foundational tokenizer covering 200 languages (for \sonarhundred{}), and an expanded omnilingual tokenizer encompassing more than \sonarlanguages{} languages (for \sonar{}).

For both tokenizers, word frequencies were computed using a balanced sample drawn equally from our parallel training data (across all target languages), and a largely mutlilingual web dataset, similar to FineWeb2. To balance the language distribution, we used the total number of characters as sampling weights and applied unimax sampling \citep{unimax}. Specifically, we squashed the proportions of the top 126 highest-resource languages to a uniform distribution and upsampled the remaining long-tail languages by a maximum factor of 100. Additionally, we manually upweighted certain languages with underrepresented scripts (such as Greek and Korean) to properly adjust their resulting token fertilities. Finally, we observed that for several languages, the primary bottleneck for token fertility was not the vocabulary itself, but rather the pre-tokenization word-splitting regular expression. To address this, we expanded the regex pattern with additional Unicode ranges and a dedicated pattern for matching intra-word diacritic marks.

\paragraph{200-Language Tokenizer:} To build the vocabulary for \sonarhundred{}, we extended the original 128k LLaMA3 vocabulary to the new 256k capacity using a Byte-Pair Encoding (BPE) ``continued training'' algorithm. This process sequentially merged the most frequently occurring consecutive pairs of tokens based on our balanced language sample. As a result, our foundational tokenizer achieved an average fertility of 44 tokens per sentence across the 200 languages in the FLORES dataset, a substantial improvement over the 79 tokens per sentence produced by the original LLaMA3 tokenizer.\footnote{The most pronounced improvements were observed in Asian languages with unique scripts (e.g., \texttt{shn\_Mymr}, \texttt{sat\_Olck}, and \texttt{dzo\_Tibt}), where fertility decreased by a factor of more than six.}

\paragraph{Omnilingual Tokenizer:} In contrast to the foundational model, the omnilingual tokenizer covering over \sonarlanguages{} languages was trained entirely from scratch. Using the exact same balanced word frequencies and regular expression enhancements described above, we built a new 256k BPE vocabulary tailored specifically for the extreme long-tail of languages. This adaptation successfully reduced average token fertilities on the BIBLE \dev set from 57.7 to 50.3, while maintaining a strong fertility of 41 on the FLORES benchmark.

\subsection{Architecture and Initialization} \label{subsection:model/Llama_init}
\paragraph{Encoder.} We employ a transformer encoder \citep{attention_is_all_you_need} built upon the Llama-3.2-1B \citep{llama3} architecture and its pretrained weights. To adapt it for encoding, the causal attention natively used in Llama is replaced by bidirectional attention. Additionally, we prepend a special \texttt{[CLS]} token to each input sequence; this serves as the pooling token whenever fixed-size sentence representations are required. The encoder’s output corresponding to this token is then projected into a $d$-dimensional space to produce the final sentence embeddings.

\paragraph{Decoder.} To build the foundational \sonarhundred{} model (\Cref{subsection:model/seq2seq_pretraining,subsection:model/contrastive_training}), we couple the encoder with a decoder, resulting in a full encoder-decoder architecture. To enable the decoder to attend to the encoder's outputs, cross-attention blocks are integrated into the standard Llama architecture. These blocks utilize grouped-query attention with $n_h$ key-value heads, matching the configuration of Llama's self-attention blocks. Drawing inspiration from recent work \citep{behnamghader2024llm2vec,zhang2025encoder}, all decoder weights are initialized directly from Llama, with the newly added cross-attention weights initialized using Llama's pretrained self-attention weights. During \Cref{subsection:model/seq2seq_pretraining} the decoder attends to all tokens outputted by the encoder, and in \Cref{subsection:model/contrastive_training} the decoder attends to the pooled sentence representations from the encoder.

\paragraph{Embedding Layer.} Initializing our model with Llama 3 weights naturally constrains us to its original vocabulary, which was trained on only a small subset of our target languages. To enhance multilingual coverage, we expanded the vocabulary for our models as described in \Cref{section:tokenizers}. To initialize the embedding matrix for the newly introduced tokens, we first tokenize each new token using the original Llama 3 tokenizer. We then compute the mean of the resulting sub-token embeddings to generate a robust initial representation for the new token \citep{gee-etal-2022-fast, moroni-etal-2025-optimizing}.

\subsection{Sequence-to-Sequence Pretraining} 
\label{subsection:model/seq2seq_pretraining}

Prior to learning the fixed-size embedding space, we introduce a sequence-to-sequence (Seq2Seq) pretraining stage. This phase serves to warm up our encoder-decoder architecture on translation tasks (Stage 1 in \Cref{fig:sonar2_method}), and increase the language coverage of the model. Importantly, the encoder outputs are not pooled at this stage; instead, the decoder cross-attends to the full non-pooled sequence of the encoder's outputs.

The model is optimized using a standard token-level cross-entropy translation objective. Given a sequence of input tokens $\mathbf{x}$ and a sequence of target tokens $\mathbf{y} = (y_1, \dots, y_T)$, we minimize the negative log-likelihood:
\[
\mathcal{L}_{\text{translation}} = -\sum_{t=1}^{T} \log P(y_t \mid y_{<t}, \mathbf{x})
\]

During this pretraining stage, we jointly optimize translation tasks across more than 5,000 directions, encompassing natural text translations for all 200 base languages, alongside code and mathematical expressions. To ensure high data quality while retaining a substantial training volume, we apply the lightweight filtering strategy detailed in \Cref{section:data_processing/filtering}. Specifically, we discard synthetic data examples that fall below a similarity score threshold of $\mu(\text{scores}_{xy})-\sigma(\text{scores}_{xy})$, as well as all other data types scoring below $\mu(\text{scores}_{xy})-3\sigma(\text{scores}_{xy})$. This filtering is exclusively applied to natural language pairs, exempting code and math data. Additionally, for forward-translated data, we restrict the target language strictly to English.

\begin{figure}
   \centering
   \includegraphics[width=0.4\linewidth]{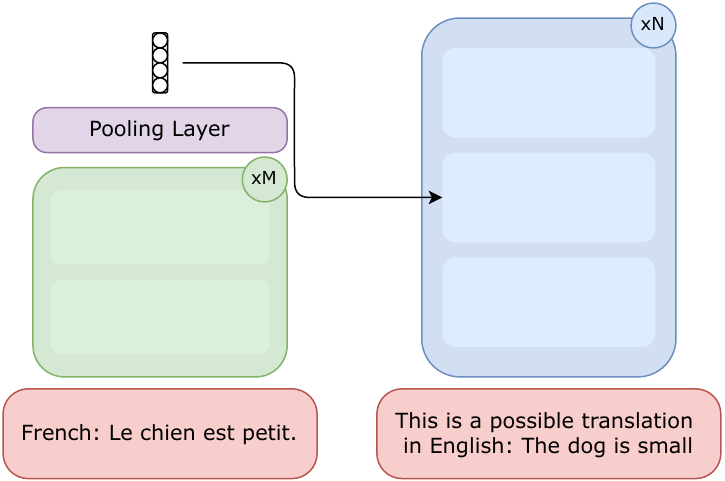}
   \caption{Example of \sonar{} natural text prompting on both encoder and decoder sides. Source sentences, which are input to the encoder, are prefixed with language identifiers. The format used is ``[language name]:''. Starting from the omnilingual extension, this prefix can be randomly replaced by ''Unspecified language:''.
Target sentences, provided to the decoder, include task specifications, output language information, and information about data provenance (indicating whether the translation is human-labeled, automatically mined, or back-translated), following \citet{nllb}.
Specifically, we use the prompts such as \texttt{This is a possible translation in [language name]:} for translation tasks and \texttt{This is a possible natural language explanation in English:} for code and math explanation tasks. This input formatting is applied consistently across all training stages.
We provide the full list of prompts in Appendix \Cref{tab:appendix/prompts}.}
   \label{fig:sonar2_prompting} 
\end{figure}

To enable effective multilingual and multitask processing, we employ natural text prompting for both the encoder and decoder inputs.
\Cref{fig:sonar2_prompting} shows how the prompts are used by the model.

\subsection{Sentence Representation Learning} \label{subsection:model/contrastive_training}

After completing the sequence-to-sequence pretraining, we fine-tune the model to learn the foundational \sonarhundred{} embedding space. This stage establishes robust cross-lingual alignment and strong downstream performance before we scale the model to omnilingual capabilities.

During this phase, we continue training the encoder-decoder architecture on translation tasks, but with a critical architectural shift: the encoder's outputs are now pooled to produce fixed-size sentence representations. Consequently, the decoder attends exclusively to these pooled embeddings rather than the full sequence of encoder outputs. We also introduce a contrastive learning objective that explicitly pulls true translation pairs closer together in the shared semantic space while pushing apart non-translation pairs.

This embedding learning stage is divided into two sequential phases, both employing joint contrastive and translation objectives:
\begin{itemize}
\item Contrastive learning relies solely on in-batch negatives.
\item Synthetic hard negatives are introduced alongside the in-batch negatives, significantly increasing the difficulty and diversity of the learning signal.
\end{itemize}

To maximize the effectiveness of the contrastive signal and the hard negatives, we restrict all training data in this stage to X-to-English directions. Because our initial experiments revealed that model performance plateaued early under standard data regimes, we transitioned to a strict filtering strategy that prioritizes data quality over sheer volume.

Building upon the filtering described in \Cref{section:data_processing/filtering}, we implement the following strict data curation rules:
\begin{itemize}
\item \textbf{Primary Sources Only:} We strictly exclude synthetic data, retaining only primary data sources.
\item \textbf{Aggressive Deduplication:} We globally deduplicate both source and target texts, ensuring that each sentence appears only once in the entire corpus—either as a source or a target. While this discards approximately 85\% of the data, it drastically reduces the incidence of false in-batch negatives during contrastive learning.
\item \textbf{Strategic Exceptions:} To prevent critical performance degradation in specific languages due to this data reduction, we introduce exceptions. For languages that exhibit weak initial cross-lingual similarity search performance (an \texttt{xsim++} error rate above 10) and possess fewer than one million training samples, we retain their original human-labeled NLLB data without deduplication.
\item \textbf{Code and Math:} For programming languages and mathematical expressions, we simply sample one million examples per translation direction.
\end{itemize}

\subsubsection{Translation and Contrastive Finetuning}
\label{subsection:model/contrastive_pretuning}

Translation and contrastive finetuning constitutes the second stage of our progressive training strategy (\Cref{fig:sonar2_method}). The weights of the encoder and decoder are initialized directly from the model obtained in the sequence-to-sequence pretraining stage (\Cref{subsection:model/seq2seq_pretraining}).

To achieve the intended cross-lingual alignment, we employ a Siamese network architecture coupled with our decoder. The encoder processes both the source and target sentences independently. The resulting representations corresponding to the \texttt{[CLS]} token are pooled to yield fixed-size sentence embeddings. These embeddings are then aligned using a contrastive loss function, which encourages the model to bring true translation pairs closer together in the shared semantic space.

Alongside this contrastive alignment, we continue to train the model on a translation task using a standard token-level cross-entropy objective, comparing the decoder’s predictions to the target sentences. However, unlike the previous pretraining stage where the decoder attended to the full unpooled encoder output, the decoder now attends exclusively to the pooled source representation. This introduces a strict informational bottleneck. Jointly training an embedding space with a decoder optimized via token-level losses has been shown to significantly enhance representation quality \citep{laser,sonar,janeiro-etal-2025-mexma}. The bottleneck compels the sentence representations to encode sufficient information for accurate decoding, resulting in embeddings that effectively balance semantic meaning with lexical content.

Our contrastive objective of \Cref{eq:model/contrastive_pretuning/contrastive_loss} leverages a modified InfoNCE loss \citep{infonce} , incorporating two key enhancements to improve the resulting embedding space.
The contrastive loss is defined as:
\begin{equation}
\mathcal{L}_\text{contrastive} = -\frac{1}{N} \sum_{i=1}^{N} \log \frac{e^{ \phi(\mathbf{x}_i, \mathbf{y}_i) - m }}{ e^{ \phi(\mathbf{x}_i, \mathbf{y}_i) - m } + \sum_{n \in \mathcal{S}_i} e^{ \phi(\mathbf{x}_i, \mathbf{y}_n) } }
\label{eq:model/contrastive_pretuning/contrastive_loss}
\end{equation}
\noindent where $\phi(\mathbf{x}_i, \mathbf{y}_i) = \tau \cdot \cos(\mathbf{x}_i, \mathbf{y}_i)$ denotes the scaled cosine similarity between the pooled sentence embedding of the source $\mathbf{x}_i$ and that of the target $\mathbf{y}_i$, with $\tau$ serving as a logit scaling hyperparameter.

First, following LaBSE \citep{feng-etal-2022-language}, we introduce an additive margin, $m \in \mathbb{R}$, that is applied to the similarity scores of source-target pairs. This margin mechanism actively encourages the representation of the correct translation to be distinctly separated from all other examples in the batch, thereby improving the model's discriminative power. 
Second, to mitigate the issue of false negatives when sampling from in-batch examples, we employ a rigorous negative filtering mechanism adapted from GISTEmbed \citep{solatorio2024gistembedguidedinsampleselection}. This is particularly crucial when mixing translations in a batch, as other valid translations or paraphrases in different languages might be inadvertently treated as negatives, which corrupts the distribution of the space. Specifically, the filtered set of negatives $\mathcal{S}_i$ for each source embedding $\mathbf{x}_i$ is defined as:
\begin{equation}
\mathcal{S}_{i} = \left\{\, j \in \{1, \ldots, N\}, j \neq i \;\middle|\; \phi(\overline{\mathbf{x}_i}, \overline{\mathbf{y}_j}) < r \cdot \phi(\overline{\mathbf{x}_i}, \overline{\mathbf{y}_i}) \,\right\}
\label{eq:model/contrastive_pretuning/negatives_filter}
\end{equation}
\noindent where $r$ defines the radius for negative removal, and $\overline{\mathbf{x}_i}$ and $\overline{\mathbf{y}_j}$ are guide embeddings generated by a frozen SONAR model \citep{sonar}. This filtering step removes any candidate negative whose similarity to the source exceeds that of the positive pair, ensuring the model does not erroneously penalize negatives that are semantically closer to the source than the true translation.

The overall training loss is the weighted sum of the contrastive and translation objectives:
\begin{equation}
\mathcal{L} = \alpha \cdot \mathcal{L}_\text{contrastive} + \beta \cdot \mathcal{L}_\text{translation}
\end{equation}
\noindent where $\alpha$ and $\beta$ are hyperparameters controlling the relative weight of each loss term.

\subsubsection{Translation and Contrastive Continued Finetuning with Hard Negatives} \label{subsection:model/contrastive_finetuning}

To further refine the model's ability to distinguish between highly similar sentences that lie close together in the representation space, we introduce a critical subsequent stage: continued finetuning with hard negatives (Stage 3 in \Cref{fig:sonar2_method}).

The generation of these synthesized hard negatives is detailed in \Cref{section:data_processing/hard_negatives_generation}. They are explicitly designed to act as challenging adversaries for the model, typically featuring only subtle lexical or syntactic alterations from the true translation. During our initial explorations, we discovered a crucial optimization dynamic: the additive margin ($m$) that significantly benefited in-batch contrastive learning proved detrimental when training with these synthesized hard negatives.

To successfully integrate both learning strategies without destabilizing training, we designed a novel combined objective: the \textit{split-softmax contrastive loss} ($\mathcal{L}_\text{contrastive\_hn}$). This formulation simultaneously optimizes the original margin-based contrastive loss over in-batch negatives (which continues to utilize the false-negative filtering defined in \Cref{eq:model/contrastive_pretuning/negatives_filter}) alongside a separate, non-margin-based contrastive term tailored specifically for the hard negatives. By decoupling the softmax denominators, we can precisely control the gradient influence of each negative set.

The resulting combined split-softmax contrastive loss is defined as:
\begin{equation}
\mathcal{L}_\text{contrastive\_hn} = (1-\gamma) \cdot \mathcal{L}_\text{contrastive} - \gamma \cdot \frac{1}{N} \sum_{i=1}^{N} \log \frac{e^{ \phi(\mathbf{x}_i, \mathbf{y}_i) }}{ e^{ \phi(\mathbf{x}_i, \mathbf{y}_i) } + \sum_{\mathbf{h}_j \in \mathcal{S}_{i}^{\text{HN}}} e^{ \phi(\mathbf{x}_i, \mathbf{h}_j) } }
\label{eq:model/contrastive_finetuning/contrastive_loss}
\end{equation}
\noindent where $\mathcal{L}_\text{contrastive}$ is the margin-based in-batch loss defined in \Cref{eq:model/contrastive_pretuning/contrastive_loss}, $\mathcal{S}_{i}^{\text{HN}}$ represents the set of hard negatives for the source embedding $\mathbf{x}_i$, $\mathbf{h}_j$ is the pooled embedding of a given hard negative, and $\gamma$ is a hyperparameter that weights the contribution of the hard negative objective.

The overall training loss for this stage is then adapted to incorporate this dual-objective contrastive loss alongside the translation objective:
\begin{equation}
\mathcal{L} = \alpha \cdot \mathcal{L}_\text{contrastive\_hn} + \beta \cdot \mathcal{L}_\text{translation}
\end{equation}

The outcome of this training stage yields our highly optimized foundational model for 200 languages including code and math, \sonarhundred{}.

\subsection{Omnilingual Extension} 
\label{subsection:model/omnilingual_extension}

After constructing the foundational embedding space covering 200 base languages, we expand its coverage to over \sonarlanguages{} languages by employing a unified teacher-student distillation framework \citep{mult_representation_distill,charsonar}. In this paradigm, the highly optimized \sonarhundred{} encoder serves as a frozen teacher, while a student model—initialized directly from the \sonarhundred{} encoder—is actively fine-tuned. Our training loop simultaneously processes all available languages of our training data including the ones supported by \sonarhundred and thousands of new ones, dynamically applying different loss configurations on a per-example basis depending on the source language type.

\subsubsection{Omnilingual Tokenizer Adaptation} \label{subsubsection:omni_extension/language_drop_tokenizer_adaptation}

The foundational tokenizer used by \sonarhundred{} natively supports only a fraction of the languages covered in our training data. To successfully extend the model to the extreme long-tail, we transition to the omnilingual tokenizer, as detailed in \Cref{section:tokenizers}.

Before injecting new languages into the embedding space, we first warm-up the student encoder to this updated tokenization scheme. This step disentangles the challenge of learning a new vocabulary representation from the challenge of learning new languages, thereby stabilizing and enhancing the subsequent omnilingual training. 

To achieve this adaptation, we minimize the Mean Squared Error (MSE) loss between the student and the frozen teacher's sentence embeddings using monolingual data exclusively for the languages supported by our foundation teacher space of \sonarhundred. The loss for a single example is defined as:
\begin{equation}
    \mathcal{L}^i_\text{MSE} = ||\mathbf{x}_i^\text{student} - \mathbf{x}_i^\text{teacher}||^2
\end{equation}
\noindent where $\mathbf{x}_i^\text{student}$ and $\mathbf{x}_i^\text{teacher}$ denote the pooled sentence embeddings produced by the student and teacher encoders, respectively, for the same input sentence. This adaptation ensures that the student encoder can faithfully reconstruct the \sonarhundred{} embedding space despite processing sequences with a radically different tokenizer.

Furthermore, as established in previous stages, we prepend a language identifier to the input sentence before encoding (e.g., \texttt{[Language Name]: \{Sentence\}}). However, explicitly knowing the source language becomes increasingly impractical for many low-resource and long-tail languages. To remove this dependency and improve the model's zero-shot robustness, we introduce a language-drop mechanism. With a probability $p_\text{unk} > 0$, we omit the true language identifier and instead prompt the student encoder with \texttt{Unspecified Language: \{Sentence\}}. In contrast, the frozen teacher encoder always receives the correct language prefix to ensure the generation of high-quality target embeddings.

\subsubsection{Omnilingual Extension Training} \label{subsec:methodology_omni_expansion}

The final stage of model development is the Omnilingual Extension Training (\Cref{fig:sonar2_method}), where we employ a teacher-student distillation approach to project the representations of over \sonarlanguages{} new language varieties into the foundational \sonarhundred{} embedding space. The student encoder is initialized directly from the tokenizer-adapted encoder described in \Cref{subsubsection:omni_extension/language_drop_tokenizer_adaptation}.

Because our method relies on extending the \sonarhundred{} teacher space, our translation data directions are strictly limited to those where the target language is one of the 200 foundational languages supported by the teacher. We filter out any training pairs that do not meet this criteria. Furthermore, to avoid degrading the structural integrity of the space, we ensure that we do not map high-performing languages to lower-performing ones during distillation. Specifically, we remove any translation direction where the target language's average representation quality (measured via xsim++ error rates in \sonarhundred{}) is worse than that of the source language. Finally, because English serves as the primary anchor language in \sonarhundred{}, we restrict English source data exclusively to monolingual examples (i.e., autoencoding) to preserve its central position in the semantic space.

For each training example $i$, we first determine the teacher's target embedding, $\mathbf{z}_i^\text{teacher}$, dynamically based on the source language type:
\begin{equation}
    \mathbf{z}_i^\text{teacher} = \begin{cases}
        \frac{1}{2}(\mathbf{x}_i^\text{teacher} + \mathbf{y}_i^\text{teacher}) & \text{if } \text{lang}(\mathbf{x}_i) \text{ is a foundational language} \\
        \mathbf{y}_i^\text{teacher} & \text{if } \text{lang}(\mathbf{x}_i) \text{ is a new language}
    \end{cases}
\end{equation}
\noindent where $\mathbf{x}_i^\text{teacher}$ and $\mathbf{y}_i^\text{teacher}$ are the frozen \sonarhundred{} embeddings of the source and target sentences, respectively. We utilize the interpolated embeddings for base languages as they provide a more stable training signal \citep{charsonar}. The student encoder then processes the source sentence to produce its embedding, $\mathbf{x}_i^\text{student}$.

We employ two complementary loss components to train the student encoder: a bidirectional contrastive loss and an MSE regularization loss.

\paragraph{Bidirectional Contrastive Loss.} This loss provides the primary cross-entropy learning signal, matching the student embedding $\mathbf{x}_i^\text{student}$ to its corresponding teacher target embedding $\mathbf{z}_i^\text{teacher}$, while simultaneously pushing it away from all other teacher embeddings in the batch. The forward component ($\text{student} \rightarrow \text{teacher}$), denoted as $\mathcal{L}^i_{\text{student} \rightarrow \text{teacher}}$, is particularly crucial for new languages, which often suffer from extreme low-resource conditions. It also mirrors the core training objective used for \sonarhundred{}, ensuring a smooth expansion without fundamentally altering the learning dynamics of the base languages. The reverse direction ($\text{teacher} \rightarrow \text{student}$), denoted as $\mathcal{L}^i_{\text{teacher} \rightarrow \text{student}}$, is also incorporated. The forward loss is implemented as a standard InfoNCE objective:
\begin{equation}
    \mathcal{L}^i_{\text{student} \rightarrow \text{teacher}} = - \log \frac{e^{\cos(\mathbf{x}_i^\text{student}, \mathbf{z}_i^\text{teacher}) \cdot \tau_i}}{\sum_{k=1}^{N} e^{\cos(\mathbf{x}_i^\text{student}, \mathbf{z}_k^\text{teacher}) \cdot \tau_i}}
\end{equation}
\noindent The per-example logit scale $\tau_i > 0$ controls the sharpness of the distribution; larger values create a smoother distribution, which can be easier to optimize under the limited training data conditions typical of new languages. This hyperparameter is thus set differently, depending on whether the language of example $i$ is foundational or new.

\paragraph{MSE Loss.} This loss acts as a geometric regularization term, ensuring that the student embeddings remain physically proximate to the teacher's target embeddings within the original \sonarhundred{} manifold.
\begin{equation}
    \mathcal{L}^i_{\text{MSE}} = ||\mathbf{x}_i^\text{student} - \mathbf{z}_i^\text{teacher}||^2
\end{equation}

The total training loss is the average of all individual example losses $\mathcal{L}_i$, which are computed as a dynamically weighted sum of the three components:
\begin{equation}
    \mathcal{L}^i = \lambda^{\text{s} \rightarrow \text{t}}_i \cdot \mathcal{L}^i_{\text{student} \rightarrow \text{teacher}} + \lambda^{\text{t} \rightarrow \text{s}}_i \cdot \mathcal{L}^i_{\text{teacher} \rightarrow \text{student}} + \lambda^{\text{MSE}}_i \cdot \mathcal{L}^i_{\text{MSE}}
\end{equation}
\noindent where the loss weights $\lambda^*_i \geq 0$ are dynamically adjusted based on the source language type (foundational versus new). This enables the fine-grained control over the training dynamics required to balance these two distinct language categories. The pseudocode detailing this dynamic weight configuration is provided in \Cref{sec:appendix/omni_algorithm}.

The highly capable, omnilingual model resulting from this final stage is our main encoder, \sonarlogoinline{}.

\subsection{Cross-Modal Speech Extension} \label{subsection:model/speech_extension}

To establish a truly comprehensive multimodal semantic space, we extend \sonar{} to the speech modality. This extension enables the extraction of fixed-size sentence embeddings directly from input speech utterances, mapping them natively into the omnilingual text space established in \Cref{subsection:model/omnilingual_extension}. 

Following the methodology introduced in the original SONAR \citep{sonar}, we employ a cross-modal teacher-student distillation framework. In this setup, our omnilingual text encoder, \sonar{}, serves as the frozen teacher, providing the target semantic representations derived from the gold transcriptions. The student is a newly introduced speech encoder tasked with matching these representations.

\paragraph{Architecture and Initialization.}
The student speech encoder is initialized using the pre-trained, massively multilingual wav2vec 2.0 model released by \citet{omni_asr}. This omnilingual acoustic model was pre-trained on approximately 4.3 million hours of unlabeled speech audio, covering more than \sonarlanguages{} languages. To compress the variable-length sequence of acoustic frames into a single fixed-size vector representation, we utilize an attention-pooling mechanism. Specifically, we use a three-layer transformer decoder that cross-attends to the outputs of the wav2vec 2.0 encoder.

\paragraph{Distillation Objective.}
The student speech encoder is trained by minimizing the Mean Squared Error (MSE) loss between its generated speech embedding and the frozen teacher's text embedding of the corresponding transcription. Given a speech utterance and its text transcription, the loss is defined as:
\begin{equation}
    \mathcal{L}^i_\text{MSE} = ||\mathbf{s}_i^\text{student} - \mathbf{x}_i^\text{teacher}||^2
\end{equation}
\noindent where $\mathbf{s}_i^\text{student}$ represents the attention-pooled embedding produced by the student speech encoder, and $\mathbf{x}_i^\text{teacher}$ is the fixed-size sentence embedding generated by the frozen \sonar{} text encoder for the target transcription.

\paragraph{Unified Multilingual Capacity.}
This approach yields a significant architectural advantage. Compared to the w2v-BERT 2.0 model \citep{seamless} used to initialize the speech encoder in the original SONAR, our chosen wav2vec 2.0 encoder covers 15 times more languages and provides empirically stronger multilingual acoustic representations \citep{omni_asr}. More importantly, while the original SONAR relied on dedicated, language-specific model checkpoints for speech processing, our \sonar{} speech encoder is entirely unified. A single model checkpoint handles all 177 supported spoken languages, projecting them seamlessly into the shared omnilingual semantic space.

\subsection{Decoder Finetuning}\label{subsection:model/decoder_finetuning}

The original SONAR architecture \citep{sonar} pairs an encoder with a decoder. While a decoder is not strictly required for extracting high-quality sentence embeddings, its presence enables the efficient decoding of fixed-size vector representations back into natural text across multiple languages. This capability is critical for emerging research paradigms, such as language modeling directly within sentence embedding spaces \citep{barrault2024large}, where predicted continuous representations must ultimately be mapped back to human-readable text.

Our foundational model, \sonarhundred{}, actively leverages a decoder during its training process, as detailed in previous sections. To optimize this decoder for downstream generative applications and explicitly adapt it to the expanded, omnilingual \sonar{} embedding space, we perform a dedicated finetuning stage. Specifically, we initialize the frozen encoder with the final omnilingual \sonar{} weights, and the active decoder with the weights from \sonarhundred{}. We then resume training the decoder using the identical translation loss function and data configurations established in \Cref{subsection:model/seq2seq_pretraining}.

\subsection{Smaller Models Distillation} \label{subsection:model/smaller_models_distillation}

To make our models more accessible to practitioners with varying computational resources, we explore the performance of smaller-scale variants of \sonar{}. A central design goal is to ensure these compact models can serve as drop-in replacements at any scale, allowing users to seamlessly switch between model sizes while maintaining compatibility with downstream components.
\paragraph{Model Pruning Strategy.} We create smaller models through structured pruning of the original 1.5B parameter encoder model. Our pruning approach encompasses multiple architectural dimensions: (1) reducing inner model dimensionality (from 2048 to 512-1792), (2) decreasing the number of encoder layers (from 16 to 8-14), (3) adjusting attention heads proportionally, and (4) scaling the feed-forward network dimensions accordingly. For layer selection, we employ a strategic sampling approach that preserves both the first and last layers while uniformly sampling intermediate layers, maintaining representational capacity across network depth. Furthermore, to reduce the number of parameters, we train new tokenizers with decreasing sizes (128k-16k) following \Cref{section:tokenizers}. Full architecture details are described in \Cref{tab:sonar2-encoder-configs}.
\paragraph{Knowledge Distillation.} Rather than training smaller models from scratch, we leverage knowledge distillation to ensure all model variants produce representations in the same aligned embedding space. We use the full 1.5B parameter \sonar{} model as the frozen teacher and train smaller student models using Mean Squared Error (MSE) loss on the output embeddings. This approach offers a critical advantage: any task-specific decoder or classifier trained on representations from one model size can be directly applied to representations from any other size, as all models produce semantically aligned embeddings in the same 1024-dimensional space. We employ the same setup as in \Cref{subsubsection:omni_extension/language_drop_tokenizer_adaptation}, using a monolingual setup covering the languages in \sonar-200.

\section{Experimental Configuration}\label{subsection:model/experimental_configuration}

\paragraph{Architecture.} Our \sonar{} architecture is based on Llama3.2-1B\footnote{\url{https://huggingface.co/meta-llama/Llama-3.2-1B}} \citep{llama3}. The encoder has 16 layers of dimensionality 2048, feed-forward dimension of 8192, swiGLU activations \citep{swiglu} and 32 attention heads. A linear layer down-projects the final encoder representation to 1024, which is then \texttt{CLS}-pooled to a single vector. Our vocabularies for the foundational space and for the omnilingual space both have a size of 256K tokens. The encoder total number of parameters is 1.5B. The decoder used during training for the translation loss, and for inference follows the same architecture, with a total of 1.8B parameters.
\paragraph{Sequence-to-Sequence.}
In this stage, we train our model for 100k steps, with 8192 tokens per GPU trained across 16 nodes of 8 GPUs each, with length bucketing.
The encoder and decoder are initialized from LLaMA3.2 1B. We trained the model with FSDP1~\citep{fsdp} and mixed precision in fp16, with a maximum gradient norm of 1.
We use the AdamW~\citep{adamw} optimizer with betas 0.9 and 0.98.
Our learning rate is set to 4e-4, with 2k warmup steps with an inverse square root learning rate scheduler. 

\paragraph{Translation and Contrastive Finetuning.}
Unless specified, the hyper-parameters are the same as the Sequence-to-Sequence configuration described above.
For this stage, we change the learning rate to 3e-4, the maximum number of tokens per GPU to 6k, and set the contrastive loss weight, $\alpha$, to 0.05, the translation loss weight, $\beta$, being 1.
We define our radius for false negatives removal, $r$, to 0.5, our margin, $m$ to 0.3 and our logit scale $\tau$ to 100.
Our model is trained for 10k steps.

\paragraph{Translation and Contrastive Finetuning with Hard Negatives.}
Unless specified otherwise, the parameters are the same as in the previous finetuning configuration.
We take 5 hard negatives per source sentence, and change the maximum number of tokens per GPU to 1.2k (6k/5).
The learning rate is changed to 1e-5, with 15k steps.
$\gamma$, the weight between the in-batch and the hard negative objectives, is defined as 0.8.

\paragraph{Omnilingual Extension.}
We use AdamW~(0.9, 0.98) with learning rate of 4e-5, with linear warm-up for 1k steps, and then reduced with a cosine annealing scheduler. Dropout is set to 0.05. We use length batching with size 16k tokens, and train with FSDP1 using 48 A100 GPUs, making the total batch 768k tokens (approx. 18k in terms of examples). Models are trained for 30k steps, which takes approximately 24 hours. For the vocabulary adaptation, we trained the model for 20k steps with a learning rate of 1e-4.

For the loss function, we use different configuration depending on whether the source language is foundational (part of the teacher \sonarhundred{} languages) or newly introduced. As we show in our ablations (\Cref{subsec:omni_ablations}), this is crucial to balance learning new languages without catastrophic forgetting.

\begin{table}[ht]
    \centering
    \resizebox{0.75\columnwidth}{!}{%
    \begin{tabular}{lcc}
        \toprule
        \textbf{Parameter} & \textbf{Foundational} & \textbf{New} \\
        \midrule
        MSE weight ($\lambda^\text{MSE}$) & $0.5$ & $0.1$ \\
        Student-Teacher Contrastive weight ($\lambda^\text{s $\rightarrow$ t}$) & $1.0$ & $1.0$ \\
        Teacher-Student Contrastive weight ($\lambda^\text{t $\rightarrow$ s}$) & $0.5$ & $0$ \\
        Logit scale ($\tau$) & $10$ & $60$ \\
        Teacher Embedding ($\mathbf{z}^\text{teacher}$) & Interpolated source-target$^\dagger$ & Target \\
        Language drop prob. ($p_\text{unk}$) & $0.25$ & $0.5$ \\
        \bottomrule
    \end{tabular}
    }
    \caption{Loss function and training parameter configurations depending on the source language type (foundational vs new) during the omnilingual extension. $^\dagger$When the source language is English, we use the source embedding of the teacher.}
    \label{tab:training_parameters}
\end{table}

\paragraph{Decoder Finetuning.}
We use same training setup as the Sequence-to-Sequence training stage except for learning rate which is set to 1e-3 and number of warmup steps which is lowered to 200.

\paragraph{Small Encoders Distillation.} We use a learning rate of 5e-4 for all model except tiny ones that use 1e-3, with linear warm-up for 1k steps, and cosine annealing for 64k steps. We set 16k maximum tokens per GPU, using 64 A100 GPUs, with an effective 768k maximum batch size.

\paragraph{Cross-modal Speech Extension.}
We train and release two sizes of \sonar{} speech encoder, one having 3B parameters and the other having 7B parameters.  The 3B one is initialized with the 3B wav2vec~2.0 model~(OmniASR-W2V-3B) from~\citet{omni_asr}, and the 7B one is initialized with the 7B wav2vec~2.0 model~(OmniASR-W2V-7B).  The \sonar{} speech encoder is unified across all supported languages, meaning it uses a single model checkpoint for every language. This differs from the original SONAR approach, where each language required its own dedicated model checkpoint. The checkpoint is selected according to the lowest average MSE loss across all languages on the validation set.  We train the 3B model with a learning rate of 1e-4, and the 7B model with a learning rate of 5e-4.  Both models are trained with the Adam optimizer for~200k steps using 128 A100 GPUs, with the first~8k steps as warm-up.  The effective batch size for training both encoders are 13.3 hours of speech.

\section{Results}
\label{section:results}
In this section, we present results obtained with \sonar{} on cross-lingual similarity search, several sentence-level downstream tasks, and decoding capabilities, for both the speech and text modalities.

\subsection{Cross-lingual Similarity Search}
We assess cross-lingual alignment by performing cross-lingual similarity search and compare \sonar{} to leading multilingual encoders. Cross-lingual similarity search evaluation on machine translation test sets involves comparing source sentence embedding to a pool of candidate translation embeddings. We then check whether the candidate embedding with the highest cosine similarity corresponds to the actual translation of the source sentence. The scores are reported as error rate, known as xsim, mining non-English sentences against their English translations. Additionally, we also report xsim++ \citep{xsimplusplus} evaluation on FLORES, which introduces English hard negative candidates for a more challenging assessment. We report results on FLORES200 \citep{nllb}, FLORES+, BOUQuET, AfroMT and BIBLE test sets.
\begin{table*}[ht]
        \centering
        \resizebox{0.98\textwidth}{!}{%
        \begin{tabular}{@{}lccccccccc@{}}
            \toprule
             & \multicolumn{2}{c}{\textbf{FLORES200}} & \multicolumn{2}{c}{\textbf{FLORES200}} & \multicolumn{2}{c}{\textbf{FLORES+}} & \textbf{BOUQuET} & \textbf{AfroMT} & \textbf{BIBLE}\\ 
            ($\#$ Languages) & \multicolumn{2}{c}{Common langs (80)} & \multicolumn{2}{c}{Full (201)} & \multicolumn{2}{c}{(212)} & (177) & (38) & (1,560) \\ 
            \cmidrule(lr){2-3} \cmidrule(lr){4-5} \cmidrule(lr){6-7} \cmidrule(lr){8-8} \cmidrule(lr){9-9} \cmidrule(lr){10-10}
            \textbf{Model} & \textbf{xsim} & \textbf{xsim++} & \textbf{xsim} & \textbf{xsim++} & \textbf{xsim} & \textbf{xsim++} & \textbf{xsim} & \textbf{xsim} & \textbf{xsim}  \\ \midrule
            \meFive{} & 0.3 & 22.3 & 7.5 & 34.9 & 28.4 & 71.8 & 31.8 & 29.9 & 72.4 \\
            \embGemma{} & 9.0 & 44.4 & 24.3 & 61.0 & 25.6 & 62.0 & 53.1 & 60.4 & 84.5 \\
            \embQwen{} & 13.3 & 60.4 & 27.0 & 71.1 & 28.4 & 71.8 & 54.8 & 65.5 & 85.4 \\
            \hline
            MEXMA & 0.1 & 7.8 & 15.9 & 35.8 & 17.5 & 37.5 & 39.3 & 58.7 & 70.2 \\
            LaBSE & 1.1 & 16.6 & 10.2 & 36.3 & 11.9 & 38.2 & 31.6 & 56.12 & 80.3\\
            SONAR & 0.2 & 9.9 & 1.4 & 15.3 & 3.3 & 19.2 & 32.2 & 33.4 & 68.7\\
            \sonarlogoinline{} & \textbf{0.1} & \textbf{3.0}  & \textbf{0.7}& \textbf{6.1} & \textbf{0.9} & \textbf{7.1} & \textbf{16.4} & \textbf{22.3} & \textbf{3.9}\\ \bottomrule
            \sonarlogoinline{} w/o Lang Tag  & 0.1 & 3.0 & 0.7 & 6.4 & 1.0 & 7.3 & 16.4 & 23.9 & 4.0
        \end{tabular}%
        }
        \caption{X-Eng cross-lingual similarity search error rates xsim/xsim++~($\downarrow$) on different multilingual test sets.}
        \label{tab:main_similarity_results}
    \end{table*}
In \Cref{tab:main_similarity_results}, we present the xsim and xsim++ results obtained for \sonar{} and several competitive baselines. On the 80 languages covered by all baseline models, the xsim++ error rate is reduced by more than 50\% compared to the best baseline model MEXMA on FLORES devtest set. When evaluating on the 200 languages covered by FLORES200, \sonar{} significantly outperforms SONAR, the previous state-of-the-art encoder, on both xsim and xsim++ metrics, also reducing the xsim++ error rate by more than 50\%. Moreover, cross-lingual evaluation on the Bible, highlights the omnilingual nature of our encoder, where it reduces the xsim error rates by 15$\times$, averaging an error rate of 3.9 across 1,560 languages. Finally, our results in three more multilingual benchmarks FLORES+, Bouquet and AfroMT confirm that our encoder is generalizable across domains.

\begin{table}[t!]
    \centering
    \resizebox{\textwidth}{!}{
    \begin{tabular}{l|cc|cccccccc}
        \toprule
        Model 
        & \makecell{GMMLU \\ all (41) } 
        & \makecell{GMMLU \\ common (30)} 
        & C%
        & C++%
        & C\#%
        & Java%
        & Javascript%
        & PHP%
        & Python%
        & All \\
        \midrule
        MEXMA   & 7.0 & 1.3 & 18.9 & 24.5 & 22.2 & 22.9 & 22.0 & 16.1 & 24.2 & 21.4 \\
        LaBSE   & 3.4 & 3.0 & 19.8 & 27.4 & 24.3 & 24.9 & 24.2 & 22.1 & 26.3 & 24.1 \\
        SONAR  & 3.2 & 3.0 & 22.0 & 29.4 & 28.3 & 29.4 & 26.0 & 22.2 & 30.8 & 26.9 \\
        \sonarlogoinline{}  & \textbf{1.2} & \textbf{1.0} & \textbf{15.4} & \textbf{19.5} & \textbf{18.2} & \textbf{18.1} & \textbf{16.0} & \textbf{12.1} & \textbf{17.3} & \textbf{16.6} \\
        \midrule
        \meFive{}    & 5.3 & 3.3 & 16.4 & 22.4 & 20.5 & 20.4 & 18.5 & 13.5 & 20.1 & 18.8 \\
        \embGemma{} & 16.5 & 5.9 & \textbf{15.4} & 20.9 & 18.9 & 19.3 & 16.1 & 12.2 & 17.6 & 17.2 \\
        \embQwen{} & 23.9 & 11.3 & 17.8 & 24.2 & 22.1 & 22.7 & 20.4 & 15.1 & 20.3 & 20.4 \\
        \midrule
        CodeSage-large-v2 & -- & -- & 19.4 & 23.0 & 21.2 & 21.5 & 18.2 & 15.5 & 20.4 & 19.9 \\
        CodeRankEmbed & -- & -- & 16.7 & 21.5 & 19.9 & 20.5 & 17.4 & 13.4 & 19.7 & 18.4 \\
        \bottomrule
    \end{tabular}
    }
    \caption{Results for GMMLU question mining (left) for all 42 languages and those covered by the baselines (common) and XLCOST (right). xsim ($\downarrow$) reported for all models.}
    \label{tab:results/mmlu_xsim_merged}
\end{table}
To further assess \sonar{} performance on more diverse domains, including mathematical expressions and programming languages, we evaluate on GMMLU paired questions \citep{singh2024globalmmluunderstandingaddressing} and XLCoST \citep{xlcost} Code2Code benchmark.

\Cref{tab:results/mmlu_xsim_merged} shows \sonar{} outperforms all systems on GMMLU. Notably, \sonar{} surpasses specialized code-embedding models like CodeSage \citep{zhang2024code} and CodeRankEmbed \citep{sureshcornstack} on XLCoST, excelling at code representation at snippet level even for unseen programming languages like C\#.

Finally, we evaluate cross-lingual similarity search for the speech modality on the FLEURS datasets.  For xsim and xsim++ on speech, the source speech embeddings are searched across candidate English text embeddings. \Cref{tab:speech_xsim} shows the result of xsim and xsim++ on 36 common languages that SONAR and \sonar{} speech models cover. For xsim, SONAR still yields the lowest error rate, but note that xsim is a metric that saturates quickly.  In fact, all models can achieve close~0 error rate over half of the languages tested, and the difference is only driven by a few languages. For xsim++, when augmenting the candidates with hard negative samples, \sonar{} then shows a significant advantage over SONAR, with a 7.6\% absolute improvement with the \sonar{}-3B model. We provide language breakdown results for xsim and xsim++ in \Cref{tab:xsim_speech_breakdown,tab:xsim_pp_speech_breakdown}.

\begin{table}[ht!]
\centering
\begin{tabular}{lcc}
\toprule
\textbf{}      & \textbf{xsim} & \textbf{xsim++} \\
\midrule
SONAR          & \textbf{0.1}       & 17.7            \\
\sonarlogoinline{}-speech-3B         & 0.4           & \textbf{10.1}   \\
\sonarlogoinline{}-speech-7B         & 0.8           & 11.6            \\
\bottomrule
\end{tabular}
\caption{Comparison of xsim and xsim++ results on FLEURS test set for the 36 languages covered by SONAR. \Cref{tab:xsim_speech_breakdown,tab:xsim_pp_speech_breakdown}.}
\label{tab:speech_xsim}
\end{table}

Overall, on cross-lingual similarity search, \sonar{} significantly outperforms previous approaches, from high-resource settings to extremely low resource ones, while covering new domains like code and math. This sets a new state of the art for cross-lingual text/speech sentence encoders.

\begin{table}[b!]
    \centering
    \begin{tabular}{l|c|c|c|c}
        \toprule
        Model & Average  & 	Classification &	Pair Classification  &	STS\\
        \midrule
        MEXMA	& 67.984 & 65.197 & 63.078 & 75.678\\
        LaBSE   & 66.782 & 61.502 & 64.662 & 74.182\\
        SONAR   &  63.338 & 58.479 & 60.712 & 70.824 \\
        \sonarlogoinline{}  & \textbf{74.114} & \textbf{71.143} & \textbf{69.037} & \textbf{82.163}\\
        \midrule
        \multicolumn{3}{l}{General-purpose models} \\
        \meFive{} & 73.234 & 63.672 & 73.527 & 82.502\\
        \embGemma{} & 77.057 & 68.108 & 78.644 & 84.418\\
        \embQwen{} & 75.263 & 64.707 & 76.358 & 84.724 \\
        \bottomrule
    \end{tabular}
    \caption{Classification and Pair Classification results from sentence-level MTEB tasks.}
    \label{tab:results/mteb_results}
\end{table}

\subsection{Downstream Tasks}
To assess the quality and generalization of our embeddings we evaluate them on several multilingual classification, pair classification and STS benchmarks under MTEB \citep{muennighoff-etal-2023-mteb}. Results are reported in \Cref{tab:results/mteb_results}.
\paragraph{Classification.}
The reported metric for classification is accuracy.
Under this setup, linear classifiers are trained on top of each model's embeddings on a held-out portion of the data, and evaluated on the rest.
Each classifier is trained and evaluated per language in this section.
Our reported numbers are first averaged over all languages in each benchmark and then over all benchmarks to create a single score.
\Cref{tab:results/mteb_results} shows how \sonar{} significantly outperforms all other models in classification tasks. This highlights the strong quality of the content captured by each individual vector. Furthermore, as we will explore in later sections, these vectors exhibit excellent interoperability across languages.
\paragraph{Pair Classification.}
This task classifies a pair of sentences, e.g. if a pair of sentences are duplicates or not based on the similary between the pair.
We report the average precision based on the cosine similarity between sentence pairs.
In this evaluation, \sonar{} continues to outperform other multilingual embedding models within its category. However, it falls short compared to the topline results of general-purpose embedding models. It is important to highlight that all our baselines along with \sonar{} are trained exclusively on translation parallel data, i.e. no task specific data is used, and the cosine distance between sentences reflects just that aspect.
\paragraph{Semantic Textual Similarity.} The STS task evaluates the model’s ability to replicate human judgments on sentence similarity.
We report the Spearman correlation based on distance.
It is possible to see that \sonar{} again outperforms all models with the exception of general-purpose models, with large margins.
Interestingly, all previous models have similar performance in STS, where it seems that translation data being the common factor across all models is a limiting factor, however \sonar{} with its improved training is able to stand out and shorten the gap with general-purpose models trained with task specific data.

\subsection{Decoding Capabilities}

Decoding sentence embeddings back into natural text provides a way to measure the text compression capabilities of an embedding model across different languages. However, the quality of the decoded results depends not only on the sentence embeddings themselves, but also on the training and capacity of the decoder. Additionally, models that generate sentence embeddings, such as Large Concept Models \citep{barrault2024large}, depend on robust text decoders to produce accurate and fluent text in multiple languages.

Therefore, we present translation results on several multilingual translation benchmarks in \Cref{tab:main_translation_results}, using \sonar{} decoder, as measured by \chrf \citep{popovic-2017-chrf} and xCOMET\footnote{\url{https://huggingface.co/Unbabel/XCOMET-XL}}~\citep{xcomet}. As the vast majority of source languages we evaluate on translation are not supported by xCOMET, we use it on reference-only mode. We compare our translation model, which only attends to the pooled representation, and not the full encoder sequence, with state-of-the-art MT systems and LLMs: NLLB~\citep{nllb}, Tower+~\citep{towerplus}, MADLAD~\citep{madlad}, Aya101~\citep{aya}, Gemma3~\citep{gemma3}, and Llama3~\citep{llama3}. Our results demonstrate the semantic richness of the \sonar{} embedding space, where it outperforms previous the SONAR sentence encoder-decoder model, and more importantly MT models and LLMs that are up to 30$\times$ larger in terms of parameters, on all evaluated benchmarks. The gap with previous models is particularly large in the Bible benchmark, which is our most multilingual test set, where \sonar{} outperforms Gemma3-27B and Llama3.3-70B by 15 \chrf points.
    
\begin{table*}[ht]
        \centering
        \resizebox{1.0\textwidth}{!}{%
        \begin{tabular}{@{}lcccccccccc@{}}
            \toprule
             & \multicolumn{2}{c}{\textbf{FLORES200}} & \multicolumn{2}{c}{\textbf{FLORES+}} & \multicolumn{2}{c}{\textbf{BOUQuET}} & \multicolumn{2}{c}{\textbf{AfroMT}} & \multicolumn{2}{c}{\textbf{BIBLE}} \\ 
            \cmidrule(lr){2-3} \cmidrule(lr){4-5} \cmidrule(lr){6-7} \cmidrule(lr){8-9} \cmidrule(lr){10-11}
            \textbf{Model} & \textbf{\chrf} & \textbf{xCOMET} & \textbf{\chrf} & \textbf{xCOMET} & \textbf{\chrf} & \textbf{xCOMET} & \textbf{\chrf} & \textbf{xCOMET} & \textbf{\chrf} & \textbf{xCOMET}  \\ \midrule
            SONAR & 52.9 & 0.817 & 31.0 & 0.450 & 40.3 & 0.705 & 34.8 & 0.619 & 18.7 & 0.315 \\
            \multicolumn{11}{l}{\textit{Translation Models and LLMs}} \\
            NLLB-600M & 52.4 & 0.799 & 31.4 & 0.441 & 42.1 & 0.726 & 38.9 & 0.665 & 21.8 & 0.335 \\
            NLLB-3B & \textbf{55.8} & 0.849 & 34.2 & 0.485 & 44.3 & 0.741 & 41.7 & 0.699 & 24.3 & 0.377  \\
            Tower+-9B & 45.1 & 0.652 & 38.3 & 0.531 & 38.4 & 0.644 & 26.7 & 0.456 & 19.4 & 0.286\\
            MADLAD-10B & 49.9 & 0.767 & 39.0 & 0.594 & 42.0 & 0.726 & 34.2 & 0.619 & 20.7 & 0.346\\
            Aya101-13B & 47.7 & 0.765 & 39.5 & 0.620 & 41.6 & 0.737 & 31.4 & 0.589 & 21.5 & 0.350\\
            Gemma3-27B & 52.4 & 0.778 & 45.7 & 0.662 & 45.5 & 0.745 & 33.7 & 0.585  & 26.3 & 0.339\\
            Llama3.3-70B & 51.0 & 0.758 & 44.8 & 0.678 & 43.8 & 0.723 & 32.3 & 0.547  & 26.2 & 0.342 \\ \midrule
            \sonarlogoinline{} & 55.4 & \textbf{0.878} & \textbf{46.1} & \textbf{0.746} & \textbf{46.0} & \textbf{0.797} & \textbf{44.0} & \textbf{0.739}   & \textbf{41.3} & \textbf{0.702} \\
            \bottomrule
        \end{tabular}%
        }
        \caption{X-Eng translation quality \chrf~($\uparrow$) / xCOMET($\uparrow$) on different multilingual test sets. FLORES+ reports only the 11 languages not in FLORES200.}
        \label{tab:main_translation_results}
    \end{table*}

\cite{duquenne2022t,duquenne2023modular} showed that it's possible to efficiently decode speech sentence embeddings into text to perform zero-shot speech-to-text translation. We follow such approach for \sonar{}, by first encoding speech utterances into fixed-sized sentence embeddings using \sonar{}-speech encoder, and then decode them with \sonar{} pre-trained text decoder. The \sonar{} text decoder was only trained to decode text sentence embeddings, while the \sonar{}-speech embeddings are distilled from the \sonar{} text embeddings.  This way, one can assess how much content information can be extracted from speech representations by a pre-trained text decoder.

We report Speech Translation performance of \sonar{}-speech by decoding speech embeddings into English text on the FLEURS test set.
In \Cref{tab:speech_xeng}, we compare the X-eng translation performance of \sonar{}-speech to SONAR and other models trained on speech-to-text translation tasks, including SeamlessM4T, and Omni-ASR. Translation results are evaluated with BLEU, using the signature in~\cite{seamless}.
Comparing the 73 languages in FLEURS covered by all models, we can see that \sonar{}-speech shows a significant improvement over SONAR, and using a 7B encoder leads to a 1\% absolute gain on the BLEU score over the 3B encoder model. Compared to other models trained on speech translation tasks, \sonar{}-speech is on par with Omni-ASR and only 1.2\% behind SeamlessM4T. When evaluated on all 101 languages in FLEURS, the gap between \sonar{} and SeamlessM4T decreases, and \sonar{}-speech-7B even yields better performance than Omni-ASR on average, across all 101 languages. 

We again highlight that the \sonar{}-speech encoder is trained only on ASR data, so speech-to-text translation is actually a zero-shot task.  This shows the strong encoding capability of the \sonar{}-speech encoder and that the speech embeddings are well-distilled from the text embeddings.

\begin{table}[hbtp!]
\centering
\begin{tabular}{lcc}
\toprule
\textbf{Encoder}      & \textbf{Full (101 languages)} & \textbf{Common (73 languages)} \\
\midrule
SONAR                & --                 & 21.7              \\
SeamlessM4T           & 22.0               & 25.6              \\
Omni-ASR 7B            & 20.8               & 24.5              \\
\midrule
\sonarlogoinline-speech-3B             & 20.4               & 23.4              \\
\sonarlogoinline-speech-7B             &21.4             & 24.4              \\
\bottomrule
\end{tabular}
\caption{Speech Translation (X-Eng) BLEU score on FLEURS, comparing \sonar{} with SONAR and other Speech Translation models.}
\label{tab:speech_xeng}
\end{table}

\subsection{Smaller Encoders Performance}
\label{subsec:distilled_encoders_performance}
\Cref{tab:smaller_models_distillation_results} summarizes the performance of our distilled models on cross-lingual similarity search, as measured by xsim and xsim++ on both the FLORES200 common (80) and full (201) language sets.
\begin{table*}[h!]
    \centering
    \resizebox{\textwidth}{!}{%
    \begin{tabular}{@{}lc|cc|cc|cccc@{}}
        \toprule
        & & \multicolumn{2}{c|}{FLORES Common (80)} & \multicolumn{2}{c|}{FLORES Full (201)} & \multicolumn{4}{c}{MTEB} \\
        \textbf{Model} & \textbf{Size} & \textbf{xsim} & \textbf{xsim++} & \textbf{xsim} & \textbf{xsim++} & \textbf{Avg} & \textbf{Class.} & \textbf{Pair} & \textbf{STS} \\
        \midrule
        \sonarlogoinline{} (Large) & 1.5B & 0.05 & 2.95 & 0.65 & 6.14 &  74.11 & 71.14 & 69.04 & 82.16 \\
        Medium & 884M & 0.05 & 3.02& 0.82 & 6.61 & 73.90 & 70.74 & 69.83 & 81.12 \\
        Small & 511M & 0.06 & 3.03 & 0.81 & 6.64 & 73.94 & 70.78 & 69.87 & 81.17 \\
        Tiny & 233M & 0.06 & 3.51 & 1.01 & 7.82 & 73.77 & 70.16 & 69.97 & 81.18 \\
        xTiny & 39M & 0.10 & 7.96& 2.70 & 16.40 & 71.70 & 65.51 & 70.89 & 78.69 \\
        \bottomrule
    \end{tabular}%
    }
    \caption{Results of smaller distilled models on cross-lingual similarity search (FLORES200) and downstream tasks (MTEB). xsim/xsim++ report error rates ($\downarrow$) and MTEB scores are higher-is-better ($\uparrow$).}
    \label{tab:smaller_models_distillation_results}
\end{table*}

Despite their reduced parameter counts, the smaller encoders preserve a large fraction of the full model’s capabilities while offering substantial computational savings.
Relative to the \sonar{} Large encoder (1.5B parameters), the Medium (884M) and Small (511M) models retain over 90\% of the full model’s performance on xsim++ across the full 201-language set.
Even the Tiny model (233M) achieves approximately 78\% of the Large model’s xsim++ score, demonstrating that effective cross-lingual representations can be maintained at significantly reduced model sizes.
This ensures that the smaller encoders can be used instead of the larger ones as a drop-in replacement in compute constrained scenarios with a minimal performance degradation, enabling the deployment of our models in all settings. Notably, the MTEB downstream performance degrades gracefully with model size. The Medium, Small, and Tiny variants all achieve average scores above 73.7, compared to 74.1 for the full model, a drop of less than 0.5 points despite 3--6$\times$ parameter reduction. The xTiny model (39M parameters) shows more noticeable degradation on classification tasks (65.51 vs.\ 71.14), but still maintains competitive performance overall (71.70 average). %

\section{Ablations}
\label{appendix:other_ablations}

\begin{table}[h!]
    \centering
    \begin{subtable}[t]{0.46\textwidth}
        \centering
        \begin{tabular}{lcc}
            \toprule
            Model & xsim & xsim++ \\
            \toprule
            LLaMA initialization & 94.57 &	99.89 \\ %
            Seq2Seq pre-training & 7.74 & 51.55 \\ %
            Contrastive Loss & 0.71 & 16.23 \\ %
            ~~~+ Decoder Loss & 0.65 & 8.95 \\ %

            ~~~~~~+ Hard negatives  & 0.76 & 7.06 \\ %

            \bottomrule
        \end{tabular}
        \caption{Full method ablation.}
        \label{tab:training_objectives/full}
    \end{subtable}
    \hspace{0.02\textwidth}
    \begin{subtable}[t]{0.5\textwidth}
        \centering
        \begin{tabular}{l|c|c}
            \toprule
            Model & xsim & xsim++ \\
            \toprule
            Decoder + MSE losses & 0.92 & 12.54 \\ %
            Decoder + Contrastive losses& 0.65 & 8.95 \\ %

            \bottomrule
        \end{tabular}
        \caption{Cross-lingual alignment objectives ablation.}
        \label{tab:training_objectives/crosslingual}
    \end{subtable}
    \caption{\textbf{Training objectives ablations}: Ablations on training objectives to learn a massively multilingual sentence embedding space on the cross-lingual similarity search task of FLORES200 dev set, as measured by xsim and xsim++.}
    \label{tab:training_objectives}
\end{table}
\label{section:analysis_ablations}

\sonar{} includes several novel design choices supported by strong performance. In this section we provide ablations for such choices in an incremental fashion, that lead to our final model reported in \Cref{section:results}. All ablations experiments are trained for 5k steps only, on an older, less filtered version of the data. 

\subsection{Training Objectives}
\sonar{} follows a multi-stage training strategy described in \Cref{section:model}. Some of these steps such as decoding loss for sentence embedding learning~\citep{sonar}, LLM re-purposing as an Encoder-Decoder~\citep{zhang2025encoder}, and contrastive learning have been explored in isolation in prior work, but \sonar{} is the first system to train an embedding model with such training strategies in a unified framework. Here, we analyze the contribution of each component to the final performance.

As shown in \Cref{tab:training_objectives/full}, each training stage yields significant improvements. After Seq2Seq pre-training, the representations are not yet optimized for sentence-level tasks, and mean-pooling over all tokens results in suboptimal performance. Nevertheless, we will later show the impact of this step as a foundation for subsequent contrastive training.

A key distinction between \sonar{} and other embedding models built on modern LLMs is the inclusion of a Decoder component. While contrastive learning alone achieves a modest xsim score, it falls short on xsim++. The addition of the cross-entropy loss from the Decoder, with its token-level language modeling signal, delivers the largest gains, highlighting its role in capturing semantic nuance beyond surface-level features. The introduction of hard negatives further reduces xsim++ scores.%

SONAR \citep{sonar} successfully leveraged a Decoder to build sentence representations. 
However, their approach combined a Mean Squared Error (MSE) objective between source and target embeddings with the translation objective. In \Cref{tab:training_objectives/crosslingual}, we show that replacing the MSE objective with a contrastive loss, as described in \Cref{subsection:model/contrastive_pretuning}, leads to a substantial improvement. This result suggests that the contrastive signal encourages a more structured embedding space by explicitly pushing apart negatives, which benefits xsim/xsim++ and helps prevent embedding space collapse.

\subsection{Contrastive Signals}
Training embedding models with Contrastive Learning requires careful choices of hyper-parameters. We analyze the effect of these options on the cross-lingual similarity search results in \Cref{tab:contrastive_ablations}. 

\begin{table}[t!]
    \centering
    \begin{subtable}[t]{0.45\textwidth}
    \centering
    \begin{tabular}{lccc}
        \toprule
         & & xsim & xsim++ \\
        \midrule
         \textbf{Margin} &  0 & 0.74 & 9.49 \\ %
           &  0.3 & 0.65 & 8.95 \\ %
          &  0.5 & 0.72 & 9.45 \\ %
        \midrule
          \textbf{Logit scale} & 1 & 1.88 & 11.90 \\ %
          &  100 & 0.65 & 8.95 \\ %
        & 150 & 0.66 & 9.07 \\ %
        \midrule
        \textbf{Gathering negatives } &  no & 0.74 & 9.44 \\
          &  yes & 0.65 & 8.95 \\ %
        \midrule
         \textbf{False negative removal} &  no & 0.69 & 9.73 \\ %
           & yes & 0.65 & 8.95 \\ %
        \bottomrule
    \end{tabular}
    \caption{Contrastive Learning hyper-parameters ablations}
    \end{subtable}
    \hspace{0.1\textwidth}
    \begin{subtable}[t]{0.4\textwidth}
        \centering
    \begin{tabular}{lcc}
        \toprule
          & xsim & xsim++ \\
         \midrule
         \multicolumn{3}{l}{\textit{\textbf{In-batch negatives only}}}\\
        ~~~One softmax & 0.65 & 8.95 \\
         \midrule
         \multicolumn{3}{l}{\textit{\textbf{In-batch + hard negatives}}}\\
        ~~~One softmax & 0.94 & 7.00 \\ %
        ~~~Split softmax & 0.76 & 7.06 \\ %
         \bottomrule
    \end{tabular}
    \caption{Ablation on the use of hard negatives in contrastive learning.}
    \end{subtable}
    \caption{\textbf{Contrastive Learning ablations:} Effect of hyper-parameters and modeling options in Contrastive Learning on cross-lingual similarity search on FLORES200 dev set.}
    \label{tab:contrastive_ablations}
\end{table}
The additive margin in the softmax improves separation between positive translations and negatives. A value of $m=0.3$ was empirically found as best for this hyper-parameter, boosting performance compared to models trained without margin. We also explore the logit scale on cosine similarity, $\tau$, and find 100 to be the best and crucial for proper contrastive learning.

The choices of negative examples is also key. By default we use all other sentences from the batch as negatives, commonly referred to as in-batch negatives. We analyze the effect of different choices of negative examples in \Cref{tab:contrastive_ablations}. First in sub-table (a), we gather negative sentence examples from other GPUs, significantly increasing the number of negatives, by a factor of number of GPUs, which in our case was 128. Such approach indeed helps reaching lower cross-lingual similarity search error rates. The increasing number of negative examples comes also at the price of higher probability of considering false negative sentences in the loss. We ablate the use of false negative removal heuristic presented in \Cref{subsection:model/contrastive_pretuning}, and validate the usefulness of such approach.

Finally, in sub-table (b), we extend the in-batch negatives with the hard negatives presented in \Cref{subsection:model/contrastive_finetuning}, either using a single contrastive learning task (one softmax) for both in-batch and hard negatives, or two contrastive learning tasks (split softmax). %
The first interesting finding is that training a model using hard negatives with a non-zero margin does not converge correctly. Therefore, we do not use any margin in the ``one softmax'' setup. This leads us to use $m=0.3$ for in-batch negatives and $m=0$ for hard negatives in the ``split softmax'' setup. We notice that hard negatives significantly lower xsim++ error rates. However, not separating the hard negatives from in-batch negatives in two different contrastive loss terms affects xsim performance. This highlights the benefits of having two contrastive learning losses, one for in-batch negatives and another for hard negatives, to better balance the two in the final loss.

\begin{table}[ht]
    \centering
    \begin{subtable}[t]{0.45\textwidth}
    \centering
    \begin{tabular}{cc}
        \toprule
        Initialization  & \chrf \\
        \midrule
        Random & 36.55 \\ %
        LLaMA & 42.71 \\ %
        \bottomrule
    \end{tabular}
    \caption{Ablation on model initialization for the Seq2Seq stage.}
    \end{subtable}
    \hspace{0.05\textwidth}
    \begin{subtable}[t]{0.45\textwidth}
    \centering
    \begin{tabular}{ccc}
        \toprule
         Initialization & xsim & xsim++ \\
         \midrule
         Random init. & 13.35 & 71.30 \\ %
         LLaMA init. & 1.02 &  11.98 \\ %
         Seq2Seq init. & 0.65 & 8.95 \\ %
         \bottomrule
    \end{tabular}
    \caption{Ablation on model initialization for Contrastive Learning stage.}
    \end{subtable}
    \caption{\textbf{Model initialization ablations}: Effect of model weight initialization for sequence-to-sequence stage as well as for contrastive learning stage on respectively decoding performance (spBLEU and \chrf) and cross-lingual similariy search (xsim and xsim++) on FLORES200 dev set.}
    \label{tab:ablations/initialization}
\end{table}
\subsection{Model Initialization}
To understand the benefits of initializing from LLaMA, we perform an ablation study by starting the Seq2Seq stage from either LLaMA or random initialization. The results of this analysis are presented in \Cref{tab:ablations/initialization}a.
The table shows that initializing from LLaMA yields significant improvements in \chrf scores compared to random initialization. Although LLaMA officially supports only 8 languages, extending it to 200 languages is still considerably easier than training a model from scratch.

To further investigate the advantage of performing an initial Seq2Seq step to adapt LLaMA for multilingual encoding and decoding, we initialize our contrastive step from LLaMA, from the Seq2Seq model, and from random initialization.
The results are shown in \Cref{tab:ablations/initialization}b, where we observe substantial improvements in xsim and xsim++ scores when initializing from the Seq2Seq model rather than from LLaMA.

\subsection{Omnilingual Extension Ablations and Analysis}
\label{subsec:omni_ablations}

Extending the embedding space of \sonar{}-200 to thousands of languages requires carefully balancing several learning signals to avoid sacrificing performance on the 200 foundational languages already covered by the teacher encoder. Here, we present the ablations that motivate our architectural and training choices. Specifically, we optimize for two competing objectives:
\begin{enumerate}
    \item \textbf{Learning New Languages:} Assessed primarily by measuring the any-to-English cross-lingual similarity (xsim) on the Bible, our most massively multilingual benchmark (1,560 languages).
    \item \textbf{Preserving the Foundational Space:} Assessed by evaluating zero-shot translation quality (\chrf) on both the BIBLE and FLORES benchmarks using the frozen \sonar{}-200 decoder. This guarantees that new languages successfully map into the original manifold, while simultaneously ensuring that the representations of the 200 foundational languages remain perfectly aligned and decodeable.
\end{enumerate}

\begin{figure}[ht]
    \centering
    \begin{subfigure}[b]{0.33\textwidth}
        \centering
        \includegraphics[width=\textwidth]{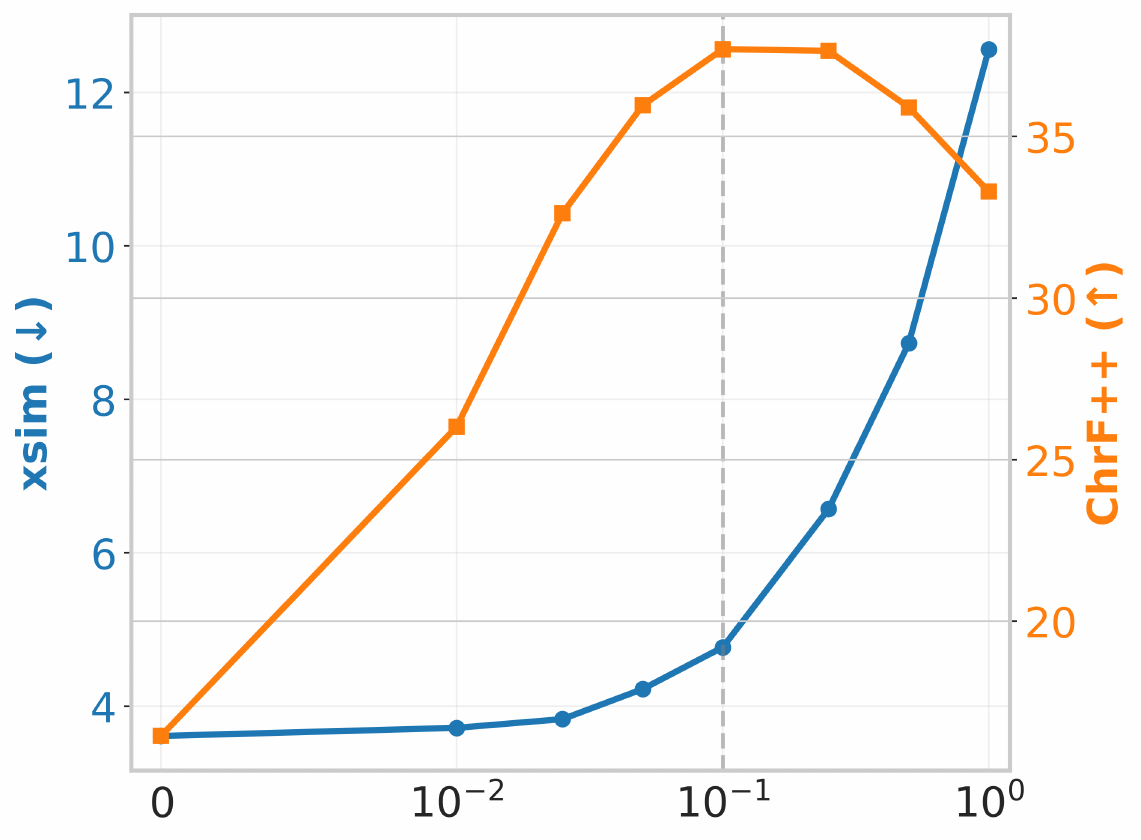}
        \caption{MSE Loss Weight}
        \label{fig:mse_weight}
    \end{subfigure}
    \hfill
    \begin{subfigure}[b]{0.325\textwidth}
        \centering
        \includegraphics[width=\textwidth]{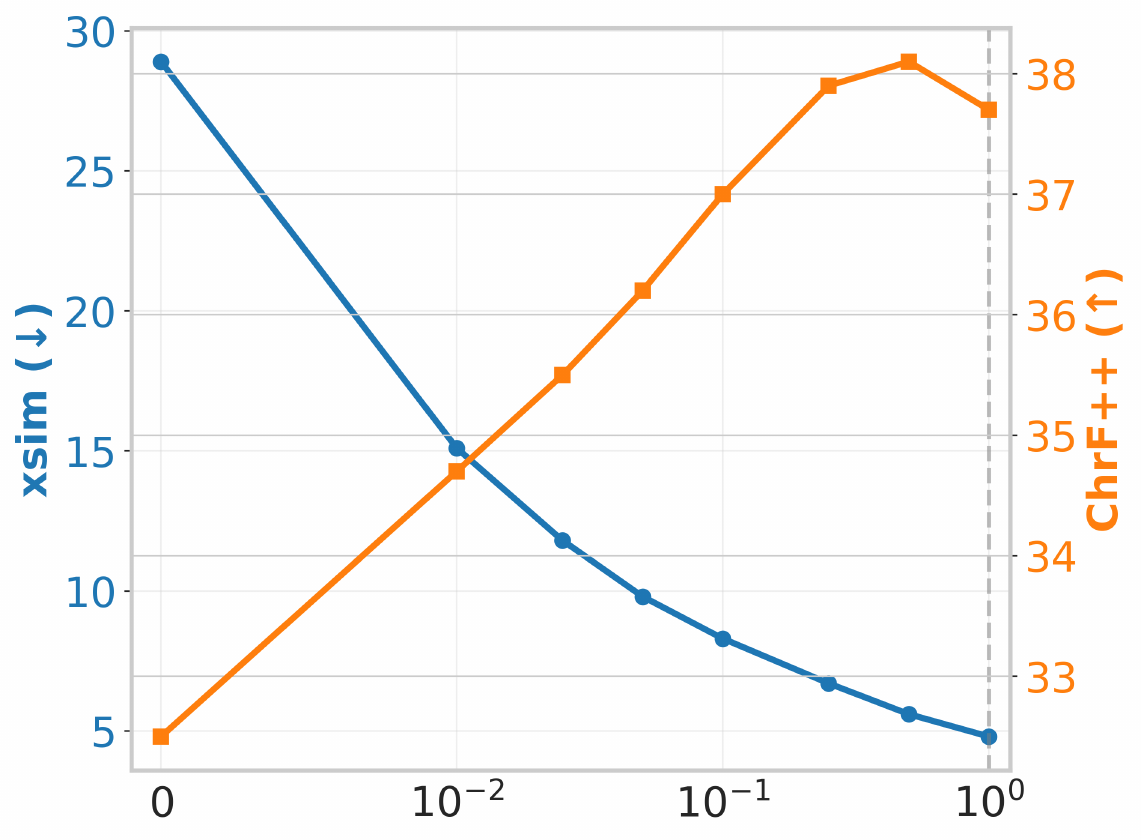}
        \caption{Contrastive Loss Weight}
        \label{fig:fc_weight}
    \end{subfigure}
    \hfill
    \begin{subfigure}[b]{0.33\textwidth}
        \centering
        \includegraphics[width=\textwidth]{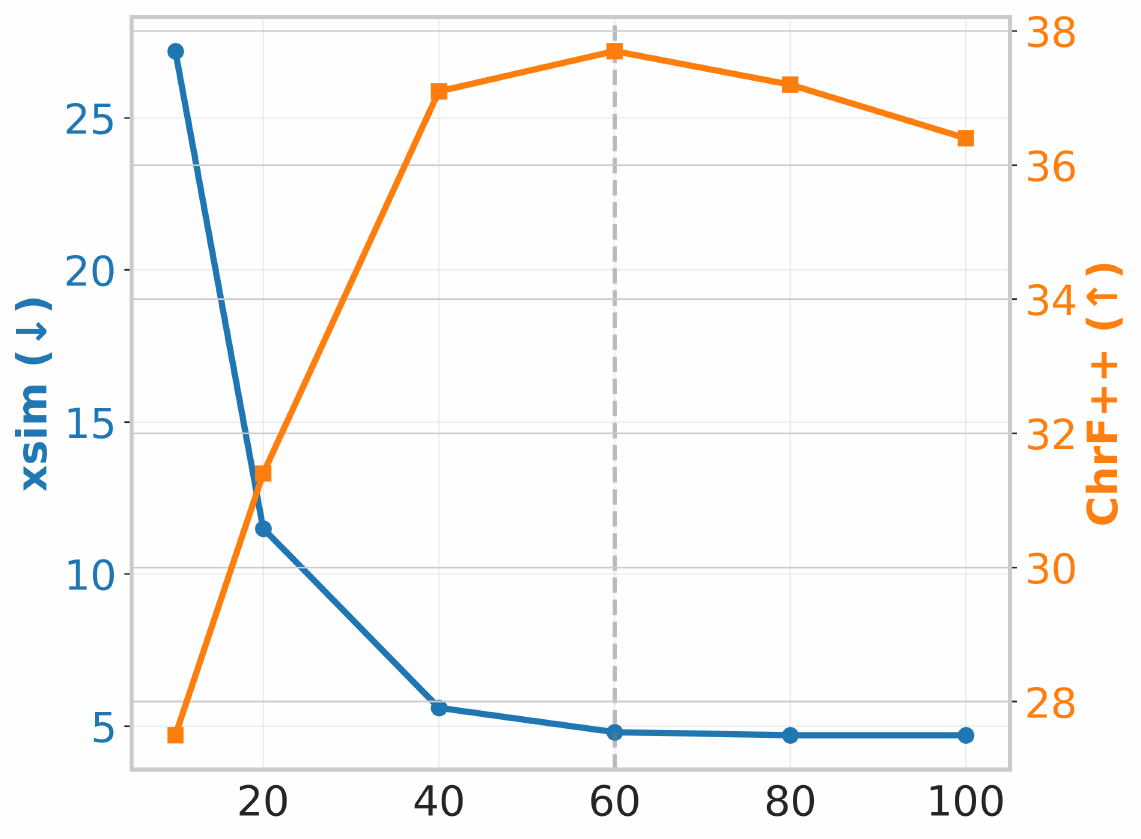}
        \caption{Contrastive Logit Scale}
        \label{fig:logit_scale}
    \end{subfigure}
    \caption{Loss function parameters for new languages. Results on Bible \dev. \chrf obtained with the \sonar{}-200 decoder. Vertical dashed lines indicate the final configuration used for \sonar{}.}
    \label{fig:hyperparameter_tuning}
\end{figure}

\paragraph{\bf Effectively learning thousands of new languages.} In \Cref{fig:hyperparameter_tuning}, we analyze three critical parameters for learning new languages: (a) the MSE loss weight, (b) the forward contrastive loss weight (student $\rightarrow$ teacher), and (c) the contrastive logit scale ($\tau$). We observe a clear trade-off in \Cref{fig:mse_weight}: minimizing the MSE weight yields the best embedding space for raw retrieval (xsim) on new languages, but detaches these representations from the original space, cratering zero-shot translation performance (\chrf). Conversely, heavily weighting the MSE loss quickly degrades xsim performance. This indicates that relying primarily on MSE leads to overfitting, as the exact geometry of the original embedding space is too complex to reconstruct given the limited data available for the long tail of new languages. \Cref{fig:fc_weight} demonstrates that the forward contrastive signal is strictly necessary to extend the space; without it, both metrics collapse. Similarly, \Cref{fig:logit_scale} shows that small logit scales fail. By increasing the logit scale, we enforce a smoother distribution for the contrastive loss, which proves much easier for the model to optimize under low-data regimes. Finally, \Cref{tab:objective_ablation}e reveals that enforcing a backward contrastive loss (teacher $\rightarrow$ student) actually harms the integration of new languages.

\paragraph{\bf Avoiding catastrophic forgetting.} To preserve the foundational languages, we apply different learning dynamics. \Cref{tab:objective_ablation}b provides evidence that, unlike for new languages, combining MSE with \textit{bidirectional} contrastive losses is optimal for the foundational languages. Removing the MSE objective completely alters the embedding space, causing a catastrophic drop in CLT. Furthermore, \Cref{tab:objective_ablation}c confirms the findings of \citet{charsonar}: interpolating the source and target embeddings to create the teacher target provides a superior, more stable training signal. Finally, \Cref{tab:objective_ablation}d shows that foundational languages—which benefit from abundant data—perform best under a sharper contrastive distribution ($\tau=10$).

\paragraph{\bf Omnilingual tokenization warm-up.} \Cref{tab:objective_ablation}a highlights the importance of our separate vocabulary adaptation stage (\Cref{subsubsection:omni_extension/language_drop_tokenizer_adaptation}). Warming-up the new tokenizer prior to extending the language set is beneficial both for boosting CLT on new languages and for preserving the performance of foundational languages. This confirms that disentangling tokenization learning from cross-lingual alignment is an advantageous step for massive language scaling.

\begin{table}[ht]
    \centering
    \resizebox{0.6\linewidth}{!}{%
    \begin{tabular}{@{} l c c c c c c @{}}
        \toprule
        \multicolumn{3}{@{}l}{} & \multicolumn{2}{c}{\textbf{BIBLE}} & \multicolumn{2}{c}{\textbf{FLORES}} \\
        \cmidrule(lr){4-5} \cmidrule(l){6-7}
        \multicolumn{3}{@{}l}{\textbf{Model}} & \textbf{xsim} & \textbf{\chrf} & \textbf{xsim++} & \textbf{\chrf} \\
        \midrule
        \multicolumn{7}{@{}l}{\textit{(a) Omnilingual Vocabulary Adaptation warm-up}} \\
        \multicolumn{3}{@{}l}{without} & \textbf{4.8} & 37.1 & 5.9 & 55.4 \\
        \rowcolor{gray!20} \multicolumn{3}{@{}l}{with} & \textbf{4.8} & \textbf{37.7} & \textbf{5.6} & \textbf{56.0} \\
        \midrule
        \multicolumn{7}{@{}l}{\textit{(b) Loss function for foundational Languages}} \\
        MSE & \begin{tabular}{@{}c@{}}Contr. \\ s $\rightarrow$ t\end{tabular} & \begin{tabular}{@{}c@{}}Contr. \\ t $\rightarrow$ s\end{tabular} & & & & \\
        \cmidrule(r){1-1} \cmidrule(lr){2-2} \cmidrule(lr){3-3}
        \ccmark & \xxmark & \xxmark & 5.1 & 37.2 & 6.1 & 55.5 \\
        \xxmark & \ccmark & \ccmark & \textbf{4.8} & 33.1 & \textbf{5.5} & 31.0 \\
        \ccmark & \ccmark & \xxmark & 4.9 & 37.5 & 6.0 & 55.7 \\
        \ccmark & \xxmark & \ccmark & 5.0 & \textbf{37.8} & 5.7 & 55.6 \\
        \rowcolor{gray!20} \ccmark & \ccmark & \ccmark & \textbf{4.8} & 37.7 & 5.6 & \textbf{56.0} \\
        \midrule
        \multicolumn{7}{@{}l}{\textit{(c) Teacher embedding type for foundational Languages}} \\
        \multicolumn{3}{@{}l}{source \phantom{iiiiiiiiiii} $\mathbf{z} = \mathbf{x}$} & \textbf{4.7} & 36.5 & 5.7 & 55.8 \\
        \multicolumn{3}{@{}l}{target \phantom{iiiiiiiiiii} $\mathbf{z} = \mathbf{y}$} & 5.0 & 37.6 & 5.9 & 55.3 \\
        \rowcolor{gray!20} \multicolumn{3}{@{}l}{interpolated \phantom{ii} $\mathbf{z} = 0.5\mathbf{x} + 0.5\mathbf{y}$} & 4.8 & \textbf{37.7} & \textbf{5.6} & \textbf{56.0} \\
        \midrule
        \multicolumn{7}{@{}l}{\textit{(d) Logit scale for foundational Languages}} \\
        \multicolumn{3}{@{}l}{large scale of $\tau = 60$} & 5.3 & 36.6 & 6.1 & 55.3 \\
        \rowcolor{gray!20} \multicolumn{3}{@{}l}{small scale of $\tau = 10$} & \textbf{4.8} & \textbf{37.7} & \textbf{5.6} & \textbf{56.0} \\
        \midrule
        \multicolumn{7}{@{}l}{\textit{(e) Loss function for new Languages}} \\
        MSE & \begin{tabular}{@{}c@{}}Contr. \\ s $\rightarrow$ t\end{tabular} & \begin{tabular}{@{}c@{}}Contr. \\ t $\rightarrow$ s\end{tabular} & & & & \\
        \cmidrule(r){1-1} \cmidrule(lr){2-2} \cmidrule(lr){3-3}
        \ccmark & \ccmark & \ccmark & 11.1 & 36.0 & 5.7 & 55.8 \\
        \rowcolor{gray!20} \ccmark & \ccmark & \xxmark & \textbf{4.8} & \textbf{37.7} & \textbf{5.6} & \textbf{56.0} \\
        \bottomrule
    \end{tabular}%
    }
    \caption{Ablations on the \sonar{} omnilingual extension evaluated on BIBLE and FLORES \dev sets. Cross-lingual similarity error rates xsim/xsim++ ($\downarrow$) and translation quality \chrf ($\uparrow$). The final \sonar{} configuration is highlighted in gray.}
    \label{tab:objective_ablation}
\end{table}

\paragraph{\bf Scaling without sacrificing foundational performance.} 
To confirm that the combination of the aforementioned training strategies successfully mitigates catastrophic forgetting, \Cref{tab:omnilingual_preservation} compares the foundational \sonar{}-200 model directly against the final omnilingual \sonar{}. The results validate our approach: expanding to 4,200+ varieties yields transformative improvements on new languages (reducing BIBLE xsim from 59.4 to 3.9 and doubling translation quality to 41.3 chrF++), without incurring any penalty on the base 200 languages. On FLORES, \sonar{} matches or marginally exceeds \sonar{}-200 across all metrics, proving that our methodology perfectly preserves the geometric alignment and expressivity of the original high-resource representations despite the massive scale-up.

\begin{table}[ht]
    \centering
    \resizebox{0.8\linewidth}{!}{
    \begin{tabular}{l ccc cccc}
        \toprule
         & \multicolumn{3}{c}{\textbf{BIBLE (1,560 languages)}} & \multicolumn{4}{c}{\textbf{FLORES (200 foundational languages)}} \\
        \cmidrule(lr){2-4} \cmidrule(lr){5-8}
        \textbf{Model} & \textbf{xsim} & \textbf{chrF++} & \textbf{xCOMET}  & \textbf{xsim} & \textbf{xsim++} & \textbf{chrF++} & \textbf{xCOMET} \\
        \midrule
        \sonar{}-200 & 59.4 & 20.9 & 0.361 & 0.70 & \textbf{6.3} & 55.2 & 0.872 \\
        \sonarlogoinline{} & \textbf{3.9} & \textbf{41.3} & \textbf{0.702} & \textbf{0.65} & 6.4 & \textbf{55.4} & \textbf{0.878} \\
        \bottomrule
    \end{tabular}
    }
    \caption{Comparison of the foundational 200-language model (\sonar{}-200) and the final omnilingual model (\sonar{}) in cross-lingual similarity search (xsim/xsim++ $\downarrow$) and translation quality (\chrf / xCOMET $\uparrow$) in Bible \test and FLORES \devtest. For translation with each model we use their dedicated decoders.}
    \label{tab:omnilingual_preservation}
\end{table}

\paragraph{\bf How important is vocabulary size for omnilinguality?} We experiment with omnilingual vocabularies ranging from 8K to 512K tokens (\Cref{tab:vocab_ablation}). Larger vocabularies consistently yield better embedding quality on new languages (Bible and FLORES+) while keeping foundational performance stable. As expected, larger vocabularies drastically improve tokenizer fertility, yielding significantly higher inference throughput at the cost of additional model parameters. We selected the 256K vocabulary as the optimal balance of representation quality, computational efficiency, and parameter count. Notably, attempting to train an omnilingual extension while retaining the original 200-language tokenizer results in moderate embedding quality losses, but suffers from major efficiency costs due to over-fragmentation (poor fertility) on the long tail of new languages.

\begin{table}[ht]
    \centering
    \resizebox{1.0\textwidth}{!}{%
    \begin{tabular}{@{}lcccccccccccccc@{}}
        \toprule
        & & \multicolumn{4}{c}{\textbf{BIBLE}} & \multicolumn{4}{c}{\textbf{FLORES}} & \multicolumn{4}{c}{\textbf{FLORES+}} \\
        \cmidrule(lr){3-6} \cmidrule(lr){7-10} \cmidrule(lr){11-14}
        \textbf{Vocabulary} & \textbf{Params} & \textbf{xsim} & \textbf{\chrf} & \textbf{Fert.} & \textbf{Thr.} & \textbf{xsim++} & \textbf{\chrf} & \textbf{Fert.} & \textbf{Thr.} & \textbf{xsim++} & \textbf{\chrf} & \textbf{Fert.} & \textbf{Thr.} \\
        \midrule
        \multicolumn{3}{@{}l}{\textit{200-lang Vocabularies}} \\
        256K & 1.5 & \cellcolor{blue!25}4.7 & 37.5 & 57.5 & 264 & \cellcolor{blue!25}5.6 & \cellcolor{blue!25}55.9 & \cellcolor{blue!15}43.2 & \cellcolor{blue!15}523 & 14.3 & 43.3 & 76.4 & 130 \\
        \multicolumn{3}{@{}l}{\textit{Omnilingual Vocabularies}} \\
        8K & 1.0 & \cellcolor{blue!40}4.6 & 37.5 & 73.2 & 263 & \cellcolor{blue!25}5.6 & \cellcolor{blue!15}55.8 & 70.3 & 321 & 14.5 & 43.7 & 78.2 & 335 \\
        16K & 1.0 & \cellcolor{blue!15}4.8 & 37.6 & 66.9 & 297 & \cellcolor{blue!25}5.6 & \cellcolor{blue!25}55.9 & 62.7 & 365 & 14.6 & 43.9 & 70.2 & 378 \\
        32K & 1.0 & \cellcolor{blue!15}4.8 & 37.5 & 61.9 & 327 & \cellcolor{blue!25}5.6 & \cellcolor{blue!25}55.9 & 56.3 & 411 & 14.1 & 44.3 & 62.7 & 414 \\
        64K & 1.1 & \cellcolor{blue!15}4.8 & \cellcolor{blue!15}37.7 & 57.3 & 362 & \cellcolor{blue!40}5.5 & \cellcolor{blue!25}55.9 & 50.6 & 454 & \cellcolor{blue!15}13.8 & 44.8 & 56.2 & 453 \\
        128K & 1.2 & \cellcolor{blue!15}4.8 & \cellcolor{blue!25}37.8 & \cellcolor{blue!15}53.5 & \cellcolor{blue!15}399 & \cellcolor{blue!40}5.5 & \cellcolor{blue!25}55.9 & 45.6 & 504 & \cellcolor{blue!25}13.5 & \cellcolor{blue!15}45.2 & \cellcolor{blue!15}50.7 & \cellcolor{blue!15}489 \\
        \textbf{256K} & 1.5 & \cellcolor{blue!15}4.8 & \cellcolor{blue!15}37.7 & \cellcolor{blue!25}50.3 & \cellcolor{blue!25}428 & \cellcolor{blue!25}5.6 & \cellcolor{blue!40}56.0 & \cellcolor{blue!25}41.3 & \cellcolor{blue!25}544 & \cellcolor{blue!40}12.8 & \cellcolor{blue!25}45.7 & \cellcolor{blue!25}45.8 & \cellcolor{blue!25}523 \\
        512K & 2.0 & \cellcolor{blue!40}4.6 & \cellcolor{blue!40}38.2 & \cellcolor{blue!40}47.4 & \cellcolor{blue!40}453 & \cellcolor{blue!25}5.6 & \cellcolor{blue!25}55.9 & \cellcolor{blue!40}37.6 & \cellcolor{blue!40}583 & \cellcolor{blue!40}12.8 & \cellcolor{blue!40}46.0 & \cellcolor{blue!40}41.6 & \cellcolor{blue!40}565 \\
        \bottomrule
    \end{tabular}%
    }
    \caption{Vocabulary size ablations. Results in X-Eng \dev sets. The 200-language vocabulary is the one used from \sonarhundred, which is extended from the Llama3 128K vocabulary. Model parameters in Billions. Metrics:  cross-lingual similarity error rates with xsim/xsim++($\downarrow$); translation quality with \chrf($\uparrow$) obtained with the \sonarhundred decoder; Tokenizer fertility($\downarrow$); and Inference Throughput($\uparrow$) in sentences per second.}
    \label{tab:vocab_ablation}
\end{table}

\begin{table}[ht]
    \centering
    \resizebox{0.75\textwidth}{!}{%
    \begin{tabular}{lcccccc}
    \toprule
    & \multicolumn{3}{c}{\textbf{\chrf (Translation Quality)}} & \multicolumn{3}{c}{\textbf{xsim (Similarity Search Error)}} \\
    \cmidrule(lr){2-4} \cmidrule(lr){5-7}
    \textbf{Feature} & \textbf{Importance} & \textbf{95\% CI} & \textbf{Rank} & \textbf{Importance} & \textbf{95\% CI} & \textbf{Rank} \\
    \midrule
    Tokenizer Fertility & 22.61 & [20.88, 24.75] & 1 & 5.12 & [4.43, 5.73] & 2 \\
    Family Examples & 14.19 & [13.23, 15.13] & 2 & 4.47 & [3.99, 5.19] & 3 \\
    Language Examples & 12.03 & [11.11, 12.99] & 3 & 6.86 & [6.07, 7.59] & 1 \\
    Latitude & 6.89 & [6.50, 7.42] & 4 & 3.63 & [3.18, 4.23] & 4 \\
    Longitude & 5.53 & [5.10, 6.09] & 5 & 3.46 & [3.06, 3.79] & 5 \\
    Script Examples & 5.23 & [4.76, 5.71] & 6 & 0.29 & [0.23, 0.38] & 8 \\
    Family ID & 1.26 & [1.12, 1.45] & 7 & 0.48 & [0.34, 0.61] & 7 \\
    Dict. Examples & 0.91 & [0.83, 0.98] & 8 & 0.86 & [0.76, 1.01] & 6 \\
    Script ID & 0.31 & [0.25, 0.38] & 9 & 0.03 & [0.02, 0.04] & 9 \\
    \midrule
    \multicolumn{7}{l}{\textbf{Model Performance}} \\
    Cross-Val RMSE & \multicolumn{3}{c}{3.92 $\pm$ 0.16} & \multicolumn{3}{c}{3.27 $\pm$ 0.87} \\
    $R^2$ & \multicolumn{3}{c}{0.928} & \multicolumn{3}{c}{0.918} \\
    Baseline Improvement & \multicolumn{3}{c}{48.5\%} & \multicolumn{3}{c}{24.8\%} \\
    \bottomrule
    \end{tabular}
    }
    \caption{Feature importance for predicting language performance on new languages (n=1,420) using Gradient Boosting with permutation-based importance and 95\% confidence intervals. Importance values indicate increase in RMSE when a feature is randomly shuffled.}
    \label{tab:feature_importance}
\end{table}

\paragraph{\bf What makes a language easy to learn?}
To identify the linguistic, geographic, and dataset properties that facilitate a language's integration into an omnilingual embedding space, we frame performance prediction as a Gradient Boosting regression problem using 1,420 new BIBLE \dev languages as data points.\footnote{We exclude the 140 Bible \dev languages that overlap with our 200 foundational languages.} Our results (\Cref{tab:feature_importance}) demonstrate high predictability (explaining over 91\% of the variance in both metrics) and reveal that feature importance is highly task-dependent. For translation tasks (\chrf), morphological complexity—measured by tokenizer fertility—is overwhelmingly the most critical feature, as severe token fragmentation severely hinders the decoder's ability to generate fluent text. In contrast, cross-lingual similarity search (xsim) is primarily dictated by the pure volume of available target language data. Furthermore, data from closely related languages (Family Examples) proves nearly as vital as target-language data, demonstrating that massive transfer learning within linguistic families actively drives omnilingual performance. While geographic proximity and dictionary data play secondary roles, they still provide a measurable positive bump for extremely low-resource languages. Ultimately, these insights offer a clear roadmap for massively multilingual scaling, proving that successfully reaching the world's linguistic long tail requires a holistic approach that balances raw data collection with optimized tokenization and targeted intra-family transfer learning.

\section{Cross-linguality Analysis}
In this section, we analyze cross-lingual transfer exhibited by \sonar{} models. First, we analyze cross-lingual transfer from the perspective of downstream tasks. Next, we examine how cross-lingual transfer occurs in the \sonar{} encoding of unseen languages for both the text and speech modalities.
\subsection{Downstream Cross-lingual Transfer}
\label{sec:results/clt}

We evaluate cross-lingual alignment across languages in the lens of classification. Namely, we train a classifier to classify French sentences from the SIB200Classification task in MTEB and apply it, in a zero-shot fashion, to the other 199 languages in SIB. We report Cross-lingual transfer (CLT) ratio in \Cref{tab:cross_lingual_transfer_combined}, which corresponds to the ratio of classification accuracy for language L with classification accuracy on French, either across the 80 common languages covered by baselines or the 200 languages of interest. 
This table highlights the strong cross-lingual transfer achieved for this classification task with \sonar{} representations, exceeding 99\% average CLT ratio on 200 languages, and over 100\% on the common 80 languages (and meaning that non-French classification results can actually exceed French classification result).

\begin{table}[htbp!]
\centering
\begin{tabular}{l|c|c}
    \toprule
    \multirow{2}{*}{model} & \multicolumn{2}{c}{SIB200 CLT ratio} \\ %
    & all & common \\ %
    \midrule
    LaBSE & 80.58\% & 91.99\% \\ %
    MEXMA & 78.38\% & 95.56\% \\ %
    mE5\textsubscript{large} & 84.76\% & 95.47\% \\ %
    SONAR & 92.34\% & 96.22\% \\ %
    \sonarlogoinline{} & \textbf{99.41\%} & \textbf{100.75\%} \\ %
    \bottomrule
\end{tabular}
\caption{Cross-lingual transfer (CLT) on SIB200Classification: Models trained on French, evaluated zero-shot on 199 languages (all) and 80 baseline-supported languages (common), reporting average relative performance to French. } %
\label{tab:cross_lingual_transfer_combined}
\end{table}

\subsection{Is Omnilinguality a Curse or a Blessing? Zero-shot Generalization on Unseen Languages}
\label{subsec:curse_of_omnilinguality}

\begin{figure}[ht]
    \centering
    \begin{subfigure}{0.495\linewidth}
        \centering
        \includegraphics[width=\linewidth]{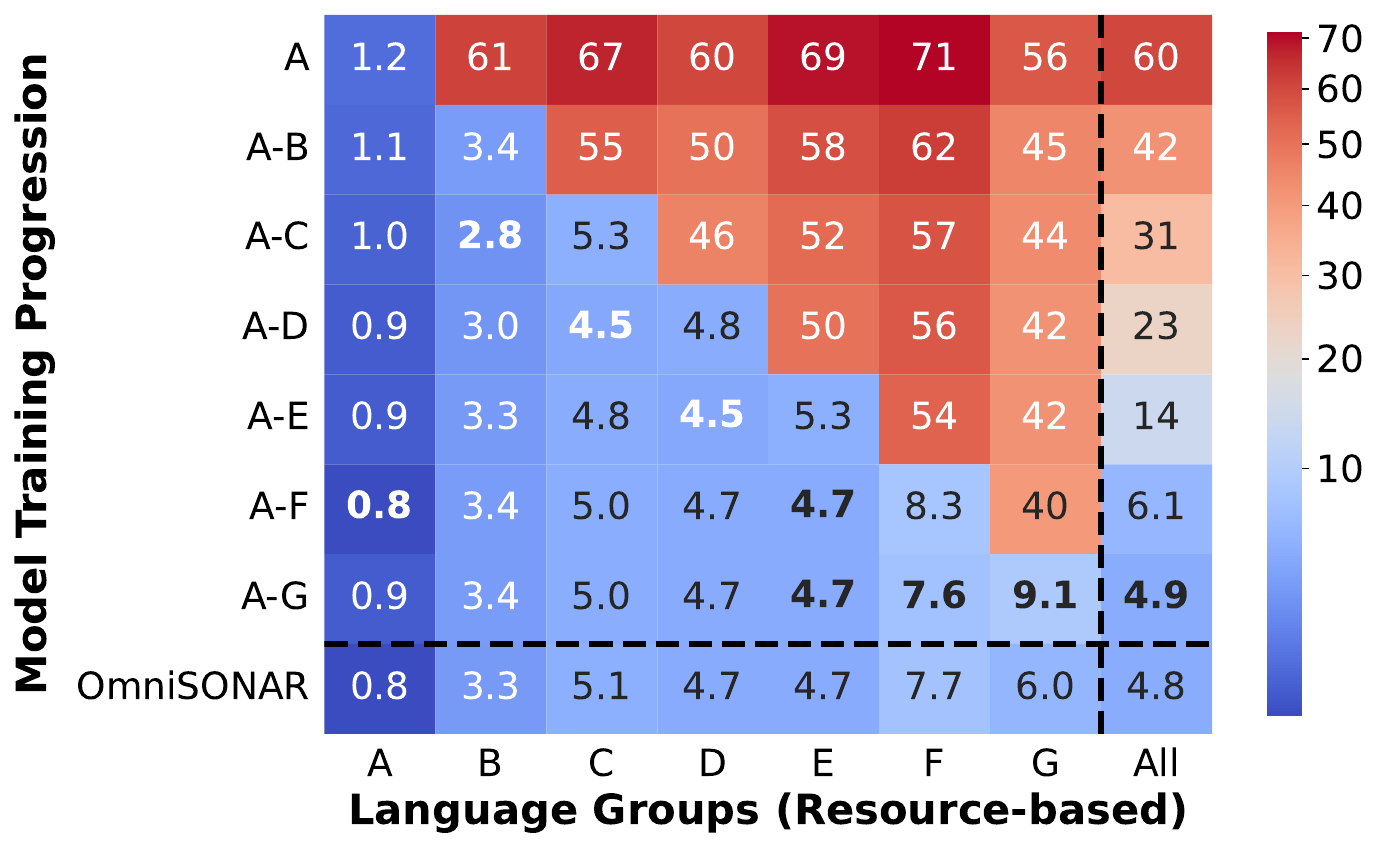}
        \caption{Resource-based: xsim ($\downarrow$)}
        \label{fig:group_heatmap_xsim}
    \end{subfigure}
    \hfill
    \begin{subfigure}{0.495\linewidth}
        \centering
        \includegraphics[width=\linewidth]{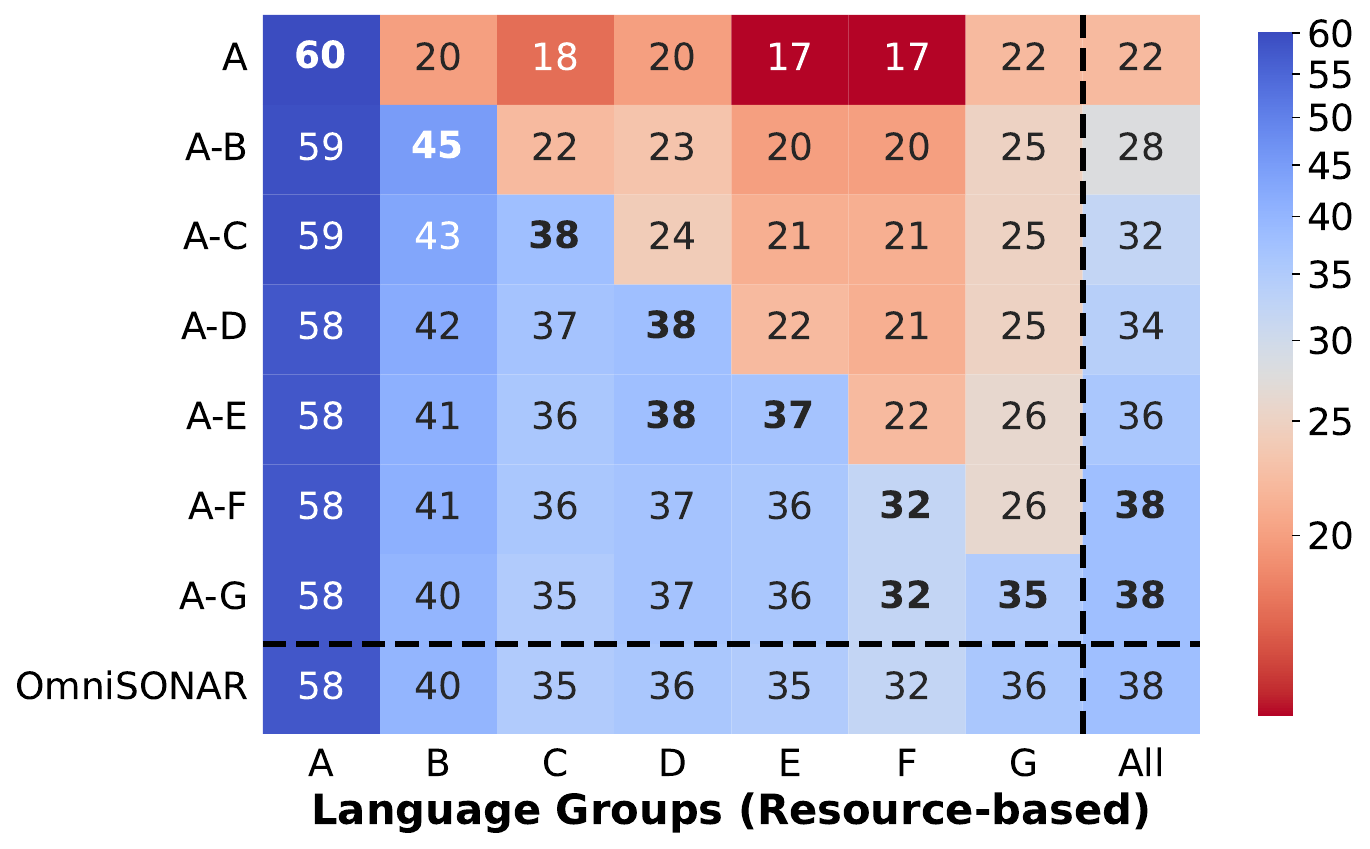}
        \caption{Resource-based: \chrf ($\uparrow$)}
        \label{fig:group_heatmap_chrf}
    \end{subfigure}
    
    \vspace{0.3cm}
    
    \begin{subfigure}{0.495\linewidth}
        \centering
        \includegraphics[width=\linewidth]{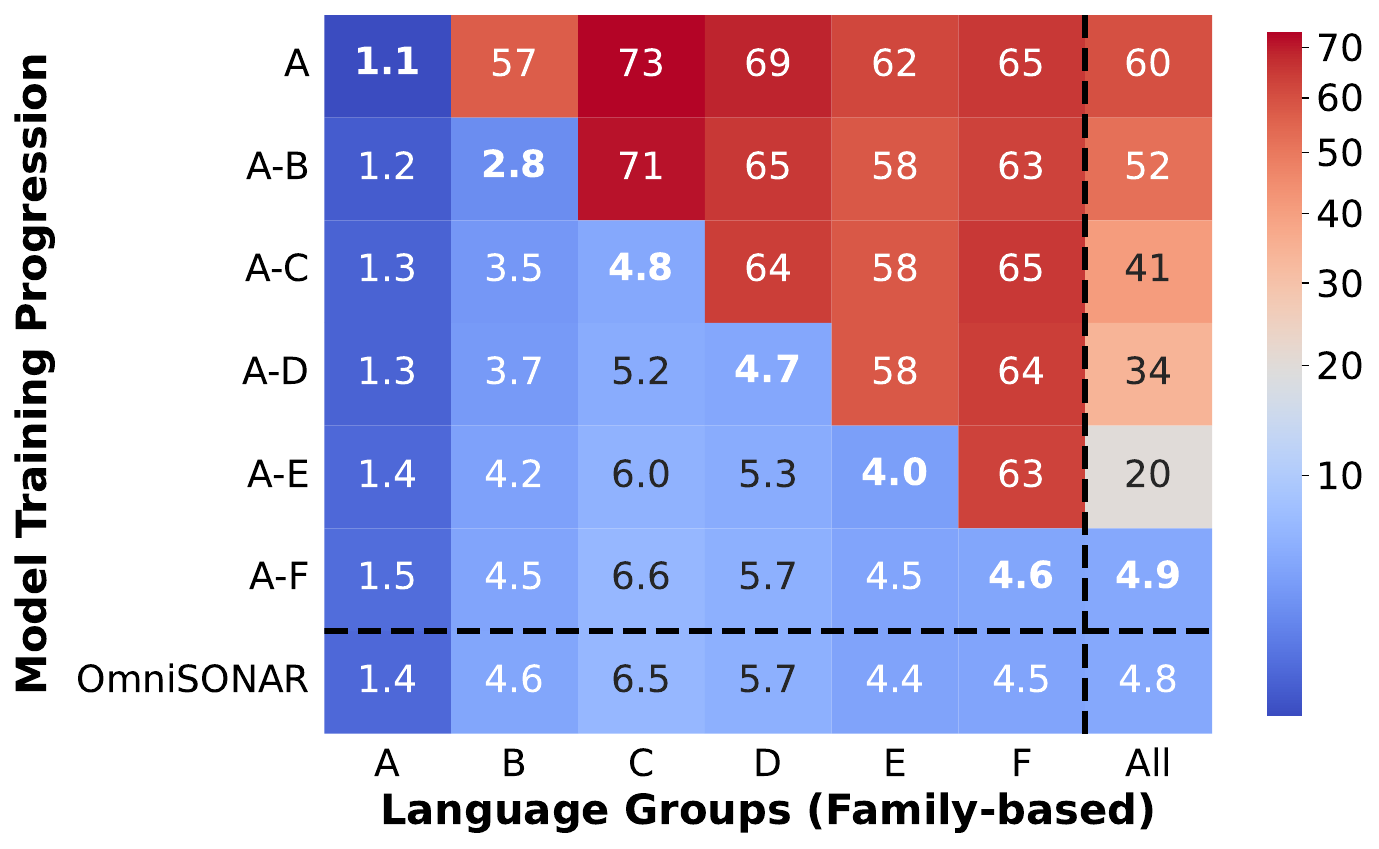}
        \caption{Family-based: xsim ($\downarrow$)}
        \label{fig:family_group_heatmap_xsim}
    \end{subfigure}
    \hfill
    \begin{subfigure}{0.495\linewidth}
        \centering
        \includegraphics[width=\linewidth]{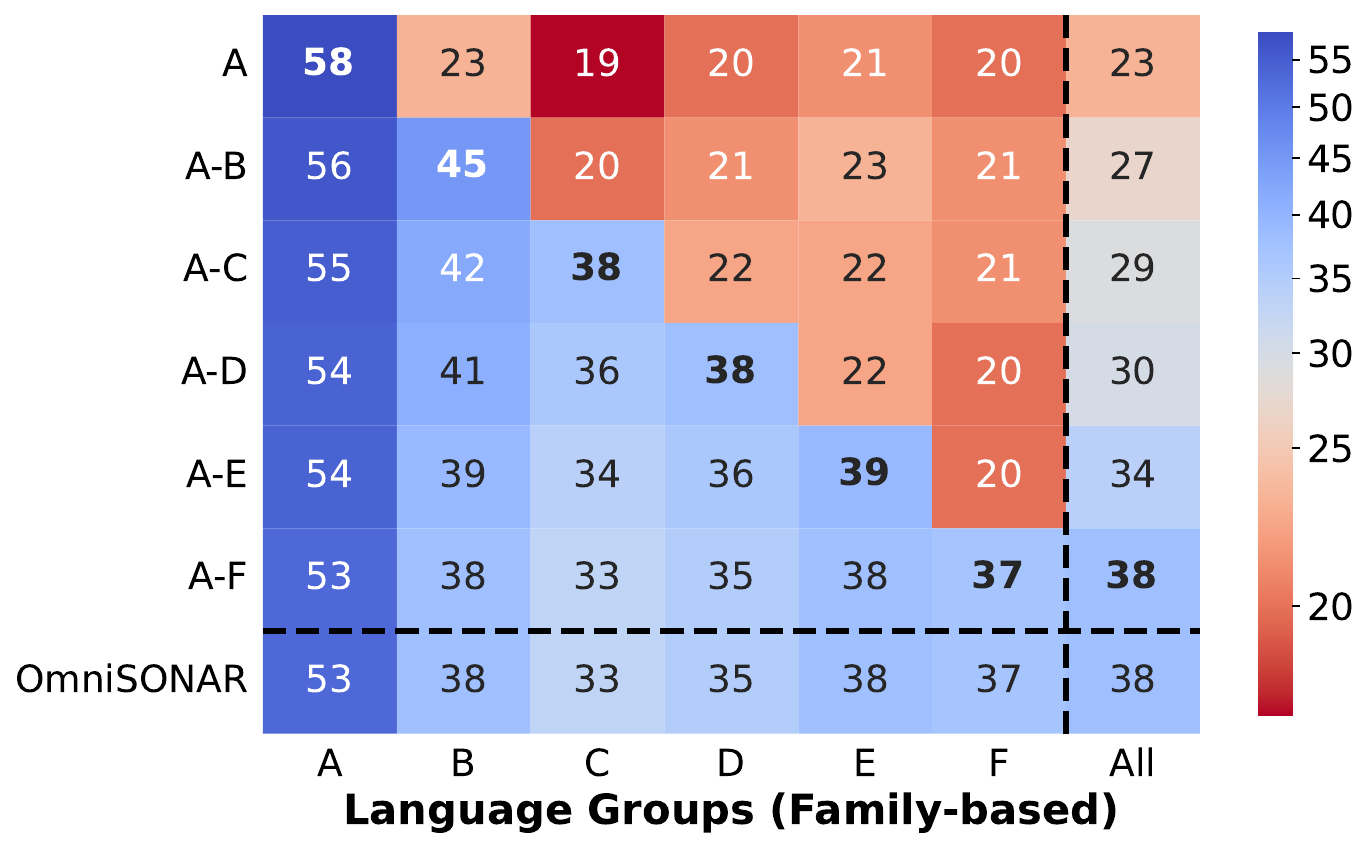}
        \caption{Family-based: \chrf ($\uparrow$)}
        \label{fig:family_group_heatmap_chrf}
    \end{subfigure}
    \caption{Cross-lingual similarity (xsim) and translation quality (\chrf) for encoders trained on progressively more language groups, evaluated on BIBLE \dev X-Eng. For translation we use the \sonarhundred decoder. In the \textbf{Top} subfigures (a) and (b), we display Resource-based grouping: Groups A-G sorted by resourcefulness. In the \textbf{Bottom} subfigures (c) and (d), we display Family-based grouping: Group A = Indo-European (largest family) and Groups B-F = other families. In bold is the best performance per column (excluding \sonar{}). The "All" column is the average across all 1,560 languages. \sonar{} is trained on all groups plus an additional group containing 1,864 extremely low resource languages.}
    \label{fig:group_performance}
\end{figure}

A well-documented limitation of massively multilingual models is the \textit{Curse of Multilinguality} \citep{lifting_the_curse_of_multilinguality,when_is_multilinguality_a_curse}, where performance on individual languages degrades as more languages are added due to capacity constraints and parameter interference \citep{multiway_multilingual_nmt,massively_multilingual_nmt,alastruey2025interferencematrixquantifyingcrosslingual}. However, a less explored corollary is the potential \textit{blessing} of omnilinguality: the increased capacity for zero-shot generalization to unseen languages via positive transfer.

To investigate this dynamic at an unprecedented scale, we train \sonar{} models on progressively larger subsets of languages. We structure this expansion under two orthogonal grouping strategies: one sorted by resource availability (Groups A through G) and another partitioned strictly by linguistic families (Group A containing Indo-European languages, and Groups B-F distributing other distinct families). We evaluate these progressive models on all 1,560 languages in the BIBLE \dev X-Eng set. To maintain a controlled experimental environment, we exclude the extreme long-tail of 1,864 lowest-resource languages (representing roughly 3M examples) from the training subsets (statistics are detailed in \Cref{tab:group_stats}).

The results reveal a stark contrast in generalization behavior depending on how the data is scaled. In the resource-based expansion (\Cref{fig:group_heatmap_xsim} and \Cref{fig:group_heatmap_chrf}), we observe massive zero-shot generalization. For instance, a model trained strictly on higher-resource groups achieves up to a 20-point improvement in xsim on completely unseen low-resource groups (e.g., dropping the error rate from 60 to 40 on Group G). We hypothesize that this strong positive transfer is driven by lexical and structural overlap \citep{lin-etal-2019-choosing,aepli-sennrich-2022-improving}, as the higher-resource groups naturally contain languages from the same families and scripts as those in the unseen lower-resource groups. To the best of our knowledge, this is the first empirical demonstration at this scale that scaling language coverage acts as a catalyst for zero-shot generalization to the world's lowest-resource languages.

Conversely, the family-based expansion (\Cref{fig:family_group_heatmap_xsim} and \Cref{fig:family_group_heatmap_chrf}) actively suppresses this transfer. When the unseen languages belong to entirely different linguistic families than the training set, the zero-shot generalization drops sharply, yielding only a marginal 2-5 point improvement in xsim. 

Crucially, evaluating both retrieval (xsim) and generation (\chrf) exposes the representational bottlenecks of omnilingual scaling. While xsim measures the topological alignment of the shared space (i.e., whether sentences map near each other), \chrf measures its information density (i.e., whether the vector retains sufficient syntactic and semantic nuance for fluent decoding). As evident in the family-based heatmaps, generation degrades faster and more severely than retrieval under capacity constraints. When unrelated families are introduced (\Cref{fig:family_group_heatmap_chrf}), the model struggles to accommodate wildly different scripts and morphological structures, bottlenecking generative expressivity long before geometric alignment (\Cref{fig:family_group_heatmap_xsim}) completely collapses. 
Furthermore, the family-based grouping clearly illustrates the underlying \textit{Curse of Multilinguality}. Across these specific heatmaps, the strongest performance for any group is typically found on or near the diagonal—when the model is evaluated on the exact subset it was trained on. As additional, less-related language groups are introduced, the performance on the initial groups reliably degrades due to parameter competition.

Collectively, these findings demonstrate that the "blessing" of omnilinguality is highly conditional on linguistic relatedness, as also confirmed by our analysis in \Cref{tab:feature_importance}. Positive transfer requires shared linguistic features; without them, scaling merely introduces interference. Our findings establish a clear rule for massively multilingual scaling: data quantity cannot overcome linguistic distance. To successfully scale sentence embeddings to cover all of the world's linguistic diversity, we must pivot from blind data scaling toward topologically-aware training strategies—such as sparse MoE architectures or family-conditioned parameter sharing \citep{share_or_not,janeiro-etal-2025-mixture}—that actively route positive transfer while limiting interference.

\subsection{Zero-shot Generalization on Unseen Languages for the Speech Modality}

\begin{table}[h!]
\centering
\begin{tabular}{lcc}
\toprule
\textbf{Language} & \textbf{\sonarlogoinline{}-speech-7B-zs} & \textbf{\sonarlogoinline{}-speech-7B}\\
\midrule
cat\_Latn & 30.3 & 39.6 \\
afr\_Latn & 39.8 & 53.0 \\
ukr\_Cyrl & 18.4 & 32.3 \\
mar\_Deva & 7.2  & 19.8 \\
bos\_Latn & 27.9 & 33.6 \\
slk\_Latn & 24.6 & 30.9 \\
ast\_Latn & 27.1 & 29.4 \\
mlt\_Latn & 7.2  & 34.4 \\
lao\_Laoo & 10.9 & 18.1 \\
jav\_Latn & 4.4  & 24.7 \\
\bottomrule
\end{tabular}
\caption{Speech Translation performance (X-Eng) on unseen languages during training.}
\label{tab:speech_xeng_zs}
\end{table}

While \sonar{} supports over 4.2k+ languages varieties for the text modality, we trained a speech encoder for only 177 languages. Extending speech modality coverage to additional languages is left for future work.
A key advantage of \sonar{}’s unified speech encoder is its ability to encode speech utterances without requiring the language to be specified. This means the model can generate embeddings from speech input even when the language is unknown. To evaluate the zero-shot cross-lingual encoding capabilities of the \sonar{} speech encoder, we conducted experiments on languages that share spoken similarities with those included in training. Specifically, we excluded 10 languages that are similar to the training set languages and assessed the encoder’s performance in a zero-shot setting by evaluating speech-to-text translation for these unseen languages.

In \Cref{tab:speech_xeng_zs}, we list the X-Eng translation results on the 10 unseen languages during training. \sonar{}-7B indicates the model trained on all 177 languages, and \sonar{}-7B-zs indicates the model trained excluding the 10 languages listed.
We can still see a non-negligible gap in all 10 languages when comparing \sonar{}-7B-zs to \sonar{}-7B. However, we can see that the zero-shot translation performance is far from being trivial. The results suggest that, despite unseen during training, the \sonar{} speech encoder can still effectively map the unseen languages into the embedding space of a closer neighbor for decoding.

\newpage

\section{\mdocu{}: Zero-shot Omnilingual Speech/Text Language Modeling with \sonar{}}

In the preceding sections, we established that \sonar{} exhibits outstanding cross-lingual and cross-modal alignment. Specifically, \sonar{} surpasses previous state-of-the-art sentence embedding spaces in cross-lingual similarity search by a factor of two, while also encompassing a significantly broader range of languages. Its capabilities are further highlighted by robust cross-modal alignment between highly multilingual speech and text. Moreover, we demonstrated strong performance on a variety of sentence-level downstream tasks, including classification, pair classification, and semantic textual similarity. While these advances unlock numerous new applications in omnilingual and cross-modal contexts, the current capabilities are still confined to the sentence level. 

This section introduces \mdocu{}, which aims to develop models capable of reasoning over longer text units, such as paragraphs and documents, while preserving broad multilingual and multimodal coverage.
It builds upon the promising results of the Large Concept Model (LCM) paradigm \citep{barrault2024large}, which investigates language modeling within sentence embedding spaces. This approach, where language modeling is learned over a language-agnostic and modality-agnostic embedding space, provides two key advantages. First, it naturally inherits the strong cross-lingual and cross-modal alignments of \sonar{} representations. Second, by operating at the sentence level instead of the token level, it enables more effective handling of long-context scenarios. In this context, the LCM has demonstrated strong zero-shot cross-lingual transfer for generative tasks such as summarization. For \mdocu{}, we transition from input-output agnostic representations to input-only agnostic representations within the language modeling backbone. Rather than employing diffusion modeling in the sentence embedding space, we leverage a Large Language Model (LLM) as a token-level predictor, while continuing to represent input text as \sonar{} embeddings.

We propose a novel architecture in which a pre-trained LLM (Llama 3.2 3B Instruct) performs cross-attention over encoder outputs that serve as contextualized representations of input speech or text, provided as \sonar{} embeddings. We introduce new pre-training and supervised finetuning strategies for this framework, training exclusively on English text. Although we operate in a zero-shot setting, our approach achieves strong performance across a range of highly multilingual speech and text benchmarks, further highlighting the seamless cross-lingual and cross-modal alignment in \sonar{} representations.

\subsection{Architecture}
\label{sec:arch}
\begin{figure}[!ht]
    \centering
    \includegraphics[width=.9\linewidth]{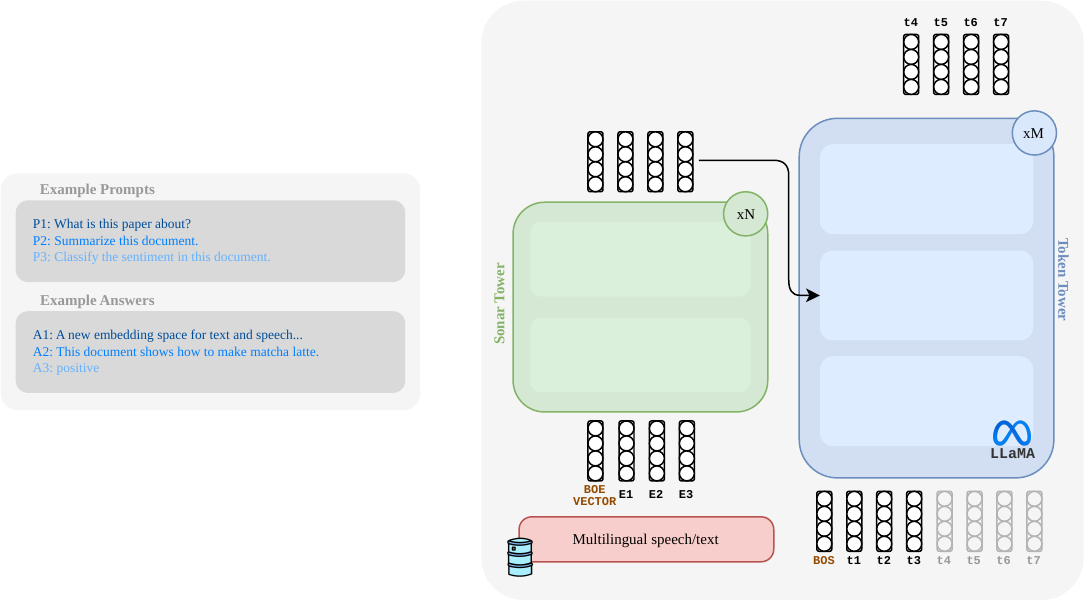}
    \caption{Architecture of the \mdocu{} model. An encoder processes a sequence of sentence-level \sonar{} embeddings, while the decoder is a token-level transformer that generates answers auto-regressively from a prefixed prompt. The decoder achieves this by performing cross-attention over the outputs of the encoder.
We refer to the encoder as the ``\sonartower{}" and the decoder as the ``\tokentower{}".}
    \label{fig:mdocu_architecture}
\end{figure}

\mdocu{} adopts an encoder-decoder architecture as depicted in \Cref{fig:mdocu_architecture}.
The input document or audio recording is first split into sentences using a segmentation model.
Each sentence is then encoded independently using the bespoke \sonar{} frozen encoder for each modality. A special vector, named BOE (Beginning of Embeddings, which in practice is the \sonar{} representation of the sentence: ``Start of text") is prepended to the input sequence.
The encoder processes this sequence of sentence-level embeddings, while the decoder is a token-level transformer that generates answers auto-regressively from a prefixed prompt. The decoder achieves this by performing cross-attention over the outputs of the encoder.
We refer to the encoder as the ``\sonartower{}" and the decoder as the ``\tokentower{}".
The \tokentower{} is initialized with Llama 3.2 3B Instruct \citep{llama3}.
The \sonartower{} is a 6-layer transformer encoder model with inner dimension of 8192, following the same architecture as Llama 3.3 70B, and a total parameter count of 5B. Including cross-attention weights the additional parameters totals 6.5B which complements the 3B parameters of the \tokentower{}.

The training can be separated into 2 phases: Pre-Training and Supervised Fine-Tuning (SFT).

\subsection*{Pre-Training}
Before training the encoder-decoder model for our target tasks, we first adapt the \sonartower{} and the cross-attention weights to the Llama-based \tokentower{}.
This adaptation is achieved through pre-training, during which the \tokentower{} remains frozen. Pre-training is composed of the following stages:

{\bf \sonartower{} Warm-up}: We first pre-train separately the \sonartower{} to predict the next sentence embedding based on the preceding sentence embeddings from the DLCM-edu raw text corpus \citep{allal2025smollm2smolgoesbig}. A causal mask is used on self-attention of this encoder to learn in parallel (i.e. teacher-forcing) to predict each sentence embedding based on the preceding ones.
We experimented with both mean squared error (MSE) regression and InfoNCE contrastive objectives between predicted and target \sonar{} embeddings. The MSE-based training is essentially similar to the MSE-LCM baseline model from \citet{barrault2024large}.
For contrastive learning, we used both in-batch and in-document negatives.
This setup provided us improved downstream performance compared to MSE-based training.

{\bf Teacher-forced Sentence-level Language Modeling}: Inspired from the LCM two-tower approach \citep{barrault2024large}, we pre-train our encoder-decoder model using a next sentence prediction task. Contrary to the LCM, this next sentence prediction task is trained with cross-entropy loss applied to the token outputs of the token tower.
Specifically, for each context of $N$ sentences represented as \sonar{} embeddings, the model predicts the tokens of the ($N+1$)$^{th}$ sentence using the \tokentower{}.
To achieve this, we keep the causal mask within the \sonartower{}. Additionally, we implement a custom cross-attention mask to ensure that tokens from the ($N+1$)$^{th}$ sentence can only attend to the contextualized \sonar{} representations of the preceding sentences. The causal self-attention mask in the \tokentower{} is also modified so that tokens can only attend to previous tokens within the same sentence, and not to tokens from earlier sentences.
This design ensures that, when predicting the next sentence in tokens, the \tokentower{} relies solely on previously generated tokens from the same sentence and on the \sonar{} representations of prior sentences. This pre-training stage is highly efficient, as the model is simultaneously trained to predict the next sentence for all possible sentence contexts within a document (i.e., contexts of size 0, 1, 2, 3, ..., up to length of the document minus one). The purpose of this pre-training stage is to train cross-attention weights, which are randomly initialized, and continue training the \sonartower{} for a language modeling task involving the frozen \tokentower{}.
This stage of pre-training is also performed on the DCLM-edu raw text corpus. 

{\bf Prefix Language Modeling}: 
While the previous pre-training stage is highly efficient, it has some limitations. 
First, the \sonartower{} is trained with a causal mask to enable sentence-level teacher-forcing. 
However, in the subsequent SFT stage, we use bi-directional self-attention on the encoder side to enhance document encoding capacity.
Introducing a final pre-training stage with bi-directional self-attention in the encoder is therefore expected to better align with the requirements of SFT.
Additionally, during the Teacher-forced Sentence-level Language Modeling stage, the \tokentower{} only predicts tokens within a single sentence, which represents a limited token context and is not fully aligned with the expectations of SFT where long token sequences can be generated.
To address these issues, we introduce a final pre-training stage based on Prefix Language Modeling (Prefix-LM) as introduced in \cite{roberts2019prefixlm}.
In this stage, each document is randomly split into two parts: the first part is processed as \sonar{} embeddings and encoded bi-directionally by the \sonartower, while the second part is predicted in tokens by the \tokentower{}. No mask is applied on cross-attention, and a simple causal mask is used in the \tokentower{}.
A prefixLM prompt is prepended to the tokens of the document continuation in this final pre-training phase to begin aligning the model with SFT setting. Backpropagation on the prompt tokens is disabled.

\begin{figure}[h!]
    \centering
     \includegraphics[width=.95\linewidth]{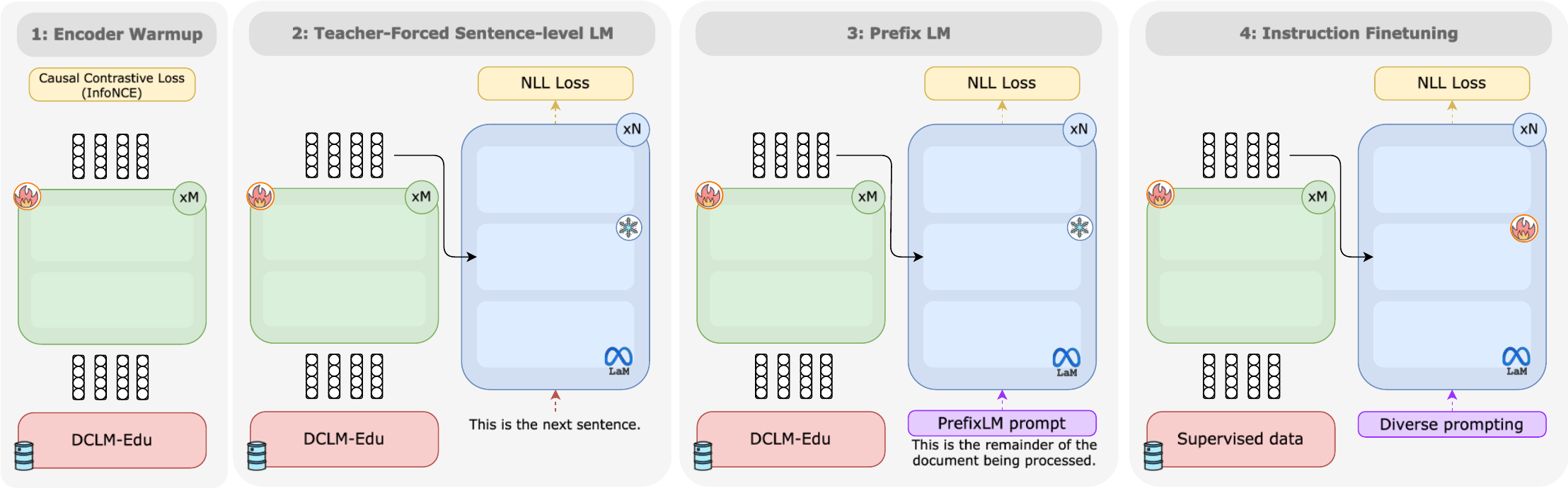}
    \caption{Training phases of the \mdocu{} model, divided in Pre-training (Encoder Warmup, Teacher-Forced Sentence-Level LM and Prefix-LM) and Instruction Finetuning.}
    \label{fig:mdocu_training}
\end{figure}

\subsection*{Supervised Fine-Tuning}
\label{sec:diverse_prompting}
Finally, we finetune our pre-trained encoder-decoder model on supervised training datasets covering a wide range of document understanding tasks. 
Given that our main motivation is to showcase the cross-lingual and cross-modal properties of \sonar{}, we train \mdocu{} exclusively on \sonar{} representations of English data. During benchmarking, we evaluate cross-lingually --- wherein the input document could be highly multilingual speech or text, but the desired target response is always in English. 
Similar to the prefixLM stage, the input document is encoded with bidirectional attention by \sonartower{}, while the \tokentower{} generates the target text in response to a token-level English prompt. To expose our model to prompt diversity, we construct a prompt template with five parts: 
\begin{enumerate}
    \item \textbf{System Prompt:} A standard prompt specifying high-level model behaviour,
    \item \textbf{Task Prompt:} A prompt describing the task, required operations and source/target languages, in varying levels of detail,
    \item \textbf{Question Block: } A prompt including a question prefix (``\textit{Question:}" or ``\textit{Q:}") and the actual question --- relevant for Multiple-Choice Questioning (MCQ) and Question-Answering (QA) tasks
    \item \textbf{Choices Block:} A prompt including a choices prefix (``\textit{Here are the choices:}", or ``\textit{Possible answers:}"), choice formatting (eg. ``\textit{1) 2) 3) 4)}" or ``\textit{A: B: C: D:}"), and the actual choices --- relevant for MCQ, Classification and Entailment tasks 
    \item \textbf{Target Format Prompt: } A prompt specifying the desired target type, including Option IDs only, Answer (in words) only, and Both. 

\end{enumerate}

For each of these parts, we construct a list of prompt templates, and dynamically sample from a cartesian product of each list per example. We format the final prompt within the Llama3 \citep{llama3} template before passing it to the model. We provide an example of a dynamically generated prompt in \Cref{fig:prompt_example}.

\begin{figure}[hbtp!]
\centering
\begin{tcolorbox}[colback=gray!10,colframe=black,title={Example of a prompt },fonttitle=\bfseries, width=.95\linewidth]
\small
\begin{promptlisting}
<|begin_of_text|>@<|start_header_id|>system<|end_header_id|>

You are a helpful assistant <|eot_id|>@
<|start_header_id|>user<|end_header_id|>

#Categorize the Indonesian document appropriately. Return only the category
label.#

£Options: "a. science/technology, b. travel, c. politics, d. sports, 
e. health, f. entertainment, g. geography". Which is correct?£
!Output the correct option by writing both the ID and the answer text.!
<|eot_id|>

<|start_header_id|>assistant<|end_header_id|>
\end{promptlisting}
\end{tcolorbox}
\caption{Example of a prompt for a Classification task from SIB200 using our diverse prompt generation approach. \textcolor{red}{System prompt} is in \textcolor{red}{red}, \textcolor{cyan}{task prompt} in \textcolor{cyan}{blue}, \textcolor{orange}{choices block} in \textcolor{orange}{orange} and the \textcolor{brown}{target format prompt} in \textcolor{brown}{brown}. The question block, used exclusively for MCQ and QA tasks, is absent here. This prompt is fed to the decoder, while the input document is passed through the encoder as usual.}
\label{fig:prompt_example}
\end{figure}

\subsection*{Data}
\label{mdocu:data}

\paragraph{\textsc{Training datasets}}
								
For the pre-training stage, we used a filtered version of the DCLM-edu pre-training corpus.
For the supervised fine-tuning stage, we used a datamix that includes many datasets covering a wide variety of tasks, such as summarization, sentiment analysis, topic classification, category prediction, toxicity classification, educational classification, rating prediction, and reading comprehension / document-based question answering. We also emphasize that our SFT datamix contains exclusively English text (both for input and output) and that it additionally includes SIB200 and Taxi1500 English training splits.
This diverse set of tasks is designed to improve the model’s capabilities across a range of document understanding and generation scenarios.

\paragraph{\textsc{Evaluation datasets}}
We evaluate our models on several multilingual speech and text benchmarks.
\begin{itemize}
    \item \textbf{XHellaswag} is a modified version of the multilingual Hellaswag corpus \citep{lai-etal-2023-okapi} where the target is in English for all languages. The task is to select the correct ending to a passage among 4 possible choices. The ending with the lowest NLL score is considered as the LLM prediction.
    \item \textbf{Spoken StoryCloze} \citep{hassid2023textually} is a spoken version of the StoryCloze \citep{mostafazadeh-etal-2016-corpus} benchmark. Given a common-sense story with four spoken sentences, the task is to distinguish the correct continuation (the positive) from an incorrect one (the negative). Accordingly, Spoken StoryCloze has two splits: \textit{SpeechStoryCloze}, wherein the negative samples follow the original StoryCloze, and \textit{TopicStoryCloze}, wherein the negatives are randomly sampled from the dataset.
    \item \textbf{SIB200} \citep{adelani-etal-2024-sib} is a large-scale open-source dataset for topic classification in 205 languages and dialects. It is based on the FLORES-200 \citep{nllb} machine translation corpus.
    \item \textbf{Taxi1500} \citep{ma-etal-2025-taxi1500} is a large-scale text classification dataset constructed using parallel translations of the Bible, which encompassed 1504 languages.
    \item \textbf{XBelebele} \citep{bandarkar-etal-2024-belebele} is a multiple choice machine reading comprehension dataset spanning 122 language variants. Given a context and a multiple-choice question, the answer is selected by choosing the option ID that has the lowest NLL score. The question and choices are given in English text.
    \item \textbf{Speech-XBelebele} \citep{costa-jussa-etal-2025-2m} is a speech comprehension dataset, covering 91 languages at the intersection of XBelebele and FLEURS \citep{conneau2023fleurs}. The setup is similar to XBelebele, except the textual context is substituted by raw audio input. The question and choices continue to be in English text.
    \item \textbf{Speech-SIB100: } This dataset has been created as part of this work. Given that SIB200 is sourced from FLORES, and so is FLEURS, we take the intersection of SIB200 and FLEURS to obtain a speech classification dataset: Speech-SIB100. In this dataset, given audio input from FLEURS, the task is to predict the corresponding class (in text) in SIB200.  
\end{itemize}

The evaluation metric used in all of these benchmarks is accuracy.

\paragraph{\textsc{Preprocessing}}

In this section, we provide more details on the preprocessing steps referred to earlier. 
Specifically, we conduct the following steps to create our benchmarks:

\textbf{Segmentation:} For sentence splitting in text, we use our custom Segment-Any-Text (SAT) model presented in \Cref{app:SaT}, which is designed for massively multilingual settings. While for speech datasets, we employ the SHAS \citep{tsiamas22_interspeech} segmentation approach. In particular, we use their publicly released multiligual speech segmentation model, that has been pretrained on English, Spanish, French, Portuguese and Italian. For other low-resource datasets like Taxi1500 and SIB200, given we found both text and speech segmentation to be unreliable, we pass unsegmented inputs to the respective \sonar{} encoders --- which we found to significantly boost performance.

\textbf{Cross-lingual preprocessing: } As mentioned earlier, we evaluate \mdocu{} in the cross-lingual X-En setting. This means that the input context to the \sonartower{} is highly multilingual (text or audio), while all inputs/outputs of the decoder (including the prompt question, prompt choices and target sequences) are English text. Some datasets like SIB200 and Taxi1500 offer this cross-lingual setting naturally, while others like Belebele have to be processed specifically for this setting by leveraging their multi-way parallel nature.

\textbf{Cross-modal preprocessing: } For speech datasets (Spoken StoryCloze, Speech-XBelebele and Speech-SIB), we perform two different evaluations, namely $S\to{}T$ and $T\to{}T$. 
$S\to{}T$ ingests speech input and generates a textual answer, while $T\to{}T$ uses text at the input and output. It should be noted that XBelebele and Speech-XBelebele are not fully aligned, so the results of Speech-XBelebele $T\to{}T$ are not comparable to those obtained on XBelebele. Finally, it is important to note that we excluded all test instances from Speech-XBelebele and Speech-SIB that also appear in the training split of FLEURS to avoid contamination from \sonar{}-speech training.

\subsection*{FLOPs Analysis}
\label{subsec:flops}
Inference efficiency is a crucial aspect for modern architectures. Recent work already explores the differences between LLM based Encoder-Decoder architectures and their counterpart LLM in terms of FLOPs \citep{zhang2025encoder}. In this section we discuss it in the context of an embedding-based Encoder as in \mdocu{}.

\paragraph{\textsc{Decoder-Only Architecture (Llama 3B/8B)}}
Decoder-only LLMs process inputs using causal attention: the prompt is encoded in a single prefill pass, then each generated token attends to all previous tokens via KV-caching. Critically, FLOPs scale with the total context length at generation time, both the input prompt and all previously generated tokens contribute to the computational cost.

\paragraph{\textsc{Encoder-Decoder Architecture (\mdocu{})}}
\mdocu{} follows the encoder-decoder paradigm but with a key innovation: instead of tokenized text, our encoder receives sentence embeddings directly from \sonar{} (a lightweight 1.5B encoder).
This design yields three major efficiency advantages:
\begin{enumerate}
    \item \textbf{One-time encoding:} The encoder processes input non-causally in a single pass. Moreover, the encoder outputs are independent from the decoder for each document, enabling us to `cache' the encoder inference for a document and re-purpose it for different decoder prompts. 
    \item \textbf{Compressed representations:} By operating on sentence embeddings rather than tokens, we dramatically reduce encoder sequence length ($\sim$20$\times$ compression assuming 20 tokens/sentence on average) and cross-attention computational cost.
    \item \textbf{Decoupled scaling:} Decoder self-attention FLOPs depend only on generated tokens, while cross-attention complexity is tied to the (compressed) encoder output length.
\end{enumerate}

\begin{figure}[h]
    \centering
    \begin{subfigure}[b]{0.49\textwidth}
        \centering
        \includegraphics[width=0.8\textwidth]{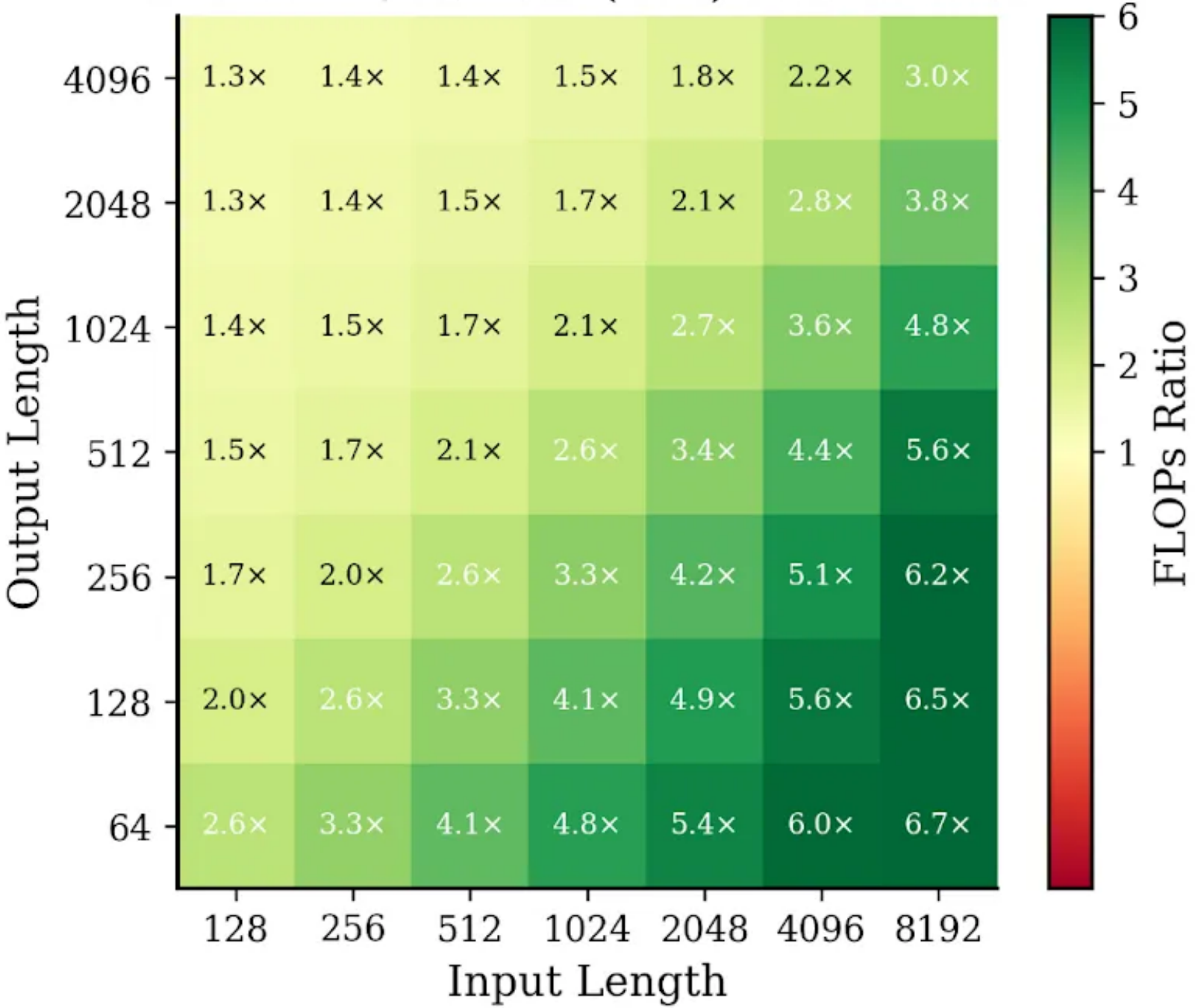}
        \caption{LLama 8B vs. \mdocu{} with \sonar{}}
        \label{fig:flops_spectrum_llama8b}
    \end{subfigure}
    \hfill
    \hfill
    \begin{subfigure}[b]{0.49\textwidth}
        \centering
        \includegraphics[width=0.8\textwidth]{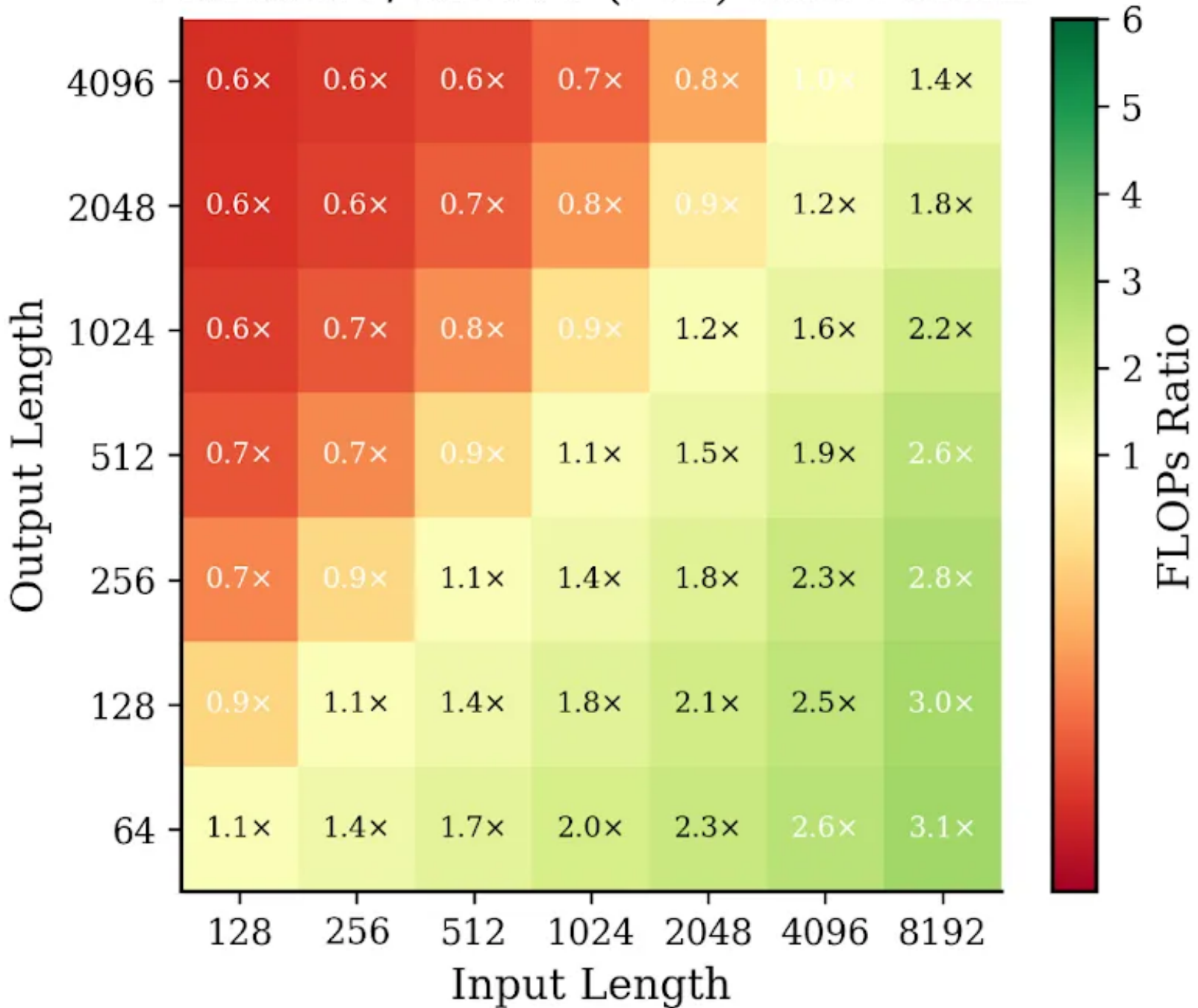}
        \caption{LLama 3B vs. \mdocu{} with \sonar{}}
        \label{fig:flops_spectrum_llama3b}
    \end{subfigure}
    \caption{FLOPs analysis of Spectrum architecture versus token-level LLM decoder-only architecture such as Llama3 models. Greener means that \mdocu{} is more compute efficient than token-level decoder-only models.}
    \label{fig:flops_spectrum_side_by_side}
\end{figure}

\paragraph{\textsc{FLOPs Comparison Methodology}}
To evaluate efficiency for document understanding tasks, we model the complete pipeline:
\sonar{} encodes the document (assuming 20 tokens/sentence average),
\mdocu{} encoder processes the resulting sentence embeddings, and
\mdocu{} decoder generates output tokens via cross-attention and causal self-attention.
\Cref{fig:flops_spectrum_side_by_side} below compares total FLOPs against Llama 3.2 3B and 3.1 8B across varying input (document) and output lengths. The heatmaps reveal substantial efficiency gains: for long documents with moderate outputs (the typical document understanding regime), \mdocu{} achieves 2-6× lower FLOPs than comparably-sized decoder-only models, despite having more total parameters. This efficiency advantage grows with input length, making \mdocu{} particularly well-suited for document-scale tasks.

\subsection*{Initial experiment}
A key requirement of our approach is the ability to extract all information from compressed document representations, since the document is encoded as a sequence of \sonar{} embeddings. To validate this, we conducted an experiment in which the model was tasked with reconstructing the full token content of a document from its \sonar{} embedding sequence.
We trained an encoder-decoder model, \textit{\mdocu{} autoencode}, on this document auto-encoding task for 35k steps using the DCLM Edu dataset, and then evaluated its performance on 100 documents from WikiLingua \citep{ladhak-etal-2020-wikilingua}. The evaluation documents have an average character length of 6,725. The results are summarized in \Cref{tab:autoencoding_results}.

\begin{table}[!ht]
\centering
\begin{tabular}{lcccc|c}
\toprule
 & AE BLEU1 & AE BLEU2 & AE BLEU3 & AE BLEU4 & AE BLEU \\
\midrule
\textit{\mdocu{} auto-encode} & 99.9 & 99.8 & 99.7 & 99.6 & \textbf{99.71} \\
\bottomrule
\end{tabular}

\caption{Auto-encoding (AE) BLEU scores obtained by splitting and encoding long documents from WikiLingua with \sonar{} and decoding them back into English with \textit{\mdocu{} auto-encode}.}
\label{tab:autoencoding_results}
\end{table}

These results demonstrate that the model can accurately reconstruct long documents from their embedding sequences, confirming the effectiveness of our approach and enabling us to proceed to more complex document understanding tasks.

\subsection{Results}

\subsubsection*{Baselines}

We compare the performance of \mdocu{} against the following baselines:

\begin{enumerate}
    \item \textbf{Llama 3.2 3B Instruct} \citep{llama3} is a multilingual large language model. While it is trained on a broader set of languages, it officially supports 8 languages\footnote{English, German, French, Italian, Portuguese, Hindi, Spanish, and Thai.}.

    \item \textbf{Llama 3.1 8B Instruct} \citep{llama3} shares a similar architecture with Llama 3.2 3B Instruct, but it is scaled to 8B parameters and trained to support a longer context window of up to 128k tokens.

    \item \textbf{SpiritLM} \citep{nguyen2025spirit} is a 7B text language model pretrained on a mixture of HuBERT \citep{hsu2021hubert} speech tokens and text sequences. It is initialized with Llama2 7B and trained on 300B text tokens and 37B speech tokens (570K hours of speech) --- making it a strong pretrained baseline compared to zero-shot \mdocu{}. We compare the two on Spoken StoryCloze (\Cref{fig:speechstorycloze}).
\end{enumerate} 

For Llama models, since our base decoder is built on top of the same chat template, we prompt them as we do \mdocu{} but adding the document in the decoder input, right after `task prompt' as \Cref{fig:prompt_example} showcases. 
\subsubsection*{Experimental Setting}

As outlined in \Cref{sec:arch}, we start training \mdocu{} by warming up the \sonartower{} for 55K steps with a learning rate of 2e-4. Next, we continue training the SONAR tower with the cross-attention modules, keeping the backbone decoder frozen, for the next sentence prediction task for 35K steps with a learning rate of 2e-4. Then, we pretrain with PrefixLM for 95K steps at a learning rate of 2e-4. Finally, for SFT, we unfreeze the decoder and train for 10K steps with an optimal learning rate of 5e-6. For the initial two stages of pre-training, we batch by sentence embeddings, and use a global batch size of 2M embeddings. For the subsequent stages of PrefixLM and SFT, given we have prompts and longer targets, we batch by tokens, with the effective size of an example being the sum of source, target and prompt tokens. Our global batch size here is 1M tokens.

\subsubsection*{Spectrum Pre-Training Evaluation}

\begin{figure}[h]
    \centering
    \begin{subfigure}[b]{0.63\textwidth}
        \centering
        \includegraphics[width=\textwidth]{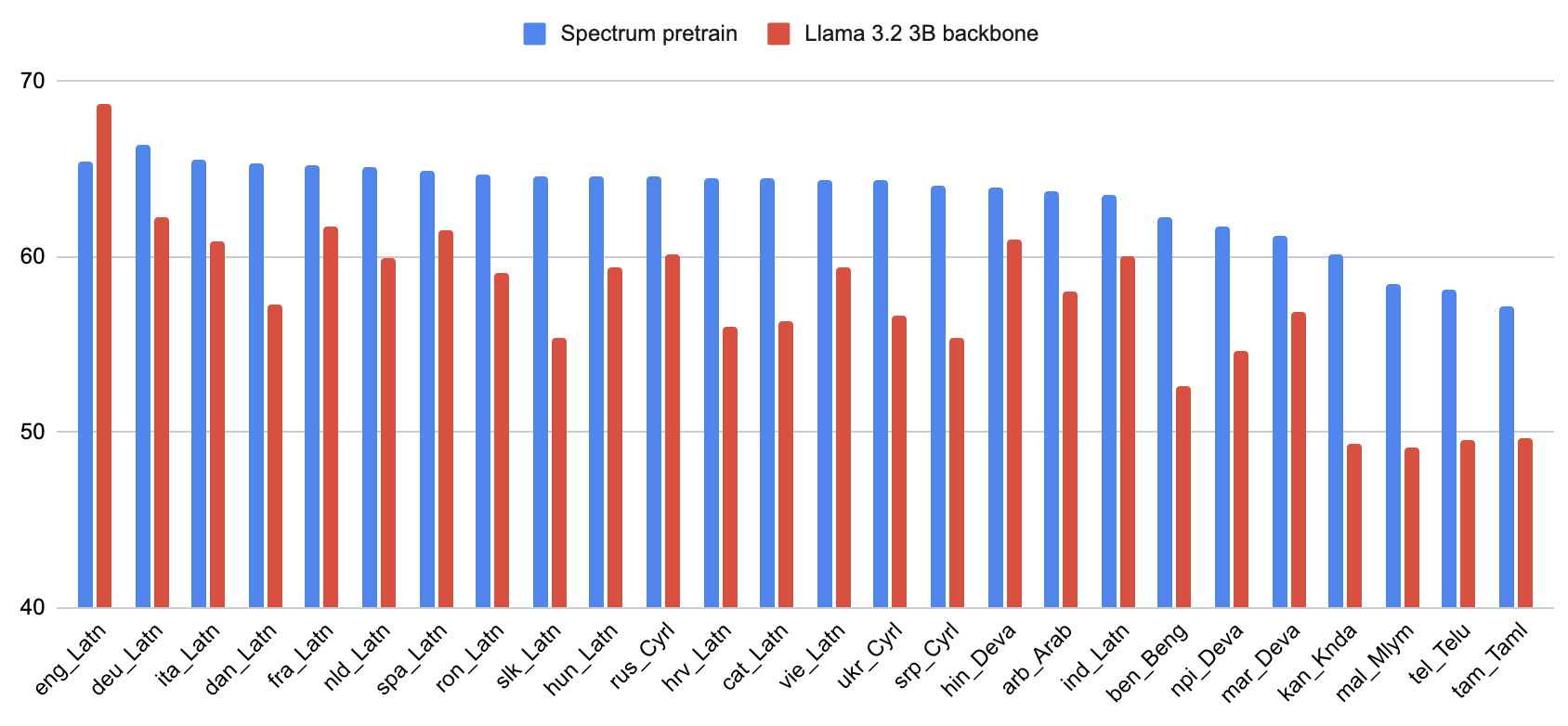}
        \caption{Zero-shot \mdocu{} accuracies on XHellaswag compared to its LLM backbone (Llama 3.2 3B Instruct) in 26 written languages.}
        \label{fig:xhellaswag}
    \end{subfigure}
    \hfill
    \hfill
    \begin{subfigure}[b]{0.35\textwidth}
        \centering
        \includegraphics[width=\textwidth]{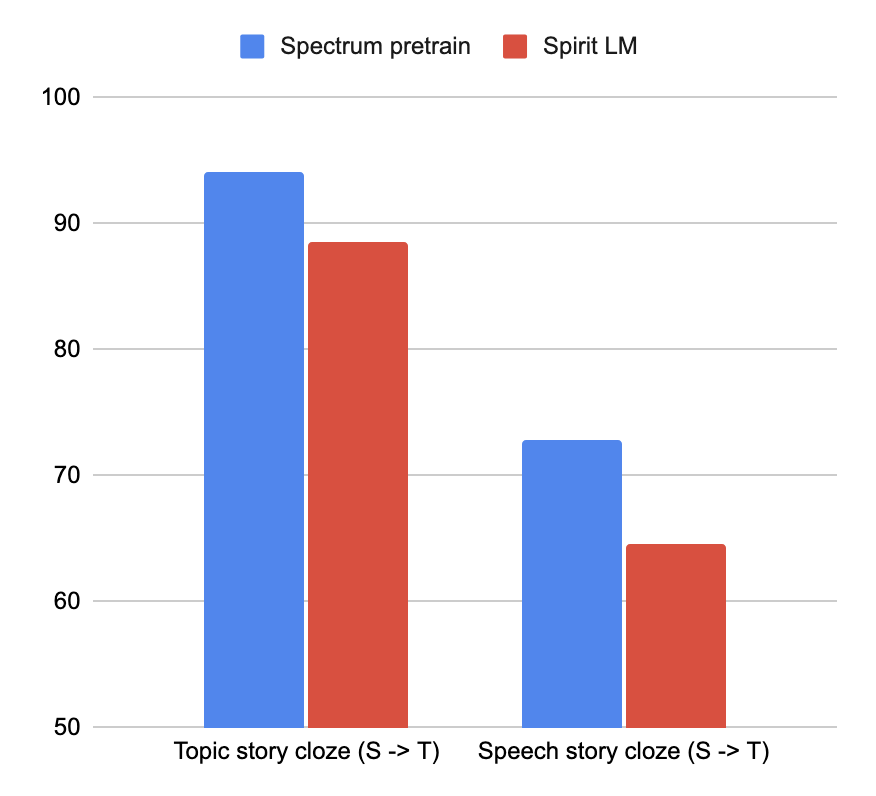}
        \caption{Pre-training results on Topic/Speech StoryCloze splits compared to SpiritLM.}
        \label{fig:speechstorycloze}
    \end{subfigure}
    \caption{\mdocu{} pre-training performance across XHellaswag and Spoken StoryCloze benchmarks.}
    \label{fig:spectrum_side_by_side}
\end{figure}

\textbf{\mdocu{} significantly outperforms Llama 3.2 3B Instruct on XHellaswag zero-shot.}
\Cref{fig:xhellaswag} shows our results on cross-lingual Hellaswag (XHellaswag) dataset across 26 languages, comparing \mdocu{} with Llama 3.2 3B Instruct --- which is the frozen LLM used to initialize the \mdocu{} decoder.
We observe some decrease in performance for English compared to Llama3.2 3B.
However, \mdocu{} displays strong performance across other languages --- it maintains near English-level performance on many high- and mid-resource languages and outperforms Llama by nearly ten points on low-resource languages. Given we trained \mdocu{} on English data alone, keeping the \tokentower{} frozen and trained only the cross-attention weights, these gains clearly demonstrate the unprecedented cross-lingual transfer facilitated by \sonar{} embeddings.

\textbf{\mdocu{} also outperforms SpiritLM zero-shot on the Spoken StoryCloze benchmark.} \Cref{fig:spectrum_side_by_side} shows the cross-modal results of \mdocu{} compared to SpiritLM. \mdocu{} scores 94.12\% on \textit{TopicStoryCloze} and 72.8\% on the harder \textit{SpeechStoryCloze} split, outperforming SpiritLM by 6.2\% on the former and 12.7\% on the latter. Similar to the results on XHellaswag, this too was achieved in a pure zero-shot setting, with \mdocu{} never being directly trained on speech data. These impressive results demonstrate the significant cross-modal transfer provided by \sonar{} embeddings.

\subsubsection*{\mdocu{} SFT Evaluation}
\begin{table}[htbp!]
\centering
\begin{tabular}{l c c c}
\toprule
\textbf{Model} & \textbf{XBelebele} & \textbf{SIB200} & \textbf{Taxi1500} \\
\midrule
Llama 3.2 3B Instruct & 53.03 & 65.69 & 19.67 \\
Llama 3.1 8B Instruct & \textbf{63.06} & 66.07 & 24.18 \\
\midrule
\mdocu{} 6B-3B & 61.85 & \textbf{79.95} & \textbf{36.9} \\
\bottomrule
\end{tabular}
\caption{Zero-shot multilingual results of \mdocu{} on XBelebele, SIB, and Taxi1500, compared with Llama models}
\label{tab:ift_results_llama_spectrum_text}
\end{table}

\begin{table}[htbp!]
\centering
\begin{tabular}{l c c}
\toprule
& \textbf{T $\to$ T} & \textbf{S $\to$ T} \\
\midrule
\textbf{Speech-XBelebele} & 63.64 & 62.74 \\
\textbf{Speech-SIB100} & 85.11 & 82.59 \\
\bottomrule
\end{tabular}
\caption{Zero-shot multilingual speech understanding results of \mdocu{} on Speech-XBelebele and Speech-SIB. We compare performance on Text $\to$ Text (T $\to$ T) and Speech $\to$ Text (S $\to$ T) splits of each benchmark to demonstrate cross-modal transfer.}
\label{tab:spectrum_ift_llama_spectrum_speech}
\end{table}

We evaluate \mdocu{} on the benchmarks listed in \Cref{mdocu:data}, and show our SFT results in \Cref{tab:ift_results_llama_spectrum_text,tab:spectrum_ift_llama_spectrum_speech}. 

\textbf{\mdocu{} consistently outperforms Llama Instruct 3B model on multilingual QA and classification benchmarks by a significant margin.} \mdocu{} achieves gains as high as 21\% on SIB200 and 52.61\% on Taxi1500 over the larger Llama 3.1 8B model, despite being trained only on the English split of these benchmarks. For XBelebele, which does not have a training split, \mdocu{} still achieves a gain of 16.63\% over Llama 3.2 3B, while achieving close performance to Llama 3.1 8B, demonstrating that the zero-shot gains of the model are not conditioned on having in-domain training data. Importantly, \mdocu{} achieves these improvements while remaining computationally efficient, requiring inference FLOPs comparable to the Llama 3B model and demonstrating even greater efficiency on longer contexts (\Cref{fig:flops_spectrum_llama3b}).

\textbf{\mdocu{} achieves near text-level performance in S $\to$ T zero-shot due to cross-modal transfer.}
To further highlight our model's cross-modal capabilities, we evaluate \mdocu{} on speech datasets (Speech-XBelebele and Speech-SIB200). We first establish a baseline using the T $\to$ T split of these benchmarks, enabling a direct comparison against S $\to$  T performance. Remarkably, transitioning from text to speech input results in only a minimal performance drop, with the model achieving 62.74\% on Speech-XBelebele and 82.59\% on Speech-SIB200. Together with our earlier results on Spoken StoryCloze, these findings underscore the powerful cross-modal transfer enabled by \sonar{}, unlocking significant gains for multilingual reasoning tasks across modalities.

\subsection{Takeaways}
In this section, we demonstrate the significant cross-lingual and cross-modal transfer benefits unlocked by language modeling over \sonar{} embeddings. Trained on English text only, Spectrum can outperform strong pretrained LLMs like SpiritLM and Llama in zero-shot cross-lingual and cross-modal settings. We hope these results encourage the community to conduct further research on omnilingual and cross-modal sentence embeddings, as a promising path towards omnilingual intelligence.

\newpage
\section{Beyond Sentence-Level Fixed-Size Representations}
In this section, we present two extensions to \sonar{} that address the level of granularity of the representations.
First, \tokensonar{} that aims to improve token-level representations while retaining sentence-level performance (\Cref{sec:tokensonar}).
Then \longcontextsonar{} which enables the encoding of longer text into fixed sized representations (\Cref{sec:longcontext}).

\subsection{\tokensonar{}: Better Cross-lingual Token Representations}
\label{sec:tokensonar}

Words are fundamental building blocks of language. While strong multilingual encoders such as XLM-R~\citep{xlmr} provide token-level representations, these representations are not explicitly aligned across languages. We introduced \sonar{} to provide well-aligned cross-lingual embeddings at the sentence level. However, robust cross-lingual token-level representations remain crucial for many downstream NLP applications, including sequence tagging, word alignment~\citep{jalili-sabet-etal-2020-simalign}, label projection~\citep{parekh-etal-2024-contextual} and machine translation quality estimation~\citep{huang-etal-2024-ottawa}. More broadly, strong cross-lingual transfer depends on balancing local and global information: token representations capture fine-grained lexical and morphosyntactic cues, while sentence representations capture global semantic context.

In this section, we report the cross-lingual characteristics of \sonar{} token representations through the word alignment task and sequence tagging. We also introduce \tokensonar{}, an adaptation of \sonar{} that provides enhanced cross-lingual token-level representations. 

\subsubsection*{The word alignment task} Given two sentences with the same meaning, the word alignment task aims to identify which words correspond to each other semantically. More formally, given a source–target sentence pair $s$, $t$ with word lengths $w_s$, $w_t$, respectively, the task of word alignment is to infer a binary matrix $M\in\{0,1\}^{w_s \times w_t}$, where $M_{ij}=1$ represents that the $i$-th word in the source sentence aligns semantically with the $j$-th word in the target sentence. We present an example of such alignment in \Cref{example_datasets}.

Many current word alignment methods derive alignments based on the similarity of token embeddings~\citep{jalili-sabet-etal-2020-simalign, dou-neubig-2021-word, azadi-etal-2023-pmi}. We define the \textit{token similarity matrix} $S\in \mathbf{R}^{l_s\times l_t}$; $S_{i,j}:=sim(\epsilon(s_i), \epsilon(t_j))$, where \textit{sim} is a similarity measure (we use cosine similarity), $s_i$ and $t_j \in \mathbb{V}$ are the $i$-th and $j$-th tokens of the source and target sentences $s$ and $t$, respectively of length $l_s$ and $l_t$, and $\epsilon:\mathbb{V} \rightarrow \mathbb{R}^d$ is a token encoder. An \textit{extraction} method is then applied to convert these similarities into a discrete alignment. The Argmax method~\citep{jalili-sabet-etal-2020-simalign} sets $M_{ij} = 1$ if and only if the $i$-th source token is the closest token to the $j$-th target one and vice-versa. That is, $$(i=\argmax_l S_{lj}) \cap (j=\argmax_l S_{il})$$

While simple, Argmax is a strong and competitive method. However, for typologically distant language pairs (e.g., English–Chinese), where correct alignments often involve many-to-many mappings between source and target tokens, Argmax tends to suffer from low recall.
Itermax addresses this by iteratively extracting new alignments while down-weighting similarity scores for tokens that have already been aligned, using a discount factor $\alpha$ \citep{jalili-sabet-etal-2020-simalign}. 

\subsubsection*{Comparison of different encoders}
We study how different sentence encoders perform the word alignment task. We adopt the same evaluation setup as in~\citet{dou-neubig-2021-word} and for multi-token words we assign a mapping if any of their respective sub-word tokens are aligned. We extract alignments with the Itermax method~\citep{jalili-sabet-etal-2020-simalign}. We report Alignment Error Rate (AER)~\citep{och-ney-2000-improved}. We study a total of 14 language pairs; we use as gold alignments the datasets XL-WA~\citep{martelli-EtAl:2023:clicit} with 10 language pairs; for en-hi, en-fr and en-cs we use the same datasets as in \citet{jalili-sabet-etal-2020-simalign}; we use \citet{neubig11kftt} for en-ja. 

\begin{figure}[hp!]
    \centering
    \includegraphics[width=.5\linewidth]{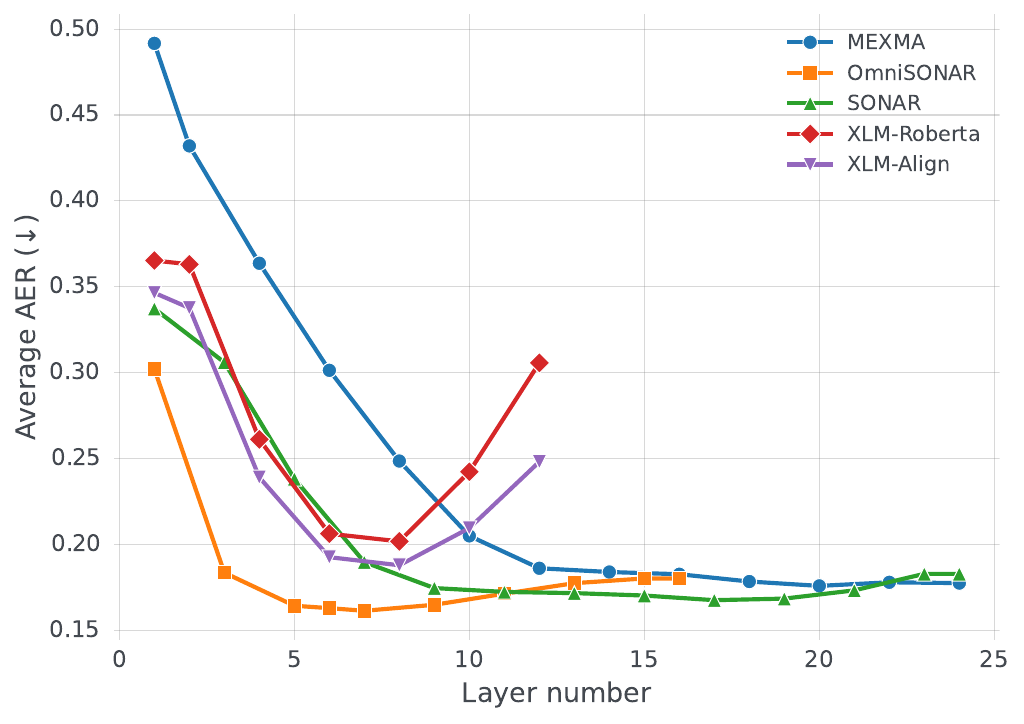}
    \caption{Average AER across language pairs for different layer numbers. We see that \sonar{} exhibits a parabolic shape, with inner layers showing better alignment than the final layers.}
    \label{fig:sonar_alignment}
\end{figure}

We compare the performance of SONAR and \sonar{} with XLM-based models XLM-Roberta~\citep{xlmr}, XLM-Align~\citep{chi-etal-2021-improving} and MEXMA~\citep{mexma}. \Cref{fig:sonar_alignment} shows the average AER across all different layers, and \Cref{tab:alignment_last} shows the performance for the last layer. 

We observe that \sonar{} exhibits a parabolic pattern: the intermediate layers exhibit a stronger alignment than the final layers. This trend is clearer than in SONAR, where AER consistently improves with increasing layer depth. We attribute this difference to the models’ training objectives; in particular, we suspect that \sonar{}’s reliance on the BOS representation for sentence encoding may degrade token-level representations in later layers. Consequently, while the 7th layer of \sonar{} provides the strongest cross-lingual representation among all encoders, its final layer is tied with MEXMA (\Cref{tab:alignment_last}).

\begin{table*}[h]
\centering
\scriptsize
\setlength{\tabcolsep}{2pt}
\begin{tabular}{lrrrrrrrrrrrrrrr}
\toprule
    \textbf{Encoder}                  & \textbf{Mean} & \textbf{en-bg} & \textbf{en-cs} & \textbf{en-da} & \textbf{en-es} & \textbf{en-et} & \textbf{en-fr} & \textbf{en-hi} & \textbf{en-hu} & \textbf{en-it} & \textbf{en-nl} & \textbf{en-pt} & \textbf{en-sl} & \textbf{ja-en} & \textbf{en-ru} \\
\midrule
\footnotesize XLM-Roberta  & 0.283 & 0.237 & 0.273 & 0.165 & 0.202 & 0.339 & 0.153 & 0.492 & 0.303 & 0.282 & 0.125 & 0.218 & 0.297 & 0.663 & 0.216 \\

\footnotesize XLM-Align    & 0.230 & 0.188 & 0.209 & 0.142 & 0.178 & 0.273 & 0.129 & 0.371 & 0.268 & 0.228 & 0.097 & 0.174 & 0.239 & 0.547 & 0.182 \\

\footnotesize MEXMA        & \textbf{0.167} & 0.148 & \textbf{0.116} & 0.093 & 0.132 & 0.170 & 0.075 & 0.288 & \textbf{0.219} & \textbf{0.172} & \textbf{0.063} & \textbf{0.140} & 0.170 & 0.418 & 0.141 \\

\footnotesize SONAR        & 0.172 & 0.153 & 0.124 & 0.095 & 0.131 & 0.179 & 0.080 & \textbf{0.276} & 0.231 & 0.179 & 0.072 & 0.150 & \textbf{0.168} & 0.430 & \textbf{0.133} \\

\footnotesize \sonarlogoinline{}     & \textbf{0.167} & \textbf{0.143} & 0.118 & \textbf{0.090} & \textbf{0.128} & \textbf{0.166} & \textbf{0.070} & 0.280 & 0.225 & 0.173 & 0.066 & 0.142 & 0.174 & \textbf{0.412} & 0.143 \\
\bottomrule
\end{tabular}
\caption{Last layer AER for different encoders.}
\label{tab:alignment_last}
\end{table*}

\subsubsection*{\tokensonar{}: repurposing \sonar{} for token embeddings}
We improve \sonar{}’s cross-lingual token representations by continuing to train the pretrained model under a new loss. This additional training stage is lightweight: we train for only 15{,}000 steps on a subset of 57 languages from our MT Primary dataset. Specifically, we optimize the following loss:
\begin{equation}
L = L_{teacher} + \lambda L_{SO}
\label{eq:so_loss}
\end{equation}
where the teacher loss $L_{teacher} = \text{MSE}(\epsilon(s) + \epsilon(t), \epsilon_{teacher}(s) + \epsilon_{teacher}(t))$; $s$ and $t$ represent a translation sentence pair, and $\epsilon_{teacher}$ is a frozen \sonar{} encoder. We set the weighting hyperparameter to $\lambda = 1$. The Self-Objective loss $L_{SO}$~\citep{dou-neubig-2021-word} is defined as follows:
\begin{equation}
    L_{SO} = \sum_{i, j} M_{ij} \frac{1}{2} \left( \frac{softmax(S_{i,*})}{n} + \frac{softmax(S_{*,j})}{m} \right)
\end{equation}
where $M_{ij} \in \{0,1\}$ is the alignment obtained with the Argmax method, $S$ is the token similarity matrix between source and target tokens. We apply a temperature scaling $\tau=500$ to control the sharpness of the softmax distribution. Intuitively, the Self-Objective is a token-level contrastive loss that treats non-aligned token pairs as negatives. It encourages semantically corresponding tokens across languages to move closer in the embedding space. The teacher loss, in turn, acts as a regularizer that preserves the sentence-level embedding space and prevents representational drift. 
This is important since it retains compatibility between \tokensonar{} and \sonar{}.

\paragraph{Evaluation} We further evaluate the cross-linguality of token representations with sequence tagging downstream tasks. Given an input sequence of tokens $x = (x_1, x_2, \dots, x_n)$, the goal is to assign a label $y_i \in \{1, \dots, N \}$ to each token $x_i$. To test how well different encoders learn cross-lingual token representations, we train a classifier head $C\in \mathbb{R}^{d \times N}$ on top of the encoder embeddings, thus keeping the encoder frozen. The head is trained only on English data and then evaluated zero-shot on all available languages.

We report performance for the following datasets: Massive (slot filling)~\citep{fitzgerald2022massive}, PAN-X (Named Entity Recognition)~\citep{pan-etal-2017-cross} and WiC (word sense disambiguation)~\citep{pilehvar-camacho-collados-2019-wic}. For WiC, which is not a sequence tagging task, we adapt the evaluation as follows: we extract the contextual representation of the target word in each sentence, concatenate the two vectors, and apply a classifier head $C\in \mathbb{R}^{2d \times 2}$. For multi-token targets, we obtain the word representation by averaging the representations of its subword tokens.

When the dataset segmentation does not perfectly align with the tokenizer (e.g., a token overlaps with multiple segments assigned to different classes), we assign the token the label of the segment with the largest character-level overlap. If there is a tie, we assign the label of the earliest overlapping segment.

\begin{table}[hbtp!]
\centering

\begin{tabular}{ccc}
\toprule
\bf{Dataset/task} & \bf{\#Test languages} & \bf{Example} \\
\midrule
Word alignment & 14 &

\begin{tikzpicture}[
  baseline=(enbase.base),
  every node/.style={inner sep=1pt, outer sep=0pt} %
]
  \node[inner sep=0pt, outer sep=0pt] (enbase) {};
  \node[anchor=base] (en1) at (0,0) {The};
  \node[anchor=base] (en2) at ([xshift=8.1mm] en1.base east) {necessary};
  \node[anchor=base] (en3) at ([xshift=9.1mm] en2.base east) {correction};
  \node[anchor=base] (en4) at ([xshift=4.1mm] en3.base east) {will};
  \node[anchor=base] (en5) at ([xshift=3.1mm] en4.base east) {be};
  \node[anchor=base] (en6) at ([xshift=6.1mm] en5.base east) {made};
  \node[anchor=base] (en7) at ([xshift=2.1mm] en6.base east) {.};
  \node[anchor=base west] (de1) at ([yshift=-10mm] en1.base west) {Eine};
  \node[anchor=base] (de2) at ([xshift=12mm] de1.base east) {entsprechende};
  \node[anchor=base] (de3) at ([xshift=8.1mm] de2.base east) {Anderung};
  \node[anchor=base] (de4) at ([xshift=4.1mm] de3.base east) {wird};
  \node[anchor=base] (de5) at ([xshift=11mm] de4.base east) {vorgenommen};
  \node[anchor=base] (de6) at ([xshift=6.1mm] de5.base east) {werden};
  \node[anchor=base] (de7) at ([xshift=4.1mm] de6.base east) {.};
  \draw[<->,thick] (de1.north) -- (en1.south);
  \draw[<->,thick] (de2.north) -- (en2.south);
  \draw[<->,thick] (de3.north) -- (en3.south);
  \draw[<->,thick] (de4.north) -- (en4.south);
  \draw[<->,thick] (de5.north) -- (en6.south);
  \draw[<->,thick] (de6.north) -- (en5.south);
  \draw[<->,thick] (de7.north) -- (en7.south);
\end{tikzpicture}
\\
\addlinespace[8pt]

Massive & 52 &
\begin{tabular}[c]{@{}l@{}}
What is today 's forecast for Berlin\\
        \makebox[0pt][l]{\hspace{3.9em}\footnotesize DATE}%
        \makebox[0pt][l]{\hspace{12.7em}\footnotesize	 PLACE}%
\end{tabular}
\\
\addlinespace[8pt]

PAN-X
& 40 &
\begin{tabular}[c]{@{}l@{}}
REDIRECCIÓN Algarrobo ( Chile ) \\

        \makebox[0pt][l]{\hspace{8.5em}\footnotesize B-LOC}%
        \makebox[0pt][l]{\hspace{13.2em}\footnotesize	 I-LOC}%

\end{tabular}
\\
\addlinespace[8pt]

WiC & 6 &
\begin{tabular}[c]{@{}l@{}}
$\left.
\begin{array}{@{}l@{}}
\text{Bolivia holds a key \underline{play} in this process of peace.} \\
\text{A musical \underline{play} on the same subject was also staged.}
\end{array}
\right\}\ 
\begin{array}{@{}l@{}}
\text{Different}\\
\text{meaning}
\end{array}$
\end{tabular}
\\
\bottomrule
\end{tabular}
\caption{Summary of the tasks used for the evaluation of cross-lingual token representations.}
\label{example_datasets}
\end{table}

\subsubsection*{Results}
We assess \tokensonar{} using both cross-lingual token representation tasks and sentence-level representation tasks, to monitor any potential degradation in sentence representations.

\begin{table}[hbtp!]
\centering
\begin{tabular}{l|c|ccc}
\toprule
\multicolumn{1}{l|}{\bf Encoder} & \multicolumn{1}{r|}{\bf AER ($\downarrow$)} & {\bf Massive ($\uparrow$)} & {\bf PAN-X ($\uparrow$)} & { \bf WiC ($\uparrow$)} \\
\midrule
XLM-Roberta     & 0.283        & 0.286       & 0.614       & 0.557       \\
XLM-Align       & 0.230        & 0.291       & 0.560       & 0.554       \\
MEXMA           & 0.167        & 0.396       & 0.593       & 0.565       \\
SONAR           & 0.172        & 0.402       & 0.582       & 0.576       \\
\sonarlogoinline{}        & 0.167        & 0.486       & 0.623       & 0.574       \\
\tokensonar{}      & {\bf 0.164}  & {\bf 0.491} & {\bf 0.630} & {\bf 0.582} \\
\bottomrule
\end{tabular}
\caption{Performance for cross-lingual token representation tasks. The first column presents the mean Alignment Error Rate (AER) for the last layer. The remaining columns display the average F1 scores achieved by a classifier that was trained using English token embeddings and then evaluated on data from other languages.}
\label{tab:token-level-eval}
\end{table}

Our results in \Cref{tab:token-level-eval} show an improved token-level representation with respect to ~\sonar{}, with better alignment across languages. \tokensonar{} shows the best results for word alignment as well as sequence tagging tasks. At the same time, the results in \Cref{tab:sentence-level-eval} show that the introduction of a specific token alignment loss has very little impact on the performance of sentence level tasks.

\begin{table}[hbtp!]
\centering
\begin{tabular}{l|ccc}
\toprule
 & \multicolumn{3}{c}{\bf FLORES (57)} \\
{\bf Encoder} & {\bf xsim ($\downarrow$)} & {\bf xsim++ ($\downarrow$)} & {\bf ChrF++ ($\uparrow$)} \\
\midrule
SONAR             & 0.167          & 9.88          & -              \\
\sonarlogoinline{}& \textbf{0.023} & \textbf{2.66} & \textbf{63.4}  \\
\tokensonar{}     & 0.025          & 2.96          & 63.0           \\
\bottomrule
\end{tabular}

\caption{Performance for sentence-level tasks. We report xsim~\citep{laser} and xsim++~\citep{xsimplusplus} for cross-lingual mining and ChrF++~\citep{chrf2} on the FLORES~\citep{nllb} dataset for translation performance.}
\label{tab:sentence-level-eval}
\end{table}

\begin{figure}[hbtp]
    \centering
    \includegraphics[width=.95\linewidth]{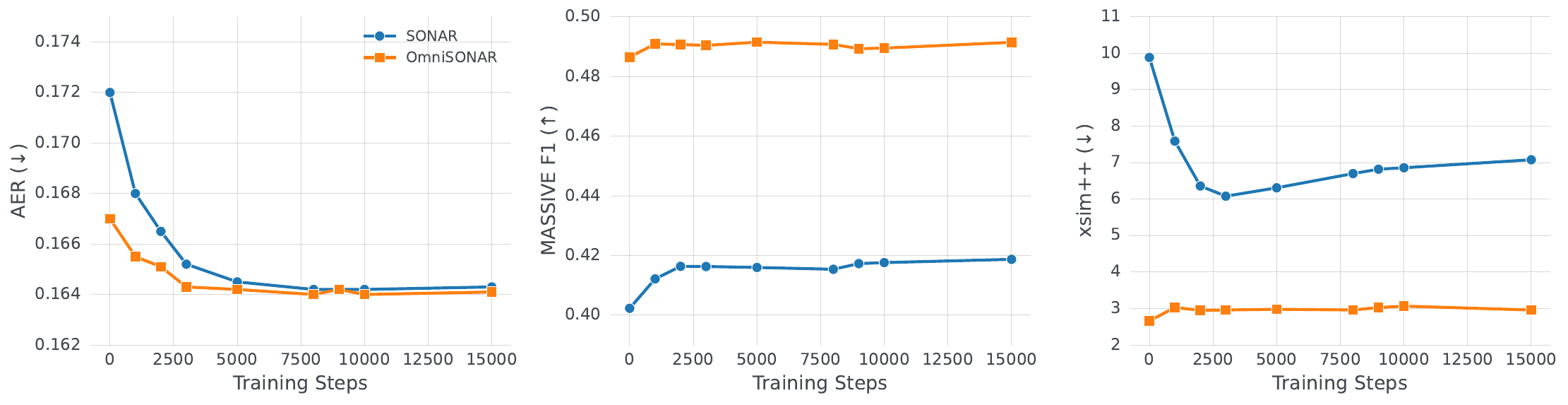}
    \caption{Training dynamics of SONAR and \sonar{} trained with the \tokensonar{} objective. \label{fig:sonar_dynamics}}
\end{figure}

\paragraph{Training dynamics} \Cref{fig:sonar_dynamics} shows the training dynamics of the \tokensonar{}'s objective (\autoref{eq:so_loss}) when applied to both \sonar{} and SONAR. We can see a similar trend for both encoders: the AER improves throughout training, mirroring the progression of the performance on the Massive task. 
While xsim++ does not improve for \sonar{}, our token-level objective improves xsim++ for SONAR.
This can be explained by the fact that SONAR is a more collapsed space, as it was not trained with a contrastive loss, contrary to \sonar{}.

In this section, we show that it is possible to further train \sonar{} with a token-level loss that improves the token cross-lingual alignment.
\tokensonar{} exhibits sentence-level performance almost identical to \sonar{} while consistently improving token-level cross-lingual performance. Notice that the resulting encoder is still compatible with \sonar{}, they both encode sequences in the same representation space.

\subsection{Extending the Context Length of \sonar{}}
\label{sec:longcontext}
Through this work we explored the use of \sonar{} for sentence representations across languages, modalities, and as Language Model inputs. In the previous subsection, \Cref{sec:tokensonar}, we explored how the learned representations of sentence models can be leveraged and improved at a smaller granularity, which naturally leads to the question: Can a cross-lingual sentence embedding model be expanded to text beyond sentence-length?

Previous work on general purpose embeddings, such as recent LLM based models \citep{qwen3embedding, embedding_gemma_2025} naturally cover multiple granularities as part of their design. From custom prompts per granularity to training data that matches queries, usually at sentence-level, with documents, once trained, their representations are flexible across query and document lengths. However, for \sonar{} and other crosslingual models discussed in this work, their reliance on sentence-level paired data at train time usually restricts their representations at that level at inference time. In this section we will explore those questions, such as whether these models already enable paragraph or document level mining, show they do not, and provide a recipe to expand \sonar{} for longer context, namely \sonar{} Long-context.

\subsubsection*{Training Data}
We leverage paragraph-level aligned data mined, using the mining methodology described in Omnilingual MT~\citep{omtbigpaper}. This corpus provides high-quality cross-lingual document alignments across multiple language pairs. Unlike sentence-level training, this data contains naturally longer text spans that better reflect real-world document mining scenarios.

\subsubsection*{Hard Negatives for Long-Context Evaluation}
Standard cross-lingual similarity search benchmarks beyond sentence-level lack hard negatives, making them poor assessors of ``mining in the wild'' performance, a simple first-sentence match can achieve high scores. To address this, we create challenging evaluation sets with hard negatives for two benchmarks:

\paragraph{Belebele Hard Negatives.} Belebele~\citep{bandarkar-etal-2024-belebele} is a multiple choice machine reading comprehension dataset spanning 122 language variants. We take their context which was built from contiguous FLORES sentences, therefore we can extend it to the 201 languages covered by FLORES, and disregard the question and answer parts. We augment this benchmark by replacing sentences within paragraphs with hard negative counterparts derived from FLORES+. These modifications include subtle semantic changes such as entity substitutions or negations, testing the model's ability to encode fine-grained semantic nuances. We report xsim++ scores on this augmented benchmark. %
\paragraph{IWSLT2017 Hard Negatives.} For document-level evaluation, we use the IWSLT2017 corpus and create five types of hard negatives per sample:
\begin{itemize}
    \item Removing the last sentence(s) from the document
    \item Replacing the last sentence(s) with content from other documents in the same language
    \item Removing a random sentence from within the document
    \item Replacing a random sentence with content from other documents
    \item Shuffling the sentence order within the document
\end{itemize}
This tests whether models capture holistic document semantics rather than relying on surface-level features like document headers.

\subsubsection*{Model Training}
Encoding longer inputs comes at higher cost due to the attention mechanism. To have a lighter memory budget than using the large counterpart we use the same architecture as \sonar{}-tiny and distill OmniSONAR into a smaller encoder as explained in  \Cref{subsec:distilled_encoders_performance}, however we extend its vocabulary from 32K to 128K vocabulary to preserve a reasonable fertility for long documents. This leads to a 343M parameter model. Starting from this distilled checkpoint, we perform lightweight continued training on the paragraph-level mined data using a contrastive loss with in-batch negatives. This approach preserves the cross-lingual alignment learned during sentence-level training while extending the model's effective context length.
The training objective follows the standard InfoNCE loss:
\begin{equation}
    \mathcal{L} = -\frac{1}{N} \sum_{i=1}^{N} \log \frac{e^{\phi(x_i, y_i) \cdot \tau}}{\sum_{j=1}^{N} e^{\phi(x_i, y_j) \cdot \tau}}
\end{equation}
where $\phi(x_i, y_i)$ denotes the cosine similarity between source document $x_i$ and target document $y_i$, and $\tau$ is the logit scale, which we set at 20. The model is trained for 9,000 steps with 128 document pairs per batch, across 8 GPUs, and a learning rate of 1e-5, with 900 linear warmup steps.

\subsubsection*{Results}
\paragraph{Sentence-Level Performance Preservation.}
A critical question is whether extending to long contexts degrades performance on sentence-level tasks. Table~\ref{tab:lc-preservation} compares \sonar{}-LC against the full \sonar{} model on both cross-lingual similarity search (FLORES) and downstream MTEB benchmarks. Despite a 5$\times$ reduction in parameters, \sonar{}-LC closely matches OmniSONAR across all metrics. On FLORES, we observe only modest increases in error rates (xsim++: 4.02\% vs.\ 3.51\% on 80 languages), compared to \sonar{}-tiny which has the same number of model parameters (the only difference is the vocabulary). Remarkably, on MTEB tasks, \sonar{}-LC improves over the original model on average score (74.43 vs.\ 74.11), classification (71.27 vs.\ 71.14), and STS (82.42 vs.\ 82.16). We hypothesize that exposure to longer, more coherent text spans during continued training provides additional semantic context that benefits sentence-level understanding. These results demonstrate that extending context length can be achieved without sacrificing short-context capabilities.
\begin{table}[t]
\centering
\caption{Sentence-level performance comparison between \sonar{} and \sonar{}-LC on FLORES (error rate $\downarrow$) and MTEB (score $\uparrow$).}
\label{tab:lc-preservation}
\begin{tabular}{l|cc|cc|cccc}
\toprule
& \multicolumn{2}{c|}{\textbf{FLORES (80)}} & \multicolumn{2}{c|}{\textbf{FLORES (200)}} & \multicolumn{4}{c}{\textbf{MTEB}} \\
\textbf{Model} & xsim & xsim++ & xsim & xsim++ & Avg & Class. & Pair & STS \\
\midrule
\sonarlogoinline{} & \textbf{0.05} & \textbf{2.95} & \textbf{0.65} & \textbf{6.14} & 74.11 & 71.14 & 69.04 & 82.16 \\
\sonarlogoinline{}-tiny & 0.06 & 3.51 & 1.01 & 7.82 & 73.77 & 70.16 & \textbf{69.97} & 81.18 \\
\sonarlogoinline{}-LC  & 0.06 & 4.02 & 1.11 & 9.20 & \textbf{74.43} & \textbf{71.27} & 69.60 & \textbf{82.42} \\
\bottomrule
\end{tabular}
\end{table}

\begin{table}[t]
\centering
\caption{Cross-lingual similarity search error rates on Belebele paragraphs.}
\label{tab:lc-belebele}
\begin{tabular}{l|cc|cc}
\toprule
& \multicolumn{2}{c|}{\textbf{xsim} ($\downarrow$)} & \multicolumn{2}{c}{\textbf{xsim++} ($\downarrow$)} \\
\textbf{Model} & Common (80) & All & Common (80) & All \\
\midrule
\embGemma{} & 2.85 & 11.23 & 31.06 & 42.80 \\
\meFive{}  & 2.61 & 6.49 & 32.52 & 35.00\\
\sonarlogoinline{} & 0.01 & \textbf{0.15} & 1.50 & \textbf{2.15} \\
\sonarlogoinline{}-tiny & 0.02 & 0.50 & 2.28 & 3.34 \\
\sonarlogoinline{}-LC & \textbf{0.00} & 0.27 & \textbf{1.49} & 5.05 \\
\bottomrule
\end{tabular}
\end{table}

\begin{table}[ht]
\centering
\caption{Cross-lingual similarity search error rates on IWSLT2017 documents (X$\rightarrow$Eng).}
\label{tab:lc-iwslt}
\begin{tabular}{l|ccccccccc|c}
\toprule
& \multicolumn{8}{c}{\textbf{xsim} ($\downarrow$)} \\
\textbf{Model} & ar & de & fr & it & jp & ko & nl & ro & zh & Avg \\
\midrule
\embGemma{} & 
3.53 & 1.55 & 1.67 & 2.19 & 4.54 & 3.98 & 1.86 & 3.83 & 3.19 & 2.93\\
\meFive{} &
3.22 & 1.52 & 1.67 & 1.59 & 5.14 & 4.72 & 1.29 & 1.94 & 3.84 & 2.77\\
\sonarlogoinline{}  &  \textbf{1.81} & \textbf{0.94} & \textbf{1.18} & 1.16 & \textbf{3.23} & \textbf{2.36} & \textbf{0.94} & \textbf{1.26} & \textbf{2.06} & \textbf{1.66}\\
\sonarlogoinline{}-tiny  & 1.99 & 1.02 & 1.21 & 1.16 & 3.43 & 2.61 & 0.97 & 1.34 & 2.21 & 1.77\\
\sonarlogoinline{}-LC & 1.90 & 1.01 & \textbf{1.18} & \textbf{1.15} & \textbf{3.23} & 2.55 &  \textbf{0.94} & 1.29 & 2.15 & 1.71 \\
\midrule
& \multicolumn{8}{c}{\textbf{xsim++} ($\downarrow$)} \\
\embGemma{} &
19.51 & 7.38 & 7.61 & 13.49 & 18.21 & 20.24 & 11.98 & 24.03 & 14.86 & 15.26\\
\meFive{} & 
36.36 & 19.01 & 21.02 & 15.26 & 41.64 & 40.59 & 16.55 & 29.81 & 34.28 & 28.28\\
\sonarlogoinline{}  & 16.80 & \textbf{1.21} & 1.57 & 1.58 & 11.69 & 17.31 & 1.35 & 1.65 & 16.01 & 7.69\\
\sonarlogoinline{}-tiny  & 28.19 & 5.66 & 5.14 & 3.80 & 13.34 & 34.40 & 4.63 & 6.57 & 19.15 & 13.43\\
\sonarlogoinline{}-LC & \textbf{2.62} & 1.22 & \textbf{1.44} & \textbf{1.42} & \textbf{4.55} & \textbf{4.05} & \textbf{1.21} & \textbf{1.47} & \textbf{2.96} & \textbf{2.33}
\\
\bottomrule
\end{tabular}
\end{table}

\paragraph{Belebele Paragraph Mining.}
\Cref{tab:lc-belebele} presents cross-lingual similarity search results on the Belebele benchmark. Without hard negatives (xsim), all models perform well on high-resource languages. However, with hard negatives (xsim++), the picture changes dramatically: general-purpose embeddings show error rates exceeding 30\%, while \longcontextsonar{} maintains robust performance across the full language spectrum. However, on this benchmark, \longcontextsonar{} does not surpass the performance of \sonar{}. This highlights that for short texts, as Belebele has an average of 3 sentences per sample, our sentence-based models preserve performance. Our long-context extension has a harder time capturing the nuanced negatives from FLORES within the paragraphs beyond the common 80 languages, probably due to the lack of hard negatives during its training and its training data not covering all 200 languages.

\paragraph{IWSLT2017 Document Mining.}
\Cref{tab:lc-iwslt} shows results on IWSLT2017 with document-level hard negatives. While all models achieve low error rates on standard xsim, the xsim++ results reveal a critical weakness: models not explicitly trained for long-context cross-lingual alignment show dramatically increased error rates when hard negatives are introduced. This points to sentence-level or English-centric training induces a bias toward initial document portions, causing failures when discriminative features appear later or within the documents. For its model size, \longcontextsonar{} vastly outperforms its similar sized tiny counterpart and keeps a 5 points lead over the full sized \sonar{}.

\newpage
\section{Conclusion}
\label{section:conclusion}

In this work, we introduce \sonar{}, a state-of-the-art cross-lingual sentence encoder covering over 4.2k text language varieties and 177 spoken languages. This work represents a major leap toward the ultimate goal of a truly language- and modality-agnostic semantic space, where sentences with identical meanings share aligned vector representations while maintaining strong and robust downstream performance. To achieve this unprecedented scale, we proposed a novel training paradigm for representation learning that couples token-level translation decoding with a contrastive loss objective, further expanded by specialized teacher-student distillation techniques.

\sonar{} establishes the first truly \textit{omnilingual} embedding space, making it by far the most massively multilingual embedding model for both the text and speech modalities. Empirically, it redefines the state of the art in cross-lingual alignment, where the error rates are more than halved for the top 200 languages and reduced by 15 times for the long-tail of 1,560 languages.
For the speech modality, cross-lingual and cross-modal similarity search is greatly improved with \sonar{}-speech compared to its predecessor SONAR, with a 43\% error rate reduction, while also supporting 2.5 times more languages within a single model.
Beyond retrieval, \sonar{} showcases impressive decoding capabilities, reaching the translation quality levels of NLLB-3B and OmniASR for text and speech translation respectively, despite relying on an intermediate fixed-size representation.
Crucially, this massive alignment does not come at the expense of general-purpose utility, with \sonar{} excelling across numerous downstream tasks, closing the gap with general-purpose embeddings that typically suffer from poor cross-lingual transfer and alignment.
Furthermore, as demonstrated by the Spectrum architecture, \sonar{} vectors contain rich semantic density which enables language modeling based on these vectors in many languages, for both speech and text, even when the model has been exclusively trained on English text \sonar{} vectors.

While our current focus centers on translation-based cross-lingual alignment, we believe future research can leverage our framework to build universal, general-purpose embedding models.
Ultimately, its combination of unprecedented linguistic coverage, state-of-the-art alignment, strong downstream performance, and cross-modal extensibility positions \sonar{} as a definitive foundational representation for the next generation of global NLP and speech applications.

\newpage
\section{Contribution Statement}
We outline the contributions from different team members and organize them into sections.

\subsection{\sonar{}}
\begin{itemize}
\item João Maria Janeiro - Core contributor. All contrastive training stages, learning objectives (split-softmax) and hard negative generation. Architecture and Evaluation.
\item Pere Lluís Huguet Cabot - Core contributor. Math and Code. Data curation. Architecture and Seq2seq training. Distilled and long-context encoders.
\item Ioannis Tsiamas - Core contributor. Omnilingual Extension: Implementation, experiments, and analysis. Omnilingual data curation.
\item Yen Meng - Core contributor. Speech Extension.
\item Belen Alastruey - SaT models.
\item Artyom Kozhevnikov - Backtranslation efforts, and data loading code.
\item David Dale - Tokenization and supervision on Omnilingual extension.
\item Alexandre Mourachko - Research Manager, helped with the overall direction, strategy and resourcing plan.
\item Christophe Ropers - Linguist support.
\item Yu-An Chung - Technical supervision on speech extension.
\item Marta R. Costa-Jussa - Technical supervision on the omnilingual extension.
\item Paul-Ambroise Duquenne - Technical Lead of OmniSONAR. Core contributor. Training and Architecture.
\end{itemize}
\subsection{\mdocu{}}
\begin{itemize}
\item Vivek Iyer - Core contributor. Instruction Tuning. Evaluation. Datasets and Benchmarks preparation. 
\item Pere Lluís Huguet Cabot - Core contributor. Pre-training. Evaluation.
\item Jaehyeong Jo - Initial contributor to pre-training experiments.
\item Kevin Heffernan - Initial contributor to pre-training experiments.
\item Artyom Kozhevnikov - Data Loading for pre-computed embeddings.
\item Alexandre Mourachko - Research Manager, helped with the overall direction, strategy and resourcing plan.
\item Holger Schwenk - Technical Supervision. 
\item Loic Barrault - Co-Technical Lead of Spectrum. Core contributor. Dataset preparation.
\item Paul-Ambroise Duquenne - Co-Technical Lead of Spectrum. Core contributor. Pretraining.
\end{itemize}
\subsection{\tokensonar{}}
\begin{itemize}
\item Guillem Ramírez - Core contributor of Token-SONAR
\item Loic Barrault - Co-Technical Lead of Token-SONAR. 
\item David Dale - Co-Technical Lead of Token-SONAR. 
\end{itemize}

\subsection{Acknowledgement}
We gratefully acknowledge the contributions of the following colleagues, whose support was essential to this work:
\begin{itemize}
\item Andrea Caciolai for his help with datamix preparation.
\item Alexander Erben and Yunchao Yang for fairseq2 update of \mdocu{} codebase.
\end{itemize}

\newpage
\bibliographystyle{assets/plainnat}
\bibliography{paper, anthology-1, anthology-2}

@string{acl = {Association for Computational Linguistics}}

@string{anth = {https://aclanthology.org/}}

@inproceedings{kocmi-etal-2025-findings,title = "Findings of the {WMT}25 General Machine Translation Shared Task: Time to Stop Evaluating on Easy Test Sets",author = "Kocmi, Tom and Artemova, Ekaterina and Avramidis, Eleftherios and Bawden, Rachel and Bojar, Ond{\v{r}}ej and Dranch, Konstantin and Dvorkovich, Anton and Dukanov, Sergey and Fishel, Mark and Freitag, Markus and Gowda, Thamme and Grundkiewicz, Roman and Haddow, Barry and Karpinska, Marzena and Koehn, Philipp and Lakougna, Howard and Lundin, Jessica and Monz, Christof and Murray, Kenton and Nagata, Masaaki and Perrella, Stefano and Proietti, Lorenzo and Popel, Martin and Popovi{\'c}, Maja and Riley, Parker and Shmatova, Mariya and Steingr{\'i}msson, Steinth{\'o}r and Yankovskaya, Lisa and Zouhar, Vil{\'e}m",editor = "Haddow, Barry and Kocmi, Tom and Koehn, Philipp and Monz, Christof",booktitle = "Proceedings of the Tenth Conference on Machine Translation",month = nov,year = "2025",address = "Suzhou, China",publisher = acl,url = anth # {2025.wmt-1.22/},doi = "10.18653/v1/2025.wmt-1.22",pages = "355--413",ISBN = "979-8-89176-341-8"}

@inproceedings{ma-etal-2025-taxi1500,title = "Taxi1500: A Dataset for Multilingual Text Classification in 1500 Languages",author = "Ma, Chunlan and Imani, Ayyoob and Ye, Haotian and Pei, Renhao and Asgari, Ehsaneddin and Schuetze, Hinrich",editor = "Chiruzzo, Luis and Ritter, Alan and Wang, Lu",booktitle = "Proceedings of the 2025 Conference of the Nations of the Americas Chapter of the Association for Computational Linguistics: Human Language Technologies (Volume 2: Short Papers)",month = apr,year = "2025",address = "Albuquerque, New Mexico",publisher = acl,url = anth # {2025.naacl-short.36/},doi = "10.18653/v1/2025.naacl-short.36",pages = "414--439",ISBN = "979-8-89176-190-2"}

@inproceedings{moroni-etal-2025-optimizing,title = "Optimizing {LLM}s for {I}talian: Reducing Token Fertility and Enhancing Efficiency Through Vocabulary Adaptation",author = "Moroni, Luca and Puccetti, Giovanni and Huguet Cabot, Pere-Llu{\'i}s and Bejgu, Andrei Stefan and Miaschi, Alessio and Barba, Edoardo and Dell{'}Orletta, Felice and Esuli, Andrea and Navigli, Roberto",editor = "Chiruzzo, Luis and Ritter, Alan and Wang, Lu",booktitle = "Findings of the Association for Computational Linguistics: NAACL 2025",month = apr,year = "2025",address = "Albuquerque, New Mexico",publisher = acl,url = anth # {2025.findings-naacl.371/},doi = "10.18653/v1/2025.findings-naacl.371",pages = "6661--6675",ISBN = "979-8-89176-195-7"}

@inproceedings{iyer-etal-2025-xl,title = "{XL}-Suite: Cross-Lingual Synthetic Training and Evaluation Data for Open-Ended Generation",author = "Iyer, Vivek and Chen, Pinzhen and Rei, Ricardo and Birch, Alexandra",editor = "Christodoulopoulos, Christos and Chakraborty, Tanmoy and Rose, Carolyn and Peng, Violet",booktitle = "Findings of the Association for Computational Linguistics: EMNLP 2025",month = nov,year = "2025",address = "Suzhou, China",publisher = acl,url = anth # {2025.findings-emnlp.550/},doi = "10.18653/v1/2025.findings-emnlp.550",pages = "10418--10432",ISBN = "979-8-89176-335-7"}

@inproceedings{costa-jussa-etal-2025-2m,title = "2{M}-{BELEBELE}: Highly Multilingual Speech and {A}merican {S}ign {L}anguage Comprehension Dataset Download {PDF}",author = "Costa-juss{\`a}, Marta R. and Yu, Bokai and Andrews, Pierre and Alastruey, Belen and Camgoz, Necati Cihan and Chuang, Joe and Maillard, Jean and Ropers, Christophe and Turkatenko, Arina and Wood, Carleigh",editor = "Che, Wanxiang and Nabende, Joyce and Shutova, Ekaterina and Pilehvar, Mohammad Taher",booktitle = "Findings of the Association for Computational Linguistics: ACL 2025",month = jul,year = "2025",address = "Vienna, Austria",publisher = acl,url = anth # {2025.findings-acl.569/},doi = "10.18653/v1/2025.findings-acl.569",pages = "10893--10904",ISBN = "979-8-89176-256-5"}

@inproceedings{janeiro-etal-2025-mixture,title = "Mixture of Languages: Improved Multilingual Encoders Through Language Grouping",author = "Janeiro, Jo{\~a}o Maria and Alastruey, Belen and Massa, Francisco and Elbayad, Maha and Piwowarski, Benjamin and Gallinari, Patrick and Barrault, Loic",editor = "Christodoulopoulos, Christos and Chakraborty, Tanmoy and Rose, Carolyn and Peng, Violet",booktitle = "Proceedings of the 2025 Conference on Empirical Methods in Natural Language Processing",month = nov,year = "2025",address = "Suzhou, China",publisher = acl,url = anth # {2025.emnlp-main.1509/},doi = "10.18653/v1/2025.emnlp-main.1509",pages = "29707--29722",ISBN = "979-8-89176-332-6"}

@inproceedings{janeiro-etal-2025-mexma,title = "{MEXMA}: Token-level objectives improve sentence representations",author = "Janeiro, Jo{\~a}o Maria and Piwowarski, Benjamin and Gallinari, Patrick and Barrault, Loic",editor = "Che, Wanxiang and Nabende, Joyce and Shutova, Ekaterina and Pilehvar, Mohammad Taher",booktitle = "Proceedings of the 63rd Annual Meeting of the Association for Computational Linguistics (Volume 1: Long Papers)",month = jul,year = "2025",address = "Vienna, Austria",publisher = acl,url = anth # {2025.acl-long.1168/},doi = "10.18653/v1/2025.acl-long.1168",pages = "23960--23995",ISBN = "979-8-89176-251-0"}

@inproceedings{parekh-etal-2024-contextual,title = "Contextual Label Projection for Cross-Lingual Structured Prediction",author = "Parekh, Tanmay and Hsu, I-Hung and Huang, Kuan-Hao and Chang, Kai-Wei and Peng, Nanyun",editor = "Duh, Kevin and Gomez, Helena and Bethard, Steven",booktitle = "Proceedings of the 2024 Conference of the North American Chapter of the Association for Computational Linguistics: Human Language Technologies (Volume 1: Long Papers)",month = jun,year = "2024",address = "Mexico City, Mexico",publisher = acl,url = anth # {2024.naacl-long.321/},doi = "10.18653/v1/2024.naacl-long.321",pages = "5738--5757"}

@inproceedings{miao-etal-2024-enhancing,title = "Enhancing Cross-lingual Sentence Embedding for Low-resource Languages with Word Alignment",author = "Miao, Zhongtao and Wu, Qiyu and Zhao, Kaiyan and Wu, Zilong and Tsuruoka, Yoshimasa",editor = "Duh, Kevin and Gomez, Helena and Bethard, Steven",booktitle = "Findings of the Association for Computational Linguistics: NAACL 2024",month = jun,year = "2024",address = "Mexico City, Mexico",publisher = acl,url = anth # {2024.findings-naacl.204/},doi = "10.18653/v1/2024.findings-naacl.204",pages = "3225--3236"}

@inproceedings{costa-jussa-etal-2024-mutox,title = "{M}u{T}ox: Universal {MU}ltilingual Audio-based {TOX}icity Dataset and Zero-shot Detector",author = "Costa-juss{\`a}, Marta R. and Meglioli, Mariano Coria and Andrews, Pierre and Dale, David and Hansanti, Prangthip and Kalbassi, Elahe and Mourachko, Alex and Ropers, Christophe and Wood, Carleigh",editor = "Ku, Lun-Wei and Martins, Andre and Srikumar, Vivek",booktitle = "Findings of the Association for Computational Linguistics: ACL 2024",month = aug,year = "2024",address = "Bangkok, Thailand",publisher = acl,url = anth # {2024.findings-acl.340/},doi = "10.18653/v1/2024.findings-acl.340",pages = "5725--5734"}

@inproceedings{huang-etal-2024-ottawa,title = "{OTTAWA}: Optimal {T}ranspor{T} Adaptive Word Aligner for Hallucination and Omission Translation Errors Detection",author = "Huang, Chenyang and Ghaddar, Abbas and Kobyzev, Ivan and Rezagholizadeh, Mehdi and Zaiane, Osmar and Chen, Boxing",editor = "Ku, Lun-Wei and Martins, Andre and Srikumar, Vivek",booktitle = "Findings of the Association for Computational Linguistics: ACL 2024",month = aug,year = "2024",address = "Bangkok, Thailand",publisher = acl,url = anth # {2024.findings-acl.377/},doi = "10.18653/v1/2024.findings-acl.377",pages = "6322--6334"}

@inproceedings{marchisio-etal-2024-understanding,title = "Understanding and Mitigating Language Confusion in {LLM}s",author = "Marchisio, Kelly and Ko, Wei-Yin and Berard, Alexandre and Dehaze, Th{\'e}o and Ruder, Sebastian",editor = "Al-Onaizan, Yaser and Bansal, Mohit and Chen, Yun-Nung",booktitle = "Proceedings of the 2024 Conference on Empirical Methods in Natural Language Processing",month = nov,year = "2024",address = "Miami, Florida, USA",publisher = acl,url = anth # {2024.emnlp-main.380/},doi = "10.18653/v1/2024.emnlp-main.380",pages = "6653--6677"}

@inproceedings{adelani-etal-2024-sib,title = "{SIB}-200: A Simple, Inclusive, and Big Evaluation Dataset for Topic Classification in 200+ Languages and Dialects",author = "Adelani, David Ifeoluwa and Liu, Hannah and Shen, Xiaoyu and Vassilyev, Nikita and Alabi, Jesujoba O. and Mao, Yanke and Gao, Haonan and Lee, En-Shiun Annie",editor = "Graham, Yvette and Purver, Matthew",booktitle = "Proceedings of the 18th Conference of the European Chapter of the Association for Computational Linguistics (Volume 1: Long Papers)",month = mar,year = "2024",address = "St. Julian{'}s, Malta",publisher = acl,url = anth # {2024.eacl-long.14/},doi = "10.18653/v1/2024.eacl-long.14",pages = "226--245"}

@inproceedings{bandarkar-etal-2024-belebele,title = "The Belebele Benchmark: a Parallel Reading Comprehension Dataset in 122 Language Variants",author = "Bandarkar, Lucas and Liang, Davis and Muller, Benjamin and Artetxe, Mikel and Shukla, Satya Narayan and Husa, Donald and Goyal, Naman and Krishnan, Abhinandan and Zettlemoyer, Luke and Khabsa, Madian",editor = "Ku, Lun-Wei and Martins, Andre and Srikumar, Vivek",booktitle = "Proceedings of the 62nd Annual Meeting of the Association for Computational Linguistics (Volume 1: Long Papers)",month = aug,year = "2024",address = "Bangkok, Thailand",publisher = acl,url = anth # {2024.acl-long.44/},doi = "10.18653/v1/2024.acl-long.44",pages = "749--775"}

@article{zhang-etal-2023-miracl,title = "{MIRACL}: A Multilingual Retrieval Dataset Covering 18 Diverse Languages",author = "Zhang, Xinyu and Thakur, Nandan and Ogundepo, Odunayo and Kamalloo, Ehsan and Alfonso-Hermelo, David and Li, Xiaoguang and Liu, Qun and Rezagholizadeh, Mehdi and Lin, Jimmy",journal = "Transactions of the Association for Computational Linguistics",volume = "11",year = "2023",address = "Cambridge, MA",publisher = "MIT Press",url = anth # {2023.tacl-1.63/},doi = "10.1162/tacl_a_00595",pages = "1114--1131"}

@inproceedings{azadi-etal-2023-pmi,title = "{PMI}-Align: Word Alignment With Point-Wise Mutual Information Without Requiring Parallel Training Data",author = "Azadi, Fatemeh and Faili, Heshaam and Dousti, Mohammad Javad",editor = "Rogers, Anna and Boyd-Graber, Jordan and Okazaki, Naoaki",booktitle = "Findings of the Association for Computational Linguistics: ACL 2023",month = jul,year = "2023",address = "Toronto, Canada",publisher = acl,url = anth # {2023.findings-acl.782/},doi = "10.18653/v1/2023.findings-acl.782",pages = "12366--12377"}

@inproceedings{lai-etal-2023-okapi,title = "Okapi: Instruction-tuned Large Language Models in Multiple Languages with Reinforcement Learning from Human Feedback",author = "Lai, Viet and Nguyen, Chien and Ngo, Nghia and Nguyen, Thuat and Dernoncourt, Franck and Rossi, Ryan and Nguyen, Thien",editor = "Feng, Yansong and Lefever, Els",booktitle = "Proceedings of the 2023 Conference on Empirical Methods in Natural Language Processing: System Demonstrations",month = dec,year = "2023",address = "Singapore",publisher = acl,url = anth # {2023.emnlp-demo.28/},doi = "10.18653/v1/2023.emnlp-demo.28",pages = "318--327"}

@inproceedings{muennighoff-etal-2023-mteb,title = "{MTEB}: Massive Text Embedding Benchmark",author = "Muennighoff, Niklas and Tazi, Nouamane and Magne, Loic and Reimers, Nils",editor = "Vlachos, Andreas and Augenstein, Isabelle",booktitle = "Proceedings of the 17th Conference of the European Chapter of the Association for Computational Linguistics",month = may,year = "2023",address = "Dubrovnik, Croatia",publisher = acl,url = anth # {2023.eacl-main.148/},doi = "10.18653/v1/2023.eacl-main.148",pages = "2014--2037"}

@inproceedings{chen-etal-2023-xsim,title = "x{SIM}++: An Improved Proxy to Bitext Mining Performance for Low-Resource Languages",author = "Chen, Mingda and Heffernan, Kevin and {\c{C}}elebi, Onur and Mourachko, Alexandre and Schwenk, Holger",editor = "Rogers, Anna and Boyd-Graber, Jordan and Okazaki, Naoaki",booktitle = "Proceedings of the 61st Annual Meeting of the Association for Computational Linguistics (Volume 2: Short Papers)",month = jul,year = "2023",address = "Toronto, Canada",publisher = acl,url = anth # {2023.acl-short.10/},doi = "10.18653/v1/2023.acl-short.10",pages = "101--109"}

@inproceedings{minixhofer-etal-2023-wheres,title = "Where{'}s the Point? Self-Supervised Multilingual Punctuation-Agnostic Sentence Segmentation",author = "Minixhofer, Benjamin and Pfeiffer, Jonas and Vuli{\'c}, Ivan",editor = "Rogers, Anna and Boyd-Graber, Jordan and Okazaki, Naoaki",booktitle = "Proceedings of the 61st Annual Meeting of the Association for Computational Linguistics (Volume 1: Long Papers)",month = jul,year = "2023",address = "Toronto, Canada",publisher = acl,url = anth # {2023.acl-long.398/},doi = "10.18653/v1/2023.acl-long.398",pages = "7215--7235"}

@inproceedings{chen-etal-2023-blaser,title = "{BLASER}: A Text-Free Speech-to-Speech Translation Evaluation Metric",author = "Chen, Mingda and Duquenne, Paul-Ambroise and Andrews, Pierre and Kao, Justine and Mourachko, Alexandre and Schwenk, Holger and Costa-juss{\`a}, Marta R.",editor = "Rogers, Anna and Boyd-Graber, Jordan and Okazaki, Naoaki",booktitle = "Proceedings of the 61st Annual Meeting of the Association for Computational Linguistics (Volume 1: Long Papers)",month = jul,year = "2023",address = "Toronto, Canada",publisher = acl,url = anth # {2023.acl-long.504/},doi = "10.18653/v1/2023.acl-long.504",pages = "9064--9079"}

@inproceedings{aepli-sennrich-2022-improving,title = "Improving Zero-Shot Cross-lingual Transfer Between Closely Related Languages by Injecting Character-Level Noise",author = {Aepli, No{\"e}mi and Sennrich, Rico},editor = "Muresan, Smaranda and Nakov, Preslav and Villavicencio, Aline",booktitle = "Findings of the Association for Computational Linguistics: ACL 2022",month = may,year = "2022",address = "Dublin, Ireland",publisher = acl,url = anth # {2022.findings-acl.321/},doi = "10.18653/v1/2022.findings-acl.321",pages = "4074--4083"}

@inproceedings{gee-etal-2022-fast,title = "Fast Vocabulary Transfer for Language Model Compression",author = "Gee, Leonidas and Zugarini, Andrea and Rigutini, Leonardo and Torroni, Paolo",editor = "Li, Yunyao and Lazaridou, Angeliki",booktitle = "Proceedings of the 2022 Conference on Empirical Methods in Natural Language Processing: Industry Track",month = dec,year = "2022",address = "Abu Dhabi, UAE",publisher = acl,url = anth # {2022.emnlp-industry.41/},doi = "10.18653/v1/2022.emnlp-industry.41",pages = "409--416"}

@inproceedings{feng-etal-2022-language,title = "Language-agnostic {BERT} Sentence Embedding",author = "Feng, Fangxiaoyu and Yang, Yinfei and Cer, Daniel and Arivazhagan, Naveen and Wang, Wei",editor = "Muresan, Smaranda and Nakov, Preslav and Villavicencio, Aline",booktitle = "Proceedings of the 60th Annual Meeting of the Association for Computational Linguistics (Volume 1: Long Papers)",month = may,year = "2022",address = "Dublin, Ireland",publisher = acl,url = anth # {2022.acl-long.62/},doi = "10.18653/v1/2022.acl-long.62",pages = "878--891"}

@inproceedings{gao-etal-2021-simcse,title = "{S}im{CSE}: Simple Contrastive Learning of Sentence Embeddings",author = "Gao, Tianyu and Yao, Xingcheng and Chen, Danqi",editor = "Moens, Marie-Francine and Huang, Xuanjing and Specia, Lucia and Yih, Scott Wen-tau",booktitle = "Proceedings of the 2021 Conference on Empirical Methods in Natural Language Processing",month = nov,year = "2021",address = "Online and Punta Cana, Dominican Republic",publisher = acl,url = anth # {2021.emnlp-main.552/},doi = "10.18653/v1/2021.emnlp-main.552",pages = "6894--6910"}

@inproceedings{oneill-etal-2021-wish,title = "{I} Wish {I} Would Have Loved This One, But {I} Didn{'}t {--} A Multilingual Dataset for Counterfactual Detection in Product Review",author = "O{'}Neill, James and Rozenshtein, Polina and Kiryo, Ryuichi and Kubota, Motoko and Bollegala, Danushka",editor = "Moens, Marie-Francine and Huang, Xuanjing and Specia, Lucia and Yih, Scott Wen-tau",booktitle = "Proceedings of the 2021 Conference on Empirical Methods in Natural Language Processing",month = nov,year = "2021",address = "Online and Punta Cana, Dominican Republic",publisher = acl,url = anth # {2021.emnlp-main.568/},doi = "10.18653/v1/2021.emnlp-main.568",pages = "7092--7108"}

@inproceedings{dou-neubig-2021-word,title = "Word Alignment by Fine-tuning Embeddings on Parallel Corpora",author = "Dou, Zi-Yi and Neubig, Graham",editor = "Merlo, Paola and Tiedemann, Jorg and Tsarfaty, Reut",booktitle = "Proceedings of the 16th Conference of the European Chapter of the Association for Computational Linguistics: Main Volume",month = apr,year = "2021",address = "Online",publisher = acl,url = anth # {2021.eacl-main.181/},doi = "10.18653/v1/2021.eacl-main.181",pages = "2112--2128"}

@inproceedings{li-etal-2021-mtop,title = "{MTOP}: A Comprehensive Multilingual Task-Oriented Semantic Parsing Benchmark",author = "Li, Haoran and Arora, Abhinav and Chen, Shuohui and Gupta, Anchit and Gupta, Sonal and Mehdad, Yashar",editor = "Merlo, Paola and Tiedemann, Jorg and Tsarfaty, Reut",booktitle = "Proceedings of the 16th Conference of the European Chapter of the Association for Computational Linguistics: Main Volume",month = apr,year = "2021",address = "Online",publisher = acl,url = anth # {2021.eacl-main.257/},doi = "10.18653/v1/2021.eacl-main.257",pages = "2950--2962"}

@inproceedings{chi-etal-2021-improving,title = "Improving Pretrained Cross-Lingual Language Models via Self-Labeled Word Alignment",author = "Chi, Zewen and Dong, Li and Zheng, Bo and Huang, Shaohan and Mao, Xian-Ling and Huang, Heyan and Wei, Furu",editor = "Zong, Chengqing and Xia, Fei and Li, Wenjie and Navigli, Roberto",booktitle = "Proceedings of the 59th Annual Meeting of the Association for Computational Linguistics and the 11th International Joint Conference on Natural Language Processing (Volume 1: Long Papers)",month = aug,year = "2021",address = "Online",publisher = acl,url = anth # {2021.acl-long.265/},doi = "10.18653/v1/2021.acl-long.265",pages = "3418--3430"}

@inproceedings{schwenk-etal-2021-ccmatrix,title = "{CCM}atrix: Mining Billions of High-Quality Parallel Sentences on the Web",author = "Schwenk, Holger and Wenzek, Guillaume and Edunov, Sergey and Grave, Edouard and Joulin, Armand and Fan, Angela",editor = "Zong, Chengqing and Xia, Fei and Li, Wenjie and Navigli, Roberto",booktitle = "Proceedings of the 59th Annual Meeting of the Association for Computational Linguistics and the 11th International Joint Conference on Natural Language Processing (Volume 1: Long Papers)",month = aug,year = "2021",address = "Online",publisher = acl,url = anth # {2021.acl-long.507/},doi = "10.18653/v1/2021.acl-long.507",pages = "6490--6500"}

@inproceedings{jalili-sabet-etal-2020-simalign,title = "{S}im{A}lign: High Quality Word Alignments Without Parallel Training Data Using Static and Contextualized Embeddings",author = {Jalili Sabet, Masoud and Dufter, Philipp and Yvon, Fran{\c{c}}ois and Sch{\"u}tze, Hinrich},editor = "Cohn, Trevor and He, Yulan and Liu, Yang",booktitle = "Findings of the Association for Computational Linguistics: EMNLP 2020",month = nov,year = "2020",address = "Online",publisher = acl,url = anth # {2020.findings-emnlp.147/},doi = "10.18653/v1/2020.findings-emnlp.147",pages = "1627--1643"}

@inproceedings{ladhak-etal-2020-wikilingua,title = "{W}iki{L}ingua: A New Benchmark Dataset for Cross-Lingual Abstractive Summarization",author = "Ladhak, Faisal and Durmus, Esin and Cardie, Claire and McKeown, Kathleen",editor = "Cohn, Trevor and He, Yulan and Liu, Yang",booktitle = "Findings of the Association for Computational Linguistics: EMNLP 2020",month = nov,year = "2020",address = "Online",publisher = acl,url = anth # {2020.findings-emnlp.360/},doi = "10.18653/v1/2020.findings-emnlp.360",pages = "4034--4048"}

@inproceedings{reimers-gurevych-2020-making,title = "Making Monolingual Sentence Embeddings Multilingual using Knowledge Distillation",author = "Reimers, Nils and Gurevych, Iryna",editor = "Webber, Bonnie and Cohn, Trevor and He, Yulan and Liu, Yang",booktitle = "Proceedings of the 2020 Conference on Empirical Methods in Natural Language Processing (EMNLP)",month = nov,year = "2020",address = "Online",publisher = acl,url = anth # {2020.emnlp-main.365/},doi = "10.18653/v1/2020.emnlp-main.365",pages = "4512--4525"}

@inproceedings{lin-etal-2019-choosing,title = "Choosing Transfer Languages for Cross-Lingual Learning",author = "Lin, Yu-Hsiang and Chen, Chian-Yu and Lee, Jean and Li, Zirui and Zhang, Yuyan and Xia, Mengzhou and Rijhwani, Shruti and He, Junxian and Zhang, Zhisong and Ma, Xuezhe and Anastasopoulos, Antonios and Littell, Patrick and Neubig, Graham",editor = "Korhonen, Anna and Traum, David and M{\`a}rquez, Llu{\'i}s",booktitle = "Proceedings of the 57th Annual Meeting of the Association for Computational Linguistics",month = jul,year = "2019",address = "Florence, Italy",publisher = acl,url = anth # {P19-1301/},doi = "10.18653/v1/P19-1301",pages = "3125--3135"}

@inproceedings{pilehvar-camacho-collados-2019-wic,title = "{W}i{C}: the Word-in-Context Dataset for Evaluating Context-Sensitive Meaning Representations",author = "Pilehvar, Mohammad Taher and Camacho-Collados, Jose",editor = "Burstein, Jill and Doran, Christy and Solorio, Thamar",booktitle = "Proceedings of the 2019 Conference of the North {A}merican Chapter of the Association for Computational Linguistics: Human Language Technologies, Volume 1 (Long and Short Papers)",month = jun,year = "2019",address = "Minneapolis, Minnesota",publisher = acl,url = anth # {N19-1128/},doi = "10.18653/v1/N19-1128",pages = "1267--1273"}

@inproceedings{och-ney-2000-improved,title = "Improved Statistical Alignment Models",author = "Och, Franz Josef and Ney, Hermann",booktitle = "Proceedings of the 38th Annual Meeting of the Association for Computational Linguistics",month = oct,year = "2000",address = "Hong Kong",publisher = acl,url = anth # {P00-1056/},doi = "10.3115/1075218.1075274",pages = "440--447"}

@inproceedings{popovic-2017-chrf,title = "chr{F}++: words helping character n-grams",author = "Popovi{\'c}, Maja",editor = "Bojar, Ond{\v{r}}ej and Buck, Christian and Chatterjee, Rajen and Federmann, Christian and Graham, Yvette and Haddow, Barry and Huck, Matthias and Yepes, Antonio Jimeno and Koehn, Philipp and Kreutzer, Julia",booktitle = "Proceedings of the Second Conference on Machine Translation",month = sep,year = "2017",address = "Copenhagen, Denmark",publisher = acl,url = anth # {W17-4770/},doi = "10.18653/v1/W17-4770",pages = "612--618"}

@inproceedings{cer-etal-2017-semeval,title = "{S}em{E}val-2017 Task 1: Semantic Textual Similarity Multilingual and Crosslingual Focused Evaluation",author = "Cer, Daniel and Diab, Mona and Agirre, Eneko and Lopez-Gazpio, I{\~n}igo and Specia, Lucia",editor = "Bethard, Steven and Carpuat, Marine and Apidianaki, Marianna and Mohammad, Saif M. and Cer, Daniel and Jurgens, David",booktitle = "Proceedings of the 11th International Workshop on Semantic Evaluation ({S}em{E}val-2017)",month = aug,year = "2017",address = "Vancouver, Canada",publisher = acl,url = anth # {S17-2001/},doi = "10.18653/v1/S17-2001",pages = "1--14"}

@inproceedings{pan-etal-2017-cross,title = "Cross-lingual Name Tagging and Linking for 282 Languages",author = "Pan, Xiaoman and Zhang, Boliang and May, Jonathan and Nothman, Joel and Knight, Kevin and Ji, Heng",editor = "Barzilay, Regina and Kan, Min-Yen",booktitle = "Proceedings of the 55th Annual Meeting of the Association for Computational Linguistics (Volume 1: Long Papers)",month = jul,year = "2017",address = "Vancouver, Canada",publisher = acl,url = anth # {P17-1178/},doi = "10.18653/v1/P17-1178",pages = "1946--1958"}

@inproceedings{mostafazadeh-etal-2016-corpus,title = "A Corpus and Cloze Evaluation for Deeper Understanding of Commonsense Stories",author = "Mostafazadeh, Nasrin and Chambers, Nathanael and He, Xiaodong and Parikh, Devi and Batra, Dhruv and Vanderwende, Lucy and Kohli, Pushmeet and Allen, James",editor = "Knight, Kevin and Nenkova, Ani and Rambow, Owen",booktitle = "Proceedings of the 2016 Conference of the North {A}merican Chapter of the Association for Computational Linguistics: Human Language Technologies",month = jun,year = "2016",address = "San Diego, California",publisher = acl,url = anth # {N16-1098/},doi = "10.18653/v1/N16-1098",pages = "839--849"}

@inproceedings{zeroswot,
    title = {{Pushing the Limits of Zero-shot End-to-End Speech Translation}},
    author = "Tsiamas, Ioannis  and
      G{\'a}llego, Gerard I.  and
      Fonollosa, Jos{\'e} A. R.  and
      Costa-juss{\`a}, Marta R.",
    editor = "Ku, Lun-Wei  and
      Martins, Andre  and
      Srikumar, Vivek",
    booktitle = "Findings of the Association for Computational Linguistics: ACL 2024",
    month = aug,
    year = "2024",
    address = "Bangkok, Thailand",
    publisher = "Association for Computational Linguistics",
    url = "https://aclanthology.org/2024.findings-acl.847/",
    doi = "10.18653/v1/2024.findings-acl.847",
    pages = "14245--14267"
}

@INPROCEEDINGS{w2v_bert,
  author={Chung, Yu-An and Zhang, Yu and Han, Wei and Chiu, Chung-Cheng and Qin, James and Pang, Ruoming and Wu, Yonghui},
  booktitle={2021 IEEE Automatic Speech Recognition and Understanding Workshop (ASRU)}, 
  title={{w2v-BERT: Combining Contrastive Learning and Masked Language Modeling for Self-Supervised Speech Pre-Training}}, 
  year={2021},
  volume={},
  number={},
  pages={244-250},
  url = {https://ieeexplore.ieee.org/document/9688253},
  doi={10.1109/ASRU51503.2021.9688253}}

@misc{solatorio2024gistembedguidedinsampleselection,
      title={{GISTEmbed: Guided In-sample Selection of Training Negatives for Text Embedding Fine-tuning}}, 
      author={Aivin V. Solatorio},
      year={2024},
      eprint={2402.16829},
      archivePrefix={arXiv},
      primaryClass={cs.LG},
      url={https://arxiv.org/abs/2402.16829}, 
}

@misc{barrault2024large,
      title={{Large Concept Models: Language Modeling in a Sentence Representation Space}}, 
      author={{LCM Team} and Loïc Barrault and Paul-Ambroise Duquenne and Maha Elbayad and Artyom Kozhevnikov and Belen Alastruey and Pierre Andrews and Mariano Coria and Guillaume Couairon and Marta R. Costa-jussà and David Dale and Hady Elsahar and Kevin Heffernan and João Maria Janeiro and Tuan Tran and Christophe Ropers and Eduardo Sánchez and Robin San Roman and Alexandre Mourachko and Safiyyah Saleem and Holger Schwenk},
      year={2024},
      eprint={2412.08821},
      archivePrefix={arXiv},
      primaryClass={cs.CL},
      url={https://arxiv.org/abs/2412.08821}, 
}

@article{chen2022wavlm,
  title={Wavlm: Large-scale self-supervised pre-training for full stack speech processing},
  author={Chen, Sanyuan and Wang, Chengyi and Chen, Zhengyang and Wu, Yu and Liu, Shujie and Chen, Zhuo and Li, Jinyu and Kanda, Naoyuki and Yoshioka, Takuya and Xiao, Xiong and others},
  journal={IEEE Journal of Selected Topics in Signal Processing},
  volume={16},
  number={6},
  pages={1505--1518},
  year={2022},
  publisher={IEEE}
}

@article{barrault2023seamlessm4t,
  title={Seamlessm4t: Massively multilingual \& multimodal machine translation},
  author={Barrault, Lo{\"\i}c and Chung, Yu-An and Meglioli, Mariano Cora and Dale, David and Dong, Ning and Duquenne, Paul-Ambroise and Elsahar, Hady and Gong, Hongyu and Heffernan, Kevin and Hoffman, John and others},
  journal={arXiv preprint arXiv:2308.11596},
  year={2023}
}

@misc{workshop2022bloom,
      title={{BLOOM: A 176B-Parameter Open-Access Multilingual Language Model}}, 
      author={{BigScience Workshop} and Teven Le Scao and Angela Fan and Christopher Akiki and Ellie Pavlick and Suzana Ilić and Daniel Hesslow and Roman Castagné and Alexandra Sasha Luccioni and François Yvon and Matthias Gallé and Jonathan Tow and Alexander M. Rush and Stella Biderman and Albert Webson and Pawan Sasanka Ammanamanchi and Thomas Wang and Benoît Sagot and Niklas Muennighoff and Albert Villanova del Moral and Olatunji Ruwase and Rachel Bawden and Stas Bekman and Angelina McMillan-Major and Iz Beltagy and Huu Nguyen and Lucile Saulnier and Samson Tan and Pedro Ortiz Suarez and Victor Sanh and Hugo Laurençon and Yacine Jernite and Julien Launay and Margaret Mitchell and Colin Raffel and Aaron Gokaslan and Adi Simhi and Aitor Soroa and Alham Fikri Aji and Amit Alfassy and Anna Rogers and Ariel Kreisberg Nitzav and Canwen Xu and Chenghao Mou and Chris Emezue and Christopher Klamm and Colin Leong and Daniel van Strien and David Ifeoluwa Adelani and Dragomir Radev and Eduardo González Ponferrada and Efrat Levkovizh and Ethan Kim and Eyal Bar Natan and Francesco De Toni and Gérard Dupont and Germán Kruszewski and Giada Pistilli and Hady Elsahar and Hamza Benyamina and Hieu Tran and Ian Yu and Idris Abdulmumin and Isaac Johnson and Itziar Gonzalez-Dios and Javier de la Rosa and Jenny Chim and Jesse Dodge and Jian Zhu and Jonathan Chang and Jörg Frohberg and Joseph Tobing and Joydeep Bhattacharjee and Khalid Almubarak and Kimbo Chen and Kyle Lo and Leandro Von Werra and Leon Weber and Long Phan and Loubna Ben Allal and Ludovic Tanguy and Manan Dey and Manuel Romero Muñoz and Maraim Masoud and María Grandury and Mario Šaško and Max Huang and Maximin Coavoux and Mayank Singh and Mike Tian-Jian Jiang and Minh Chien Vu and Mohammad A. Jauhar and Mustafa Ghaleb and Nishant Subramani and Nora Kassner and Nurulaqilla Khamis and Olivier Nguyen and Omar Espejel and Ona de Gibert and Paulo Villegas and Peter Henderson and Pierre Colombo and Priscilla Amuok and Quentin Lhoest and Rheza Harliman and Rishi Bommasani and Roberto Luis López and Rui Ribeiro and Salomey Osei and Sampo Pyysalo and Sebastian Nagel and Shamik Bose and Shamsuddeen Hassan Muhammad and Shanya Sharma and Shayne Longpre and Somaieh Nikpoor and Stanislav Silberberg and Suhas Pai and Sydney Zink and Tiago Timponi Torrent and Timo Schick and Tristan Thrush and Valentin Danchev and Vassilina Nikoulina and Veronika Laippala and Violette Lepercq and Vrinda Prabhu and Zaid Alyafeai and Zeerak Talat and Arun Raja and Benjamin Heinzerling and Chenglei Si and Davut Emre Taşar and Elizabeth Salesky and Sabrina J. Mielke and Wilson Y. Lee and Abheesht Sharma and Andrea Santilli and Antoine Chaffin and Arnaud Stiegler and Debajyoti Datta and Eliza Szczechla and Gunjan Chhablani and Han Wang and Harshit Pandey and Hendrik Strobelt and Jason Alan Fries and Jos Rozen and Leo Gao and Lintang Sutawika and M Saiful Bari and Maged S. Al-shaibani and Matteo Manica and Nihal Nayak and Ryan Teehan and Samuel Albanie and Sheng Shen and Srulik Ben-David and Stephen H. Bach and Taewoon Kim and Tali Bers and Thibault Fevry and Trishala Neeraj and Urmish Thakker and Vikas Raunak and Xiangru Tang and Zheng-Xin Yong and Zhiqing Sun and Shaked Brody and Yallow Uri and Hadar Tojarieh and Adam Roberts and Hyung Won Chung and Jaesung Tae and Jason Phang and Ofir Press and Conglong Li and Deepak Narayanan and Hatim Bourfoune and Jared Casper and Jeff Rasley and Max Ryabinin and Mayank Mishra and Minjia Zhang and Mohammad Shoeybi and Myriam Peyrounette and Nicolas Patry and Nouamane Tazi and Omar Sanseviero and Patrick von Platen and Pierre Cornette and Pierre François Lavallée and Rémi Lacroix and Samyam Rajbhandari and Sanchit Gandhi and Shaden Smith and Stéphane Requena and Suraj Patil and Tim Dettmers and Ahmed Baruwa and Amanpreet Singh and Anastasia Cheveleva and Anne-Laure Ligozat and Arjun Subramonian and Aurélie Névéol and Charles Lovering and Dan Garrette and Deepak Tunuguntla and Ehud Reiter and Ekaterina Taktasheva and Ekaterina Voloshina and Eli Bogdanov and Genta Indra Winata and Hailey Schoelkopf and Jan-Christoph Kalo and Jekaterina Novikova and Jessica Zosa Forde and Jordan Clive and Jungo Kasai and Ken Kawamura and Liam Hazan and Marine Carpuat and Miruna Clinciu and Najoung Kim and Newton Cheng and Oleg Serikov and Omer Antverg and Oskar van der Wal and Rui Zhang and Ruochen Zhang and Sebastian Gehrmann and Shachar Mirkin and Shani Pais and Tatiana Shavrina and Thomas Scialom and Tian Yun and Tomasz Limisiewicz and Verena Rieser and Vitaly Protasov and Vladislav Mikhailov and Yada Pruksachatkun and Yonatan Belinkov and Zachary Bamberger and Zdeněk Kasner and Alice Rueda and Amanda Pestana and Amir Feizpour and Ammar Khan and Amy Faranak and Ana Santos and Anthony Hevia and Antigona Unldreaj and Arash Aghagol and Arezoo Abdollahi and Aycha Tammour and Azadeh HajiHosseini and Bahareh Behroozi and Benjamin Ajibade and Bharat Saxena and Carlos Muñoz Ferrandis and Daniel McDuff and Danish Contractor and David Lansky and Davis David and Douwe Kiela and Duong A. Nguyen and Edward Tan and Emi Baylor and Ezinwanne Ozoani and Fatima Mirza and Frankline Ononiwu and Habib Rezanejad and Hessie Jones and Indrani Bhattacharya and Irene Solaiman and Irina Sedenko and Isar Nejadgholi and Jesse Passmore and Josh Seltzer and Julio Bonis Sanz and Livia Dutra and Mairon Samagaio and Maraim Elbadri and Margot Mieskes and Marissa Gerchick and Martha Akinlolu and Michael McKenna and Mike Qiu and Muhammed Ghauri and Mykola Burynok and Nafis Abrar and Nazneen Rajani and Nour Elkott and Nour Fahmy and Olanrewaju Samuel and Ran An and Rasmus Kromann and Ryan Hao and Samira Alizadeh and Sarmad Shubber and Silas Wang and Sourav Roy and Sylvain Viguier and Thanh Le and Tobi Oyebade and Trieu Le and Yoyo Yang and Zach Nguyen and Abhinav Ramesh Kashyap and Alfredo Palasciano and Alison Callahan and Anima Shukla and Antonio Miranda-Escalada and Ayush Singh and Benjamin Beilharz and Bo Wang and Caio Brito and Chenxi Zhou and Chirag Jain and Chuxin Xu and Clémentine Fourrier and Daniel León Periñán and Daniel Molano and Dian Yu and Enrique Manjavacas and Fabio Barth and Florian Fuhrimann and Gabriel Altay and Giyaseddin Bayrak and Gully Burns and Helena U. Vrabec and Imane Bello and Ishani Dash and Jihyun Kang and John Giorgi and Jonas Golde and Jose David Posada and Karthik Rangasai Sivaraman and Lokesh Bulchandani and Lu Liu and Luisa Shinzato and Madeleine Hahn de Bykhovetz and Maiko Takeuchi and Marc Pàmies and Maria A Castillo and Marianna Nezhurina and Mario Sänger and Matthias Samwald and Michael Cullan and Michael Weinberg and Michiel De Wolf and Mina Mihaljcic and Minna Liu and Moritz Freidank and Myungsun Kang and Natasha Seelam and Nathan Dahlberg and Nicholas Michio Broad and Nikolaus Muellner and Pascale Fung and Patrick Haller and Ramya Chandrasekhar and Renata Eisenberg and Robert Martin and Rodrigo Canalli and Rosaline Su and Ruisi Su and Samuel Cahyawijaya and Samuele Garda and Shlok S Deshmukh and Shubhanshu Mishra and Sid Kiblawi and Simon Ott and Sinee Sang-aroonsiri and Srishti Kumar and Stefan Schweter and Sushil Bharati and Tanmay Laud and Théo Gigant and Tomoya Kainuma and Wojciech Kusa and Yanis Labrak and Yash Shailesh Bajaj and Yash Venkatraman and Yifan Xu and Yingxin Xu and Yu Xu and Zhe Tan and Zhongli Xie and Zifan Ye and Mathilde Bras and Younes Belkada and Thomas Wolf},
      year={2023},
      eprint={2211.05100},
      archivePrefix={arXiv},
      primaryClass={cs.CL},
      url={https://arxiv.org/abs/2211.05100}, 
}

@inproceedings{
behnamghader2024llm2vec,
title={{LLM2Vec: Large Language Models Are Secretly Powerful Text Encoders}},
author={Parishad BehnamGhader and Vaibhav Adlakha and Marius Mosbach and Dzmitry Bahdanau and Nicolas Chapados and Siva Reddy},
booktitle={First Conference on Language Modeling},
year={2024},
url={https://openreview.net/forum?id=IW1PR7vEBf}
}

@misc{zhang2025encoder,
      title={{Encoder-Decoder Gemma: Improving the Quality-Efficiency Trade-Off via Adaptation}}, 
      author={Biao Zhang and Fedor Moiseev and Joshua Ainslie and Paul Suganthan and Min Ma and Surya Bhupatiraju and Fede Lebron and Orhan Firat and Armand Joulin and Zhe Dong},
      year={2025},
      eprint={2504.06225},
      archivePrefix={arXiv},
      primaryClass={cs.CL},
      url={https://arxiv.org/abs/2504.06225}, 
}

@misc{gong2025,
      title={{Structure-Aware Fill-in-the-Middle Pretraining for Code}}, 
      author={Linyuan Gong and Alvin Cheung and Mostafa Elhoushi and Sida Wang},
      year={2025},
      eprint={2506.00204},
      archivePrefix={arXiv},
      primaryClass={cs.CL},
      url={https://arxiv.org/abs/2506.00204}, 
}

@misc{wang2024multilingual,
      title={{Multilingual E5 Text Embeddings: A Technical Report}}, 
      author={Liang Wang and Nan Yang and Xiaolong Huang and Linjun Yang and Rangan Majumder and Furu Wei},
      year={2024},
      eprint={2402.05672},
      archivePrefix={arXiv},
      primaryClass={cs.CL},
      url={https://arxiv.org/abs/2402.05672}, 
}

@article{nussbaum2025trainingsparsemixtureexperts,
  publtype={informal},
  author={Zach Nussbaum and Brandon Duderstadt},
  title={{Training Sparse Mixture Of Experts Text Embedding Models}},
  year={2025},
  month={February},
  cdate={1738368000000},
  journal={CoRR},
  volume={abs/2502.07972},
  url={https://doi.org/10.48550/arXiv.2502.07972}
}

@inproceedings{
    zhang2024code,
    title={{Code Representation Learning At Scale}},
    author={Dejiao Zhang and Wasi Uddin Ahmad and Ming Tan and Hantian Ding and Ramesh Nallapati and Dan Roth and Xiaofei Ma and Bing Xiang},
    booktitle={{The Twelfth International Conference on Learning Representations}},
    year={2024},
    url={https://openreview.net/forum?id=vfzRRjumpX}
}

@inproceedings{sureshcornstack,
  title={{CoRNStack: High-Quality Contrastive Data for Better Code Retrieval and Reranking}},
  author={Suresh, Tarun and Reddy, Revanth Gangi and Xu, Yifei and Nussbaum, Zach and Mulyar, Andriy and Duderstadt, Brandon and Ji, Heng},
  booktitle={{The Thirteenth International Conference on Learning Representations}},
  url = {https://openreview.net/forum?id=iyJOUELYir},
  year={2025}
}

@inproceedings{
liu2025codexembed,
title={{CodeXEmbed: A Generalist Embedding Model Family for Multilingual and Multi-task Code Retrieval}},
author={Ye Liu and Rui Meng and Shafiq Joty and silvio savarese and Caiming Xiong and Yingbo Zhou and Semih Yavuz},
booktitle={Second Conference on Language Modeling},
year={2025},
url={https://openreview.net/forum?id=z3lG70Azbg}
}

@misc{Husain2019CodeSearchNetCE,
      title={{CodeSearchNet Challenge: Evaluating the State of Semantic Code Search}}, 
      author={Hamel Husain and Ho-Hsiang Wu and Tiferet Gazit and Miltiadis Allamanis and Marc Brockschmidt},
      year={2020},
      eprint={1909.09436},
      archivePrefix={arXiv},
      primaryClass={cs.LG},
      url={https://arxiv.org/abs/1909.09436}, 
}

@inproceedings{fitzgerald2022massive,
    title = {{MASSIVE: A 1M-Example Multilingual Natural Language Understanding Dataset with 51 Typologically-Diverse Languages}},
    author = "FitzGerald, Jack  and
      Hench, Christopher  and
      Peris, Charith  and
      Mackie, Scott  and
      Rottmann, Kay  and
      Sanchez, Ana  and
      Nash, Aaron  and
      Urbach, Liam  and
      Kakarala, Vishesh  and
      Singh, Richa  and
      Ranganath, Swetha  and
      Crist, Laurie  and
      Britan, Misha  and
      Leeuwis, Wouter  and
      Tur, Gokhan  and
      Natarajan, Prem",
    editor = "Rogers, Anna  and
      Boyd-Graber, Jordan  and
      Okazaki, Naoaki",
    booktitle = "Proceedings of the 61st Annual Meeting of the Association for Computational Linguistics (Volume 1: Long Papers)",
    month = jul,
    year = "2023",
    address = "Toronto, Canada",
    publisher = "Association for Computational Linguistics",
    url = "https://aclanthology.org/2023.acl-long.235/",
    doi = "10.18653/v1/2023.acl-long.235",
    pages = "4277--4302"
}

@inproceedings{conneau2018xnli,
  author = {Conneau, Alexis
and Rinott, Ruty
and Lample, Guillaume
and Williams, Adina
and Bowman, Samuel R.
and Schwenk, Holger
and Stoyanov, Veselin},
  booktitle = {{Proceedings of the 2018 Conference on Empirical Methods
in Natural Language Processing}},
  location = {Brussels, Belgium},
  publisher = {Association for Computational Linguistics},
  title = {{XNLI: Evaluating Cross-lingual Sentence Representations}},
  year = {2018},
}

@misc{upadhyay2023xnli,
      title={{XNLI 2.0: Improving XNLI dataset and performance on Cross Lingual Understanding (XLU)}}, 
      author={Ankit Kumar Upadhyay and Harsit Kumar Upadhya},
      year={2023},
      eprint={2301.06527},
      archivePrefix={arXiv},
      primaryClass={cs.CL},
      url={https://arxiv.org/abs/2301.06527}, 
}

@inproceedings{
allal2025smollm2smolgoesbig,
title={{SmolLM2: When Smol Goes Big Data-Centric Training of a Fully Open Small Language Model}},
author={Loubna Ben Allal and Anton Lozhkov and Elie Bakouch and Gabriel Martin Blazquez and Guilherme Penedo and Lewis Tunstall and Andr{\'e}s Marafioti and Agust{\'\i}n Piqueres Lajar{\'\i}n and Hynek Kydl{\'\i}{\v{c}}ek and Vaibhav Srivastav and Joshua Lochner and Caleb Fahlgren and Xuan Son NGUYEN and Ben Burtenshaw and Cl{\'e}mentine Fourrier and Haojun Zhao and Hugo Larcher and Mathieu Morlon and Cyril Zakka and Colin Raffel and Leandro Von Werra and Thomas Wolf},
booktitle={Second Conference on Language Modeling},
year={2025},
url={https://openreview.net/forum?id=3JiCl2A14H}
}

@inproceedings{singh2024globalmmluunderstandingaddressing,
    title = {{Global MMLU: Understanding and Addressing Cultural and Linguistic Biases in Multilingual Evaluation}},
    author = "Singh, Shivalika  and
      Romanou, Angelika  and
      Fourrier, Cl{\'e}mentine  and
      Adelani, David Ifeoluwa  and
      Ngui, Jian Gang  and
      Vila-Suero, Daniel  and
      Limkonchotiwat, Peerat  and
      Marchisio, Kelly  and
      Leong, Wei Qi  and
      Susanto, Yosephine  and
      Ng, Raymond  and
      Longpre, Shayne  and
      Ruder, Sebastian  and
      Ko, Wei-Yin  and
      Bosselut, Antoine  and
      Oh, Alice  and
      Martins, Andre  and
      Choshen, Leshem  and
      Ippolito, Daphne  and
      Ferrante, Enzo  and
      Fadaee, Marzieh  and
      Ermis, Beyza  and
      Hooker, Sara",
    editor = "Che, Wanxiang  and
      Nabende, Joyce  and
      Shutova, Ekaterina  and
      Pilehvar, Mohammad Taher",
    booktitle = "Proceedings of the 63rd Annual Meeting of the Association for Computational Linguistics (Volume 1: Long Papers)",
    month = jul,
    year = "2025",
    address = "Vienna, Austria",
    publisher = "Association for Computational Linguistics",
    url = "https://aclanthology.org/2025.acl-long.919/",
    doi = "10.18653/v1/2025.acl-long.919",
    pages = "18761--18799",
    ISBN = "979-8-89176-251-0"
}

@inproceedings{yang2019improving,
  title     = {{Improving Multilingual Sentence Embedding using Bi-directional Dual Encoder with Additive Margin Softmax}},
  author    = {Yang, Yinfei and Hernandez Abrego, Gustavo and Yuan, Steve and Guo, Mandy and Shen, Qinlan and Cer, Daniel and Sung, Yun-hsuan and Strope, Brian and Kurzweil, Ray},
  booktitle = {Proceedings of the Twenty-Eighth International Joint Conference on
               Artificial Intelligence, {IJCAI-19}},
  publisher = {International Joint Conferences on Artificial Intelligence Organization},
  pages     = {5370--5378},
  year      = {2019},
  month     = {7},
  doi       = {10.24963/ijcai.2019/746},
  url       = {https://doi.org/10.24963/ijcai.2019/746},
}

@inproceedings{sturua2024jinaembeddingsv3multilingualembeddingstask,
author = {Sturua, Saba and Mohr, Isabelle and Kalim Akram, Mohammad and G\"{u}nther, Michael and Wang, Bo and Krimmel, Markus and Wang, Feng and Mastrapas, Georgios and Koukounas, Andreas and Wang, Nan and Xiao, Han},
title = {{Jina Embeddings V3: Multilingual Text Encoder with Low-Rank Adaptations}},
year = {2025},
isbn = {978-3-031-88719-2},
publisher = {Springer-Verlag},
address = {Berlin, Heidelberg},
url = {https://doi.org/10.1007/978-3-031-88720-8_21},
doi = {10.1007/978-3-031-88720-8_21},
booktitle = {Advances in Information Retrieval: 47th European Conference on Information Retrieval, ECIR 2025, Lucca, Italy, April 6–10, 2025, Proceedings, Part V},
pages = {123–129},
numpages = {7},
location = {Lucca, Italy}
}

@article{qwen3embedding,
  title={{Qwen3 Embedding: Advancing Text Embedding and Reranking Through Foundation Models}},
  author={Zhang, Yanzhao and Li, Mingxin and Long, Dingkun and Zhang, Xin and Lin, Huan and Yang, Baosong and Xie, Pengjun and Yang, An and Liu, Dayiheng and Lin, Junyang and Huang, Fei and Zhou, Jingren},
  journal={arXiv preprint arXiv:2506.05176},
  year={2025}
}

@article{mcinnes2020umapuniformmanifoldapproximation, doi = {10.21105/joss.00861}, url = {https://doi.org/10.21105/joss.00861}, year = {2018}, publisher = {The Open Journal}, volume = {3}, number = {29}, pages = {861}, author = {McInnes, Leland and Healy, John and Saul, Nathaniel and Großberger, Lukas}, title = {{UMAP: Uniform Manifold Approximation and Projection}}, journal = {Journal of Open Source Software} }

@misc{alastruey2025interferencematrixquantifyingcrosslingual,
      title={{Interference Matrix: Quantifying Cross-Lingual Interference in Transformer Encoders}}, 
      author={Belen Alastruey and João Maria Janeiro and Alexandre Allauzen and Maha Elbayad and Loïc Barrault and Marta R. Costa-jussà},
      year={2025},
      eprint={2508.02256},
      archivePrefix={arXiv},
      primaryClass={cs.CL},
      url={https://arxiv.org/abs/2508.02256}, 
}

@inproceedings{attention_is_all_you_need,
    author = {Vaswani, Ashish and Shazeer, Noam and Parmar, Niki and Uszkoreit, Jakob and Jones, Llion and Gomez, Aidan N and Kaiser, \L ukasz and Polosukhin, Illia},
    booktitle = {{Advances in Neural Information Processing Systems}},
    editor = {I. Guyon and U. Von Luxburg and S. Bengio and H. Wallach and R. Fergus and S. Vishwanathan and R. Garnett},
    pages = {},
    publisher = {Curran Associates, Inc.},
    title = {{Attention is All you Need}},
    url = {https://proceedings.neurips.cc/paper_files/paper/2017/file/3f5ee243547dee91fbd053c1c4a845aa-Paper.pdf},
    volume = {30},
    year = {2017}
}

@article{mms,
author = {Pratap, Vineel and Tjandra, Andros and Shi, Bowen and Tomasello, Paden and Babu, Arun and Kundu, Sayani and Elkahky, Ali and Ni, Zhaoheng and Vyas, Apoorv and Fazel-Zarandi, Maryam and Baevski, Alexei and Adi, Yossi and Zhang, Xiaohui and Hsu, Wei-Ning and Conneau, Alexis and Auli, Michael},
title = {{Scaling Speech Technology to 1,000+ Languages}},
year = {2024},
issue_date = {January 2024},
publisher = {JMLR.org},
volume = {25},
number = {1},
issn = {1532-4435},
journal = {J. Mach. Learn. Res.},
month = jan,
articleno = {97},
numpages = {52}
}

@inproceedings{tatoeba,
    title = {{The Tatoeba Translation Challenge -- Realistic Data Sets for Low Resource and Multilingual MT}},
    author = {Tiedemann, J{\"o}rg},
    editor = {Barrault, Lo{\"i}c  and
      Bojar, Ond{\v{r}}ej  and
      Bougares, Fethi  and
      Chatterjee, Rajen  and
      Costa-juss{\`a}, Marta R.  and
      Federmann, Christian  and
      Fishel, Mark  and
      Fraser, Alexander  and
      Graham, Yvette  and
      Guzman, Paco  and
      Haddow, Barry  and
      Huck, Matthias  and
      Yepes, Antonio Jimeno  and
      Koehn, Philipp  and
      Martins, Andr{\'e}  and
      Morishita, Makoto  and
      Monz, Christof  and
      Nagata, Masaaki  and
      Nakazawa, Toshiaki  and
      Negri, Matteo},
    booktitle = {{Proceedings of the Fifth Conference on Machine Translation}},
    month = nov,
    year = "2020",
    address = "Online",
    publisher = "Association for Computational Linguistics",
    url = "https://aclanthology.org/2020.wmt-1.139/",
    pages = "1174--1182"
}

@misc{xlcost,
     title = {{XLCoST: A Benchmark Dataset for Cross-lingual Code Intelligence}},
     url = {https://arxiv.org/abs/2206.08474},
     author = {Zhu, Ming and Jain, Aneesh and Suresh, Karthik and Ravindran, Roshan and Tipirneni, Sindhu and Reddy, Chandan K.},
     year = {2022},
     eprint={2206.08474},
     archivePrefix={arXiv}
}

@inproceedings{glotlid,
    title = {{GlotLID: Language Identification for Low-Resource Languages}},
    author = "Kargaran, Amir Hossein  and
      Imani, Ayyoob  and
      Yvon, Fran{\c{c}}ois  and
      Schuetze, Hinrich",
    editor = "Bouamor, Houda  and
      Pino, Juan  and
      Bali, Kalika",
    booktitle = {{Findings of the Association for Computational Linguistics: EMNLP 2023}},
    month = dec,
    year = "2023",
    address = "Singapore",
    publisher = "Association for Computational Linguistics",
    url = "https://aclanthology.org/2023.findings-emnlp.410/",
    doi = "10.18653/v1/2023.findings-emnlp.410",
    pages = "6155--6218"
}

@inproceedings{mexma,
    title = {{MEXMA Token-level Objectives improve Sentence Representations}},
    author = "Janeiro, Jo{\~a}o Maria  and
      Piwowarski, Benjamin  and
      Gallinari, Patrick  and
      Barrault, Loic",
    editor = "Che, Wanxiang  and
      Nabende, Joyce  and
      Shutova, Ekaterina  and
      Pilehvar, Mohammad Taher",
    booktitle = {{Proceedings of the 63rd Annual Meeting of the Association for Computational Linguistics (Volume 1: Long Papers)}},
    month = jul,
    year = "2025",
    address = "Vienna, Austria",
    publisher = "Association for Computational Linguistics",
    url = "https://aclanthology.org/2025.acl-long.1168/",
    doi = "10.18653/v1/2025.acl-long.1168",
    pages = "23960--23995",
    ISBN = "979-8-89176-251-0"
}

@inproceedings{labse,
    title = {{Language-agnostic BERT Sentence Embedding}},
    author = "Feng, Fangxiaoyu  and
      Yang, Yinfei  and
      Cer, Daniel  and
      Arivazhagan, Naveen  and
      Wang, Wei",
    editor = "Muresan, Smaranda  and
      Nakov, Preslav  and
      Villavicencio, Aline",
    booktitle = {{Proceedings of the 60th Annual Meeting of the Association for Computational Linguistics (Volume 1: Long Papers)}},
    month = may,
    year = "2022",
    address = "Dublin, Ireland",
    publisher = "Association for Computational Linguistics",
    url = "https://aclanthology.org/2022.acl-long.62/",
    doi = "10.18653/v1/2022.acl-long.62",
    pages = "878--891"
}

@article{laser,
    title = {{Massively Multilingual Sentence Embeddings for Zero-Shot Cross-Lingual Transfer and Beyond}},
    author = "Artetxe, Mikel  and
      Schwenk, Holger",
    editor = "Lee, Lillian  and
      Johnson, Mark  and
      Roark, Brian  and
      Nenkova, Ani",
    journal = "Transactions of the Association for Computational Linguistics",
    volume = "7",
    year = "2019",
    address = "Cambridge, MA",
    publisher = "MIT Press",
    url = "https://aclanthology.org/Q19-1038/",
    pages = "597--610"
}

@inproceedings{adelanietal2024sib,
    title = {{SIB-200: A Simple, Inclusive, and Big Evaluation Dataset for Topic Classification in 200+ Languages and Dialects}},
    author = "Adelani, David Ifeoluwa  and
      Liu, Hannah  and
      Shen, Xiaoyu  and
      Vassilyev, Nikita  and
      Alabi, Jesujoba O.  and
      Mao, Yanke  and
      Gao, Haonan  and
      Lee, En-Shiun Annie",
    editor = "Graham, Yvette  and
      Purver, Matthew",
    booktitle = "Proceedings of the 18th Conference of the European Chapter of the Association for Computational Linguistics (Volume 1: Long Papers)",
    month = mar,
    year = "2024",
    address = "St. Julian{'}s, Malta",
    publisher = "Association for Computational Linguistics",
    url = "https://aclanthology.org/2024.eacl-long.14/",
    doi = "10.18653/v1/2024.eacl-long.14",
    pages = "226--245"
}

@inproceedings{xlmr,
    title = {{Unsupervised Cross-lingual Representation Learning at Scale}},
    author = "Conneau, Alexis  and
      Khandelwal, Kartikay  and
      Goyal, Naman  and
      Chaudhary, Vishrav  and
      Wenzek, Guillaume  and
      Guzm{\'a}n, Francisco  and
      Grave, Edouard  and
      Ott, Myle  and
      Zettlemoyer, Luke  and
      Stoyanov, Veselin",
    editor = "Jurafsky, Dan  and
      Chai, Joyce  and
      Schluter, Natalie  and
      Tetreault, Joel",
    booktitle = {{Proceedings of the 58th Annual Meeting of the Association for Computational Linguistics}},
    month = jul,
    year = "2020",
    address = "Online",
    publisher = "Association for Computational Linguistics",
    url = "https://aclanthology.org/2020.acl-main.747/",
    doi = "10.18653/v1/2020.acl-main.747",
    pages = "8440--8451"
}

@inproceedings{blaser2,
    title = {{BLASER 2.0: A Metric for Evaluation and Quality Estimation of Massively Multilingual Speech and Text Translation}},
    author = "Dale, David  and
      Costa-juss{\`a}, Marta R.",
    editor = "Al-Onaizan, Yaser  and
      Bansal, Mohit  and
      Chen, Yun-Nung",
    booktitle = {{Findings of the Association for Computational Linguistics: EMNLP 2024}},
    month = nov,
    year = "2024",
    address = "Miami, Florida, USA",
    publisher = "Association for Computational Linguistics",
    url = "https://aclanthology.org/2024.findings-emnlp.943/",
    doi = "10.18653/v1/2024.findings-emnlp.943",
    pages = "16075--16085"
}

@misc{neubig11kftt,
  author = {Graham Neubig},
  title = {The Kyoto Free Translation Task},
  howpublished = {http://www.phontron.com/kftt},
  year = {2011}
}

@misc{omni_asr,
      title={{Omnilingual ASR: Open-Source Multilingual Speech Recognition for 1600+ Languages}}, 
      author={{Omnilingual ASR Team}  and Gil Keren and Artyom Kozhevnikov and Yen Meng and Christophe Ropers and Matthew Setzler and Skyler Wang and Ifeoluwanimi Adebara and Michael Auli and Can Balioglu and Kevin Chan and Chierh Cheng and Joe Chuang and Caley Droof and Mark Duppenthaler and Paul-Ambroise Duquenne and Alexander Erben and Cynthia Gao and Gabriel Mejia Gonzalez and Kehan Lyu and Sagar Miglani and Vineel Pratap and Kaushik Ram Sadagopan and Safiyyah Saleem and Arina Turkatenko and Albert Ventayol-Boada and Zheng-Xin Yong and Yu-An Chung and Jean Maillard and Rashel Moritz and Alexandre Mourachko and Mary Williamson and Shireen Yates},
      year={2025},
      eprint={2511.09690},
      archivePrefix={arXiv},
      primaryClass={cs.CL},
      url={https://arxiv.org/abs/2511.09690}, 
}

@inproceedings{charsonar,
    title = {{Improving Language and Modality Transfer in Translation by Character-level Modeling}},
    author = "Tsiamas, Ioannis  and
      Dale, David  and
      Costa-juss{\`a}, Marta R.",
    editor = "Che, Wanxiang  and
      Nabende, Joyce  and
      Shutova, Ekaterina  and
      Pilehvar, Mohammad Taher",
    booktitle = {{Proceedings of the 63rd Annual Meeting of the Association for Computational Linguistics (Volume 1: Long Papers)}},
    month = jul,
    year = "2025",
    address = "Vienna, Austria",
    publisher = "Association for Computational Linguistics",
    url = "https://aclanthology.org/2025.acl-long.988/",
    doi = "10.18653/v1/2025.acl-long.988",
    pages = "20171--20187",
    ISBN = "979-8-89176-251-0"
}

@article{nllb,
    title={{Scaling Neural Machine Translation to 200 Languages}}, 
    author={{NLLB Team}},
    year={2024},
    journal={Nature},
    volume={630}, 
    pages={841–846},
    doi={10.1038/s41586-024-07335-x},
    url={https://www.nature.com/articles/s41586-024-07335-x},
}

@article{seamless,
    title={{Joint Speech and Text Machine Translation for up to 100 Languages}}, 
    author={{SEAMLESS Communication Team}},
    year={2025},
    journal={Nature},
    volume={637}, 
    pages={587–593},
    doi={10.1038/s41586-024-08359-z},
    url={https://www.nature.com/articles/s41586-024-08359-z}, 
}

@inproceedings{wmt_oldi_24,
    title = {{Findings of the WMT 2024 Shared Task of the Open Language Data Initiative}},
    author = "Burchell, Laurie  and
      Maillard, Jean  and
      Anastasopoulos, Antonios  and
      Federmann, Christian  and
      Koehn, Philipp  and
      Wang, Skyler",
    editor = "Haddow, Barry  and
      Kocmi, Tom  and
      Koehn, Philipp  and
      Monz, Christof",
    booktitle = {{Proceedings of the Ninth Conference on Machine Translation}},
    month = nov,
    year = "2024",
    address = "Miami, Florida, USA",
    publisher = "Association for Computational Linguistics",
    url = "https://aclanthology.org/2024.wmt-1.4/",
    doi = "10.18653/v1/2024.wmt-1.4",
    pages = "110--117"
}

@inproceedings{madlad,
 author = {Kudugunta, Sneha and Caswell, Isaac and Zhang, Biao and Garcia, Xavier and Xin, Derrick and Kusupati, Aditya and Stella, Romi and Bapna, Ankur and Firat, Orhan},
 booktitle = {{Advances in Neural Information Processing Systems}},
 editor = {A. Oh and T. Naumann and A. Globerson and K. Saenko and M. Hardt and S. Levine},
 pages = {67284--67296},
 publisher = {Curran Associates, Inc.},
 title = {{MADLAD-400: A Multilingual And Document-Level Large Audited Dataset}},
 url = {https://proceedings.neurips.cc/paper_files/paper/2023/file/d49042a5d49818711c401d34172f9900-Paper-Datasets_and_Benchmarks.pdf},
 volume = {36},
 year = {2023}
}

@misc{gemma3,
      title={{Gemma 3 Technical Report}}, 
      author={{Gemma Team} and Aishwarya Kamath and Johan Ferret and Shreya Pathak and Nino Vieillard and Ramona Merhej and Sarah Perrin and Tatiana Matejovicova and Alexandre Ramé and Morgane Rivière and Louis Rouillard and Thomas Mesnard and Geoffrey Cideron and Jean-bastien Grill and Sabela Ramos and Edouard Yvinec and Michelle Casbon and Etienne Pot and Ivo Penchev and Gaël Liu and Francesco Visin and Kathleen Kenealy and Lucas Beyer and Xiaohai Zhai and Anton Tsitsulin and Robert Busa-Fekete and Alex Feng and Noveen Sachdeva and Benjamin Coleman and Yi Gao and Basil Mustafa and Iain Barr and Emilio Parisotto and David Tian and Matan Eyal and Colin Cherry and Jan-Thorsten Peter and Danila Sinopalnikov and Surya Bhupatiraju and Rishabh Agarwal and Mehran Kazemi and Dan Malkin and Ravin Kumar and David Vilar and Idan Brusilovsky and Jiaming Luo and Andreas Steiner and Abe Friesen and Abhanshu Sharma and Abheesht Sharma and Adi Mayrav Gilady and Adrian Goedeckemeyer and Alaa Saade and Alex Feng and Alexander Kolesnikov and Alexei Bendebury and Alvin Abdagic and Amit Vadi and András György and André Susano Pinto and Anil Das and Ankur Bapna and Antoine Miech and Antoine Yang and Antonia Paterson and Ashish Shenoy and Ayan Chakrabarti and Bilal Piot and Bo Wu and Bobak Shahriari and Bryce Petrini and Charlie Chen and Charline Le Lan and Christopher A. Choquette-Choo and CJ Carey and Cormac Brick and Daniel Deutsch and Danielle Eisenbud and Dee Cattle and Derek Cheng and Dimitris Paparas and Divyashree Shivakumar Sreepathihalli and Doug Reid and Dustin Tran and Dustin Zelle and Eric Noland and Erwin Huizenga and Eugene Kharitonov and Frederick Liu and Gagik Amirkhanyan and Glenn Cameron and Hadi Hashemi and Hanna Klimczak-Plucińska and Harman Singh and Harsh Mehta and Harshal Tushar Lehri and Hussein Hazimeh and Ian Ballantyne and Idan Szpektor and Ivan Nardini and Jean Pouget-Abadie and Jetha Chan and Joe Stanton and John Wieting and Jonathan Lai and Jordi Orbay and Joseph Fernandez and Josh Newlan and Ju-yeong Ji and Jyotinder Singh and Kat Black and Kathy Yu and Kevin Hui and Kiran Vodrahalli and Klaus Greff and Linhai Qiu and Marcella Valentine and Marina Coelho and Marvin Ritter and Matt Hoffman and Matthew Watson and Mayank Chaturvedi and Michael Moynihan and Min Ma and Nabila Babar and Natasha Noy and Nathan Byrd and Nick Roy and Nikola Momchev and Nilay Chauhan and Noveen Sachdeva and Oskar Bunyan and Pankil Botarda and Paul Caron and Paul Kishan Rubenstein and Phil Culliton and Philipp Schmid and Pier Giuseppe Sessa and Pingmei Xu and Piotr Stanczyk and Pouya Tafti and Rakesh Shivanna and Renjie Wu and Renke Pan and Reza Rokni and Rob Willoughby and Rohith Vallu and Ryan Mullins and Sammy Jerome and Sara Smoot and Sertan Girgin and Shariq Iqbal and Shashir Reddy and Shruti Sheth and Siim Põder and Sijal Bhatnagar and Sindhu Raghuram Panyam and Sivan Eiger and Susan Zhang and Tianqi Liu and Trevor Yacovone and Tyler Liechty and Uday Kalra and Utku Evci and Vedant Misra and Vincent Roseberry and Vlad Feinberg and Vlad Kolesnikov and Woohyun Han and Woosuk Kwon and Xi Chen and Yinlam Chow and Yuvein Zhu and Zichuan Wei and Zoltan Egyed and Victor Cotruta and Minh Giang and Phoebe Kirk and Anand Rao and Kat Black and Nabila Babar and Jessica Lo and Erica Moreira and Luiz Gustavo Martins and Omar Sanseviero and Lucas Gonzalez and Zach Gleicher and Tris Warkentin and Vahab Mirrokni and Evan Senter and Eli Collins and Joelle Barral and Zoubin Ghahramani and Raia Hadsell and Yossi Matias and D. Sculley and Slav Petrov and Noah Fiedel and Noam Shazeer and Oriol Vinyals and Jeff Dean and Demis Hassabis and Koray Kavukcuoglu and Clement Farabet and Elena Buchatskaya and Jean-Baptiste Alayrac and Rohan Anil and Dmitry and Lepikhin and Sebastian Borgeaud and Olivier Bachem and Armand Joulin and Alek Andreev and Cassidy Hardin and Robert Dadashi and Léonard Hussenot},
      year={2025},
      eprint={2503.19786},
      archivePrefix={arXiv},
      primaryClass={cs.CL},
      url={https://arxiv.org/abs/2503.19786}, 
}

@misc{aya,
      title={{Aya Model: An Instruction Finetuned Open-Access Multilingual Language Model}}, 
      author={Ahmet Üstün and Viraat Aryabumi and Zheng-Xin Yong and Wei-Yin Ko and Daniel D'souza and Gbemileke Onilude and Neel Bhandari and Shivalika Singh and Hui-Lee Ooi and Amr Kayid and Freddie Vargus and Phil Blunsom and Shayne Longpre and Niklas Muennighoff and Marzieh Fadaee and Julia Kreutzer and Sara Hooker},
      year={2024},
      eprint={2402.07827},
      archivePrefix={arXiv},
      primaryClass={cs.CL},
      url={https://arxiv.org/abs/2402.07827}, 
}

@misc{towerplus,
      title={{Tower+: Bridging Generality and Translation Specialization in Multilingual LLMs}}, 
      author={Ricardo Rei and Nuno M. Guerreiro and José Pombal and João Alves and Pedro Teixeirinha and Amin Farajian and André F. T. Martins},
      year={2025},
      eprint={2506.17080},
      archivePrefix={arXiv},
      primaryClass={cs.CL},
      url={https://arxiv.org/abs/2506.17080}, 
}

@article{xcomet,
    title = {{xCOMET: Transparent Machine Translation Evaluation through Fine-grained Error Detection}},
    author = "Guerreiro, Nuno M.  and
      Rei, Ricardo  and
      Stigt, Daan van  and
      Coheur, Luisa  and
      Colombo, Pierre  and
      Martins, Andr{\'e} F. T.",
    journal = "Transactions of the Association for Computational Linguistics",
    volume = "12",
    year = "2024",
    address = "Cambridge, MA",
    publisher = "MIT Press",
    url = "https://aclanthology.org/2024.tacl-1.54/",
    doi = "10.1162/tacl_a_00683",
    pages = "979--995"
}

@inproceedings{chrf2,
    title = {{chrF++: Words Helping Character N-grams}},
    author = "Popovi{\'c}, Maja",
    editor = "Bojar, Ond{\v{r}}ej  and
      Buck, Christian  and
      Chatterjee, Rajen  and
      Federmann, Christian  and
      Graham, Yvette  and
      Haddow, Barry  and
      Huck, Matthias  and
      Yepes, Antonio Jimeno  and
      Koehn, Philipp  and
      Kreutzer, Julia",
    booktitle = {{Proceedings of the Second Conference on Machine Translation}},
    month = sep,
    year = "2017",
    address = "Copenhagen, Denmark",
    publisher = "Association for Computational Linguistics",
    url = "https://aclanthology.org/W17-4770/",
    doi = "10.18653/v1/W17-4770",
    pages = "612--618"
}

@inproceedings{xsimplusplus,
    title = {{xSIM++: An Improved Proxy to Bitext Mining Performance for Low-Resource Languages}},
    author = "Chen, Mingda  and
      Heffernan, Kevin  and
      {\c{C}}elebi, Onur  and
      Mourachko, Alexandre  and
      Schwenk, Holger",
    editor = "Rogers, Anna  and
      Boyd-Graber, Jordan  and
      Okazaki, Naoaki",
    booktitle = {{Proceedings of the 61st Annual Meeting of the Association for Computational Linguistics (Volume 2: Short Papers)}},
    month = jul,
    year = "2023",
    address = "Toronto, Canada",
    publisher = "Association for Computational Linguistics",
    url = "https://aclanthology.org/2023.acl-short.10/",
    doi = "10.18653/v1/2023.acl-short.10",
    pages = "101--109"
}

@article{fsdp,
author = {Zhao, Yanli and Gu, Andrew and Varma, Rohan and Luo, Liang and Huang, Chien-Chin and Xu, Min and Wright, Less and Shojanazeri, Hamid and Ott, Myle and Shleifer, Sam and Desmaison, Alban and Balioglu, Can and Damania, Pritam and Nguyen, Bernard and Chauhan, Geeta and Hao, Yuchen and Mathews, Ajit and Li, Shen},
title = {{PyTorch FSDP: Experiences on Scaling Fully Sharded Data Parallel}},
year = {2023},
issue_date = {August 2023},
publisher = {VLDB Endowment},
volume = {16},
number = {12},
issn = {2150-8097},
url = {https://doi.org/10.14778/3611540.3611569},
doi = {10.14778/3611540.3611569},
journal = {Proc. VLDB Endow.},
month = aug,
pages = {3848–3860},
numpages = {13}
}

@inproceedings{
adamw,
title={{Decoupled Weight Decay Regularization}},
author={Ilya Loshchilov and Frank Hutter},
booktitle={{International Conference on Learning Representations}},
year={2019},
url={https://openreview.net/forum?id=Bkg6RiCqY7},
}

@misc{swiglu,
      title={{GLU Variants Improve Transformer}}, 
      author={Noam Shazeer},
      year={2020},
      eprint={2002.05202},
      archivePrefix={arXiv},
      primaryClass={cs.LG},
      url={https://arxiv.org/abs/2002.05202}, 
}

@inproceedings{afrolingumt,
    title = {{Toucan: Many-to-Many Translation for 150 African Language Pairs}},
    author = "Elmadany, AbdelRahim  and
      Adebara, Ife  and
      Abdul-Mageed, Muhammad",
    editor = "Ku, Lun-Wei  and
      Martins, Andre  and
      Srikumar, Vivek",
    booktitle = {{Findings of the Association for Computational Linguistics: ACL 2024}},
    month = aug,
    year = "2024",
    address = "Bangkok, Thailand",
    publisher = "Association for Computational Linguistics",
    url = "https://aclanthology.org/2024.findings-acl.781/",
    doi = "10.18653/v1/2024.findings-acl.781",
    pages = "13189--13206"
}

@inproceedings{wmt_oldi_25,
    title = {{Findings of the WMT 2025 Shared Task of the Open Language Data Initiative}},
    author = "Dale, David  and
      Burchell, Laurie  and
      Maillard, Jean  and
      Abdulmumin, Idris  and
      Anastasopoulos, Antonios  and
      Caswell, Isaac  and
      Koehn, Philipp",
    editor = "Haddow, Barry  and
      Kocmi, Tom  and
      Koehn, Philipp  and
      Monz, Christof",
    booktitle = {{Proceedings of the Tenth Conference on Machine Translation}},
    month = nov,
    year = "2025",
    address = "Suzhou, China",
    publisher = "Association for Computational Linguistics",
    url = "https://aclanthology.org/2025.wmt-1.26/",
    doi = "10.18653/v1/2025.wmt-1.26",
    pages = "495--502",
    ISBN = "979-8-89176-341-8"
}

@misc{sonar,
      title={{SONAR: Sentence-Level Multimodal and Language-Agnostic Representations}}, 
      author={Paul-Ambroise Duquenne and Holger Schwenk and Benoît Sagot},
      year={2023},
      eprint={2308.11466},
      archivePrefix={arXiv},
      primaryClass={cs.CL},
      url={https://arxiv.org/abs/2308.11466}, 
}

@inproceedings{
share_or_not,
title={{Share or Not? Learning to Schedule Language-Specific Capacity for Multilingual Translation}},
author={Biao Zhang and Ankur Bapna and Rico Sennrich and Orhan Firat},
booktitle={{International Conference on Learning Representations}},
year={2021},
url={https://openreview.net/forum?id=Wj4ODo0uyCF}
}

@InProceedings{panlex,
  author = {David Kamholz and Jonathan Pool and Susan Colowick},
  title = {{PanLex: Building a Resource for Panlingual Lexical Translation}},
  booktitle = {{Proceedings of the Ninth International Conference on Language Resources and Evaluation (LREC'14)}},
  year = {2014},
  month = {may},
  date = {26-31},
  address = {Reykjavik, Iceland},
  editor = {Nicoletta Calzolari (Conference Chair) and Khalid Choukri and Thierry Declerck and Hrafn Loftsson and Bente Maegaard and Joseph Mariani and Asuncion Moreno and Jan Odijk and Stelios Piperidis},
  publisher = {European Language Resources Association (ELRA)},
  isbn = {978-2-9517408-8-4},
  language = {english}
 }

@inproceedings{bouquet,
    title = {{BOUQuET : Dataset, Benchmark and Open initiative for Universal Quality Evaluation in Translation}},
    author = "{Omnilingual MT Team} and Andrews, Pierre  and
      Artetxe, Mikel  and
      Meglioli, Mariano Coria  and
      Costa-juss{\`a}, Marta R.  and
      Chuang, Joe  and
      Dale, David  and
      Duppenthaler, Mark  and
      Ekberg, Nathanial Paul  and
      Gao, Cynthia  and
      Licht, Daniel Edward  and
      Maillard, Jean  and
      Mourachko, Alexandre  and
      Ropers, Christophe  and
      Saleem, Safiyyah  and
      S{\'a}nchez, Eduardo  and
      Tsiamas, Ioannis  and
      Turkatenko, Arina  and
      Ventayol-Boada, Albert  and
      Yates, Shireen",
    editor = "Christodoulopoulos, Christos  and
      Chakraborty, Tanmoy  and
      Rose, Carolyn  and
      Peng, Violet",
    booktitle = {{Proceedings of the 2025 Conference on Empirical Methods in Natural Language Processing}},
    month = nov,
    year = "2025",
    address = "Suzhou, China",
    publisher = "Association for Computational Linguistics",
    url = "https://aclanthology.org/2025.emnlp-main.1400/",
    doi = "10.18653/v1/2025.emnlp-main.1400",
    pages = "27503--27523",
    ISBN = "979-8-89176-332-6"
}

@inproceedings{kreyol,
    title = {Krey{\`o}l-{MT}: Building {MT} for {L}atin {A}merican, {C}aribbean and Colonial {A}frican Creole Languages},
    author = {Robinson, Nathaniel  and
      Dabre, Raj  and
      Shurtz, Ammon  and
      Dent, Rasul  and
      Onesi, Onenamiyi  and
      Monroc, Claire  and
      Grobol, Lo{\"i}c  and
      Muhammad, Hasan  and
      Garg, Ashi  and
      Etori, Naome  and
      Tiyyala, Vijay Murari  and
      Samuel, Olanrewaju  and
      Stutzman, Matthew  and
      Odoom, Bismarck  and
      Khudanpur, Sanjeev  and
      Richardson, Stephen  and
      Murray, Kenton},
    editor = "Duh, Kevin  and
      Gomez, Helena  and
      Bethard, Steven",
    booktitle = {{Proceedings of the 2024 Conference of the North American Chapter of the Association for Computational Linguistics: Human Language Technologies (Volume 1: Long Papers)}},
    month = jun,
    year = "2024",
    address = "Mexico City, Mexico",
    publisher = "Association for Computational Linguistics",
    url = "https://aclanthology.org/2024.naacl-long.170/",
    doi = "10.18653/v1/2024.naacl-long.170",
    pages = "3083--3110"
}

@inproceedings{americasnlp,
    title = {{Findings of the AmericasNLP 2025 Shared Tasks on Machine Translation, Creation of Educational Material, and Translation Metrics for Indigenous Languages of the Americas}},
    author = {De Gibert, Ona  and
      Pugh, Robert  and
      Marashian, Ali  and
      Vazquez, Raul  and
      Ebrahimi, Abteen  and
      Denisov, Pavel  and
      Rice, Enora  and
      Gow-Smith, Edward  and
      Prieto, Juan  and
      Robles, Melissa  and
      Manrique, Rub{\'e}n  and
      Moreno, Oscar  and
      Lino, Angel  and
      Coto-Solano, Rolando  and
      Alvarez, Aldo  and
      Ag{\"u}ero-Torales, Marvin  and
      Ortega, John E.  and
      Chiruzzo, Luis  and
      Oncevay, Arturo  and
      Rijhwani, Shruti  and
      Von Der Wense, Katharina  and
      Mager, Manuel},
    editor = "Mager, Manuel  and
      Ebrahimi, Abteen  and
      Pugh, Robert  and
      Rijhwani, Shruti  and
      Von Der Wense, Katharina  and
      Chiruzzo, Luis  and
      Coto-Solano, Rolando  and
      Oncevay, Arturo",
    booktitle = {{Proceedings of the Fifth Workshop on NLP for Indigenous Languages of the Americas (AmericasNLP)}},
    month = may,
    year = "2025",
    address = "Albuquerque, New Mexico",
    publisher = "Association for Computational Linguistics",
    url = "https://aclanthology.org/2025.americasnlp-1.16/",
    doi = "10.18653/v1/2025.americasnlp-1.16",
    pages = "134--152",
    ISBN = "979-8-89176-236-7"
}

@article{bpcc,
title={{IndicTrans2: Towards High-Quality and Accessible Machine Translation Models for all 22 Scheduled Indian Languages}},
author={Jay Gala and Pranjal A Chitale and A K Raghavan and Varun Gumma and Sumanth Doddapaneni and Aswanth Kumar M and Janki Atul Nawale and Anupama Sujatha and Ratish Puduppully and Vivek Raghavan and Pratyush Kumar and Mitesh M Khapra and Raj Dabre and Anoop Kunchukuttan},
journal={Transactions on Machine Learning Research},
issn={2835-8856},
year={2023},
url={https://openreview.net/forum?id=vfT4YuzAYA},
note={}
}

@inproceedings{smol,
    title = {{SMOL: Professionally Translated Parallel Data for 115 Under-represented Languages}},
    author = "Caswell, Isaac  and
      Nielsen, Elizabeth  and
      Luo, Jiaming  and
      Cherry, Colin  and
      Kovacs, Geza  and
      Shemtov, Hadar  and
      Talukdar, Partha  and
      Tewari, Dinesh  and
      Diane, Baba Mamadi  and
      Diane, Djibrila  and
      Ciss{\'e}, Solo Farabado  and
      Doumbouya, Koulako Moussa  and
      Ferrante, Edoardo  and
      Guasoni, Alessandro  and
      Homan, Christopher  and
      Keita, Mamadou K.  and
      DebBarma, Sudhamoy  and
      Kuzhuget, Ali  and
      Anugraha, David  and
      Shulthan Habibi, Muhammad Ravi  and
      Ahmadi, Sina  and
      Munthali, Anthony  and
      Liu, Jonathan Mingfei  and
      Eng, Jonathan",
    editor = "Haddow, Barry  and
      Kocmi, Tom  and
      Koehn, Philipp  and
      Monz, Christof",
    booktitle = {{Proceedings of the Tenth Conference on Machine Translation}},
    month = nov,
    year = "2025",
    address = "Suzhou, China",
    publisher = "Association for Computational Linguistics",
    url = "https://aclanthology.org/2025.wmt-1.85/",
    doi = "10.18653/v1/2025.wmt-1.85",
    pages = "1103--1123",
    ISBN = "979-8-89176-341-8"
}

@inproceedings{gatitos,
    title = {{GATITOS: Using a New Multilingual Lexicon for Low-resource Machine Translation}},
    author = "Jones, Alexander  and
      Caswell, Isaac  and
      Firat, Orhan  and
      Saxena, Ishank",
    editor = "Bouamor, Houda  and
      Pino, Juan  and
      Bali, Kalika",
    booktitle = {{Proceedings of the 2023 Conference on Empirical Methods in Natural Language Processing}},
    month = dec,
    year = "2023",
    address = "Singapore",
    publisher = "Association for Computational Linguistics",
    url = "https://aclanthology.org/2023.emnlp-main.26/",
    doi = "10.18653/v1/2023.emnlp-main.26",
    pages = "371--405"
}

@inproceedings{glot500,
    title = {{Glot500: Scaling Multilingual Corpora and Language Models to 500 Languages}},
    author = {Imani, Ayyoob  and
      Lin, Peiqin  and
      Kargaran, Amir Hossein  and
      Severini, Silvia  and
      Jalili Sabet, Masoud  and
      Kassner, Nora  and
      Ma, Chunlan  and
      Schmid, Helmut  and
      Martins, Andr{\'e}  and
      Yvon, Fran{\c{c}}ois  and
      Sch{\"u}tze, Hinrich},
    editor = "Rogers, Anna  and
      Boyd-Graber, Jordan  and
      Okazaki, Naoaki",
    booktitle = {{Proceedings of the 61st Annual Meeting of the Association for Computational Linguistics (Volume 1: Long Papers)}},
    month = jul,
    year = "2023",
    address = "Toronto, Canada",
    publisher = "Association for Computational Linguistics",
    url = "https://aclanthology.org/2023.acl-long.61/",
    doi = "10.18653/v1/2023.acl-long.61",
    pages = "1082--1117"
}

@inproceedings{mult_representation_distill,
    title = {{Multilingual Representation Distillation with Contrastive Learning}},
    author = "Tan, Weiting  and
      Heffernan, Kevin  and
      Schwenk, Holger  and
      Koehn, Philipp",
    editor = "Vlachos, Andreas  and
      Augenstein, Isabelle",
    booktitle = {{Proceedings of the 17th Conference of the European Chapter of the Association for Computational Linguistics}},
    month = may,
    year = "2023",
    address = "Dubrovnik, Croatia",
    publisher = "Association for Computational Linguistics",
    url = "https://aclanthology.org/2023.eacl-main.108/",
    doi = "10.18653/v1/2023.eacl-main.108",
    pages = "1477--1490"
}

@inproceedings{multiway_multilingual_nmt,
    title = {{Multi-Way, Multilingual Neural Machine Translation with a Shared Attention Mechanism}},
    author = "Firat, Orhan  and
      Cho, Kyunghyun  and
      Bengio, Yoshua",
    editor = "Knight, Kevin  and
      Nenkova, Ani  and
      Rambow, Owen",
    booktitle = {{Proceedings of the 2016 Conference of the North {A}merican Chapter of the Association for Computational Linguistics: Human Language Technologies}},
    month = jun,
    year = "2016",
    address = "San Diego, California",
    publisher = "Association for Computational Linguistics",
    url = "https://aclanthology.org/N16-1101/",
    doi = "10.18653/v1/N16-1101",
    pages = "866--875"
}

@inproceedings{massively_multilingual_nmt,
    title = {{Massively Multilingual Neural Machine Translation}},
    author = "Aharoni, Roee  and
      Johnson, Melvin  and
      Firat, Orhan",
    editor = "Burstein, Jill  and
      Doran, Christy  and
      Solorio, Thamar",
    booktitle = {{Proceedings of the 2019 Conference of the North {A}merican Chapter of the Association for Computational Linguistics: Human Language Technologies, Volume 1 (Long and Short Papers)}},
    month = jun,
    year = "2019",
    address = "Minneapolis, Minnesota",
    publisher = "Association for Computational Linguistics",
    url = "https://aclanthology.org/N19-1388/",
    doi = "10.18653/v1/N19-1388",
    pages = "3874--3884"
}

@inproceedings{lifting_the_curse_of_multilinguality,
    title = {{Lifting the Curse of Multilinguality by Pre-training Modular Transformers}},
    author = "Pfeiffer, Jonas  and
      Goyal, Naman  and
      Lin, Xi  and
      Li, Xian  and
      Cross, James  and
      Riedel, Sebastian  and
      Artetxe, Mikel",
    editor = "Carpuat, Marine  and
      de Marneffe, Marie-Catherine  and
      Meza Ruiz, Ivan Vladimir",
    booktitle = {{Proceedings of the 2022 Conference of the North American Chapter of the Association for Computational Linguistics: Human Language Technologies}},
    month = jul,
    year = "2022",
    address = "Seattle, United States",
    publisher = "Association for Computational Linguistics",
    url = "https://aclanthology.org/2022.naacl-main.255/",
    doi = "10.18653/v1/2022.naacl-main.255",
    pages = "3479--3495"
}

@inproceedings{when_is_multilinguality_a_curse,
    title = {{When Is Multilinguality a Curse? Language Modeling for 250 High- and Low-Resource Languages}},
    author = "Chang, Tyler A.  and
      Arnett, Catherine  and
      Tu, Zhuowen  and
      Bergen, Ben",
    editor = "Al-Onaizan, Yaser  and
      Bansal, Mohit  and
      Chen, Yun-Nung",
    booktitle = {{Proceedings of the 2024 Conference on Empirical Methods in Natural Language Processing}},
    month = nov,
    year = "2024",
    address = "Miami, Florida, USA",
    publisher = "Association for Computational Linguistics",
    url = "https://aclanthology.org/2024.emnlp-main.236/",
    doi = "10.18653/v1/2024.emnlp-main.236",
    pages = "4074--4096"
}

@inproceedings{unimax,
    title={{UniMax: Fairer and More Effective Language Sampling for Large-Scale Multilingual Pretraining}},
    author={Hyung Won Chung and Xavier Garcia and Adam Roberts and Yi Tay and Orhan Firat and Sharan Narang and Noah Constant},
    booktitle={{The Eleventh International Conference on Learning Representations }},
    year={2023},
    url={https://openreview.net/forum?id=kXwdL1cWOAi}
}

@inproceedings{infonce,
    author = {Chen, Ting and Kornblith, Simon and Norouzi, Mohammad and Hinton, Geoffrey},
    title = {{A Simple Framework for Contrastive Learning of Visual Representations}},
    year = {2020},
    publisher = {JMLR.org},
    booktitle = {{Proceedings of the 37th International Conference on Machine Learning}},
    articleno = {149},
    numpages = {11},
    series = {ICML'20}
}

@misc{llama3,
      title={{The Llama 3 Herd of Models}}, 
      author={Llama Team, AI @ Meta},
      year={2024},
      eprint={2407.21783},
      archivePrefix={arXiv},
      primaryClass={cs.AI},
      url={https://arxiv.org/abs/2407.21783}, 
}

@article{duquenne2021multimodal,
  title={{Multimodal and Multilingual Embeddings for Large-Scale Speech Mining}},
  author={Duquenne, Paul-Ambroise and Gong, Hongyu and Schwenk, Holger},
  journal={Advances in Neural Information Processing Systems},
  volume={34},
  url={https://proceedings.neurips.cc/paper/2021/hash/8466f9ace6a9acbe71f75762ffc890f1-Abstract.html},
  pages={15748--15761},
  year={2021}
}

@inproceedings{duquenne2023modular,
  author    = {Paul-Ambroise Duquenne and Holger Schwenk and Benoît Sagot},
  title     = {{Modular Speech-to-Text Translation for Zero-Shot Cross-Modal Transfer}},
  booktitle = {Proc. Interspeech 2023},
  pages     = {32--36},
  year      = {2023},
  doi       = {10.21437/Interspeech.2023-2484}
}

@inproceedings{duquenne2022t,
    title = {{T-Modules: Translation Modules for Zero-Shot Cross-Modal Machine Translation}},
    author = "Duquenne, Paul-Ambroise  and
      Gong, Hongyu  and
      Sagot, Beno{\^i}t  and
      Schwenk, Holger",
    editor = "Goldberg, Yoav  and
      Kozareva, Zornitsa  and
      Zhang, Yue",
    booktitle = "Proceedings of the 2022 Conference on Empirical Methods in Natural Language Processing",
    month = dec,
    year = "2022",
    address = "Abu Dhabi, United Arab Emirates",
    publisher = "Association for Computational Linguistics",
    url = "https://aclanthology.org/2022.emnlp-main.391/",
    doi = "10.18653/v1/2022.emnlp-main.391",
    pages = "5794--5806"
}

@ARTICLE{khurana2022samu,
  author={Khurana, Sameer and Laurent, Antoine and Glass, James},
  journal={IEEE Journal of Selected Topics in Signal Processing}, 
  title={{SAMU-XLSR: Semantically-Aligned Multimodal Utterance-Level Cross-Lingual Speech Representation}}, 
  year={2022},
  volume={16},
  number={6},
  pages={1493-1504},
  doi={10.1109/JSTSP.2022.3192714}}

@inproceedings{duquenne2023speechmatrix,
    title = {{SpeechMatrix: A Large-Scale Mined Corpus of Multilingual Speech-to-Speech Translations}},
    author = "Duquenne, Paul-Ambroise  and
      Gong, Hongyu  and
      Dong, Ning  and
      Du, Jingfei  and
      Lee, Ann  and
      Goswami, Vedanuj  and
      Wang, Changhan  and
      Pino, Juan  and
      Sagot, Beno{\^i}t  and
      Schwenk, Holger",
    editor = "Rogers, Anna  and
      Boyd-Graber, Jordan  and
      Okazaki, Naoaki",
    booktitle = "Proceedings of the 61st Annual Meeting of the Association for Computational Linguistics (Volume 1: Long Papers)",
    month = jul,
    year = "2023",
    address = "Toronto, Canada",
    publisher = "Association for Computational Linguistics",
    url = "https://aclanthology.org/2023.acl-long.899/",
    doi = "10.18653/v1/2023.acl-long.899",
    pages = "16251--16269"
}

@inproceedings{wang2021voxpopuli,
    title = {{VoxPopuli: A Large-Scale Multilingual Speech Corpus for Representation Learning, Semi-Supervised Learning and Interpretation}},
    author = "Wang, Changhan  and
      Riviere, Morgane  and
      Lee, Ann  and
      Wu, Anne  and
      Talnikar, Chaitanya  and
      Haziza, Daniel  and
      Williamson, Mary  and
      Pino, Juan  and
      Dupoux, Emmanuel",
    editor = "Zong, Chengqing  and
      Xia, Fei  and
      Li, Wenjie  and
      Navigli, Roberto",
    booktitle = "Proceedings of the 59th Annual Meeting of the Association for Computational Linguistics and the 11th International Joint Conference on Natural Language Processing (Volume 1: Long Papers)",
    month = aug,
    year = "2021",
    address = "Online",
    publisher = "Association for Computational Linguistics",
    url = "https://aclanthology.org/2021.acl-long.80/",
    doi = "10.18653/v1/2021.acl-long.80",
    pages = "993--1003"
}

@inproceedings{mitsui2024pslm,
  title={Pslm: Parallel generation of text and speech with llms for low-latency spoken dialogue systems},
  author={Mitsui, Kentaro and Mitsuda, Koh and Wakatsuki, Toshiaki and Hono, Yukiya and Sawada, Kei},
  booktitle={Findings of the Association for Computational Linguistics: EMNLP 2024},
  pages={2692--2700},
  year={2024}
}

@misc{omtbigpaper,
      title={Omnilingual {MT}: Machine Translation for 1,600 Languages}, 
      author={{Omnilingual MT Team} and Belen Alastruey and Niyati Bafna and Andrea Caciolai and Kevin Heffernan and Artyom Kozhevnikov and Christophe Ropers and Eduardo S{\'a}nchez and Charles-Eric Saint-James and Ioannis Tsiamas and Chierh Cheng and Joe Chuang and Paul-Ambroise Duquenne and Mark Duppenthaler and Nate Ekberg and Cynthia Gao and Pere Llu{\'i}s Huguet Cabot and Jo{\~a}o Maria Janeiro and Jean Maillard and Gabriel Mejia Gonzalez and Holger Schwenk and Edan Toledo and Arina Turkatenko and Albert Ventayol-Boada and Rashel Moritz and Alexandre Mourachko and Surya Parimi and Mary Williamson and Shireen Yates and David Dale and Marta R. Costa-juss{\`a}},
      year={2026},
      eprint={2603.16309},
      archivePrefix={arXiv},
      primaryClass={cs.CL},
      url={https://arxiv.org/abs/2603.16309}, 
}

@article{wang2025mats,
  title={MATS: An Audio Language Model under Text-only Supervision},
  author={Wang, Wen and Hou, Ruibing and Chang, Hong and Shan, Shiguang and Chen, Xilin},
  journal={arXiv preprint arXiv:2502.13433},
  year={2025}
}

@article{agostinelli2023musiclm,
  title={Musiclm: Generating music from text},
  author={Agostinelli, Andrea and Denk, Timo I and Borsos, Zal{\'a}n and Engel, Jesse and Verzetti, Mauro and Caillon, Antoine and Huang, Qingqing and Jansen, Aren and Roberts, Adam and Tagliasacchi, Marco and others},
  journal={arXiv preprint arXiv:2301.11325},
  year={2023}
}

@inproceedings{lee2022textless,
    title = {{Textless Speech-to-Speech Translation on Real Data}},
    author = "Lee, Ann  and
      Gong, Hongyu  and
      Duquenne, Paul-Ambroise  and
      Schwenk, Holger  and
      Chen, Peng-Jen  and
      Wang, Changhan  and
      Popuri, Sravya  and
      Adi, Yossi  and
      Pino, Juan  and
      Gu, Jiatao  and
      Hsu, Wei-Ning",
    editor = "Carpuat, Marine  and
      de Marneffe, Marie-Catherine  and
      Meza Ruiz, Ivan Vladimir",
    booktitle = "Proceedings of the 2022 Conference of the North American Chapter of the Association for Computational Linguistics: Human Language Technologies",
    month = jul,
    year = "2022",
    address = "Seattle, United States",
    publisher = "Association for Computational Linguistics",
    url = "https://aclanthology.org/2022.naacl-main.63/",
    doi = "10.18653/v1/2022.naacl-main.63",
    pages = "860--872"
}

@unpublished{duquenne:hal-04629427,
  TITLE = {{SONAR EXPRESSIVE: Zero-shot Expressive Speech-to-Speech Translation}},
  AUTHOR = {Duquenne, Paul-Ambroise and Heffernan, Kevin and Mourachko, Alexandre and Sagot, Beno{\^i}t and Schwenk, Holger},
  URL = {https://hal.science/hal-04629427},
  NOTE = {working paper or preprint},
  YEAR = {2023},
  MONTH = Nov,
  HAL_ID = {hal-04629427},
  HAL_VERSION = {v1},
}

@inproceedings{chen2023speech,
    title = {{Speech-to-Speech Translation for a Real-world Unwritten Language}},
    author = "Chen, Peng-Jen  and
      Tran, Kevin  and
      Yang, Yilin  and
      Du, Jingfei  and
      Kao, Justine  and
      Chung, Yu-An  and
      Tomasello, Paden  and
      Duquenne, Paul-Ambroise  and
      Schwenk, Holger  and
      Gong, Hongyu  and
      Inaguma, Hirofumi  and
      Popuri, Sravya  and
      Wang, Changhan  and
      Pino, Juan  and
      Hsu, Wei-Ning  and
      Lee, Ann",
    editor = "Rogers, Anna  and
      Boyd-Graber, Jordan  and
      Okazaki, Naoaki",
    booktitle = "Findings of the Association for Computational Linguistics: ACL 2023",
    month = jul,
    year = "2023",
    address = "Toronto, Canada",
    publisher = "Association for Computational Linguistics",
    url = "https://aclanthology.org/2023.findings-acl.307/",
    doi = "10.18653/v1/2023.findings-acl.307",
    pages = "4969--4983"
}

@inproceedings{abate2005alffa,
  title     = {{An Amharic Speech Corpus for Large Vocabulary Continuous Speech Recognition}},
  author    = {Solomon Teferra Abate and Wolfgang Menzel and Bairu Tafila},
  year      = {2005},
  booktitle = {{Interspeech 2005}},
  pages     = {1601--1604},
  doi       = {10.21437/Interspeech.2005-467},
  issn      = {2958-1796},
  url = {https://www.isca-archive.org/interspeech_2005/abate05_interspeech.html}
}

@inproceedings{gelas2012alffa,
  title     = {{Developments of Swahili resources for an automatic speech recognition system}},
  author    = {Hadrien Gelas and Laurent Besacier and François Pellegrino},
  year      = {2012},
  booktitle = {{3rd Workshop on Spoken Language Technologies for Under-Resourced Languages (SLTU 2012)}},
  pages     = {94--101},
  url = {https://www.isca-archive.org/sltu_2012/gelas12_sltu.html}
}

@inproceedings{gauthier2016alffa,
    title = {{Collecting Resources in Sub-Saharan African Languages for Automatic Speech Recognition: a Case Study of Wolof}},
    author = "Gauthier, Elodie  and
      Besacier, Laurent  and
      Voisin, Sylvie  and
      Melese, Michael  and
      Elingui, Uriel Pascal",
    editor = "Calzolari, Nicoletta  and
      Choukri, Khalid  and
      Declerck, Thierry  and
      Goggi, Sara  and
      Grobelnik, Marko  and
      Maegaard, Bente  and
      Mariani, Joseph  and
      Mazo, Helene  and
      Moreno, Asuncion  and
      Odijk, Jan  and
      Piperidis, Stelios",
    booktitle = "Proceedings of the Tenth International Conference on Language Resources and Evaluation ({LREC}'16)",
    month = may,
    year = "2016",
    address = "Portoro{\v{z}}, Slovenia",
    publisher = "European Language Resources Association (ELRA)",
    url = "https://aclanthology.org/L16-1611/",
    pages = "3863--3867"
}

@inproceedings{vanniekerk2017rapid,
title	= {{Rapid Development of TTS corpora for Four South African Languages}},author	= {Daniel van Niekerk and Charl van Heerden and Marelie Davel and Neil Kleynhans and Oddur Kjartansson and Martin Jansche and Linne Ha},year	= {2017},URL	= {http://dx.doi.org/10.21437/Interspeech.2017-1139},booktitle	= {Proc. Interspeech 2017},pages	= {2178--2182}}

@inproceedings{kjartansson2018crowd,
title = {{Crowd-Sourced Speech Corpora for Javanese, Sundanese,  Sinhala, Nepali, and Bangladeshi Bengali}},
author = {Oddur Kjartansson and Supheakmungkol Sarin and Knot Pipatsrisawat and Martin Jansche and Linne Ha},
booktitle = {Proc. The 6th Intl. Workshop on Spoken Language Technologies for Under-Resourced Languages (SLTU)},
year  = {2018},
}

@inproceedings{kjartansson2018tts,
title = {{A Step-by-Step Process for Building TTS Voices Using Open Source Data and Framework for Bangla, Javanese, Khmer, Nepali, Sinhala, and Sundanese}},
author = {Keshan Sodimana and Knot Pipatsrisawat and Linne Ha and Martin Jansche and Oddur Kjartansson and Pasindu De Silva and Supheakmungkol Sarin},
booktitle = {Proc. The 6th Intl. Workshop on Spoken Language Technologies for Under-Resourced Languages (SLTU)},
year  = {2018}
}

@inproceedings{he2020open,
title = {{Open-source Multi-speaker Speech Corpora for Building Gujarati, Kannada, Malayalam, Marathi, Tamil and Telugu Speech Synthesis Systems}},
author = {He, Fei and Chu, Shan-Hui Cathy and Kjartansson, Oddur and Rivera, Clara and Katanova, Anna and Gutkin, Alexander and Demirsahin, Isin and Johny, Cibu and Jansche, Martin and Sarin, Supheakmungkol and Pipatsrisawat, Knot},
booktitle = {Proceedings of The 12th Language Resources and Evaluation Conference (LREC)},
year = {2020}
}

@inproceedings{oo2020burmese,
    title = {{Burmese Speech Corpus, Finite-State Text Normalization and Pronunciation Grammars with an Application to Text-to-Speech}},
    author = {Oo, Yin May and Wattanavekin, Theeraphol and Li, Chenfang and De Silva, Pasindu and Sarin, Supheakmungkol and Pipatsrisawat, Knot and Jansche, Martin and Kjartansson, Oddur and Gutkin, Alexander},
    booktitle = {Proceedings of The 12th Language Resources and Evaluation Conference (LREC)},
    year = {2020}
}

@article{park2019css10,
  title={CSS10: A Collection of Single Speaker Speech Datasets for 10 Languages},
  author={Park, Kyubyong and Mulc, Thomas},
  journal={Interspeech},
  year={2019}
}

@misc{fosd,
  author = {Tran, Duc Chung},
  title = {{FPT} {O}pen {S}peech {D}ataset ({FOSD}) - {V}ietnamese},
  url = {https://data.mendeley.com/datasets/k9sxg2twv4/4},
  year = {2020},
  doi = {10.17632/k9sxg2twv4.4},
  journal = {Mendeley Data, V4},
}

@article{gopinath2022imascic,
      title={{IMaSC} -- {ICFOSS} Malayalam Speech Corpus}, 
      author={Deepa P Gopinath and Thennal D K and Vrinda V Nair and Swaraj K S and Sachin G},
      year={2022},
      journal={arXiv preprint arXiv:2211.12796},
      url={https://arxiv.org/abs/2211.12796}, 
}

@inproceedings{ljubesic2022parlaspeech,
    title = "{P}arla{S}peech-{HR} - a Freely Available {ASR} Dataset for {C}roatian Bootstrapped from the {P}arla{M}int Corpus",
    author = "Ljube{\v{s}}i{\'c}, Nikola  and
      Kor{\v{z}}inek, Danijel  and
      Rupnik, Peter  and
      Jazbec, Ivo-Pavao",
    booktitle = "Proceedings of the Workshop ParlaCLARIN III within the 13th Language Resources and Evaluation Conference",
    year = "2022"
}

@inproceedings{khassanov2021crowdsourced,
  title = "A Crowdsourced Open-Source {K}azakh Speech Corpus and Initial Speech Recognition Baseline",
  author={Yerbolat Khassanov and Saida Mussakhojayeva and Almas Mirzakhmetov and Alen Adiyev and Mukhamet Nurpeiissov and Huseyin Atakan Varol},
  booktitle = "Proceedings of the 16th Conference of the European Chapter of the Association for Computational Linguistics: Main Volume",
  year = "2021"
}

@inproceedings{mollberg2020samromur,
    title = "{S}amr{\'o}mur: Crowd-sourcing Data Collection for {I}celandic Speech Recognition",
    author = "Mollberg, David Erik  and
      J{\'o}nsson, {\'O}lafur Helgi  and
      {\TH}orsteinsd{\'o}ttir, Sunneva  and
      Steingr{\'\i}msson, Stein{\th}{\'o}r  and
      Magn{\'u}sd{\'o}ttir, Eyd{\'\i}s Huld  and
      Gudnason, Jon",
    booktitle = "Proceedings of the 12th Language Resources and Evaluation Conference",
    year = "2020"
}

@inproceedings{solberg2022norwegian,
    title = "The {N}orwegian Parliamentary Speech Corpus",
    author = "Solberg, Per Erik  and
      Ortiz, Pablo",
    booktitle = "Proceedings of the Thirteenth Language Resources and Evaluation Conference",
    year = "2022"
}

@article{emezue2025naijavoices,
  title={The NaijaVoices Dataset: Cultivating Large-Scale, High-Quality, Culturally-Rich Speech Data for African Languages},
  author={Emezue, Chris and Community, NaijaVoices and Awobade, Busayo and Owodunni, Abraham and Emezue, Handel and Emezue, Gloria Monica Tobechukwu and Emezue, Nefertiti Nneoma and Ogun, Sewade and Akinremi, Bunmi and Adelani, David Ifeoluwa and others},
  journal={arXiv preprint arXiv:2505.20564},
  year={2025}
}

@inproceedings{panayotov2015librispeech,
  title={Librispeech: an asr corpus based on public domain audio books},
  author={Panayotov, Vassil and Chen, Guoguo and Povey, Daniel and Khudanpur, Sanjeev},
  booktitle={2015 IEEE international conference on acoustics, speech and signal processing (ICASSP)},
  pages={5206--5210},
  year={2015},
  organization={IEEE}
}

@inproceedings{valk2021slt,
  title={{VoxLingua107}: a Dataset for Spoken Language Recognition},
  author={J{\"o}rgen Valk and Tanel Alum{\"a}e},
  booktitle={Proc. IEEE SLT Workshop},
  year={2021},
}

@article{pratap2020mls,
  title={MLS: A Large-Scale Multilingual Dataset for Speech Research},
  author={Pratap, Vineel and Xu, Qiantong and Sriram, Anuroop and Synnaeve, Gabriel and Collobert, Ronan},
  journal={Interspeech 2020},
  year={2020},
  publisher={ISCA}
}

@inproceedings{ardila2020common,
  title={Common Voice: A Massively-Multilingual Speech Corpus},
  author={Ardila, Rosana and Branson, Megan and Davis, Kelly and Kohler, Michael and Meyer, Josh and Henretty, Michael and Morais, Reuben and Saunders, Lindsay and Tyers, Francis and Weber, Gregor},
  booktitle={Proceedings of the Twelfth Language Resources and Evaluation Conference},
  pages={4218--4222},
  year={2020}
}

@inproceedings{conneau2023fleurs,
  title={Fleurs: Few-shot learning evaluation of universal representations of speech},
  author={Conneau, Alexis and Ma, Min and Khanuja, Simran and Zhang, Yu and Axelrod, Vera and Dalmia, Siddharth and Riesa, Jason and Rivera, Clara and Bapna, Ankur},
  booktitle={2022 IEEE Spoken Language Technology Workshop (SLT)},
  pages={798--805},
  year={2023},
  organization={IEEE}
}

@inproceedings{seed-23,
    title = {Small Data, Big Impact: Leveraging Minimal Data for Effective Machine Translation},
    author = {Maillard, Jean and Gao, Cynthia and Kalbassi, Elahe and Sadagopan, Kaushik Ram and Goswami, Vedanuj and Koehn, Philipp and Fan, Angela and Guzmán, Francisco},
    booktitle = {Proceedings of the 61st Annual Meeting of the Association for Computational Linguistics (Volume 1: Long Papers)},
    year = {2023},
    address = {Toronto, Canada},
    publisher = {Association for Computational Linguistics},
    pages = {2740--2756},
    url = {https://aclanthology.org/2023.acl-long.154},
}

@inproceedings{heffernan2022bitext,
    title = {{Bitext Mining Using Distilled Sentence Representations for Low-Resource Languages}},
    author = "Heffernan, Kevin  and
      {\c{C}}elebi, Onur  and
      Schwenk, Holger",
    editor = "Goldberg, Yoav  and
      Kozareva, Zornitsa  and
      Zhang, Yue",
    booktitle = "Findings of the Association for Computational Linguistics: EMNLP 2022",
    month = dec,
    year = "2022",
    address = "Abu Dhabi, United Arab Emirates",
    publisher = "Association for Computational Linguistics",
    url = "https://aclanthology.org/2022.findings-emnlp.154/",
    doi = "10.18653/v1/2022.findings-emnlp.154",
    pages = "2101--2112"
}

@article{goyal2022flores,
    title = "The {F}lores-101 Evaluation Benchmark for Low-Resource and Multilingual Machine Translation",
    author = "Goyal, Naman  and
      Gao, Cynthia  and
      Chaudhary, Vishrav  and
      Chen, Peng-Jen  and
      Wenzek, Guillaume  and
      Ju, Da  and
      Krishnan, Sanjana  and
      Ranzato, Marc{'}Aurelio  and
      Guzm{\'a}n, Francisco  and
      Fan, Angela",
    editor = "Roark, Brian  and
      Nenkova, Ani",
    journal = "Transactions of the Association for Computational Linguistics",
    volume = "10",
    year = "2022",
    address = "Cambridge, MA",
    publisher = "MIT Press",
    url = "https://aclanthology.org/2022.tacl-1.30/",
    doi = "10.1162/tacl_a_00474",
    pages = "522--538"
}

@article{roberts2019prefixlm,
author = {Raffel, Colin and Shazeer, Noam and Roberts, Adam and Lee, Katherine and Narang, Sharan and Matena, Michael and Zhou, Yanqi and Li, Wei and Liu, Peter J.},
title = {{Exploring the Limits of Transfer Learning with a Unified Text-to-Text Transformer}},
year = {2020},
issue_date = {January 2020},
publisher = {JMLR.org},
volume = {21},
number = {1},
issn = {1532-4435},
journal = {J. Mach. Learn. Res.},
month = jan,
articleno = {140},
numpages = {67}
}

@InProceedings{martelli-EtAl:2023:clicit,
  author    = {Martelli, Federico  and  Bejgu, Andrei Stefan  and  Campagnano, Cesare  and  Čibej, Jaka  and  Costa, Rute  and  Gantar, Apolonija  and  Kallas, Jelena  and  Koeva, Svetla  and  Koppel, Kristina  and  Krek, Simon  and  Langemets, Margit  and  Lipp, Veronika  and  Nimb, Sanni  and  Olsen, Sussi  and  Pedersen, Bolette Sandford  and  Quochi, Valeria  and  Salgado, Ana  and  Simon, László  and  Tiberius, Carole  and  Ureña-Ruiz, Rafael-J  and  Navigli, Roberto},
  title     = {{XL-WA: a Gold Evaluation Benchmark for Word Alignment in 14 Language Pairs}},
  booktitle      = {Procedings of the Ninth Italian Conference on Computational Linguistics (CLiC-it 2023)},
  month          = {November},
  year           = {2023}
}

@article{yang2025qwen3,
  title={Qwen3 technical report},
  author={Yang, An and Li, Anfeng and Yang, Baosong and Zhang, Beichen and Hui, Binyuan and Zheng, Bo and Yu, Bowen and Gao, Chang and Huang, Chengen and Lv, Chenxu and others},
  journal={arXiv preprint arXiv:2505.09388},
  year={2025}
}

@techreport{salamanca2025tinyaya,
  title={Tiny Aya: Bridging Scale and Multilingual Depth},
  author={Salamanca, Alejandro R. and Abagyan, Diana and D'souza, Daniel and Khairi, Ammar and Mora, David and Dash, Saurabh and Aryabumi, Viraat and Rajaee, Sara and Mofakhami, Mehrnaz and Sahu, Ananya and Euyang, Thomas and Prince, Brittawnya and Smith, Madeline and Lin, Hangyu and Locatelli, Acyr and Hooker, Sara and Kocmi, Tom and Gomez, Aidan and Zhang, Ivan and Blunsom, Phil and Frosst, Nick and Pineau, Joelle and Ermis, Beyza and {\"U}st{\"u}n, Ahmet and Kreutzer, Julia and Fadaee, Marzieh},
  year={2025},
  month={January},
  institution={Cohere Labs},
  url={https://github.com/Cohere-Labs/tiny-aya-tech-report/blob/main/tiny_aya_tech_report.pdf},
  note={Accessed: 2025-01-23}
}

@article{nguyen2025spirit,
  title={Spirit-lm: Interleaved spoken and written language model},
  author={Nguyen, Tu Anh and Muller, Benjamin and Yu, Bokai and Costa-Jussa, Marta R and Elbayad, Maha and Popuri, Sravya and Ropers, Christophe and Duquenne, Paul-Ambroise and Algayres, Robin and Mavlyutov, Ruslan and others},
  journal={Transactions of the Association for Computational Linguistics},
  volume={13},
  pages={30--52},
  year={2025},
  publisher={MIT Press 255 Main Street, 9th Floor, Cambridge, Massachusetts 02142, USA~…}
}

@article{hassid2023textually,
  title={Textually pretrained speech language models},
  author={Hassid, Michael and Remez, Tal and Nguyen, Tu Anh and Gat, Itai and Conneau, Alexis and Kreuk, Felix and Copet, Jade and Defossez, Alexandre and Synnaeve, Gabriel and Dupoux, Emmanuel and others},
  journal={Advances in Neural Information Processing Systems},
  volume={36},
  pages={63483--63501},
  year={2023}
}

@article{hsu2021hubert,
  title={Hubert: Self-supervised speech representation learning by masked prediction of hidden units},
  author={Hsu, Wei-Ning and Bolte, Benjamin and Tsai, Yao-Hung Hubert and Lakhotia, Kushal and Salakhutdinov, Ruslan and Mohamed, Abdelrahman},
  journal={IEEE/ACM transactions on audio, speech, and language processing},
  volume={29},
  pages={3451--3460},
  year={2021},
  publisher={IEEE}
}

@misc{embedding_gemma_2025,
    title={{EmbeddingGemma: Powerful and Lightweight Text Representations}},
    author={Schechter Vera, Henrique and Dua, Sahil and Zhang, Biao and Salz, Daniel and Mullins, Ryan and Raghuram Panyam, Sindhu and Smoot, Sara and Naim, Iftekhar and Zou, Joe and Chen, Feiyang and Cer, Daniel and Lisak, Alice and Choi, Min and Gonzalez, Lucas and Sanseviero, Omar and Cameron, Glenn and Ballantyne, Ian and Black, Kat and Chen, Kaifeng and Wang, Weiyi and Li, Zhe and Martins, Gus and Lee, Jinhyuk and Sherwood, Mark and Ji, Juyeong and Wu, Renjie and Zheng, Jingxiao and Singh, Jyotinder and Sharma, Abheesht and Sreepat, Divya and Jain, Aashi and Elarabawy, Adham and Co, AJ and Doumanoglou, Andreas and Samari, Babak and Hora, Ben and Potetz, Brian and Kim, Dahun and Alfonseca, Enrique and Moiseev, Fedor and Han, Feng and Palma Gomez, Frank and Hernández Ábrego, Gustavo and Zhang, Hesen and Hui, Hui and Han, Jay and Gill, Karan and Chen, Ke and Chen, Koert and Shanbhogue, Madhuri and Boratko, Michael and Suganthan, Paul and Duddu, Sai Meher Karthik and Mariserla, Sandeep and Ariafar, Setareh and Zhang, Shanfeng and Zhang, Shijie and Baumgartner, Simon and Goenka, Sonam and Qiu, Steve and Dabral, Tanmaya and Walker, Trevor and Rao, Vikram and Khawaja, Waleed and Zhou, Wenlei and Ren, Xiaoqi and Xia, Ye and Chen, Yichang and Chen, Yi-Ting and Dong, Zhe and Ding, Zhongli and Visin, Francesco and Liu, Gaël and Zhang, Jiageng and Kenealy, Kathleen and Casbon, Michelle and Kumar, Ravin and Mesnard, Thomas and Gleicher, Zach and Brick, Cormac and Lacombe, Olivier and Roberts, Adam and Sung, Yunhsuan and Hoffmann, Raphael and Warkentin, Tris and Joulin, Armand and Duerig, Tom and Seyedhosseini, Mojtaba},
    publisher={Google DeepMind},
    year={2025},
    url={https://arxiv.org/abs/2509.20354}
}

@inproceedings{tsiamas22_interspeech,
  title     = {{SHAS: Approaching Optimal Segmentation for End-to-End Speech Translation}},
  author    = {Ioannis Tsiamas and Gerard I. Gállego and José A. R. Fonollosa and Marta R. Costa-jussà},
  year      = {2022},
  booktitle = {{Interspeech 2022}},
  pages     = {106--110},
  doi       = {10.21437/Interspeech.2022-59},
  issn      = {2958-1796},
}

@misc{moshi,
      title={{Moshi: A Speech-Text Foundation Model for Real-time Dialogue}},
      author={Alexandre D\'efossez and Laurent Mazar\'e and Manu Orsini and
      Am\'elie Royer and Patrick P\'erez and Herv\'e J\'egou and Edouard Grave and Neil Zeghidour},
      year={2024},
      eprint={2410.00037},
      archivePrefix={arXiv},
      primaryClass={eess.AS},
      url={https://arxiv.org/abs/2410.00037},
}

@misc{clip,
      title={{Learning Transferable Visual Models From Natural Language Supervision}}, 
      author={Alec Radford and Jong Wook Kim and Chris Hallacy and Aditya Ramesh and Gabriel Goh and Sandhini Agarwal and Girish Sastry and Amanda Askell and Pamela Mishkin and Jack Clark and Gretchen Krueger and Ilya Sutskever},
      year={2021},
      eprint={2103.00020},
      archivePrefix={arXiv},
      primaryClass={cs.CV},
      url={https://arxiv.org/abs/2103.00020}, 
}

@inproceedings{zhang2023speechgpt,
  title={Speechgpt: Empowering large language models with intrinsic cross-modal conversational abilities},
  author={Zhang, Dong and Li, Shimin and Zhang, Xin and Zhan, Jun and Wang, Pengyu and Zhou, Yaqian and Qiu, Xipeng},
  booktitle={Findings of the Association for Computational Linguistics: EMNLP 2023},
  pages={15757--15773},
  year={2023}
}

@article{roy2026personaplex,
  title={PersonaPlex: Voice and Role Control for Full Duplex Conversational Speech Models},
  author={Roy, Rajarshi and Raiman, Jonathan and Lee, Sang-gil and Ene, Teodor-Dumitru and Kirby, Robert and Kim, Sungwon and Kim, Jaehyeon and Catanzaro, Bryan},
  journal={arXiv preprint arXiv:2602.06053},
  year={2026}
}

\newpage
\beginappendix

\addcontentsline{toc}{section}{Appendices}
\section{Data Processing} \label{appendix:data_processing}

\subsection{Custom Segment-Any-Text model}
\label{app:SaT}
To leverage long-sequence data to train our sentence embedding system, we require robust tools capable of segmenting texts into sentences. While prior work has relied on the Segment-Any-Text (SaT) model \cite{minixhofer-etal-2023-wheres}, it covers only 85 languages and lacks support for code segmentation. To address these limitations and meet our data processing requirements, we develop a custom SaT model with broader language coverage.
Our objective is to build a sentence segmentation model that generalizes across all NLLB \citep{nllb} languages, as well as programming languages. We design a scalable pipeline that generates synthetic training data for these domains and fine-tune the original SaT model on this expanded dataset.
\subsubsection*{Synthetic Multilingual Data Generation}
We develop a two-stage pipeline that combines high-quality English sentence segmentation with machine translation to generate sentence-level annotations across all NLLB-supported languages.
\paragraph{Stage 1: English Segmentation.} We process a subset of the DCLM-EDU dataset using the SaT-12L-sm model\footnote{\url{https://huggingface.co/segment-any-text/sat-12l-sm}} to perform sentence segmentation. From this segmented corpus, we extract 10k consecutive sentences for each target language in our training set.
\paragraph{Stage 2: Multilingual Translation.} We translate the selected English sentences into each target language using the NLLB 3B translation model. The translated sentences are then concatenated to form synthetic multilingual documents that preserve sentence boundary information from the English source.
This pipeline enables us to generate sentence segmentation annotations for low-resource languages where manually annotated data is scarce or unavailable. The key insight is that translation preserves sentence boundaries, allowing us to transfer segmentation annotations across languages.
For code segmentation, we leverage the strategy described in the next section \ref{app:code_math_data_generation}, which naturally produces segmented scripts at our desired granularity. We sample 10k consecutive code snippets for each target programming language.

\subsubsection*{Model Training}
We fine-tune the SaT-12L-sm model on the collected synthetic data using the original training codebase. Training proceeds for 5k steps with a batch size of 128, dropout rate of 0.1, and label smoothing of 0.001.
\subsubsection*{Evaluation}
\begin{figure}[hbtp!]
    \centering
    \begin{subfigure}[b]{0.49\linewidth}
        \centering
        \includegraphics[width=\linewidth]{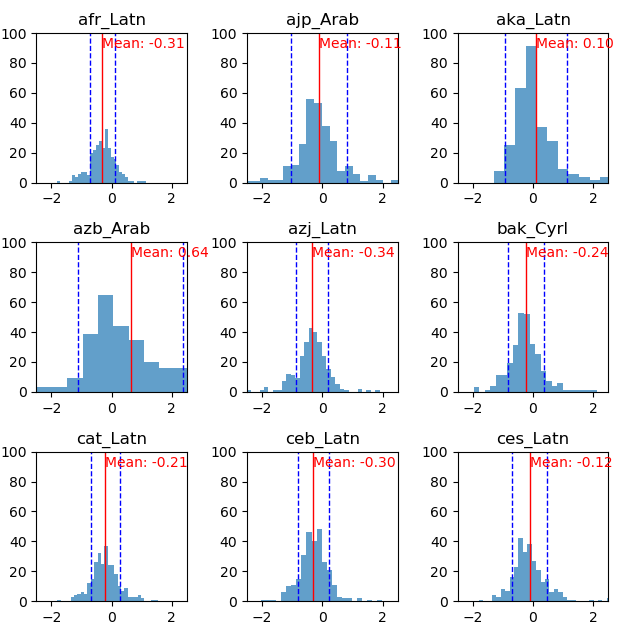}
        \caption{SaT-12L-sm}
        \label{fig:sat_baseline}
    \end{subfigure}    
    \begin{subfigure}[b]{0.49\linewidth}
        \centering
        \includegraphics[width=\linewidth]{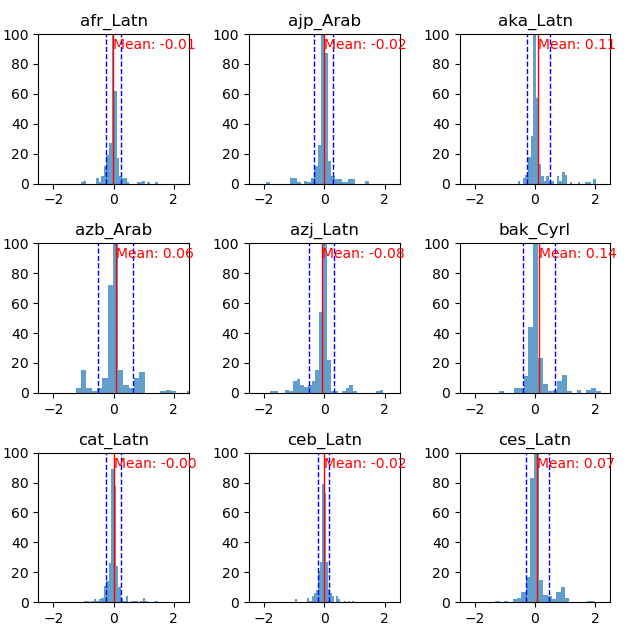}
        \caption{Our custom SaT model}
        \label{fig:sat_ours}
    \end{subfigure}
    \caption{Distribution of sentence count differences from the mean across 9 languages from FLORES paragraphs. Each subplot shows the histogram of (language sentence count - average sentence count) across all paragraphs. Red dashed lines indicate zero (perfect consistency) and the mean deviation for each language. \textbf{(a)} The baseline SaT-12L-sm model exhibits substantial deviation from zero, with means ranging from -0.34 to 0.64 and an overall standard deviation of 1.48 across all 201 languages, indicating inconsistent segmentation. \textbf{(b)} Our custom model produces distributions tightly centered around zero, with means ranging from -0.09 to 0.13 and an overall standard deviation of 0.34.}
    \label{fig:sat_comparison}
\end{figure}

We validate our custom SaT model through two extrinsic evaluation methodologies that assess segmentation quality without requiring manually annotated test data.
\paragraph{Round-trip Translation Quality.}
We hypothesize that superior sentence segmentation should improve long-form translation quality when documents are segmented before translation. To test this, we conduct a round-trip translation experiment on Belebele \citep{bandarkar-etal-2024-belebele} paragraphs across 116 languages. For each paragraph, we: (1) segment it using either our model or the baseline SaT-12L-sm, (2) translate each segment to English using the NLLB model, and (3) translate back to the source language. We then compute BLEU scores between the round-trip translation and the original text. Higher BLEU scores indicate better preservation of meaning, which correlates with segmentation quality. Our model achieves an average BLEU score of 32.84 compared to 32.48 for the baseline, with improvements observed in 85 out of 116 languages. This consistent improvement demonstrates that our enhanced segmentation better preserves semantic boundaries during translation.
\paragraph{Cross-lingual Segmentation Consistency.} Good sentence segmentation should produce similar numbers of sentences across languages for parallel text. We evaluate this property using FLORES paragraph-level data across 201 languages, where sentences were merged to create parallel paragraphs. For each paragraph, we compute the difference between the number of sentences produced by our model in each language and the average sentence count across all languages. We then measure the standard deviation of these differences across languages. Lower values indicate more consistent segmentation. Our model achieves a standard deviation of 0.34 compared to 1.48 for SaT-12L-sm. \Cref{fig:sat_baseline} and \Cref{fig:sat_ours} illustrate this improvement for a subset of 9 languages, showing that our model produces distributions tightly centered around zero (perfect consistency), while the baseline exhibits substantial deviation.

\subsection{Translation Data Sources}
\label{app:translation_data}

We utilize several massively multilingual data sources to construct our parallel training corpus covering 4,200 language varieties:

\begin{itemize}
    \item \textbf{Tatoeba} \citep{tatoeba}: Provides sentence-level translations across 500 languages with 25M examples spanning 24K directions, including many low-resource pairs (we retain the 15 most resourceful directions per source language).
    \item \textbf{PanLex} \citep{panlex}: A large-scale dictionary covering 6,800 languages with word- and phrase-level translations across over 100K directions (we use the 2 most resourceful directions per source language).
    \item \textbf{SMOL} \citep{smol,gatitos}: Contains manually translated words, sentences, and short documents from English into 200 low-resource languages.
    \item \textbf{Weblate}\footnote{\url{https://huggingface.co/datasets/ayymen/Weblate-Translations}}: Offers IT interface translations across hundreds of languages.
    \item \textbf{BPCC} \citep{bpcc}: Provides Indic language translations paired with English.
    \item \textbf{OLDI Seed} \citep{seed-23}: Designed to kickstart machine translation for under-resourced language directions.
    \item \textbf{Pontoon}\footnote{\url{https://huggingface.co/datasets/ayymen/Pontoon-Translations}}: Features English sentences from Mozilla projects translated into 200 languages.
    \item \textbf{The Bible dataset}: Provides aligned texts for over \sonarlanguages{} languages. These are filtered and aligned with five high-resource languages, yielding approximately 23M examples per target language (see \Cref{subsec:appendix_bible}).
    \item \textbf{JW}\footnote{\url{https://jw.org}}: Contains parallel data for 700 languages with 100M examples across 6K translation directions.
    \item \textbf{MS Terms}\footnote{\url{https://huggingface.co/datasets/microsoft/ms_terms}}: Offers IT terminology translated from English into 100 languages.
    \item \textbf{AmericasNLP} \citep{americasnlp}: Focuses on American indigenous languages paired primarily with Spanish.
    \item \textbf{KreyoMT} \citep{kreyol}: Provides parallel data for Creole languages.
    \item \textbf{AfroLingu-MT} \citep{afrolingumt}: Includes sentences translated between English, French, and 44 African languages.
\end{itemize}

\subsection{Code and Math Translation data generation}
\label{app:code_math_data_generation}
\subsubsection*{Code Snippet Segmentation}
To construct sentence-level code snippets suitable for embeddings, it is essential to define what constitutes a \textit{sentence} in the context of programming languages. Unlike natural language, where sentences are typically delimited by punctuation, code structure is governed by syntax and semantics, making naive approaches, such as splitting at line breaks, insufficient and potentially misaligned with real-world coding practices.

To address this, we adopt a syntax-aware segmentation strategy similar to \citet{gong2025}, leveraging Abstract Syntax Trees (ASTs) to identify meaningful breakpoints within code. This approach allows us to segment code in a way that respects its logical and syntactic boundaries, rather than relying on superficial heuristics. For our experiments, we use code from seven programming languages (Python, Java, JavaScript, Go, C, C++, and Ruby) sourced from publicly licensed GitHub repositories.

Our segmentation process begins by parsing source code into an AST using the Tree-sitter library\footnote{\url{https://github.com/tree-sitter/tree-sitter}}. We then traverse the tree in reverse Breadth-First Search (BFS) order, starting from the leaf nodes and progressing bottom-up. For each node, if it is a leaf with non-empty text and has not yet been visited, we initiate a snippet. We classify the snippet as either \textit{code} or \textit{text} based on the node type (e.g., comments and strings are labeled as \textit{text}).

To form coherent and contextually meaningful snippets, we recursively expand each snippet upward by merging the parent statement or declaration and its unvisited children, provided that the combined size does not exceed a maximum threshold of 100 non-whitespace characters. This ensures that each snippet remains concise and suitable for sentence-level representation. The process continues until all nodes have been visited, resulting in a comprehensive set of segmented code snippets.

The full segmentation procedure is detailed in \Cref{alg:code-segmentation}, which outlines the AST traversal, snippet formation, classification, and postprocessing steps. This method enables us to extract sentence-level code snippets that are both syntactically coherent and semantically meaningful, facilitating their integration into our modality-agnostic embedding space.

\begin{algorithm}
\caption{Code Segmentation via Abstract Syntax Tree Traversal}
\label{alg:code-segmentation}
\begin{algorithmic}[1]
\Require Source code, language parser, segmentation parameters (max size, depth, etc.)
\Ensure List of code segments (character ranges, types)

\State \textbf{Parse} source code $\rightarrow$ syntax tree (using \url{https://github.com/tree-sitter/tree-sitter})

\State \textbf{Initialize} empty list of snippets, visited node set

\For{each tree level (BFS order), processed in reverse order (bottom-up)}
    \For{each node at this level}
        \If{node is a leaf, has non-empty text, and is not visited}
            \State snippet $\gets$ \{node\}
            \State \textbf{Classify} snippet type:
                \If{node type is comment or string}
                    \State snippet\_type $\gets$ "text"
                \Else
                    \State snippet\_type $\gets$ "code"
                \EndIf
            \While{expansion upward is allowed (size and depth constraints not exceeded)}
                \If{parent node is a statement/declaration and adding it (and its children) keeps snippet size within allowed maximum}
                    \State snippet $\gets$ snippet $\cup$ parent node $\cup$ eligible siblings
                    \State \textbf{Update} snippet\_type if parent changes classification
                \Else
                    \State \textbf{break}
                \EndIf
            \EndWhile
            \State Mark included nodes as visited
            \State Add (snippet, snippet\_type) to output list
        \EndIf
    \EndFor
\EndFor

\State \textbf{Postprocess:}
    \State Merge adjacent snippets if their combined size is below the threshold and they are contiguous
    \State Adjust segment boundaries to snap to whitespace or newlines as configured

\For{each snippet}
    \State Compute snippet's character range in source code
\EndFor

\State \Return list of snippet ranges, snippet types

\end{algorithmic}
\end{algorithm}
\subsubsection*{Math expressions gathering}
To build a high-quality dataset of mathematical expressions, we extract LaTeX math content from large-scale scientific corpora such as FineMath~\citep{allal2025smollm2smolgoesbig} and arXiv. Our extraction process is designed to capture both inline and display math, reflecting the diversity of mathematical notation found in scientific writing. The expressions used can be found in \Cref{fig:regex-math-extraction}.
We use a comprehensive set of regular expressions to identify a wide range of LaTeX math environments. To ensure the quality and relevance of the extracted expressions, we apply the following filters:
\begin{itemize}
    \item Expressions between 20 and 150 characters.
    \item Expressions where more than 90\% of non-whitespace characters are alphabetic are discarded, except for in-line math.
\end{itemize}

The resulting dataset consists of unique LaTeX mathematical expressions, both in isolation and within their natural language in-line context, providing a rich resource for training and evaluating modality-agnostic sentence-level embeddings.

\begin{table}[hbtp!]
    \centering
    \begin{tabular}{|l|l|}
        \hline
        \textbf{Pattern} & \textbf{Description} \\
        \hline
        \verb|$(.*?)$| & Inline math (e.g., \verb|$a^2 + b^2 = c^2$|) \\
        \verb|$$(.*?)$$| & Display math with double dollar signs \\
        \verb|\\[(.*?)\\]| & Display math with \verb|\[ ... \]| \\
        \verb|\\begin{equation}(.*?)\\end{equation}| & Equation environment \\
        \verb|\\begin{align}(.*?)\\end{align}| & Align environment \\
        \verb|\\begin{align*}(.*?)\\end{align*}| & Align* environment \\
        \verb|\\begin{multline}(.*?)\\end{multline}| & Multline environment \\
        \verb|\\begin{multline*}(.*?)\\end{multline*}| & Multline* environment \\
        \verb|\\begin{gather}(.*?)\\end{gather}| & Gather environment \\
        \verb|\\begin{gather*}(.*?)\\end{gather*}| & Gather* environment \\
        \verb|\\begin{eqnarray}(.*?)\\end{eqnarray}| & Eqnarray environment \\
        \verb|\\begin{eqnarray*}(.*?)\\end{eqnarray*}| & Eqnarray* environment \\
        \hline
        \verb|(?<=[.!?])\s+| & Sentence splitting after \verb|.|, \verb|!|, or \verb|?| \\
        \verb|(?<!\$)\$[^$]+\$(?!\$)| & Short inline LaTeX expressions \\
        \hline
    \end{tabular}
    \caption{Summary of regular expressions used for extracting LaTeX math expressions and splitting sentences.}
    \label{fig:regex-math-extraction}
\end{table}

\begin{figure}[hbtp!]
    \centering
    \begin{tcolorbox}[colback=gray!10, colframe=black, title=\textbf{Math Text Translation}, fonttitle=\bfseries, boxrule=1pt, width=0.95\linewidth, arc=2mm]
    \small
    \textbf{System prompt:}\\
    \texttt{You are a helpful translation assistant. You respond only with the translation, without additional comments, context, or explanation.}

    \medskip
    \textbf{User prompt:}\\
    \texttt{Translate the following text from English into \{target\_lang\}. Don't produce any other output outside of the translation.}\\
    \texttt{\{example\}}
    \end{tcolorbox}
    
    \vspace{2em}

    \begin{tcolorbox}[colback=gray!10, colframe=black, title=\textbf{Code Snippet Translation}, fonttitle=\bfseries, boxrule=1pt, width=0.95\linewidth, arc=2mm]
    \small
    \texttt{Translate the following \{programming\_language\} snippet to a single sentence, ensuring that all elements and operations in the code are included. The sentence should convey the semantic meaning of the code, effectively translating it into a clear and concise lexical explanation without making any assumptions or inferences beyond what is explicitly stated in the code. Describe only and exactly its explicit elements and operations, without any additional context or explanation. Use a single, direct sentence that includes all elements and operations in the code, avoiding introductory words or additional context. Please provide only the sentence and nothing else:} \\
    \texttt{\{example\}}
    \end{tcolorbox}
    
    \vspace{2em}

    \begin{tcolorbox}[colback=gray!10, colframe=black, title=\textbf{Math Formula Translation}, fonttitle=\bfseries, boxrule=1pt, width=0.95\linewidth, arc=2mm]
    \small
    \texttt{Describe the following mathematical text in a single sentence. The sentence should convey the semantic meaning of the mathematical notation, effectively translating the mathematical notation into a clear and concise lexical explanation. Please provide only the sentence and nothing else.}\\
    \texttt{\{example\}}
    \end{tcolorbox}

    \caption{Three prompt templates for translation, code snippet semantic description, and mathematical notation explanation.}
    \label{fig:three-prompts}
\end{figure}

\subsubsection*{Natural Language Description Generation}

We leverage Llama-3.3-70B-Instruct to generate natural language descriptions for both code snippets and mathematical expressions. The model's extensive training on code and mathematical content enables effective paraphrasing of technical content into clear English descriptions. Importantly, this task involves paraphrasing existing content rather than generating new information.
The prompts used for this generation process are shown in \Cref{fig:three-prompts}.

\subsubsection*{Multilingual Back-translation}

To expand coverage of mixed-modality data, particularly sentences with inline expressions, we generate back-translations using Llama-3.3-70B-Instruct. We translate English descriptions and mixed-mode sentences into seven target languages: French, German, Hindi, Italian, Portuguese, Spanish, and Thai. This process creates a comprehensive multilingual dataset that enhances the diversity and utility of our training data while maintaining semantic consistency across languages.

\subsubsection*{Consistency Filtering}

To validate the quality of our synthetic code-to-text pairs, we implement a consistency check using the CodeRankEmbed embedding model \citep{sureshcornstack}. This process verifies that generated English descriptions accurately capture the semantics of their corresponding code snippets.

For each generated English description, we use it as a query to retrieve the most semantically similar code snippet from a pool of 100,000 candidates within the same programming language. If our synthetic data generation is effective, the English description should retrieve its original corresponding code snippet as the top match.

We find that in 99\% of cases, the English description successfully retrieves its original code snippet as the top-1 match. This high retrieval accuracy indicates strong semantic alignment between code snippets and their generated natural language descriptions, demonstrating the reliability and fidelity of our synthetic data generation approach.

\subsection{Hard negatives generation}
\label{app:data_processing/hard_negatives}

\begin{figure}[hbtp!]
\centering

\begin{tcolorbox}[
    colback=blue!10,
    colframe=blue!70!black,
    title={Python Example},
    fonttitle=\bfseries,
    width=\textwidth
]
\small
\textbf{Source:} \texttt{if input\_event["enable\_points"] or event\_info.get("bonus"):}

\vspace{0.1em}
\textbf{Target:} \textit{The script determines if the \texttt{"enable\_points"} in the \texttt{input\_event} dictionary is set to True, or if the \texttt{event\_info} dictionary includes a key called \texttt{"bonus"} with any associated value.}

\vspace{0.2em}
\textbf{Hard Negatives:}
\begin{enumerate}
    \item The script checks whether \texttt{'ExitSignal'} exists within the \texttt{'Actions'} list of the dictionary linked to the \texttt{'session'} key in the \texttt{'raw\_action\_logs'} dictionary, and if so, assigns the result of a numpy operation (after squeezing) to the variable \texttt{'exit\_signals\_processed'}.
    \item The script verifies if the \texttt{event\_label} is \texttt{'Position'} and the \texttt{'Anchored'} value in the entry dictionary is \texttt{True}, then updates \texttt{self.disable\_anchor} to \texttt{True}.
    \item The script checks if the \texttt{event\_category} property of the \texttt{trip\_event} object matches \texttt{TripEventCategory.POSITION}.
\end{enumerate}
\end{tcolorbox}

\vspace{0.3em}

\begin{tcolorbox}[
    colback=orange!10,
    colframe=orange!70!black,
    title={JavaScript Example},
    fonttitle=\bfseries,
    width=\textwidth
]
\small
\textbf{Source:} \texttt{const DataHandler = require('./lib/dataHandler.js'); let windowRef; const dataHandler = new DataHandler(\{\});}

\vspace{0.1em}
\textbf{Target:} \textit{A constant named \texttt{DataHandler} is initialized by importing a module from a file called \texttt{dataHandler.js} located in a folder named \texttt{lib}, then a variable \texttt{windowRef} is declared, and a constant \texttt{dataHandler} is created as a new instance of \texttt{DataHandler}, passing an empty object to its constructor.}

\vspace{0.2em}
\textbf{Hard Negatives:}
\begin{enumerate}
    \item The script imports a function or variable called \texttt{buildHandler} from a file at \texttt{'../lib/dataHandler'}.
    \item The script imports \texttt{HandlerData} from a file named \texttt{dataHandler} and uses it to instantiate a new \texttt{Vuex Handler}, assigning the result to a constant called \texttt{handler}.
    \item The script imports the \texttt{initializeHandler} function from the \texttt{'redux'} package and also imports the reducer from \texttt{'./handlerReducer'} to set up a handler.
\end{enumerate}
\end{tcolorbox}

\vspace{0.3em}

\begin{tcolorbox}[
    colback=green!10,
    colframe=green!70!black,
    title={Math Example},
    fonttitle=\bfseries,
    width=\textwidth
]
\small
\textbf{Source:} $G^n \to \mathcal{X}$

\vspace{0.1em}
\textbf{Target:} \textit{A function is defined that maps an element from a group $G$ raised to the power of $n$ to a set or space denoted as $\mathcal{X}$.}

\vspace{0.2em}
\textbf{Hard Negatives:}
\begin{enumerate}
    \item The function $\mathcal{G}_x$ maps every element of the set $E$ to a corresponding element in the set $\mathcal{S}$.
    \item There exists a mapping or function from the set or space $\mathcal{G}$ to the set or space $\mathcal{H}$.
    \item A function $G$, parameterized by $\lambda$, maps to or transforms into a space or set denoted as $X$, also parameterized by $\lambda$.
\end{enumerate}
\end{tcolorbox}

\caption{Examples of hard negatives generation across Python code (blue), JavaScript code (orange), and mathematical expressions (green).}
\label{fig:code_hard_negs_example}
\end{figure}
For hard negatives generation we follow two strategies:
\paragraph{Natural Language} For natural language (i.e. no code or math) translation, we generate hard negatives using Llama 3.3 70B Instruct. We follow an approach inspired by xsim++ negatives \cite{chen-etal-2023-xsim}, where they crafted hard-to-distinguish negative examples for translation pairs.  We use the prompt described in \Cref{fig:xsimpp_prompt} and generate up to 5 hard negatives per sample, examples are provided in \Cref{tab:appendix/data/hard_negs_examples}.
\paragraph{Code and math} Here we follow a more straightforward approach and mine hard negatives using the \sonar{} checkpoint trained in \Cref{subsection:model/contrastive_pretuning} before hard negatives are introduced. We mine the top 5 negatives over a pool of 200k candidates for each sample. An example is provided in \Cref{fig:code_hard_negs_example}.

\begin{table}[hbtp!]
    \centering
    \begin{tabular}{p{0.35\textwidth}|p{0.55\textwidth}}
        \toprule
        Original sentence & Generated hard negatives \\
        \midrule
        \multirow{5}{=}{Tonight will be long.} & Tonight will not be short \\
        & Yesterday may be long \\
        & Tomorrow will be short \\
        & Tonight may not be long \\
        & Last night was short \\
        \midrule
        \multirow{5}{=}{You got the archaeologist and the psychologist wrong.} &
            You did not get the archaeologist and the psychologist right \\
        & She got the anthropologist and the sociologist wrong \\
        & They got the archaeologist and the psychologist correct \\
        & We did not get the biologist and the physicist wrong \\
        & He got the economist and the politician right \\
        \bottomrule
    \end{tabular}
    \caption{Natural language hard negative examples}
    \label{tab:appendix/data/hard_negs_examples}
\end{table}

\begin{figure}[hbtp!]
\centering
\begin{tcolorbox}[colback=gray!10,colframe=black,title={Hard Negatives Generation},fonttitle=\bfseries]
\small
\textbf{You are a text transformation specialist. Generate \underline{ONLY} valid xsim++ transformations using these \underline{EXCLUSIVE} methods:}
\textbf{1. CAUSALITY ALTERATION:}
\begin{itemize}
    \item Add/remove negations (\emph{``did not'', ``was not''})
    \item Replace adjectives with antonyms (\emph{``good'' $\rightarrow$ ``bad''})
    \item Change modal verbs (\emph{``may'' $\rightarrow$ ``will''})
\end{itemize}
\textbf{2. ENTITY REPLACEMENT:}
\begin{itemize}
    \item Swap proper nouns (people, locations, organizations)
    \item Replace pronouns (\emph{he $\rightarrow$ she, they $\rightarrow$ we})
\end{itemize}
\textbf{3. NUMBER ALTERATION:}
\begin{itemize}
    \item Change quantities (\emph{5 $\rightarrow$ 12})
    \item Modify dates/times (\emph{2023 $\rightarrow$ 2019})
    \item Alter percentages (\emph{15\% $\rightarrow$ 22\%})
\end{itemize}
\textbf{Follow these patterns from training examples:}\\
\texttt{\{few-shot examples\}}\\
\textbf{Now transform THIS SPECIFIC INPUT SENTENCE using the above patterns. Output ONLY a Python list of 1-5 modified sentences in this exact format:}
\begin{verbatim}
[
"Transformed sentence 1",
"Transformed sentence 2",
...
]
\end{verbatim}
\textbf{Key requirements:}
\begin{enumerate}
    \item Create 1-5 unique modified sentences
    \item Maximize difference from original text
    \item Mix transformation types where possible
    \item Maintain grammatical correctness
    \item Do NOT generate paraphrases, or synonyms
    \item NEVER output empty strings
    \item Output ONLY a Python list of strings
    \item No explanations, headers, or additional text
\end{enumerate}
\textbf{Input sentence to transform:} \texttt{\{example\}}
\end{tcolorbox}
\caption{Prompt for generating xsim++ transformations with clear instructions and structure.}
\label{fig:xsimpp_prompt}
\end{figure}

\subsection{Language code correspondence}
\definecolor{cadmiumgreen}{rgb}{0.0, 0.42, 0.24}
\definecolor{carnelian}{rgb}{0.7, 0.11, 0.11}
\definecolor{gold}{rgb}{0.72, 0.53, 0.04}

\newcommand{\newfloreslanguage}[1]{#1}
\begin{table*}[h!]
    \centering
    \scriptsize
    \begin{tabular}{lllllccl}
    \toprule
    \bf Code & \bf Language & \bf Script & \bf Family & \bf Subgrouping
  & \bf Res.  \\
    \midrule 
\newfloreslanguage{ace\_Arab}	&	\bf	Acehnese 	&	Arabic	&	Austronesian	&	Malayo-Polynesian	&	 	Low	\\
\newfloreslanguage{ace\_Latn}	&	\bf	Acehnese 	&	Latin	&	Austronesian	&	Malayo-Polynesian	&	 	Low	\\
\newfloreslanguage{acm\_Arab}	&	\bf	Mesopotamian Arabic	&	Arabic	&	Afro-Asiatic	&	Semitic	&	 	Low	\\
\newfloreslanguage{acq\_Arab}	&	\bf	Ta'izzi-Adeni Arabic	&	Arabic	&	Afro-Asiatic	&	Semitic	&	 	Low	\\
\newfloreslanguage{aeb\_Arab}	&	\bf	Tunisian Arabic	&	Arabic	&	Afro-Asiatic	&	Semitic	&	 	Low	\\
afr\_Latn	&	\bf	Afrikaans	&	Latin	&	Indo-European	&	Germanic	&	 	High\\
\newfloreslanguage{ajp\_Arab}	&	\bf	South Levantine Arabic	&	Arabic	&	Afro-Asiatic	&	Semitic	&	 	Low	\\
\newfloreslanguage{aka\_Latn}	&	\bf	Akan	&	Latin	&	Atlantic-Congo	&	Kwa Volta-Congo	&	 	Low	\\
amh\_Ethi	&	\bf	Amharic	&	Ge'ez	&	Afro-Asiatic	&	Semitic	&	 	Low	\\
\newfloreslanguage{apc\_Arab}	&	\bf	North Levantine Arabic	&	Arabic	&	Afro-Asiatic	&	Semitic	&	 	Low	\\
arb\_Arab	&	\bf	Modern Standard Arabic	&	Arabic	&	Afro-Asiatic	&	Semitic	&	 	High	&		\\
\newfloreslanguage{arb\_Latn}	&	\bf	Modern Standard Arabic 	&	Latin	&	Afro-Asiatic	&	Semitic	&	 	Low	\\
\newfloreslanguage{ars\_Arab}	&	\bf	Najdi Arabic	&	Arabic	&	Afro-Asiatic	&	Semitic	&	 	Low	\\
\newfloreslanguage{ary\_Arab}	&	\bf	Moroccan Arabic	&	Arabic	&	Afro-Asiatic	&	Semitic	&	 	Low		\\
\newfloreslanguage{arz\_Arab}	&	\bf	Egyptian Arabic	&	Arabic	&	Afro-Asiatic	&	Semitic	&	 	Low		\\
asm\_Beng	&	\bf	Assamese	&	Bengali	&	Indo-European	&	Indo-Aryan	&	 	Low		\\
ast\_Latn	&	\bf	Asturian	&	Latin	&	Indo-European	&	Italic	&	 	Low		\\
\newfloreslanguage{awa\_Deva}	&	\bf	Awadhi	&	Devanagari	&	Indo-European	&	Indo-Aryan	&	 	Low		\\
\newfloreslanguage{ayr\_Latn}	&	\bf	Central Aymara	&	Latin	&	Aymaran	&	Central Southern Aymara	&	 	Low		\\
\newfloreslanguage{azb\_Arab}	&	\bf	South Azerbaijani	&	Arabic	&	Turkic	&	Common Turkic	&	 	Low		\\
azj\_Latn	&	\bf	North Azerbaijani	&	Latin	&	Turkic	&	Common Turkic	&	 	Low	\\
\newfloreslanguage{bak\_Cyrl}	&	\bf	Bashkir	&	Cyrillic	&	Turkic	&	Common Turkic	&	 	Low	\\
\newfloreslanguage{bam\_Latn}	&	\bf	Bambara	&	Latin	&	Mande	&	Western Mande	&	 	Low	\\
\newfloreslanguage{ban\_Latn}	&	\bf	Balinese	&	Latin	&	Austronesian	&	Malayo-Polynesian	&	 	Low			\\
bel\_Cyrl	&	\bf	Belarusian	&	Cyrillic	&	Indo-European	&	Balto-Slavic	&	 	Low		\\
\newfloreslanguage{bem\_Latn}	&	\bf	Bemba	&	Latin	&	Atlantic-Congo	&	Benue-Congo	&	 	Low		\\
ben\_Beng	&	\bf	Bengali	&	Bengali	&	Indo-European	&	Indo-Aryan	&	 	High	\\
\newfloreslanguage{bho\_Deva}	&	\bf	Bhojpuri	&	Devanagari	&	Indo-European	&	Indo-Aryan	&	 	Low		\\
\newfloreslanguage{bjn\_Arab}	&	\bf	Banjar 	&	Arabic	&	Austronesian	&	Malayo-Polynesian	&	 	Low	\\
\newfloreslanguage{bjn\_Latn}	&	\bf	Banjar 	&	Latin	&	Austronesian	&	Malayo-Polynesian	&	 	Low		\\
\newfloreslanguage{bod\_Tibt}	&	\bf	Standard Tibetan	&	Tibetan	&	Sino-Tibetan	&	Bodic	&	 	Low	\\
bos\_Latn	&	\bf	Bosnian	&	Latin	&	Indo-European	&	Balto-Slavic	&	 	High	\\
\newfloreslanguage{bug\_Latn}	&	\bf	Buginese	&	Latin	&	Austronesian	&	Malayo-Polynesian	&	 	Low		\\
bul\_Cyrl	&	\bf	Bulgarian	&	Cyrillic	&	Indo-European	&	Balto-Slavic	&	 	High	\\
cat\_Latn	&	\bf	Catalan	&	Latin	&	Indo-European	&	Italic	&	 	High	\\
ceb\_Latn	&	\bf	Cebuano	&	Latin	&	Austronesian	&	Malayo-Polynesian	&	 	Low	\\
ces\_Latn	&	\bf	Czech	&	Latin	&	Indo-European	&	Balto-Slavic	&	 	High	\\
\newfloreslanguage{cjk\_Latn}	&	\bf	Chokwe	&	Latin	&	Atlantic-Congo	&	Benue-Congo	&	 	Low		\\
ckb\_Arab	&	\bf	Central Kurdish	&	Arabic	&	Indo-European	&	Iranian	&	 	Low	\\
\newfloreslanguage{crh\_Latn}	&	\bf	Crimean Tatar	&	Latin	&	Turkic	&	Common Turkic	&	 	Low		\\
cym\_Latn	&	\bf	Welsh	&	Latin	&	Indo-European	&	Celtic	&	 	Low	\\
dan\_Latn	&	\bf	Danish	&	Latin	&	Indo-European	&	Germanic	&	 	High	\\
deu\_Latn	&	\bf	German	&	Latin	&	Indo-European	&	Germanic	&	 	High	\\
\newfloreslanguage{dik\_Latn}	&	\bf	Southwestern Dinka	&	Latin	&	Nilotic	&	Western Nilotic	&	 	Low	\\
\newfloreslanguage{dyu\_Latn}	&	\bf	Dyula	&	Latin	&	Mande	&	Western Mande	&	 	Low	\\
\newfloreslanguage{dzo\_Tibt}	&	\bf	Dzongkha	&	Tibetan	&	Sino-Tibetan	&	Bodic	&	 	Low	\\
ell\_Grek	&	\bf	Greek	&	Greek	&	Indo-European	&	Graeco-Phrygian	&	 	High		\\
eng\_Latn	&	\bf	English	&	Latin	&	Indo-European	&	Germanic	&	 	High	\\
\newfloreslanguage{epo\_Latn}	&	\bf	Esperanto	&	Latin	&	Constructed	&	Esperantic	&	 	Low		\\
est\_Latn	&	\bf	Estonian	&	Latin	&	Uralic	&	Finnic	&	 	High	\\
\newfloreslanguage{eus\_Latn}	&	\bf	Basque	&	Latin	&	Basque	&	--	&	 	High	\\
\newfloreslanguage{ewe\_Latn}	&	\bf	Ewe	&	Latin	&	Atlantic-Congo	&	Kwa Volta-Congo	&	 	Low	\\
\newfloreslanguage{fao\_Latn}	&	\bf	Faroese	&	Latin	&	Indo-European	&	Germanic	&	 	Low	\\
\newfloreslanguage{fij\_Latn}	&	\bf	Fijian	&	Latin	&	Austronesian	&	Malayo-Polynesian	&	 	Low	\\
fin\_Latn	&	\bf	Finnish	&	Latin	&	Uralic	&	Finnic	&	 	High\\
\newfloreslanguage{fon\_Latn}	&	\bf	Fon	&	Latin	&	Atlantic-Congo	&	Kwa Volta-Congo	&	 	Low	\\
fra\_Latn	&	\bf	French	&	Latin	&	Indo-European	&	Italic	&	 	High	\\
\newfloreslanguage{fur\_Latn}	&	\bf	Friulian	&	Latin	&	Indo-European	&	Italic	&	 	Low	\\
fuv\_Latn	&	\bf	Nigerian Fulfulde	&	Latin	&	Atlantic-Congo	&	North-Central Atlantic	&	 	Low	\\
\newfloreslanguage{gla\_Latn}	&	\bf	Scottish Gaelic	&	Latin	&	Indo-European	&	Celtic	&	 	Low	\\
gle\_Latn	&	\bf	Irish	&	Latin	&	Indo-European	&	Celtic	&	 	Low	\\
glg\_Latn	&	\bf	Galician	&	Latin	&	Indo-European	&	Italic	&	 	Low	\\
\newfloreslanguage{grn\_Latn}	&	\bf	Guarani	&	Latin	&	Tupian	&	Maweti-Guarani	&	 	Low	\\
guj\_Gujr	&	\bf	Gujarati	&	Gujarati	&	Indo-European	&	Indo-Aryan	&	 	Low	\\
\newfloreslanguage{hat\_Latn}	&	\bf	Haitian Creole	&	Latin	&	Indo-European	&	Italic	&	 	Low	\\
hau\_Latn	&	\bf	Hausa	&	Latin	&	Afro-Asiatic	&	Chadic	&	 	Low	\\
heb\_Hebr	&	\bf	Hebrew	&	Hebrew	&	Afro-Asiatic	&	Semitic	&	 	High	\\
hin\_Deva	&	\bf	Hindi	&	Devanagari	&	Indo-European	&	Indo-Aryan	&	 	High	\\
\newfloreslanguage{hne\_Deva}	&	\bf	Chhattisgarhi	&	Devanagari	&	Indo-European	&	Indo-Aryan	&	 	Low	\\
hrv\_Latn	&	\bf	Croatian	&	Latin	&	Indo-European	&	Balto-Slavic	&	 	High\\
hun\_Latn	&	\bf	Hungarian	&	Latin	&	Uralic	&	--	&	 	High	\\
hye\_Armn	&	\bf	Armenian	&	Armenian	&	Indo-European	&	Armenic	&	 	Low	\\
ibo\_Latn	&	\bf	Igbo	&	Latin	&	Atlantic-Congo	&	Benue-Congo	&	 	Low		\\
\newfloreslanguage{ilo\_Latn}	&	\bf	Ilocano	&	Latin	&	Austronesian	&	Malayo-Polynesian	&	 	Low	\\
ind\_Latn	&	\bf	Indonesian	&	Latin	&	Austronesian	&	Malayo-Polynesian	&	 	High	\\
isl\_Latn	&	\bf	Icelandic	&	Latin	&	Indo-European	&	Germanic	&	 	High\\
    \bottomrule
    \end{tabular}
\end{table*}

\begin{table*}[h!]
    \centering
    \scriptsize
    \begin{tabular}{lllllccl}
    \toprule
    \bf Code & \bf Language & \bf Script & \bf Family & \bf Subgrouping
  & \bf Res.   \\
    \midrule 
ita\_Latn	&	\bf	Italian	&	Latin	&	Indo-European	&	Italic	&	 	High	\\
jav\_Latn	&	\bf	Javanese	&	Latin	&	Austronesian	&	Malayo-Polynesian	&	 	Low	\\
jpn\_Jpan	&	\bf	Japanese	&	Japanese	&	Japonic	&	Japanesic	&	 	High	\\
\newfloreslanguage{kab\_Latn}	&	\bf	Kabyle	&	Latin	&	Afro-Asiatic	&	Berber	&	 	Low	\\
\newfloreslanguage{kac\_Latn}	&	\bf	Jingpho	&	Latin	&	Sino-Tibetan	&	Brahmaputran	&	 	Low	\\
kam\_Latn	&	\bf	Kamba	&	Latin	&	Atlantic-Congo	&	Benue-Congo	&	 	Low	\\
kan\_Knda	&	\bf	Kannada	&	Kannada	&	Dravidian	&	South Dravidian	&	 	Low	\\
\newfloreslanguage{kas\_Arab}	&	\bf	Kashmiri 	&	Arabic	&	Indo-European	&	Indo-Aryan	&	 	Low	\\
\newfloreslanguage{kas\_Deva}	&	\bf	Kashmiri 	&	Devanagari	&	Indo-European	&	Indo-Aryan	&	 	Low	\\
kat\_Geor	&	\bf	Georgian	&	Georgian	&	Kartvelian	&	Georgian-Zan	&	 	Low	\\
\newfloreslanguage{knc\_Arab}	&	\bf	Central Kanuri 	&	Arabic	&	Saharan	&	Western Saharan	&	 	Low		\\
\newfloreslanguage{knc\_Latn}	&	\bf	Central Kanuri 	&	Latin	&	Saharan	&	Western Saharan	&	 	Low		\\
kaz\_Cyrl	&	\bf	Kazakh	&	Cyrillic	&	Turkic	&	Common Turkic	&	 	High		\\
\newfloreslanguage{kbp\_Latn}	&	\bf	Kabiyè	&	Latin	&	Atlantic-Congo	&	North Volta-Congo	&	 	Low	\\
\newfloreslanguage{kea\_Latn}	&	\bf	Kabuverdianu	&	Latin	&	Indo-European	&	Italic	&	 	Low	\\
khm\_Khmr	&	\bf	Khmer	&	Khmer	&	Austroasiatic	&	Khmeric	&	 	Low	\\
\newfloreslanguage{kik\_Latn}	&	\bf	Kikuyu	&	Latin	&	Atlantic-Congo	&	Benue-Congo	&	 	Low		\\
\newfloreslanguage{kin\_Latn}	&	\bf	Kinyarwanda	&	Latin	&	Atlantic-Congo	&	Benue-Congo	&	 	Low	\\
kir\_Cyrl	&	\bf	Kyrgyz	&	Cyrillic	&	Turkic	&	Common Turkic	&	 	Low	\\
\newfloreslanguage{kmb\_Latn}	&	\bf	Kimbundu	&	Latin	&	Atlantic-Congo	&	Benue-Congo	&	 	Low	\\
\newfloreslanguage{kmr\_Latn}	&	\bf	Northern Kurdish	&	Latin	&	Indo-European	&	Iranian	&	 	Low	\\
\newfloreslanguage{kon\_Latn}	&	\bf	Kikongo	&	Latin	&	Atlantic-Congo	&	Benue-Congo	&	 	Low	\\
kor\_Hang	&	\bf	Korean	&	Hangul	&	Koreanic	&	Korean	&	 	High\\
lao\_Laoo	&	\bf	Lao	&	Lao	&	Tai-Kadai	&	Kam-Tai	&	 	Low	\\
\newfloreslanguage{lij\_Latn}	&	\bf	Ligurian	&	Latin	&	Indo-European	&	Italic	&	 	Low	\\
\newfloreslanguage{lim\_Latn}	&	\bf	Limburgish	&	Latin	&	Indo-European	&	Germanic	&	 	Low		\\
lin\_Latn	&	\bf	Lingala	&	Latin	&	Atlantic-Congo	&	Benue-Congo	&	 	Low		\\
lit\_Latn	&	\bf	Lithuanian	&	Latin	&	Indo-European	&	Balto-Slavic	&	 	High		\\
\newfloreslanguage{lmo\_Latn}	&	\bf	Lombard	&	Latin	&	Indo-European	&	Italic	&	 	Low	\\
\newfloreslanguage{ltg\_Latn}	&	\bf	Latgalian	&	Latin	&	Indo-European	&	Balto-Slavic	&	 	Low	\\
ltz\_Latn	&	\bf	Luxembourgish	&	Latin	&	Indo-European	&	Germanic	&	 	Low		\\
\newfloreslanguage{lua\_Latn}	&	\bf	Luba-Kasai	&	Latin	&	Atlantic-Congo	&	Benue-Congo	&	 	Low	\\
lug\_Latn	&	\bf	Ganda	&	Latin	&	Atlantic-Congo	&	Benue-Congo	&	 	Low	\\
luo\_Latn	&	\bf	Luo	&	Latin	&	Nilotic	&	Western Nilotic	&	 	Low		\\
\newfloreslanguage{lus\_Latn}	&	\bf	Mizo	&	Latin	&	Sino-Tibetan	&	Kuki-Chin-Naga	&	 	Low	\\
lvs\_Latn	&	\bf	Standard Latvian	&	Latin	&	Indo-European	&	Balto-Slavic	&	 	High	\\
\newfloreslanguage{mag\_Deva}	&	\bf	Magahi	&	Devanagari	&	Indo-European	&	Indo-Aryan	&	 	Low		\\
\newfloreslanguage{mai\_Deva}	&	\bf	Maithili	&	Devanagari	&	Indo-European	&	Indo-Aryan	&	 	Low		\\
mal\_Mlym	&	\bf	Malayalam	&	Malayalam	&	Dravidian	&	South Dravidian	&	 	Low		\\
mar\_Deva	&	\bf	Marathi	&	Devanagari	&	Indo-European	&	Indo-Aryan	&	 	Low	\\
\newfloreslanguage{min\_Arab}	&	\bf	Minangkabau 	&	Arabic	&	Austronesian	&	Malayo-Polynesian	&	 	Low		\\
\newfloreslanguage{min\_Latn}	&	\bf	Minangkabau 	&	Latin	&	Austronesian	&	Malayo-Polynesian	&	 	Low	\\
mkd\_Cyrl	&	\bf	Macedonian	&	Cyrillic	&	Indo-European	&	Balto-Slavic	&	 	High	\\
\newfloreslanguage{plt\_Latn}	&	\bf	Plateau Malagasy	&	Latin	&	Austronesian	&	Malayo-Polynesian	&	 	Low	\\
mlt\_Latn	&	\bf	Maltese	&	Latin	&	Afro-Asiatic	&	Semitic	&	 	High	\\
\newfloreslanguage{mni\_Beng}	&	\bf	Meitei 	&	Bengali	&	Sino-Tibetan	&	Kuki-Chin-Naga	&	 	Low	\\
khk\_Cyrl	&	\bf	Halh Mongolian	&	Cyrillic	&	Mongolic-Khitan	&	Mongolic	&	 	Low	\\
\newfloreslanguage{mos\_Latn}	&	\bf	Mossi	&	Latin	&	Atlantic-Congo	&	North Volta-Congo	&	 	Low	\\
mri\_Latn	&	\bf	Maori	&	Latin	&	Austronesian	&	Malayo-Polynesian	&	 	Low	\\
mya\_Mymr	&	\bf	Burmese	&	Myanmar	&	Sino-Tibetan	&	\tiny Burmo-Qiangic	&	 	Low		\\
nld\_Latn	&	\bf	Dutch	&	Latin	&	Indo-European	&	Germanic	&	 	High	\\
\newfloreslanguage{nno\_Latn}	&	\bf	Norwegian Nynorsk	&	Latin	&	Indo-European	&	Germanic	&	 	Low		\\
nob\_Latn	&	\bf	Norwegian Bokmål	&	Latin	&	Indo-European	&	Germanic	&	 	Low		\\
npi\_Deva	&	\bf	Nepali	&	Devanagari	&	Indo-European	&	Indo-Aryan	&	 	Low		\\
nso\_Latn	&	\bf	Northern Sotho	&	Latin	&	Atlantic-Congo	&	Benue-Congo	&	 	Low		\\
\newfloreslanguage{nus\_Latn}	&	\bf	Nuer	&	Latin	&	Nilotic	&	Western Nilotic	&	 	Low		\\
nya\_Latn	&	\bf	Nyanja	&	Latin	&	Atlantic-Congo	&	Benue-Congo	&	 	Low	\\
oci\_Latn	&	\bf	Occitan	&	Latin	&	Indo-European	&	Italic	&	 	Low		\\
\newfloreslanguage{gaz\_Latn}	&	\bf	West Central Oromo	&	Latin	&	Afro-Asiatic	&	Cushitic	&	 	Low		\\
ory\_Orya	&	\bf	Odia	&	Oriya	&	Indo-European	&	Indo-Aryan	&	 	Low	\\
\newfloreslanguage{pag\_Latn}	&	\bf	Pangasinan	&	Latin	&	Austronesian	&	Malayo-Polynesian	&	 	Low		\\
pan\_Guru	&	\bf	Eastern Panjabi	&	Gurmukhi	&	Indo-European	&	Indo-Aryan	&	 	Low	\\
\newfloreslanguage{pap\_Latn}	&	\bf	Papiamento	&	Latin	&	Indo-European	&	Italic	&	 	Low	\\
pes\_Arab	&	\bf	Western Persian	&	Arabic	&	Indo-European	&	Iranian	&	 	High	\\
pol\_Latn	&	\bf	Polish	&	Latin	&	Indo-European	&	Balto-Slavic	&	 	High	\\
por\_Latn	&	\bf	Portuguese	&	Latin	&	Indo-European	&	Italic	&	 	High	\\
\newfloreslanguage{prs\_Arab}	&	\bf	Dari	&	Arabic	&	Indo-European	&	Iranian	&	 	Low		\\
pbt\_Arab	&	\bf	Southern Pashto	&	Arabic	&	Indo-European	&	Iranian	&	 	Low		\\
\newfloreslanguage{quy\_Latn}	&	\bf	Ayacucho Quechua	&	Latin	&	Quechuan	&	Chinchay	&	 	Low		\\
ron\_Latn	&	\bf	Romanian	&	Latin	&	Indo-European	&	Italic	&	 	High	\\
\newfloreslanguage{run\_Latn}	&	\bf	Rundi	&	Latin	&	Atlantic-Congo	&	Benue-Congo	&	 	Low	\\
rus\_Cyrl	&	\bf	Russian	&	Cyrillic	&	Indo-European	&	Balto-Slavic	&	 	High	\\
\newfloreslanguage{sag\_Latn}	&	\bf	Sango	&	Latin	&	Atlantic-Congo	&	North Volta-Congo	&	 	Low		\\
\newfloreslanguage{san\_Deva}	&	\bf	Sanskrit	&	Devanagari	&	Indo-European	&	Indo-Aryan	&	 	Low		\\
\newfloreslanguage{sat\_Olck}	&	\bf	Santali	&	Ol Chiki	&	Austroasiatic	&	Mundaic	&	 	Low	\\
\newfloreslanguage{scn\_Latn}	&	\bf	Sicilian	&	Latin	&	Indo-European	&	Italic	&	 	Low	\\
\newfloreslanguage{shn\_Mymr}	&	\bf	Shan	&	Myanmar	&	Tai-Kadai	&	Kam-Tai	&	 	Low	\\
\newfloreslanguage{sin\_Sinh}	&	\bf	Sinhala	&	Sinhala	&	Indo-European	&	Indo-Aryan	&	 	Low	\\
slk\_Latn	&	\bf	Slovak	&	Latin	&	Indo-European	&	Balto-Slavic	&	 	High	\\
\newfloreslanguage{slv\_Latn}	&	\bf	Slovenian	&	Latin	&	Indo-European	&	Balto-Slavic	&	 	High	\\
    \bottomrule
    \end{tabular}
\end{table*}    
\begin{table*}[h!] 
    \centering
    \scriptsize
    \begin{tabular}{lllllccl}
    \toprule
    \bf Code & \bf Language & \bf Script & \bf Family & \bf Subgrouping
  & \bf Res.  \\
    \midrule 
\newfloreslanguage{smo\_Latn}	&	\bf	Samoan	&	Latin	&	Austronesian	&	Malayo-Polynesian	&	 	Low	\\
sna\_Latn	&	\bf	Shona	&	Latin	&	Atlantic-Congo	&	Benue-Congo	&	 	Low	\\
snd\_Arab	&	\bf	Sindhi	&	Arabic	&	Indo-European	&	Indo-Aryan	&	 	Low		\\
som\_Latn	&	\bf	Somali	&	Latin	&	Afro-Asiatic	&	Cushitic	&	 	Low		\\
\newfloreslanguage{sot\_Latn}	&	\bf	Southern Sotho	&	Latin	&	Atlantic-Congo	&	Benue-Congo	&	 	High	\\
spa\_Latn	&	\bf	Spanish	&	Latin	&	Indo-European	&	Italic	&	 	High	\\
\newfloreslanguage{als\_Latn}	&	\bf	Tosk Albanian	&	Latin	&	Indo-European	&	Albanian	&	 	High	\\
\newfloreslanguage{srd\_Latn}	&	\bf	Sardinian	&	Latin	&	Indo-European	&	Italic	&	 	Low		\\
srp\_Cyrl	&	\bf	Serbian	&	Cyrillic	&	Indo-European	&	Balto-Slavic	&	 	Low		\\
\newfloreslanguage{ssw\_Latn}	&	\bf	Swati	&	Latin	&	Atlantic-Congo	&	Benue-Congo	&	 	Low	\\
\newfloreslanguage{sun\_Latn}	&	\bf	Sundanese	&	Latin	&	Austronesian	&	Malayo-Polynesian	&	 	Low	\\
swe\_Latn	&	\bf	Swedish	&	Latin	&	Indo-European	&	Germanic	&	 	High\\
swh\_Latn	&	\bf	Swahili	&	Latin	&	Atlantic-Congo	&	Benue-Congo	&	 	High	\\
\newfloreslanguage{szl\_Latn}	&	\bf	Silesian	&	Latin	&	Indo-European	&	Balto-Slavic	&	 	Low	\\
tam\_Taml	&	\bf	Tamil	&	Tamil	&	Dravidian	&	South Dravidian	&	 	Low		\\
\newfloreslanguage{tat\_Cyrl}	&	\bf	Tatar	&	Cyrillic	&	Turkic	&	Common Turkic	&	 	Low	\\
tel\_Telu	&	\bf	Telugu	&	Telugu	&	Dravidian	&	South Dravidian	&	 	Low		\\
tgk\_Cyrl	&	\bf	Tajik	&	Cyrillic	&	Indo-European	&	Iranian	&	 	Low		\\
tgl\_Latn	&	\bf	Tagalog	&	Latin	&	Austronesian	&	Malayo-Polynesian	&	 	High	\\
tha\_Thai	&	\bf	Thai	&	Thai	&	Tai-Kadai	&	Kam-Tai	&	 	High	\\
\newfloreslanguage{tir\_Ethi}	&	\bf	Tigrinya	&	Ge'ez	&	Afro-Asiatic	&	Semitic	&	 	Low		\\
\newfloreslanguage{taq\_Latn}	&	\bf	Tamasheq 	&	Latin	&	Afro-Asiatic	&	Berber	&	 	Low	\\
\newfloreslanguage{taq\_Tfng}	&	\bf	Tamasheq 	&	Tifinagh	&	Afro-Asiatic	&	Berber	&	 	Low	\\
\newfloreslanguage{tpi\_Latn}	&	\bf	Tok Pisin	&	Latin	&	Indo-European	&	Germanic	&	 	Low	\\
\newfloreslanguage{tsn\_Latn}	&	\bf	Tswana	&	Latin	&	Atlantic-Congo	&	Benue-Congo	&	 	High	\\
\newfloreslanguage{tso\_Latn}	&	\bf	Tsonga	&	Latin	&	Atlantic-Congo	&	Benue-Congo	&	 	Low		\\
\newfloreslanguage{tuk\_Latn}	&	\bf	Turkmen	&	Latin	&	Turkic	&	Common Turkic	&	 	Low	\\
\newfloreslanguage{tum\_Latn}	&	\bf	Tumbuka	&	Latin	&	Atlantic-Congo	&	Benue-Congo	&	 	Low	\\
tur\_Latn	&	\bf	Turkish	&	Latin	&	Turkic	&	Common Turkic	&	 	High	\\
\newfloreslanguage{twi\_Latn}	&	\bf	Twi	&	Latin	&	Atlantic-Congo	&	Kwa Volta-Congo	&	 	Low		\\
\newfloreslanguage{tzm\_Tfng}	&	\bf	Central Atlas Tamazight	&	Tifinagh	&	Afro-Asiatic	&	Berber	&	 	Low		\\
\newfloreslanguage{uig\_Arab}	&	\bf	Uyghur	&	Arabic	&	Turkic	&	Common Turkic	&	 	Low		\\
ukr\_Cyrl	&	\bf	Ukrainian	&	Cyrillic	&	Indo-European	&	Balto-Slavic	&	 	High	\\
umb\_Latn	&	\bf	Umbundu	&	Latin	&	Atlantic-Congo	&	Benue-Congo	&	 	Low	\\
urd\_Arab	&	\bf	Urdu	&	Arabic	&	Indo-European	&	Indo-Aryan	&	 	Low	\\
uzn\_Latn	&	\bf	Northern Uzbek	&	Latin	&	Turkic	&	Common Turkic	&	 	High	\\
\newfloreslanguage{vec\_Latn}	&	\bf	Venetian	&	Latin	&	Indo-European	&	Italic	&	 	Low		\\
vie\_Latn	&	\bf	Vietnamese	&	Latin	&	Austroasiatic	&	Vietic	&	 	High	\\
\newfloreslanguage{war\_Latn}	&	\bf	Waray	&	Latin	&	Austronesian	&	Malayo-Polynesian	&	 	Low		\\
wol\_Latn	&	\bf	Wolof	&	Latin	&	Atlantic-Congo	&	North-Central Atlantic	&	 	Low	\\
xho\_Latn	&	\bf	Xhosa	&	Latin	&	Atlantic-Congo	&	Benue-Congo	&	 	High	\\
\newfloreslanguage{ydd\_Hebr}	&	\bf	Eastern Yiddish	&	Hebrew	&	Indo-European	&	Germanic	&	 	Low	\\
yor\_Latn	&	\bf	Yoruba	&	Latin	&	Atlantic-Congo	&	Benue-Congo	&	 	Low	\\
\newfloreslanguage{yue\_Hant}	&	\bf	Yue Chinese	&	Han (Traditional)	&	Sino-Tibetan	&	Sinitic	&	 	Low		\\
zho\_Hans	&	\bf	Chinese 	&	Han (Simplified)	&	Sino-Tibetan	&	Sinitic	&	 	High	\\
zho\_Hant	&	\bf	Chinese 	&	Han (Traditional)	&	Sino-Tibetan	&	Sinitic	&	 	High	\\
zsm\_Latn	&	\bf	Standard Malay	&	Latin	&	Austronesian	&	Malayo-Polynesian	&	 	High\\
zul\_Latn	&	\bf	Zulu	&	Latin	&	Atlantic-Congo	&	Benue-Congo	&	 	High	\\
    \bottomrule
    \end{tabular}
    \captionsetup{margin={0cm, 0cm}}
    \caption{
    \label{tab:all_languages}
    Correspondence table from No Language Left Behind project for \sonar{}-200 languages. This table was adapted from \citep{nllb}.
    }
\end{table*}

\subsection{Languages breakdown} \label{appendix:data/languages_breakdown}

\autoref{tab:all_languages_coverage} lists all the languages supported by \sonar{}-200 and common for all baseline models.  It also lists all the languages supported by \sonar{}-speech.

\begin{table*}[hbtp!]

\centering

\begin{tabular}{lllllll}

\toprule

\textbf{Languages} & & & & & & \\

\midrule

ace\_Arab & \underline{ace\_Latn} & \textbf{\underline{acm\_Arab}} & acq\_Arab & \textbf{\underline{aeb\_Arab}} & \textbf{\underline{afr\_Latn}} & ajp\_Arab \\

\underline{aka\_Latn} & \underline{als\_Latn} & \textbf{\underline{amh\_Ethi}} & \underline{apc\_Arab} & \underline{arb\_Arab} & \underline{ars\_Arab} & \textbf{\underline{ary\_Arab}} \\

\textbf{\underline{arz\_Arab}} & \textbf{\underline{asm\_Beng}} & \underline{ast\_Latn} & \underline{awa\_Deva} & \underline{ayr\_Latn} & \textbf{azb\_Arab} & \textbf{azj\_Latn} \\

\underline{bak\_Cyrl} & \underline{bam\_Latn} & \underline{ban\_Latn} & \textbf{\underline{bel\_Cyrl}} & \underline{bem\_Latn} & \textbf{\underline{ben\_Beng}} & \underline{bho\_Deva} \\

bjn\_Arab & \underline{bjn\_Latn} & \underline{bod\_Tibt} & \textbf{\underline{bos\_Latn}} & \underline{bug\_Latn} & \textbf{\underline{bul\_Cyrl}} & \textbf{\underline{cat\_Latn}} \\

\underline{ceb\_Latn} & \textbf{\underline{ces\_Latn}} & \underline{cjk\_Latn} & \textbf{\underline{ckb\_Arab}} & crh\_Latn & \textbf{\underline{cym\_Latn}} & \textbf{\underline{dan\_Latn}} \\

\textbf{\underline{deu\_Latn}} & \underline{dik\_Latn} & \underline{dyu\_Latn} & \underline{dzo\_Tibt} & \textbf{\underline{ell\_Grek}} & \textbf{\underline{eng\_Latn}} & \textbf{\underline{epo\_Latn}} \\

\textbf{est\_Latn} & \textbf{\underline{eus\_Latn}} & \underline{ewe\_Latn} & \underline{fao\_Latn} & \underline{fij\_Latn} & \textbf{\underline{fin\_Latn}} & \underline{fon\_Latn} \\

\textbf{\underline{fra\_Latn}} & fur\_Latn & \underline{fuv\_Latn} & gaz\_Latn & \textbf{gla\_Latn} & \textbf{\underline{gle\_Latn}} & \textbf{\underline{glg\_Latn}} \\

\underline{grn\_Latn} & \textbf{\underline{guj\_Gujr}} & \underline{hat\_Latn} & \textbf{\underline{hau\_Latn}} & \textbf{\underline{heb\_Hebr}} & \textbf{\underline{hin\_Deva}} & \underline{hne\_Deva} \\

\textbf{\underline{hrv\_Latn}} & \textbf{\underline{hun\_Latn}} & \textbf{\underline{hye\_Armn}} & \underline{ibo\_Latn} & \underline{ilo\_Latn} & \textbf{\underline{ind\_Latn}} & \textbf{\underline{isl\_Latn}} \\

\textbf{\underline{ita\_Latn}} & \textbf{\underline{jav\_Latn}} & \textbf{\underline{jpn\_Jpan}} & \underline{kab\_Latn} & \underline{kac\_Latn} & \underline{kam\_Latn} & \textbf{\underline{kan\_Knda}} \\

\underline{kas\_Arab} & kas\_Deva & \textbf{\underline{kat\_Geor}} & \textbf{\underline{kaz\_Cyrl}} & \underline{kbp\_Latn} & \underline{kea\_Latn} & \underline{khk\_Cyrl} \\

\textbf{\underline{khm\_Khmr}} & \underline{kik\_Latn} & \underline{kin\_Latn} & \textbf{\underline{kir\_Cyrl}} & kmb\_Latn & \underline{kmr\_Latn} & knc\_Arab \\

\underline{knc\_Latn} & kon\_Latn & \textbf{\underline{kor\_Hang}} & \textbf{\underline{lao\_Laoo}} & \underline{lij\_Latn} & lim\_Latn & \underline{lin\_Latn} \\

\underline{lit\_Latn} & lmo\_Latn & \underline{ltg\_Latn} & \underline{ltz\_Latn} & \underline{lua\_Latn} & \underline{lug\_Latn} & \underline{luo\_Latn} \\

\underline{lus\_Latn} & lvs\_Latn & \underline{mag\_Deva} & \underline{mai\_Deva} & \textbf{\underline{mal\_Mlym}} & \textbf{\underline{mar\_Deva}} & \underline{min\_Latn} \\

\textbf{\underline{mkd\_Cyrl}} & \underline{mlt\_Latn} & \underline{mni\_Beng} & \underline{mos\_Latn} & \underline{mri\_Latn} & \textbf{\underline{mya\_Mymr}} & \textbf{\underline{nld\_Latn}} \\

\textbf{\underline{nno\_Latn}} & \textbf{\underline{nob\_Latn}} & \textbf{\underline{npi\_Deva}} & \underline{nso\_Latn} & \underline{nus\_Latn} & \underline{nya\_Latn} & \underline{oci\_Latn} \\

\underline{ory\_Orya} & \underline{pag\_Latn} & \underline{pan\_Guru} & \underline{pap\_Latn} & \underline{pbt\_Arab} & pes\_Arab & \underline{plt\_Latn} \\

\textbf{\underline{pol\_Latn}} & \textbf{\underline{por\_Latn}} & prs\_Arab & \underline{quy\_Latn} & \textbf{\underline{ron\_Latn}} & \underline{run\_Latn} & \textbf{\underline{rus\_Cyrl}} \\

\underline{sag\_Latn} & \textbf{san\_Deva} & sat\_Beng & \underline{scn\_Latn} & \underline{shn\_Mymr} & \textbf{\underline{sin\_Sinh}} & \textbf{\underline{slk\_Latn}} \\

\textbf{\underline{slv\_Latn}} & \underline{smo\_Latn} & \underline{sna\_Latn} & \textbf{\underline{snd\_Arab}} & \textbf{\underline{som\_Latn}} & sot\_Latn & \textbf{\underline{spa\_Latn}} \\

\underline{srd\_Latn} & \textbf{\underline{srp\_Cyrl}} & ssw\_Latn & \textbf{\underline{sun\_Latn}} & \textbf{\underline{swe\_Latn}} & \textbf{\underline{swh\_Latn}} & szl\_Latn \\

\textbf{\underline{tam\_Taml}} & \underline{taq\_Latn} & taq\_Tfng & \underline{tat\_Cyrl} & \textbf{\underline{tel\_Telu}} & \underline{tgk\_Cyrl} & \underline{tgl\_Latn} \\

\textbf{\underline{tha\_Thai}} & \underline{tir\_Ethi} & \underline{tpi\_Latn} & \underline{tsn\_Latn} & \underline{tso\_Latn} & \underline{tuk\_Latn} & tum\_Latn \\

\textbf{\underline{tur\_Latn}} & \underline{twi\_Latn} & tzm\_Tfng & \textbf{\underline{uig\_Arab}} & \textbf{\underline{ukr\_Cyrl}} & \underline{umb\_Latn} & \textbf{\underline{urd\_Arab}} \\

\underline{uzn\_Latn} & vec\_Latn & \textbf{\underline{vie\_Latn}} & \underline{war\_Latn} & \underline{wol\_Latn} & \textbf{\underline{xho\_Latn}} & \underline{ydd\_Hebr} \\

\underline{yor\_Latn} & \underline{yue\_Hant} & zho\_Hans & \textbf{zho\_Hant} & \underline{zsm\_Latn} & \underline{zul\_Latn} & \\

\bottomrule

\end{tabular}

\caption{Complete list of languages covered by \sonar{}-200. Languages shown in \textbf{bold} are supported by all models in our comparison. \underline{Underlined} languages are supported by \sonar{}-speech 3B and 7B models. Our initial training covers 202 languages total, with 81 languages supported across all compared models.}

\label{tab:all_languages_coverage}

\end{table*}

\subsection{Data Statistics}
We present the statistics of our training data for both pre-training and fine-tuning stages in \autoref{tab:appendix/data/seq2seq_contrastive}.
\begin{table}[ht]
\centering
\begin{tabular}{l|rrrr|rrrr}
\toprule
& \multicolumn{4}{c|}{Seq2Seq} & \multicolumn{4}{c}{Contrastive} \\
\midrule
Dataset & pairs & dirs & source & target & pairs & dirs & source & target \\
\midrule
BT Math              & 445M  & 56   & 8   & 8   & --     & --  & --  & --  \\
Dictionary           & 153M  & 3.3K & 182 & 97 & --     & --  & --  & --  \\
Code/Math $\to$ Eng  & 1B  & 9    & 9   & 1   & 9.0M   & 9   & 9   & 1   \\
Eng $\to$ Code       & 941M   & 8    & 1   & 8   & --     & --  & --  & --  \\
Eng $\to$ Math       & 80M   & 1    & 1   & 1   & --     & --  & --  & --  \\
\midrule
Mined data           & 1.3B  & 5.1k & 186 & 186 & -  & - & - & -   \\
Backward Translation         & 1.1B   & 392  & 201 & 193 & -   & - & - & -   \\
Forward Translation         & 448M   &  192  & 192  & 1  & -   & -  & -  & -   \\
MT Primary data         & 5.7B   & 6.9k & 199 & 199 & 183M  & 196 & 196 & 1   \\
\bottomrule
\end{tabular}
\caption{Training Data Statistics for Seq2Seq and Contrastive Learning Approaches. 
Each dataset shows: \textit{pairs} (number of translation pairs), \textit{dirs} (number of translation directions, i.e., language X to Y), \textit{source} (number of source languages), and \textit{target} (number of target languages).}
\label{tab:appendix/data/seq2seq_contrastive}
\end{table}

\subsection{Bible}
\label{subsec:appendix_bible}

We construct the BIBLE benchmark by combining Ebible\footnote{\url{https://github.com/BibleNLP/ebible}}, which contains 833 languages and MMS-1000~\citep{mms}, which contains 1100+ languages. We align sentences by verse with high-resource target languages. To ensure that our data is of high-quality, we apply several stages of filtering based on LID, length-based signals, Machine Translation Quality Estimation (MTQE). The resulting corpus covers \sonarlanguages{} languages. The detailed process is describe below.

We assign the Gospel of John for evaluation which is the most translated part of the Bible: chapters 1--10 as \dev (479 verses) and chapters 11--21 as \test (400 verses). All remaining books are assigned to the training set. Most sources average around 7,000 verses, but there are exceptions with up to 40,000 verses (usually high-resource languages), since there is not always a common agreement of what is actually part of the Bible.

In our combined dataset, languages usually have different versions (sources) of the bible. Thus the dataset is initially organized by \texttt{LANG}/\texttt{SOURCE}/\texttt{VERSE}. To standardize the language codes to ISO 639-3 + script format (e.g., \texttt{fra\_Latn}) and identify mislabeled data, we apply language identification (LID) and script identification with  GlotLID~\citep{glotlid}, and label each source according to the most frequent predictions for its examples. 

Usually sources contain a significant amount of noise since the data are mostly automatically produced, may contain OCR artifacts, and verses might have been spitted inarticulately. 

We remove empty verses and normalize spaces, and apply monolingual and cross-lingual text-based filtering. (1) Monolingual: For each \texttt{LANG}/\texttt{VERSE} we compute the average length of the sources and remove that deviate significantly from it. (2) Cross-lingual: For \texttt{VERSE} we compute the lengths stats across languages and sources, and after normalizing by the expected length for each language (to account for languages like Mandarin Chinese) we exclude examples that deviate from the mean of their verse length by more than 3 standard deviations. 

To find the best source and remove very noisy ones in an automatic way, for each language we compute the \chrf score of each source with the rest of the sources. Then we compute the Quality Score of the source as the average of each \chrf scores. The motivation is that good quality sources will on average have high overall with the rest. We then normalize these scores by dividing with the best quality score for that language. We also compute a completeness according to the verses in the 3 splits. 
Then we pick the source with highest quality/completeness score as the best one, and furthermore completely remove sources with quality scores below 0.75.

For the \dev and \test splits, we only use the best source for each language, while for the \train split we use all sources.

Then we proceed to aligning all language with 5 high-resource target languages (English, French, Spanish, German, Russian) by verse, in order to create translation data. For those 5 target languages we want to have some extra verification of their quality, thus we apply the same idea as in the cross-lingual text based filtering, but now with MTQE. We use BLASER2~\cite{blaser2} and for each verse we compute pairwise QE scores, for all each available sources, and then pick the one with he highest average as the best translation for this verse.

\subsection{Details on omnilingual language groups}
We provide in \Cref{tab:group_stats} the language group statistics for resource-based and family-based grouping experiments presented in \autoref{subsec:curse_of_omnilinguality}.
\begin{table}[hbtp!]
    \centering
    \resizebox{\width}{!}{%
    \begin{tabular}{@{}lccccc@{}}
        \toprule
        & \multicolumn{2}{c}{\textbf{Langs}} & \multicolumn{2}{c}{\textbf{Examples}} & \\
        \cmidrule(lr){2-3} \cmidrule(lr){4-5} 
        \textbf{Group} & \textbf{Train} & \textbf{Dev} & \textbf{Total (M)} & \textbf{Median (K)} & \textbf{Note} \\
        \midrule
        \multicolumn{6}{l}{\textit{Resource-based grouping (\Cref{fig:group_heatmap_xsim,fig:group_heatmap_chrf})}} \\
        \midrule
        A & 211 & 140 & 2,764 & 541 & High-resource \\
        B & 300 & 246 & 93 & 104 & \\
        C & 300 & 269 & 17 & 55 & \\
        D & 300 & 275 & 11 & 37 & \\
        E & 300 & 299 & 11 & 35 & \\
        F & 300 & 270 & 10 & 34 & \\
        G & 705 & 61 & 6 & 5 & Low-resource \\
        \midrule
        \textbf{Total} & \textbf{2,416} & \textbf{1,560} & \textbf{2,912} & -- & \\
        \midrule
        \multicolumn{6}{l}{\textit{Family-based grouping (\Cref{fig:family_group_heatmap_xsim,fig:family_group_heatmap_chrf})}} \\
        \midrule
        A & 298 & 122 & 2,074 & 34 & Indo-European only \\
        B & 276 & 166 & 285 & 35 & 30 families \\
        C & 370 & 266 & 146 & 35 & 30 families \\
        D & 285 & 193 & 141 & 35 & 30 families \\
        E & 630 & 404 & 90 & 35 & 30 families \\
        F & 547 & 409 & 79 & 35 & 29 families \\
        \midrule
        \textbf{Total} & \textbf{2,406} & \textbf{1,560} & \textbf{2,816} & -- & \textbf{150 families} \\
        \bottomrule
    \end{tabular}
    }
    \caption{Language group statistics for resource-based (top) and family-based (bottom) grouping experiments. The resource-based grouping includes all languages with sufficient parallel data, sorted by resourcefulness. The family-based grouping assigns Indo-European to Group A and distributes other families equally across Groups B-F.}
    \label{tab:group_stats}
\end{table}

\newpage
\section{Experimental Configuration}
\label{app:experimental_config}

\begin{table}[h]
\centering
\begin{tabular}{l|ccccc}
\toprule
\textbf{Parameter} & \textbf{Large} & \textbf{Medium} & \textbf{Small} & \textbf{Tiny} & \textbf{xTiny} \\
\midrule
\texttt{vocab\_size}           & 256,232 & 128,232 & 64,232  & 32,232  & 16,232  \\
\texttt{hidden\_size}          & 2,048   & 1,792   & 1,536   & 1,152   & 512     \\
\texttt{intermediate\_size}    & 8,192   & 7,168   & 6,144   & 4,608   & 2,048   \\
\texttt{num\_hidden\_layers}   & 16      & 14      & 12      & 10      & 8       \\
\texttt{num\_attention\_heads} & 32      & 28      & 24      & 18      & 16      \\
\texttt{num\_key\_value\_heads}& 8       & 7       & 6       & 6       & 4       \\
\texttt{head\_dim}             & 64      & 64      & 64      & 64      & 32      \\
\texttt{embedding\_size}       & 1,024   & 1,024   & 1,024   & 1,024   & 1,024   \\
\bottomrule
\end{tabular}
\caption{Differences across the model sizes for \sonar{} as explained in \Cref{subsection:model/smaller_models_distillation}}
\label{tab:sonar2-encoder-configs}
\end{table}

\newpage
\section{Other Ablations and Analysis}
\label{appendix:sonar2_200_other_ablations}

\subsection{Pooling}
It is a common debate whether to use mean pooling or CLS pooling, with SONAR \citep{sonar} reporting better result with mean pooling, while MEXMA \citep{janeiro-etal-2025-mexma} reported better results with CLS pooling.
Intuitively, CLS pooling should work better, since it has the freedom to attend differently to each token.
In \Cref{tab:appendix/pooling} we experiment with both pooling methods and find that our model performs best with CLS pooling.
\begin{table}[h!]
    \centering
    \begin{tabular}{c|c|c}
        \toprule
         Model & xsim & xsim++  \\
         \midrule
         Mean & 0.68 & 9.25 \\
         CLS & 0.64 & 8.77 \\ %
         \bottomrule
    \end{tabular}
    \caption{Ablation on different pooling strategies, evaluated on FLORES200 dev set.}
    \label{tab:appendix/pooling}
\end{table}

\subsection{Model Representation Collapse}
An often overlooked aspect of learned representations is how much of the embedding space they actually utilize, that is, whether their representations are collapsed within the space. \citet{sonar} have already highlighted this issue, which is especially pronounced when training with MSE regression signals, as models may exploit collapse to minimize the loss. This issue can be significant in the deployment of embeddings in current production systems that leverage mixed precision to reduce the memory footprint, as collapse can largely affect performance at lower precision. In \Cref{tab:collapse} we see how \sonar{} successfully avoids collapse compared to other models like SONAR, with a healthy standard deviation on its features, similar to widely used models such as mE5\textsubscript{large}.
\begin{table}[h!]
    \centering
    \begin{tabular}{l|l|r}
        \toprule
        model & std & mean \\
        \midrule
        MEXMA & 0.0312 & -0.0011 \\
        mE5\textsubscript{large} & 0.0312 & -0.0008 \\
        LaBSE & 0.0358 & 0.0049 \\
        SONAR & 0.0074 & 0.0000 \\
        \sonar{} & 0.0356 & -0.0006 \\ %
        \bottomrule
    \end{tabular}
    \caption{Standard deviation (std) and mean of embedding features for different models when encoding FLORES200 dev set on all common languages.}
    \label{tab:collapse}
\end{table}

\subsection{Embedding Dimension Informativeness}
\begin{figure}[t!]
    \centering
    \includegraphics[width=0.5\linewidth]{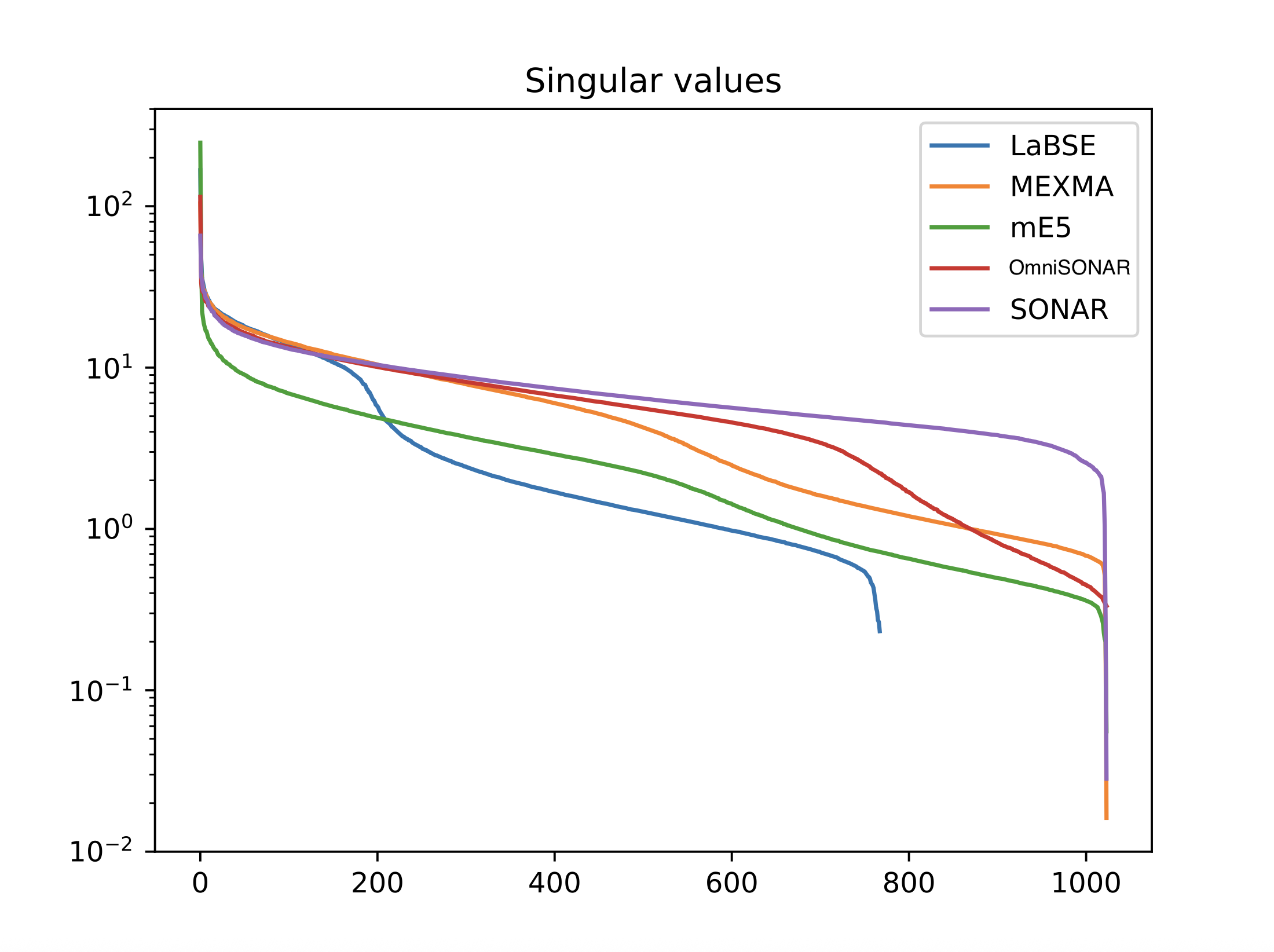}
    \caption{Singular values of embeddings from different models.}
    \label{fig:SVD}
\end{figure}

Singular Value Decomposition (SVD) provides a principled approach to analyze the intrinsic dimensionality and information distribution in embeddings. By examining the decay pattern of singular values, we can assess how different models utilize their feature space and identify potential dimensional collapse, where models concentrate information in fewer dimensions than their nominal embedding size. 

\Cref{fig:SVD} plots the SVD of our baselines on the FLORES dev set. From it, it can be inferred that \sonar{} showcases a stable decay pattern reaching up to 800 dimensions, while other models decay earlier, with the sole exception of SONAR.

\subsection{Analyzing examples to understand where models fail}

In this section, we perform some qualitative analysis of the errors of \sonar{}, and other models.
By inspection, \sonar{}'s mistakes look to be related to unit conversion, matching with the sentence where the actual number matches, i.e. matching "15 cm" to "15 inches" instead of "6 inches".
This is likely due to our hard negatives, which focused on matching the actual numbers, but lead to errors when the translation transforms the units.
Meanwhile we see SONAR and MEXMA make mistakes related to both values and semantics (may/will, white/black), such as the examples provided below.
Examples are provided in \Cref{tab:negatives_comparison}.

\begin{table}[ht]
\small
\setlength{\extrarowheight}{2pt}
\centering
\begin{tabularx}{\textwidth}{l p{4cm} p{4cm} p{4cm}}
\toprule
System & Source Sentence & Desired Retrieved & Actual Retrieved\\
\midrule
\sonar{} & O Corpo de Engenheiros dos EUA estimou que 15 cm de chuva podem romper os diques anteriormente danificados. 
& The U.S. Corps of Engineers estimated that \colorbox{green!15}{6} inches of rainfall could breach the previously damaged levees. 
& The U.S. Corps of Engineers estimated that \colorbox{red!15}{15} inches of rainfall could breach the previously damaged levees.\\

\sonar{} & Os limites de velocidade anunciados sao visivelmente mais baixos do que nas secoes anteriores e subsequentes - comumente 55-65 km/h - e a estrita obediencia a eles e ainda mais importante do que o contrario. 
& Posted speed limits are noticeably lower than in previous and subsequent sections  commonly \colorbox{green!15}{35-40} mph (56-64 km/h)  and strict obedience to them is even more important than otherwise. 
& Posted speed limits are noticeably lower than in previous and subsequent sections  commonly \colorbox{red!15}{35-90} mph (56-64 km/h)  and strict obedience to them is even more important than otherwise. \\
\midrule

MEXMA & O Corpo de Engenheiros dos EUA estimou que 15 cm de chuva podem romper os diques anteriormente danificados. 
& The U.S. Corps of Engineers estimated that \colorbox{green!15}{6} inches of rainfall could breach the previously damaged levees. 
& The U.S. Corps of Engineers estimated that \colorbox{red!15}{15} inches of rainfall could breach the previously damaged levees.  \\

MEXMA & Reportagens televisivas divulgam a fumaca esbranquicada saindo da planta. 
& Television reports show \colorbox{green!15}{white} smoke coming from the plant. 
& Television reports show \colorbox{red!15}{black} smoke coming from the plant.  \\
\midrule

SONAR & No periodo de um ano, uma pessoa infectada pode infectar entre 10 e 15 contatos proximos. 
& In one year's time, an infected person \colorbox{green!15}{may} infect 10 to 15 close contacts. 
& In one year's time, an infected person \colorbox{red!15}{will} infect 10 to 15 close contacts.  \\

SONAR & Aconteceu novamente no mesmo mes em Mashhad, outro aviao comercial entrou em uma pista e atingiu uma parede, matando dezessete pessoas. 
& The same month \colorbox{green!15}{saw} another airliner overrun a runway at Mashhad and strike a wall, killing seventeen. 
& The same month \colorbox{red!15}{did not saw} another airliner overrun a runway at Mashhad and strike a wall, killing seventeen.  \\
\bottomrule
\end{tabularx}
\caption{
    Comparison of three systems (\sonar{}, MEXMA, SONAR) on two examples each. 
    For each example, the table shows the original source sentence (in Portuguese) and the desired retrieved English sentence, the actual retrieved English sentence.
}
\label{tab:negatives_comparison}
\end{table}

\newpage
\newpage
\section{Prompts}

\Cref{tab:appendix/prompts} shows the prompts we used when tokenizing the input for both the Encoder and the Decoder, as explained in \Cref{subsection:model/seq2seq_pretraining}.
\begin{table}[ht!]
\centering
\begin{tabular}{l|p{10cm}}
\toprule
\textbf{Source} & \textbf{Prompt Template} \\
\midrule
\multicolumn{2}{c}{\textbf{Encoder}} \\
\midrule
Source/Anchor & \texttt{"<CLS><s> [LANGUAGE]:<SEP> [INPUT SENTENCE] </s>"} \\
\midrule
\multicolumn{2}{c}{\textbf{Decoder}} \\
\midrule
Primary data & \texttt{"<s> This is a possible translation in [LANGUAGE]:<SEP> [INPUT SENTENCE] </s>"} \\
\addlinespace[0.2em]
Mined data & \texttt{"<s> This is a possible mined translation in [LANGUAGE]:<SEP> [INPUT SENTENCE] </s>"} \\
\addlinespace[0.2em]
Back-translations & \texttt{"<s> This is a possible back-translation in [LANGUAGE]:<SEP> [INPUT SENTENCE] </s>"} \\
\addlinespace[0.2em]
Eng $\rightarrow$ Code & \texttt{"<s> This is a corresponding code snippet in [LANGUAGE]:<SEP> [INPUT SENTENCE] </s>"} \\
\addlinespace[0.2em]
Eng $\rightarrow$ Math & \texttt{"<s> This is a corresponding math formula:<SEP> [INPUT SENTENCE] </s>"} \\
\addlinespace[0.2em]
Code/Math $\rightarrow$ Eng & \texttt{"<s> This is a possible natural language explanation in [LANGUAGE]:<SEP> [INPUT SENTENCE] </s>"} \\
\bottomrule
\end{tabular}
\caption{Prompt Templates for Encoder and Decoder Components. The encoder uses classification prompts to identify language and content, while the decoder uses descriptive prompts tailored to different data sources and translation types. Placeholders \texttt{[LANGUAGE]} and \texttt{[INPUT SENTENCE]} are replaced with actual values during training.}
\label{tab:appendix/prompts}
\end{table}

\newpage
\section{Full Results}
\label{appendix:full_results}
We present a breakdown of the cross-lingual similarity search results for our 200 focus languages in \autoref{tab:xsim_all_languages} and \autoref{tab:xsim_pp_all_languages}.
We also present results broken down per family in \cref{tab:lang_families}.

\begin{landscape}
\begin{table}[h]
    \tiny
    \setlength{\tabcolsep}{1pt}
    \centering
    \begin{tabular}{l|c|c|c|c|c||l|c|c|c|c|c||l|c|c|c|c|c||l|c|c|c|c|c}
        \hline
        Lang & SONAR & LaBSE & MEXMA & \sonar{} & mE5 &
        Lang & SONAR & LaBSE & MEXMA & \sonar{} & mE5 &
        Lang & SONAR & LaBSE & MEXMA & \sonar{} & mE5 &
        Lang & SONAR & LaBSE & MEXMA & \sonar{} & mE5 \\
        \hline
ace\_Arab & 6.23 & 96.44 & 57.91 & 8.60 & 77.77 & ace\_Latn & 0.30 & 32.61 & 8.60 & 0.10 & 8.70 & acm\_Arab & 0.10 & 0.30 & 0.00 & 0.00 & 0.10 & acq\_Arab & 0.00 & 0.10 & 0.00 & 0.00 & 0.10 \\
aeb\_Arab & 0.40 & 4.94 & 0.20 & 0.20 & 0.99 & afr\_Latn & 0.00 & 0.00 & 0.00 & 0.00 & 0.00 & ajp\_Arab & 0.10 & 0.49 & 0.00 & 0.00 & 0.20 & aka\_Latn & 0.30 & 53.46 & 44.86 & 0.10 & 7.21 \\
als\_Latn & 0.00 & 0.00 & 0.00 & 0.00 & 0.00 & amh\_Ethi & 0.00 & 0.00 & 0.00 & 0.00 & 0.40 & apc\_Arab & 0.10 & 1.09 & 0.00 & 0.00 & 0.10 & arb\_Arab & 0.00 & 0.00 & 0.00 & 0.00 & 0.00 \\
ars\_Arab & 0.10 & 0.00 & 0.00 & 0.00 & 0.10 & ary\_Arab & 0.99 & 13.14 & 0.99 & 0.89 & 2.77 & arz\_Arab & 0.20 & 0.69 & 0.10 & 0.00 & 0.40 & asm\_Beng & 0.10 & 1.78 & 0.00 & 0.00 & 0.69 \\
ast\_Latn & 0.00 & 0.20 & 0.00 & 0.00 & 0.00 & awa\_Deva & 0.99 & 1.09 & 0.89 & 0.89 & 0.99 & ayr\_Latn & 3.85 & 72.13 & 54.25 & 1.68 & 43.97 & azb\_Arab & 2.96 & 44.37 & 1.68 & 0.69 & 11.36 \\
azj\_Latn & 0.30 & 0.30 & 0.20 & 0.20 & 0.20 & bak\_Cyrl & 0.00 & 41.90 & 11.36 & 0.00 & 1.68 & bam\_Latn & 4.05 & 65.61 & 52.37 & 2.17 & 14.62 & ban\_Latn & 0.40 & 8.40 & 1.09 & 0.30 & 2.57 \\
bel\_Cyrl & 0.49 & 0.00 & 0.00 & 0.00 & 0.20 & bem\_Latn & 0.00 & 44.07 & 36.66 & 0.10 & 14.23 & ben\_Beng & 0.00 & 0.00 & 0.00 & 0.00 & 0.10 & bho\_Deva & 0.20 & 2.57 & 0.30 & 0.00 & 0.79 \\
bjn\_Arab & 4.84 & 95.36 & 69.96 & 6.23 & 82.41 & bjn\_Latn & 0.10 & 8.60 & 0.30 & 0.10 & 1.68 & bod\_Tibt & 1.28 & 14.13 & 88.93 & 0.49 & 92.19 & bos\_Latn & 0.00 & 0.00 & 0.00 & 0.00 & 0.00 \\
bug\_Latn & 0.79 & 40.61 & 13.14 & 0.49 & 12.65 & bul\_Cyrl & 0.10 & 0.00 & 0.00 & 0.00 & 0.10 & cat\_Latn & 0.00 & 0.00 & 0.00 & 0.00 & 0.00 & ceb\_Latn & 0.00 & 0.00 & 6.82 & 0.00 & 0.10 \\
ces\_Latn & 0.00 & 0.00 & 0.00 & 0.00 & 0.00 & cjk\_Latn & 12.55 & 62.45 & 42.69 & 7.02 & 43.97 & ckb\_Arab & 0.10 & 88.83 & 0.10 & 0.00 & 3.26 & crh\_Latn & 0.10 & 2.67 & 0.10 & 0.00 & 0.30 \\
cym\_Latn & 0.00 & 0.00 & 0.00 & 0.00 & 0.49 & dan\_Latn & 0.00 & 0.00 & 0.00 & 0.00 & 0.00 & deu\_Latn & 0.00 & 0.00 & 0.00 & 0.00 & 0.00 & dik\_Latn & 11.26 & 62.25 & 46.15 & 8.89 & 46.34 \\
dyu\_Latn & 21.34 & 74.51 & 53.36 & 13.54 & 50.30 & dzo\_Tibt & 1.19 & 67.19 & 99.41 & 0.49 & 99.51 & ell\_Grek & 0.00 & 0.00 & 0.00 & 0.00 & 0.00 & epo\_Latn & 0.00 & 0.00 & 0.00 & 0.00 & 0.10 \\
est\_Latn & 0.00 & 0.00 & 0.00 & 0.00 & 0.10 & eus\_Latn & 0.00 & 0.10 & 0.00 & 0.00 & 0.00 & ewe\_Latn & 1.19 & 64.53 & 53.16 & 0.89 & 16.21 & fao\_Latn & 0.10 & 0.49 & 0.00 & 0.00 & 2.47 \\
fij\_Latn & 0.49 & 60.77 & 52.27 & 0.30 & 13.24 & fin\_Latn & 0.30 & 0.10 & 0.10 & 0.10 & 0.10 & fon\_Latn & 5.83 & 70.16 & 57.41 & 4.64 & 19.66 & fra\_Latn & 0.00 & 0.00 & 0.00 & 0.00 & 0.00 \\
fur\_Latn & 0.00 & 12.06 & 0.20 & 0.00 & 0.89 & fuv\_Latn & 10.97 & 65.02 & 43.38 & 4.55 & 36.86 & gaz\_Latn & 0.20 & 81.72 & 47.92 & 0.10 & 12.15 & gla\_Latn & 0.10 & 0.20 & 0.10 & 0.10 & 4.25 \\
gle\_Latn & 0.00 & 0.00 & 0.00 & 0.00 & 1.19 & glg\_Latn & 0.00 & 0.00 & 0.00 & 0.00 & 0.00 & grn\_Latn & 0.30 & 47.92 & 27.37 & 0.40 & 10.87 & guj\_Gujr & 0.00 & 0.00 & 0.00 & 0.00 & 0.00 \\
hat\_Latn & 0.59 & 0.59 & 13.83 & 0.59 & 1.28 & hau\_Latn & 0.40 & 0.30 & 0.30 & 0.30 & 2.67 & heb\_Hebr & 0.00 & 0.00 & 0.00 & 0.00 & 0.00 & hin\_Deva & 0.10 & 0.00 & 0.00 & 0.00 & 0.00 \\
hne\_Deva & 0.40 & 1.78 & 0.40 & 0.40 & 0.69 & hrv\_Latn & 0.00 & 0.00 & 0.00 & 0.00 & 0.00 & hun\_Latn & 0.10 & 0.00 & 0.00 & 0.00 & 0.00 & hye\_Armn & 0.00 & 0.00 & 0.00 & 0.00 & 0.00 \\
ibo\_Latn & 0.10 & 1.09 & 48.81 & 0.00 & 4.45 & ilo\_Latn & 0.00 & 30.24 & 16.30 & 0.00 & 1.68 & ind\_Latn & 0.00 & 0.00 & 0.00 & 0.00 & 0.30 & isl\_Latn & 0.20 & 0.10 & 0.10 & 0.10 & 0.10 \\
ita\_Latn & 0.10 & 0.00 & 0.00 & 0.00 & 0.00 & jav\_Latn & 0.00 & 0.00 & 0.00 & 0.00 & 0.00 & jpn\_Jpan & 0.20 & 0.00 & 0.10 & 0.00 & 0.00 & kab\_Latn & 0.10 & 82.41 & 67.19 & 0.00 & 37.35 \\
kac\_Latn & 1.78 & 67.98 & 51.09 & 0.10 & 41.40 & kam\_Latn & 3.36 & 54.45 & 38.74 & 2.17 & 29.25 & kan\_Knda & 0.00 & 0.00 & 0.00 & 0.00 & 0.30 & kas\_Arab & 0.20 & 34.88 & 3.06 & 0.20 & 4.84 \\
kas\_Deva & 1.88 & 56.72 & 15.91 & 0.59 & 16.60 & kat\_Geor & 0.40 & 0.00 & 0.00 & 0.00 & 0.10 & kaz\_Cyrl & 0.30 & 0.20 & 0.20 & 0.20 & 0.30 & kbp\_Latn & 4.94 & 67.79 & 55.34 & 4.35 & 39.33 \\
kea\_Latn & 0.00 & 14.82 & 1.09 & 0.00 & 0.79 & khk\_Cyrl & 0.30 & 0.00 & 0.10 & 0.00 & 0.59 & khm\_Khmr & 0.00 & 2.37 & 0.00 & 0.69 & 0.79 & kik\_Latn & 0.89 & 52.37 & 43.18 & 0.59 & 6.72 \\
kin\_Latn & 0.30 & 0.30 & 49.51 & 0.20 & 2.87 & kir\_Cyrl & 0.30 & 0.10 & 0.00 & 0.00 & 0.59 & kmb\_Latn & 0.89 & 61.66 & 48.02 & 1.28 & 36.76 & kmr\_Latn & 0.20 & 0.30 & 3.66 & 0.00 & 2.17 \\
knc\_Arab & 63.74 & 96.74 & 80.14 & 50.89 & 79.55 & knc\_Latn & 7.81 & 65.22 & 42.39 & 0.99 & 45.45 & kon\_Latn & 0.40 & 52.47 & 40.42 & 0.20 & 9.29 & kor\_Hang & 0.10 & 0.00 & 0.00 & 0.00 & 0.20 \\
lao\_Laoo & 0.00 & 3.46 & 0.00 & 0.00 & 0.79 & lij\_Latn & 0.10 & 10.57 & 0.59 & 0.10 & 1.38 & lim\_Latn & 0.20 & 9.09 & 0.30 & 0.00 & 3.56 & lin\_Latn & 0.20 & 50.69 & 40.71 & 0.20 & 3.85 \\
lit\_Latn & 0.49 & 0.40 & 0.49 & 0.40 & 0.40 & lmo\_Latn & 0.30 & 16.40 & 0.69 & 0.00 & 2.77 & ltg\_Latn & 0.10 & 25.20 & 12.65 & 0.10 & 5.34 & ltz\_Latn & 0.00 & 0.00 & 4.55 & 0.00 & 0.89 \\
lua\_Latn & 1.28 & 50.49 & 38.04 & 0.49 & 16.80 & lug\_Latn & 0.20 & 45.65 & 41.90 & 0.30 & 9.78 & luo\_Latn & 0.00 & 64.43 & 49.70 & 0.10 & 23.91 & lus\_Latn & 1.48 & 52.47 & 36.36 & 0.49 & 15.81 \\
lvs\_Latn & 0.20 & 0.00 & 0.00 & 0.00 & 0.00 & mag\_Deva & 0.10 & 0.30 & 0.00 & 0.10 & 0.00 & mai\_Deva & 0.00 & 0.20 & 0.10 & 0.00 & 0.10 & mal\_Mlym & 0.10 & 0.10 & 0.10 & 0.10 & 0.10 \\
mar\_Deva & 0.00 & 0.00 & 0.00 & 0.00 & 0.10 & min\_Latn & 0.10 & 12.85 & 0.89 & 0.10 & 1.98 & mkd\_Cyrl & 0.00 & 0.00 & 0.00 & 0.00 & 0.00 & mlt\_Latn & 0.00 & 0.00 & 15.71 & 0.00 & 0.79 \\
mni\_Beng & 0.00 & 90.02 & 72.13 & 0.30 & 46.84 & mos\_Latn & 10.67 & 70.36 & 51.19 & 5.73 & 45.16 & mri\_Latn & 0.10 & 2.47 & 57.91 & 0.00 & 11.56 & mya\_Mymr & 0.69 & 0.30 & 0.20 & 0.20 & 0.69 \\
nld\_Latn & 0.40 & 0.00 & 0.00 & 0.00 & 0.00 & nno\_Latn & 0.10 & 0.10 & 0.10 & 0.10 & 0.10 & nob\_Latn & 0.20 & 0.10 & 0.10 & 0.10 & 0.10 & npi\_Deva & 0.59 & 0.30 & 0.30 & 0.30 & 0.40 \\
nso\_Latn & 0.10 & 7.02 & 44.66 & 0.10 & 2.96 & nus\_Latn & 2.27 & 79.45 & 64.92 & 1.98 & 49.41 & nya\_Latn & 0.10 & 0.79 & 37.55 & 0.20 & 3.85 & oci\_Latn & 0.00 & 0.49 & 0.10 & 0.00 & 0.10 \\
ory\_Orya & 0.20 & 0.00 & 0.00 & 0.00 & 0.10 & pag\_Latn & 0.89 & 30.43 & 17.39 & 0.30 & 4.35 & pan\_Guru & 0.00 & 0.00 & 0.00 & 0.00 & 0.10 & pap\_Latn & 0.00 & 11.26 & 1.09 & 0.00 & 0.30 \\
pbt\_Arab & 0.10 & 1.09 & 0.10 & 0.00 & 0.49 & pes\_Arab & 0.20 & 0.00 & 0.00 & 0.00 & 0.00 & plt\_Latn & 0.00 & 0.49 & 15.32 & 0.00 & 1.19 & pol\_Latn & 0.00 & 0.00 & 0.00 & 0.00 & 0.20 \\
por\_Latn & 0.00 & 0.00 & 0.00 & 0.00 & 0.59 & prs\_Arab & 0.10 & 0.00 & 0.00 & 0.00 & 0.00 & quy\_Latn & 3.95 & 67.39 & 42.49 & 3.16 & 30.83 & ron\_Latn & 0.00 & 0.00 & 0.00 & 0.00 & 0.00 \\
run\_Latn & 0.20 & 2.87 & 48.32 & 0.10 & 3.85 & rus\_Cyrl & 0.20 & 0.00 & 0.00 & 0.00 & 0.00 & sag\_Latn & 3.16 & 60.97 & 43.28 & 1.78 & 33.89 & san\_Deva & 0.69 & 19.17 & 0.40 & 0.40 & 2.08 \\
scn\_Latn & 0.30 & 8.30 & 1.09 & 0.00 & 1.68 & shn\_Mymr & 0.49 & 71.54 & 53.06 & 0.00 & 42.19 & sin\_Sinh & 0.30 & 0.00 & 0.00 & 0.10 & 0.20 & slk\_Latn & 0.10 & 0.00 & 0.00 & 0.00 & 0.00 \\
slv\_Latn & 0.10 & 0.00 & 0.00 & 0.00 & 0.10 & smo\_Latn & 0.10 & 1.38 & 49.41 & 0.10 & 4.55 & sna\_Latn & 0.20 & 2.37 & 43.18 & 0.20 & 3.16 & snd\_Arab & 0.00 & 0.00 & 0.00 & 0.00 & 0.49 \\
som\_Latn & 0.10 & 1.09 & 0.10 & 0.10 & 4.64 & sot\_Latn & 0.00 & 0.59 & 46.94 & 0.00 & 1.78 & spa\_Latn & 0.10 & 0.10 & 0.10 & 0.10 & 0.10 & srd\_Latn & 0.00 & 9.09 & 0.49 & 0.00 & 0.79 \\
srp\_Cyrl & 0.00 & 0.00 & 0.00 & 0.00 & 0.00 & ssw\_Latn & 0.49 & 16.70 & 6.32 & 0.30 & 6.52 & sun\_Latn & 0.10 & 0.20 & 0.10 & 0.10 & 0.30 & swe\_Latn & 0.00 & 0.00 & 0.00 & 0.00 & 0.00 \\
swh\_Latn & 0.00 & 0.00 & 0.00 & 0.00 & 0.69 & szl\_Latn & 0.69 & 4.94 & 0.79 & 0.69 & 0.79 & tam\_Taml & 0.00 & 0.00 & 0.00 & 0.00 & 0.10 & taq\_Latn & 22.33 & 66.80 & 48.81 & 16.90 & 48.42 \\
taq\_Tfng & 21.34 & 95.55 & 86.07 & 25.79 & 87.55 & tat\_Cyrl & 0.00 & 0.00 & 5.83 & 0.00 & 0.59 & tel\_Telu & 0.20 & 0.00 & 0.00 & 0.00 & 0.00 & tgk\_Cyrl & 0.20 & 0.30 & 49.01 & 0.20 & 1.48 \\
tgl\_Latn & 0.00 & 0.00 & 0.30 & 0.00 & 0.10 & tha\_Thai & 0.10 & 6.62 & 0.10 & 0.10 & 0.40 & tir\_Ethi & 0.40 & 77.27 & 16.70 & 0.00 & 6.32 & tpi\_Latn & 0.00 & 46.84 & 17.39 & 0.00 & 3.36 \\
tsn\_Latn & 1.09 & 8.50 & 51.58 & 1.09 & 4.25 & tso\_Latn & 0.49 & 55.34 & 43.08 & 0.40 & 5.14 & tuk\_Latn & 0.10 & 0.69 & 5.04 & 0.00 & 19.47 & tum\_Latn & 0.79 & 24.21 & 41.40 & 0.20 & 6.13 \\
tur\_Latn & 0.00 & 0.00 & 0.00 & 0.00 & 0.00 & twi\_Latn & 0.40 & 49.60 & 42.49 & 0.10 & 8.00 & tzm\_Tfng & 0.79 & 95.55 & 89.43 & 0.99 & 90.42 & uig\_Arab & 0.40 & 0.20 & 0.10 & 0.10 & 2.47 \\
ukr\_Cyrl & 0.00 & 0.00 & 0.00 & 0.00 & 0.00 & umb\_Latn & 5.43 & 64.82 & 46.34 & 4.64 & 38.04 & urd\_Arab & 0.20 & 0.10 & 0.10 & 0.10 & 0.40 & uzn\_Latn & 0.10 & 0.10 & 0.10 & 0.10 & 0.20 \\
vec\_Latn & 0.00 & 4.15 & 0.10 & 0.00 & 0.59 & vie\_Latn & 0.00 & 0.00 & 0.00 & 0.00 & 0.10 & war\_Latn & 0.00 & 0.49 & 6.52 & 0.00 & 0.20 & wol\_Latn & 0.99 & 54.84 & 42.09 & 1.09 & 17.19 \\
xho\_Latn & 0.10 & 0.99 & 0.10 & 0.10 & 1.98 & ydd\_Hebr & 0.00 & 0.89 & 0.20 & 0.00 & 2.87 & yor\_Latn & 0.20 & 13.14 & 51.88 & 0.00 & 11.76 & yue\_Hant & 0.20 & 0.10 & 0.00 & 0.00 & 0.00 \\
zho\_Hans & 0.00 & 0.00 & 0.10 & 0.00 & 0.00 & zho\_Hant & 0.30 & 0.40 & 0.10 & 0.00 & 0.20 & zsm\_Latn & 0.00 & 0.00 & 0.00 & 0.00 & 0.10 & zul\_Latn & 0.10 & 0.20 & 0.30 & 0.10 & 1.38 \\
        \hline
    \end{tabular}
    \caption{xsim results for all models in all languages, x-eng in FLORES200 devtest set.}
    \label{tab:xsim_all_languages}
\end{table}
\end{landscape}

\begin{landscape}
\begin{table}[h]
    \tiny
    \setlength{\tabcolsep}{1pt}
    \centering
    \begin{tabular}{l|c|c|c|c|c||l|c|c|c|c|c||l|c|c|c|c|c||l|c|c|c|c|c}
        \hline
        Lang & SONAR & LaBSE & MEXMA & \sonar{} & mE5 &
        Lang & SONAR & LaBSE & MEXMA & \sonar{} & mE5 &
        Lang & SONAR & LaBSE & MEXMA & \sonar{} & mE5 &
        Lang & SONAR & LaBSE & MEXMA & \sonar{} & mE5 \\
        \hline
ace\_Arab & 55.34 & 100.00 & 92.09 & 36.76 & 98.91 & ace\_Latn & 16.30 & 82.21 & 49.41 & 8.10 & 48.22 & acm\_Arab & 11.17 & 52.67 & 9.39 & 5.83 & 27.37 & acq\_Arab & 8.30 & 46.54 & 8.20 & 10.87 & 23.02 \\
aeb\_Arab & 11.76 & 66.60 & 13.64 & 6.72 & 31.72 & afr\_Latn & 5.14 & 9.49 & 4.64 & 1.19 & 13.34 & ajp\_Arab & 7.61 & 54.15 & 9.19 & 5.24 & 23.81 & aka\_Latn & 19.47 & 92.39 & 85.18 & 10.77 & 42.79 \\
als\_Latn & 5.14 & 10.97 & 9.39 & 1.78 & 18.48 & amh\_Ethi & 8.60 & 28.85 & 6.23 & 3.36 & 43.48 & apc\_Arab & 9.58 & 58.00 & 12.65 & 5.14 & 28.46 & arb\_Arab & 6.82 & 35.87 & 6.72 & 2.37 & 19.27 \\
ars\_Arab & 13.93 & 39.72 & 12.65 & 8.50 & 23.52 & ary\_Arab & 14.03 & 77.08 & 22.23 & 12.45 & 39.03 & arz\_Arab & 10.87 & 58.99 & 11.46 & 4.84 & 24.51 & asm\_Beng & 14.92 & 62.15 & 11.17 & 6.42 & 41.50 \\
ast\_Latn & 9.68 & 38.04 & 14.13 & 6.82 & 19.86 & awa\_Deva & 11.07 & 26.98 & 12.06 & 3.75 & 27.67 & ayr\_Latn & 34.78 & 97.43 & 83.10 & 19.27 & 85.77 & azb\_Arab & 42.59 & 94.76 & 32.61 & 21.25 & 65.32 \\
azj\_Latn & 14.03 & 17.98 & 10.77 & 6.13 & 29.45 & bak\_Cyrl & 11.46 & 91.90 & 51.09 & 3.56 & 40.71 & bam\_Latn & 30.14 & 95.26 & 85.67 & 17.09 & 61.56 & ban\_Latn & 13.04 & 60.38 & 29.15 & 6.13 & 36.26 \\
bel\_Cyrl & 17.19 & 26.88 & 11.36 & 5.53 & 24.90 & bem\_Latn & 14.82 & 87.35 & 73.02 & 6.92 & 54.05 & ben\_Beng & 10.77 & 19.76 & 6.92 & 5.34 & 26.09 & bho\_Deva & 12.45 & 45.95 & 20.45 & 4.94 & 36.86 \\
bjn\_Arab & 42.69 & 100.00 & 95.06 & 31.13 & 98.62 & bjn\_Latn & 11.76 & 58.30 & 20.06 & 4.94 & 33.20 & bod\_Tibt & 25.99 & 82.51 & 97.23 & 17.89 & 99.21 & bos\_Latn & 5.93 & 9.29 & 3.56 & 1.19 & 11.66 \\
bug\_Latn & 24.11 & 85.57 & 52.27 & 12.25 & 56.62 & bul\_Cyrl & 7.81 & 9.88 & 4.74 & 2.08 & 9.68 & cat\_Latn & 4.84 & 12.94 & 3.95 & 1.88 & 8.10 & ceb\_Latn & 9.29 & 16.60 & 48.42 & 3.16 & 26.19 \\
ces\_Latn & 7.02 & 15.32 & 5.04 & 1.98 & 9.88 & cjk\_Latn & 63.04 & 95.65 & 83.50 & 31.72 & 86.17 & ckb\_Arab & 10.97 & 99.31 & 11.36 & 4.64 & 53.75 & crh\_Latn & 9.19 & 50.59 & 21.64 & 3.95 & 37.94 \\
cym\_Latn & 5.34 & 12.94 & 3.85 & 0.99 & 31.03 & dan\_Latn & 4.84 & 7.21 & 3.75 & 1.09 & 9.78 & deu\_Latn & 4.84 & 7.61 & 4.64 & 1.58 & 7.41 & dik\_Latn & 46.94 & 94.66 & 79.05 & 34.19 & 79.74 \\
dyu\_Latn & 65.51 & 97.83 & 83.79 & 41.21 & 86.66 & dzo\_Tibt & 24.31 & 98.42 & 99.80 & 15.71 & 99.90 & ell\_Grek & 9.19 & 18.77 & 7.41 & 2.87 & 13.14 & epo\_Latn & 4.55 & 8.79 & 4.15 & 1.38 & 18.77 \\
est\_Latn & 6.82 & 11.56 & 4.05 & 2.08 & 12.45 & eus\_Latn & 9.88 & 14.13 & 7.11 & 2.96 & 20.45 & ewe\_Latn & 22.63 & 96.34 & 83.40 & 13.83 & 59.98 & fao\_Latn & 11.36 & 38.14 & 22.33 & 4.05 & 39.82 \\
fij\_Latn & 16.01 & 94.66 & 84.39 & 8.30 & 53.95 & fin\_Latn & 7.51 & 15.32 & 7.02 & 3.36 & 11.96 & fon\_Latn & 35.08 & 96.05 & 87.15 & 26.38 & 62.85 & fra\_Latn & 4.84 & 9.19 & 4.64 & 1.78 & 7.81 \\
fur\_Latn & 5.83 & 71.05 & 25.59 & 4.45 & 34.09 & fuv\_Latn & 49.51 & 96.15 & 81.62 & 27.57 & 76.38 & gaz\_Latn & 16.30 & 98.72 & 83.10 & 8.70 & 60.47 & gla\_Latn & 13.74 & 27.67 & 10.57 & 3.66 & 48.22 \\
gle\_Latn & 8.70 & 17.49 & 8.70 & 3.26 & 39.03 & glg\_Latn & 6.13 & 7.71 & 4.55 & 2.37 & 11.96 & grn\_Latn & 18.87 & 91.50 & 68.08 & 9.58 & 55.83 & guj\_Gujr & 8.50 & 15.02 & 6.62 & 3.06 & 31.72 \\
hat\_Latn & 8.79 & 26.28 & 62.85 & 4.55 & 39.92 & hau\_Latn & 11.26 & 28.16 & 11.36 & 5.14 & 37.94 & heb\_Hebr & 5.43 & 17.00 & 6.52 & 2.77 & 18.68 & hin\_Deva & 7.51 & 10.87 & 5.24 & 2.57 & 17.00 \\
hne\_Deva & 9.58 & 39.92 & 16.21 & 4.15 & 31.52 & hrv\_Latn & 7.02 & 9.88 & 4.64 & 2.96 & 13.24 & hun\_Latn & 7.02 & 13.34 & 6.32 & 2.77 & 11.07 & hye\_Armn & 6.32 & 11.86 & 6.92 & 2.67 & 32.51 \\
ibo\_Latn & 12.06 & 45.95 & 79.84 & 6.52 & 43.28 & ilo\_Latn & 10.18 & 82.81 & 55.93 & 4.64 & 35.28 & ind\_Latn & 6.23 & 8.00 & 4.74 & 2.77 & 14.92 & isl\_Latn & 8.50 & 14.43 & 6.72 & 3.46 & 19.86 \\
ita\_Latn & 6.72 & 12.15 & 4.64 & 2.27 & 8.30 & jav\_Latn & 10.77 & 19.07 & 7.81 & 3.85 & 23.91 & jpn\_Jpan & 13.44 & 20.85 & 8.30 & 3.46 & 14.53 & kab\_Latn & 22.23 & 98.81 & 93.08 & 17.00 & 86.46 \\
kac\_Latn & 27.27 & 97.33 & 85.38 & 17.59 & 82.71 & kam\_Latn & 34.98 & 92.98 & 74.90 & 22.92 & 72.73 & kan\_Knda & 11.17 & 20.16 & 8.60 & 4.45 & 29.55 & kas\_Arab & 16.90 & 90.61 & 45.45 & 9.88 & 52.17 \\
kas\_Deva & 34.98 & 94.37 & 62.65 & 22.92 & 71.44 & kat\_Geor & 12.94 & 24.11 & 8.89 & 4.74 & 32.51 & kaz\_Cyrl & 10.77 & 13.83 & 7.41 & 3.95 & 29.55 & kbp\_Latn & 29.15 & 95.65 & 89.53 & 18.18 & 83.50 \\
kea\_Latn & 20.65 & 75.89 & 32.61 & 4.35 & 33.50 & khk\_Cyrl & 13.34 & 24.01 & 17.29 & 5.83 & 38.54 & khm\_Khmr & 11.86 & 24.11 & 8.00 & 8.70 & 45.45 & kik\_Latn & 22.83 & 92.09 & 78.75 & 11.07 & 48.02 \\
kin\_Latn & 9.49 & 29.64 & 82.41 & 4.35 & 36.36 & kir\_Cyrl & 13.83 & 27.96 & 11.46 & 6.42 & 33.10 & kmb\_Latn & 27.77 & 95.16 & 80.53 & 20.65 & 78.36 & kmr\_Latn & 15.32 & 36.46 & 35.57 & 7.61 & 45.65 \\
knc\_Arab & 89.82 & 100.00 & 95.55 & 77.57 & 97.33 & knc\_Latn & 47.04 & 96.25 & 76.38 & 19.96 & 80.14 & kon\_Latn & 17.69 & 93.58 & 76.78 & 10.57 & 50.20 & kor\_Hang & 10.57 & 21.64 & 7.41 & 3.95 & 17.98 \\
lao\_Laoo & 9.39 & 23.42 & 6.03 & 3.16 & 40.32 & lij\_Latn & 8.79 & 68.38 & 24.60 & 3.66 & 30.73 & lim\_Latn & 12.35 & 61.46 & 25.30 & 5.14 & 47.23 & lin\_Latn & 10.18 & 91.21 & 76.68 & 5.53 & 41.70 \\
lit\_Latn & 10.18 & 14.43 & 17.09 & 4.25 & 15.12 & lmo\_Latn & 17.59 & 75.40 & 30.34 & 8.79 & 37.75 & ltg\_Latn & 9.09 & 83.10 & 55.14 & 5.63 & 52.47 & ltz\_Latn & 8.40 & 20.95 & 39.72 & 3.06 & 35.18 \\
lua\_Latn & 32.51 & 91.60 & 75.40 & 16.50 & 59.78 & lug\_Latn & 19.86 & 89.62 & 78.46 & 13.04 & 56.32 & luo\_Latn & 12.65 & 95.75 & 83.30 & 7.31 & 65.81 & lus\_Latn & 24.41 & 91.60 & 70.95 & 12.25 & 56.82 \\
lvs\_Latn & 7.71 & 9.98 & 11.26 & 2.37 & 14.13 & mag\_Deva & 8.70 & 32.51 & 16.60 & 4.25 & 28.66 & mai\_Deva & 10.38 & 38.83 & 17.79 & 2.37 & 31.82 & mal\_Mlym & 10.57 & 25.69 & 8.50 & 4.74 & 26.38 \\
mar\_Deva & 9.29 & 18.87 & 6.42 & 3.66 & 26.19 & min\_Latn & 9.49 & 64.53 & 23.52 & 3.85 & 36.46 & mkd\_Cyrl & 6.62 & 9.98 & 5.14 & 2.27 & 11.86 & mlt\_Latn & 5.04 & 8.89 & 60.08 & 1.68 & 27.77 \\
mni\_Beng & 19.07 & 99.80 & 95.36 & 12.65 & 91.50 & mos\_Latn & 41.60 & 96.64 & 83.30 & 26.38 & 86.07 & mri\_Latn & 13.54 & 46.94 & 84.29 & 8.79 & 57.11 & mya\_Mymr & 17.69 & 41.70 & 12.06 & 6.42 & 47.43 \\
nld\_Latn & 10.87 & 12.55 & 7.11 & 3.56 & 9.78 & nno\_Latn & 14.03 & 11.56 & 6.72 & 3.06 & 12.85 & nob\_Latn & 11.76 & 10.18 & 5.73 & 2.67 & 8.79 & npi\_Deva & 11.36 & 14.33 & 5.53 & 3.16 & 29.74 \\
nso\_Latn & 9.88 & 59.98 & 75.99 & 4.64 & 36.26 & nus\_Latn & 29.15 & 98.62 & 92.79 & 20.06 & 87.85 & nya\_Latn & 13.34 & 43.58 & 71.64 & 7.91 & 36.17 & oci\_Latn & 5.53 & 36.46 & 15.51 & 2.96 & 24.41 \\
ory\_Orya & 9.78 & 18.18 & 10.77 & 2.67 & 28.75 & pag\_Latn & 16.01 & 86.36 & 60.77 & 9.19 & 45.06 & pan\_Guru & 9.58 & 21.54 & 11.46 & 3.06 & 31.03 & pap\_Latn & 7.11 & 67.49 & 29.84 & 1.28 & 23.81 \\
pbt\_Arab & 13.04 & 52.08 & 19.57 & 5.43 & 38.34 & pes\_Arab & 8.70 & 11.86 & 6.42 & 2.77 & 16.60 & plt\_Latn & 7.21 & 31.13 & 62.25 & 3.06 & 32.11 & pol\_Latn & 8.70 & 12.35 & 6.03 & 3.16 & 9.19 \\
por\_Latn & 5.43 & 8.20 & 5.04 & 1.68 & 17.59 & prs\_Arab & 7.71 & 14.92 & 7.21 & 2.96 & 21.94 & quy\_Latn & 28.85 & 96.15 & 80.43 & 17.29 & 77.08 & ron\_Latn & 5.83 & 7.02 & 3.46 & 1.68 & 7.41 \\
run\_Latn & 11.07 & 51.38 & 80.43 & 4.84 & 42.98 & rus\_Cyrl & 6.52 & 10.28 & 6.13 & 2.57 & 9.88 & sag\_Latn & 39.23 & 95.75 & 80.04 & 26.09 & 76.98 & san\_Deva & 19.96 & 80.14 & 19.86 & 8.40 & 45.55 \\
scn\_Latn & 12.25 & 62.85 & 33.30 & 6.42 & 41.11 & shn\_Mymr & 18.97 & 96.44 & 76.58 & 10.18 & 83.60 & sin\_Sinh & 9.09 & 18.18 & 6.13 & 4.25 & 34.78 & slk\_Latn & 8.10 & 9.88 & 5.53 & 2.67 & 12.55 \\
slv\_Latn & 7.91 & 12.75 & 5.34 & 2.27 & 13.14 & smo\_Latn & 11.96 & 41.50 & 83.40 & 5.53 & 44.57 & sna\_Latn & 11.76 & 49.11 & 77.27 & 4.05 & 40.91 & snd\_Arab & 11.17 & 43.87 & 8.10 & 4.74 & 45.85 \\
som\_Latn & 12.15 & 41.70 & 13.04 & 8.70 & 45.06 & sot\_Latn & 7.91 & 43.18 & 79.64 & 4.35 & 34.98 & spa\_Latn & 8.00 & 14.92 & 5.43 & 2.67 & 9.98 & srd\_Latn & 10.47 & 66.70 & 26.78 & 6.13 & 33.79 \\
srp\_Cyrl & 5.43 & 10.38 & 3.66 & 1.38 & 10.08 & ssw\_Latn & 12.06 & 74.21 & 47.92 & 6.62 & 45.55 & sun\_Latn & 10.87 & 18.18 & 8.10 & 3.85 & 29.64 & swe\_Latn & 5.83 & 8.30 & 4.84 & 1.28 & 8.99 \\
swh\_Latn & 7.11 & 16.80 & 7.71 & 2.77 & 28.75 & szl\_Latn & 6.72 & 57.61 & 18.68 & 3.56 & 32.51 & tam\_Taml & 14.23 & 18.68 & 9.29 & 4.05 & 31.92 & taq\_Latn & 57.61 & 96.05 & 79.84 & 39.43 & 86.17 \\
taq\_Tfng & 62.35 & 100.00 & 96.64 & 53.46 & 98.02 & tat\_Cyrl & 7.91 & 23.62 & 43.18 & 3.46 & 38.34 & tel\_Telu & 12.06 & 16.01 & 8.40 & 3.85 & 26.38 & tgk\_Cyrl & 8.40 & 23.81 & 82.91 & 3.66 & 45.26 \\
tgl\_Latn & 6.62 & 12.75 & 25.49 & 2.67 & 22.43 & tha\_Thai & 8.30 & 39.43 & 6.23 & 3.06 & 14.33 & tir\_Ethi & 14.82 & 98.52 & 64.62 & 7.11 & 58.00 & tpi\_Latn & 13.64 & 94.47 & 61.56 & 7.91 & 42.98 \\
tsn\_Latn & 13.54 & 61.07 & 82.71 & 6.03 & 40.91 & tso\_Latn & 13.14 & 91.80 & 74.80 & 5.34 & 40.91 & tuk\_Latn & 9.49 & 40.51 & 42.59 & 3.85 & 76.98 & tum\_Latn & 18.28 & 78.06 & 73.12 & 9.68 & 44.07 \\
tur\_Latn & 6.23 & 10.67 & 5.04 & 2.37 & 12.55 & twi\_Latn & 18.28 & 91.60 & 85.47 & 9.68 & 45.75 & tzm\_Tfng & 26.88 & 100.00 & 97.33 & 18.08 & 98.72 & uig\_Arab & 13.83 & 28.56 & 11.07 & 6.82 & 54.25 \\
ukr\_Cyrl & 7.91 & 11.96 & 6.42 & 3.16 & 10.08 & umb\_Latn & 36.56 & 95.06 & 77.57 & 26.98 & 77.47 & urd\_Arab & 9.88 & 17.79 & 6.82 & 4.15 & 30.43 & uzn\_Latn & 8.50 & 16.60 & 16.01 & 3.66 & 30.34 \\
vec\_Latn & 7.81 & 53.46 & 14.03 & 2.67 & 28.66 & vie\_Latn & 5.63 & 11.56 & 5.53 & 2.27 & 12.06 & war\_Latn & 7.11 & 32.51 & 47.04 & 2.87 & 24.60 & wol\_Latn & 28.56 & 93.87 & 77.77 & 16.90 & 66.60 \\
xho\_Latn & 10.18 & 41.01 & 12.94 & 4.74 & 34.68 & ydd\_Hebr & 8.60 & 52.67 & 31.62 & 3.36 & 57.61 & yor\_Latn & 22.73 & 71.25 & 84.49 & 16.80 & 58.79 & yue\_Hant & 10.67 & 58.70 & 8.30 & 3.95 & 17.59 \\
zho\_Hans & 9.98 & 50.69 & 7.41 & 3.06 & 14.43 & zho\_Hant & 14.23 & 58.30 & 9.78 & 4.15 & 17.89 & zsm\_Latn & 5.93 & 7.11 & 4.55 & 2.08 & 12.25 & zul\_Latn & 8.70 & 33.10 & 21.64 & 4.05 & 34.49 \\
        \hline
    \end{tabular}
    \caption{xsim++ results for all models in all languages, x-eng in FLORES200 devtest set.}
    \label{tab:xsim_pp_all_languages}
\end{table}
\end{landscape}

\begin{table}[htbp!]
    \centering
    \setlength{\tabcolsep}{1pt}
    \begin{subtable}[t]{0.48\textwidth}
        \centering
        \begin{tabular}{cccc}
            \toprule
            Lang & SONAR & \sonar & \sonar{} \\
             &  & -speech-3B & -speech-7B \\
            \midrule
            deu\_Latn & 0.0 & 0.0 & 0.0 \\
            fra\_Latn & 0.0 & 0.0 & 0.0 \\
            ron\_Latn & 0.6 & 0.6 & 0.6 \\
            rus\_Cyrl & 0.0 & 0.0 & 0.0 \\
            nld\_Latn & 0.0 & 0.0 & 0.0 \\
            fin\_Latn & 0.6 & 0.3 & 0.6 \\
            fin\_Latn & 0.0 & 0.0 & 0.0 \\
            spa\_Latn & 0.0 & 0.0 & 0.0 \\
            swh\_Latn & 0.0 & 1.6 & 2.6 \\
            por\_Latn & 0.0 & 0.0 & 0.0 \\
            pol\_Latn & 0.0 & 0.0 & 0.0 \\
            ita\_Latn & 0.0 & 0.0 & 0.0 \\
            cat\_Latn & 0.0 & 0.0 & 0.0 \\
            ben\_Beng & 0.0 & 0.0 & 0.0 \\
            kor\_Hang & 0.0 & 0.0 & 0.0 \\
            hin\_Deva & 0.4 & 1.5 & 1.5 \\
            bul\_Cyrl & 0.3 & 0.0 & 0.0 \\
            vie\_Latn & 0.3 & 0.3 & 0.0 \\
            urd\_Arab & 0.4 & 0.4 & 1.3 \\
            ukr\_Cyrl & 0.0 & 0.0 & 0.0 \\
            tha\_Thai & 0.3 & 0.3 & 2.0 \\
            tel\_Telu & 0.3 & 0.3 & 2.0 \\
            tgl\_Latn & 0.0 & 0.3 & 3.7 \\
            tam\_Taml & 0.0 & 1.2 & 4.2 \\
            ind\_Latn & 0.0 & 0.0 & 0.6 \\
            cym\_Latn & 0.3 & 1.4 & 0.3 \\
            arb\_Arab & 0.0 & 0.0 & 0.0 \\
            cmn\_Hans & 0.0 & 0.6 & 0.0 \\
            fas\_Arab & 0.6 & 0.0 & 0.0 \\
            uzn\_Latn & 0.3 & 2.6 & 7.5 \\
            jpn\_Jpan & 0.3 & 0.0 & 0.0 \\
            kan\_Knda & 0.3 & 0.0 & 0.3 \\
            dan\_Latn & 0.0 & 0.6 & 1.0 \\
            slk\_Latn & 0.3 & 0.3 & 0.9 \\
            ces\_Latn & 0.0 & 0.0 & 0.0 \\
            mlt\_Latn & 0.0 & 0.3 & 0.0 \\
            \midrule
            average & 0.1 & 0.4 & 0.8 \\
            \bottomrule
        \end{tabular}
        \caption{xsim}
        \label{tab:xsim_speech_breakdown}
    \end{subtable}
    \hfill
    \begin{subtable}[t]{0.48\textwidth}
        \centering
        \begin{tabular}{cccc}
            \toprule
            Lang & SONAR & \sonar & \sonar{} \\
             &  & -speech-3B & -speech-7B \\
            \midrule
            deu\_Latn & 11.2 & 5.2 & 4.9 \\
            fra\_Latn & 13.0 & 5.4 & 6.2 \\
            ron\_Latn & 13.8 & 8.9 & 9.8 \\
            rus\_Cyrl & 12.5 & 4.4 & 5.8 \\
            nld\_Latn & 16.4 & 6.4 & 5.2 \\
            fin\_Latn & 17.0 & 7.5 & 7.2 \\
            fin\_Latn & 14.0 & 4.9 & 8.2 \\
            spa\_Latn & 15.2 & 4.6 & 5.5 \\
            swh\_Latn & 19.6 & 17.0 & 16.7 \\
            por\_Latn & 13.8 & 2.6 & 4.9 \\
            pol\_Latn & 15.2 & 5.5 & 7.9 \\
            ita\_Latn & 17.7 & 3.5 & 4.6 \\
            cat\_Latn & 9.4 & 5.7 & 4.0 \\
            ben\_Beng & 17.5 & 11.8 & 10.9 \\
            kor\_Hang & 17.0 & 7.0 & 10.7 \\
            hin\_Deva & 15.9 & 9.8 & 12.5 \\
            bul\_Cyrl & 16.9 & 8.7 & 9.0 \\
            vie\_Latn & 25.3 & 15.6 & 15.3 \\
            urd\_Arab & 21.7 & 18.3 & 20.9 \\
            ukr\_Cyrl & 15.2 & 6.1 & 6.7 \\
            tha\_Thai & 21.2 & 12.3 & 15.2 \\
            tel\_Telu & 20.9 & 17.6 & 23.8 \\
            tgl\_Latn & 26.9 & 25.2 & 32.7 \\
            tam\_Taml & 22.7 & 12.7 & 25.1 \\
            ind\_Latn & 18.9 & 11.6 & 13.1 \\
            cym\_Latn & 39.1 & 24.9 & 19.7 \\
            arb\_Arab & 14.8 & 11.3 & 7.4 \\
            cmn\_Hans & 16.6 & 8.6 & 5.7 \\
            fas\_Arab & 19.4 & 7.7 & 10.2 \\
            uzn\_Latn & 16.2 & 17.4 & 28.7 \\
            jpn\_Jpan & 18.1 & 9.4 & 10.6 \\
            kan\_Knda & 21.2 & 7.0 & 11.9 \\
            dan\_Latn & 13.7 & 9.7 & 14.2 \\
            slk\_Latn & 14.7 & 6.2 & 7.4 \\
            ces\_Latn & 14.5 & 5.9 & 7.1 \\
            mlt\_Latn & 18.7 & 11.8 & 11.2 \\
            \midrule
            average & 17.7 & 10.1 & 11.6 \\
            \bottomrule
        \end{tabular}
        \caption{xsim++}
        \label{tab:xsim_pp_speech_breakdown}
    \end{subtable}
    
    \caption{\sonar{}-speech modality results on 36 common languages SONAR and \sonar-speech models cover, x-eng on FLEURS dev set. SONAR results are for 36 different language-specific encoders, while \sonar-speech are one model across all languages.}
    \label{tab:speech_comparison}
\end{table}

\begin{table*}[h!]
    \centering
    \resizebox{1.0\textwidth}{!}{%
    \begin{tabular}{@{}lcccc clcccc@{}}
        \toprule
        \textbf{Family} & \textbf{Langs} & \textbf{Examples} & \textbf{xsim} & \textbf{\chrf} & \phantom{sp} & \textbf{Family} & \textbf{Langs} & \textbf{Examples} & \textbf{xsim} & \textbf{\chrf} \\
        \cmidrule(r){1-5} \cmidrule(l){7-11}
        Indo-European & 535 & 6,329,570 & 0.9 & 56.3 & & Huitotoan & 3 & 107 & 6.1 & 32.3 \\
        Atlantic-Congo & 652 & 464,728 & 2.8 & 43.3 & & Border & 2 & 105 & 6.1 & 31.4 \\
        Afroasiatic & 196 & 419,370 & 4.2 & 44.0 & & Eastern Daly & 3 & 102 & 7.5 & 31.1 \\
        Turkic & 82 & 410,656 & 0.6 & 55.4 & & Kunimaipan & 3 & 96 & 9.3 & 29.9 \\
        Dravidian & 39 & 369,957 & 1.9 & 49.7 & & Araucanian & 3 & 90 & 4.5 & 38.9 \\
        Austronesian & 795 & 315,373 & 3.0 & 41.8 & & Lengua-Mascoy & 6 & 84 & 12.2 & 31.4 \\
        Uralic & 56 & 239,799 & 0.4 & 54.1 & & Boran & 2 & 82 & 9.8 & 30.8 \\
        Artificial Language & 28 & 72,436 & 0.2 & 69.2 & & Guaicuruan & 4 & 78 & 7.1 & 31.6 \\
        Tai-Kadai & 46 & 31,125 & 1.7 & 51.3 & & Zamucoan & 2 & 74 & 8.4 & 30.8 \\
        Japonic & 36 & 29,761 & 0.3 & 65.1 & & Yanomamic & 4 & 73 & 8.2 & 29.4 \\
        Sino-Tibetan & 217 & 29,272 & 2.6 & 44.3 & & Uru-Chipaya & 1 & 72 & 5.8 & 35.6 \\
        Austroasiatic & 102 & 25,720 & 1.8 & 45.9 & & Heiban & 2 & 71 & 2.8 & 35.8 \\
        Mongolic & 15 & 16,335 & 1.2 & 52.0 & & Chiquitano & 1 & 71 & 7.3 & 33.9 \\
        Koreanic & 4 & 16,146 & 0.2 & 63.6 & & South-Central Papuan & 1 & 70 & 7.8 & 34.4 \\
        Saharan & 4 & 14,351 & 2.5 & 40.2 & & Baining & 3 & 70 & 3.6 & 36.6 \\
        Nilotic & 28 & 12,853 & 2.5 & 44.5 & & Ticuna-Yuri & 1 & 68 & 5.3 & 33.6 \\
        Quechuan & 34 & 6,784 & 2.6 & 43.5 & & Tarascan & 2 & 67 & 2.8 & 37.4 \\
        Mande & 40 & 6,437 & 2.9 & 42.2 & & Cahuapanan & 1 & 65 & 11.9 & 27.7 \\
        Nuclear Trans New Guinea & 112 & 4,434 & 6.6 & 32.5 & & Yareban & 2 & 61 & 10.5 & 28.0 \\
        Otomanguean & 110 & 4,195 & 3.6 & 37.5 & & Iroquoian & 9 & 60 & 3.5 & 29.0 \\
        Mayan & 32 & 2,858 & 3.6 & 38.7 & & Eastern Trans-Fly & 2 & 59 & 6.0 & 32.5 \\
        Aymaran & 2 & 1,870 & 0.2 & 54.1 & & Jicaquean & 1 & 59 & 9.3 & 30.2 \\
        Uto-Aztecan & 38 & 1,767 & 5.0 & 39.2 & & Koman & 1 & 58 & 1.0 & 45.1 \\
        Arawakan & 31 & 1,218 & 7.3 & 31.8 & & Khoe-Kwadi & 7 & 56 & 3.0 & 40.9 \\
        Central Sudanic & 33 & 1,013 & 4.0 & 39.0 & & Misumalpan & 4 & 52 & 3.8 & 38.7 \\
        Tupian & 32 & 948 & 6.4 & 34.7 & & Bororoan & 1 & 48 & 16.1 & 28.6 \\
        Tucanoan & 18 & 831 & 6.5 & 33.2 & & Yawanawá & 1 & 47 & 7.5 & 31.7 \\
        Gongduk & 11 & 735 & 1.9 & 44.8 & & Southern Daly & 1 & 45 & 3.0 & 36.8 \\
        Nakh-Daghestanian & 34 & 728 & 4.1 & 39.7 & & Kamula & 1 & 44 & 4.8 & 33.9 \\
        Pama-Nyungan & 83 & 678 & 9.2 & 29.1 & & Torricelli & 1 & 41 & 10.4 & 29.4 \\
        Chibchan & 12 & 576 & 7.6 & 32.7 & & Maningrida & 1 & 41 & 5.8 & 31.9 \\
        Panoan & 20 & 555 & 11.9 & 29.7 & & Zaparoan & 2 & 40 & 7.3 & 29.8 \\
        Cariban & 19 & 508 & 7.0 & 31.4 & & Peba-Yagua & 2 & 37 & 8.7 & 32.1 \\
        Totonacan & 10 & 481 & 5.5 & 36.2 & & Nadahup & 2 & 37 & 9.8 & 30.9 \\
        Kru & 14 & 460 & 4.8 & 37.0 & & Huavean & 3 & 36 & 5.2 & 33.0 \\
        Algic & 43 & 395 & 3.5 & 37.6 & & Kadu & 1 & 36 & 3.2 & 38.4 \\
        Mixe-Zoque & 14 & 387 & 5.0 & 35.1 & & Anim & 1 & 36 & 6.0 & 34.5 \\
        Songhay & 7 & 369 & 0.8 & 50.3 & & Manubaran & 1 & 36 & 1.8 & 38.8 \\
        Nuclear Torricelli & 9 & 337 & 4.9 & 33.5 & & Kresh-Aja & 1 & 35 & 4.5 & 36.7 \\
        Eskimo-Aleut & 14 & 333 & 3.9 & 33.6 & & Hatam-Mansim & 1 & 35 & 5.8 & 33.1 \\
        Sepik & 8 & 321 & 8.0 & 32.1 & & Gumuz & 1 & 35 & 3.3 & 37.0 \\
        Dogon & 12 & 315 & 5.6 & 35.6 & & Left May & 1 & 35 & 11.4 & 28.6 \\
        Nuclear Macro-Je & 11 & 313 & 13.2 & 26.7 & & North Bougainville & 1 & 35 & 2.8 & 33.4 \\
        Bookkeeping & 23 & 310 & 4.6 & 40.5 & & Tequistlatecan & 1 & 35 & 11.0 & 29.9 \\
        Athabaskan-Eyak-Tlingit & 38 & 296 & 5.5 & 33.8 & & Somahai & 1 & 35 & 6.2 & 32.3 \\
        Jivaroan & 4 & 295 & 5.3 & 33.3 & & Pauwasi & 1 & 35 & 4.0 & 35.6 \\
        Chocoan & 5 & 252 & 5.9 & 33.4 & & Sentani & 2 & 35 & 6.5 & 31.3 \\
        Ramu-Lower Sepik & 4 & 252 & 3.2 & 36.8 & & Maybrat & 1 & 34 & 9.1 & 30.7 \\
        Bosavi & 4 & 234 & 6.8 & 34.0 & & Senagi & 1 & 34 & 5.8 & 35.0 \\
        Koiarian & 5 & 219 & 6.8 & 32.2 & & Piawi & 1 & 34 & 8.9 & 30.8 \\
        Angan & 6 & 209 & 7.0 & 31.4 & & Kakua-Nukak & 2 & 34 & 11.2 & 29.0 \\
        Barbacoan & 5 & 206 & 5.8 & 33.0 & & Harakmbut & 2 & 34 & 14.6 & 28.0 \\
        Pidgin & 5 & 203 & 0.5 & 50.8 & & Nambiquaran & 1 & 34 & 10.3 & 29.6 \\
        Teberan & 2 & 178 & 3.9 & 35.5 & & Geelvink Bay & 1 & 33 & 10.9 & 27.3 \\
        Ndu & 4 & 174 & 9.3 & 29.3 & & Kiowa-Tanoan & 4 & 30 & 10.0 & 31.2 \\
        North Halmahera & 5 & 157 & 4.6 & 36.8 & & Narrinyeri & 1 & 28 & 5.5 & 31.3 \\
        Dagan & 4 & 156 & 6.2 & 32.9 & & Garrwan & 2 & 21 & 14.0 & 26.1 \\
        Matacoan & 7 & 140 & 10.8 & 29.6 & & South Bougainville & 1 & 17 & 15.5 & 27.0 \\
        Guahiboan & 3 & 139 & 9.6 & 29.3 & & Iwaidjan Proper & 1 & 1 & 0.0 & 35.0 \\
        East Bird's Head & 4 & 130 & 2.8 & 36.4 & & & & & & \\
        Gunwinyguan & 3 & 119 & 7.7 & 30.0 & & & & & & \\
        Surmic & 3 & 113 & 5.7 & 34.9 & & & & & & \\
        Arauan & 3 & 112 & 11.1 & 26.8 & & & & & & \\
        \bottomrule
    \end{tabular}%
    }
    \caption{Detailed statistics and results per language family. Results in BIBLE \test. Number of examples is in thousands. Families sorted by number of examples. We removed 55 families that are absent from BIBLE \test.}
    \label{tab:lang_families}
\end{table*}

\newpage
\section{Embedding visualization}
\label{app:umap_vis}
\begin{figure}[htbp]
    \centering

    \begin{subfigure}{0.32\linewidth}
        \centering
        \includegraphics[width=\linewidth]{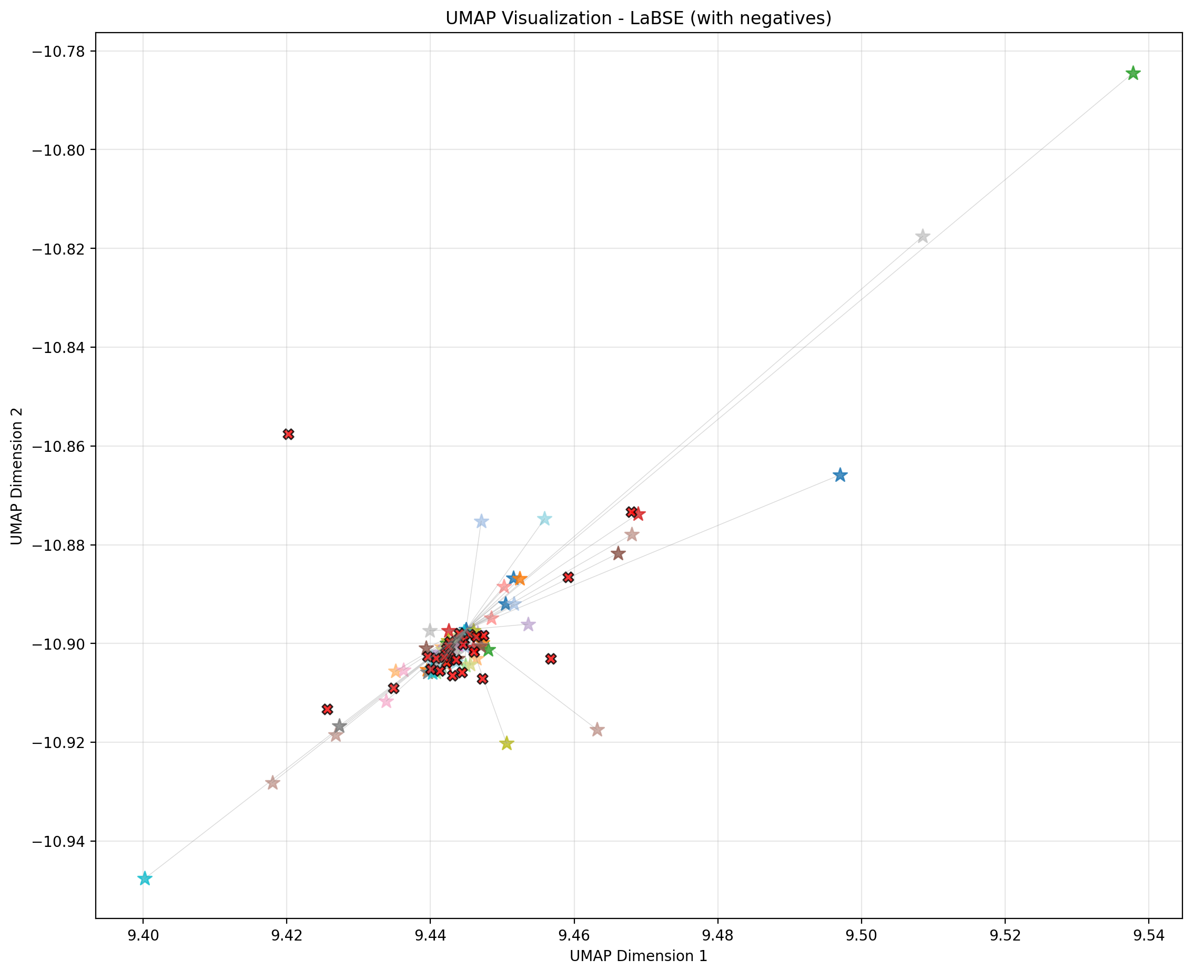}
        \caption{LaBSE}
    \end{subfigure}
    \hfill
    \begin{subfigure}{0.32\linewidth}
        \centering
        \includegraphics[width=\linewidth]{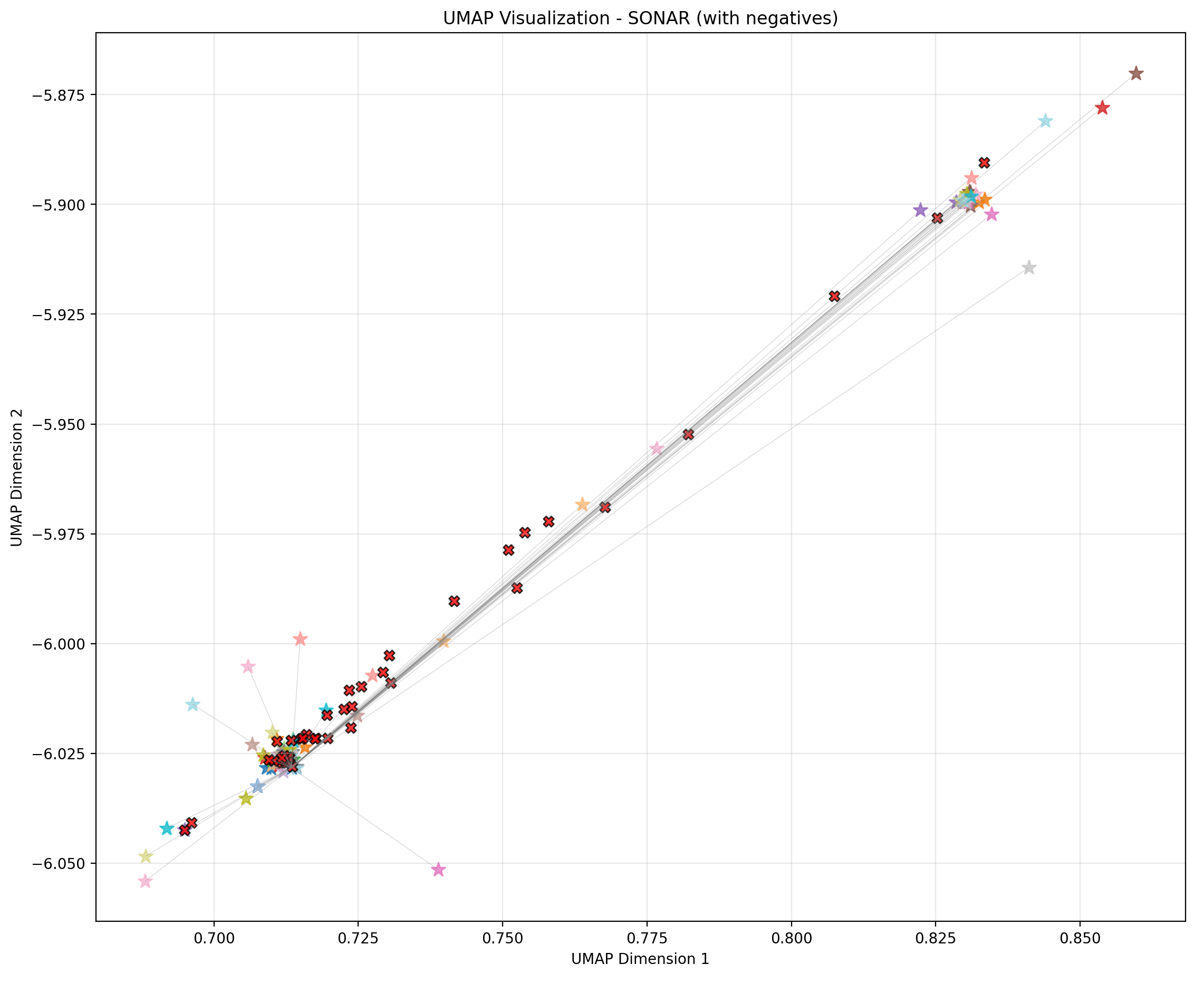}
        \caption{SONAR}
    \end{subfigure}
    \hfill
    \begin{subfigure}{0.32\linewidth}
        \centering
        \includegraphics[width=\linewidth]{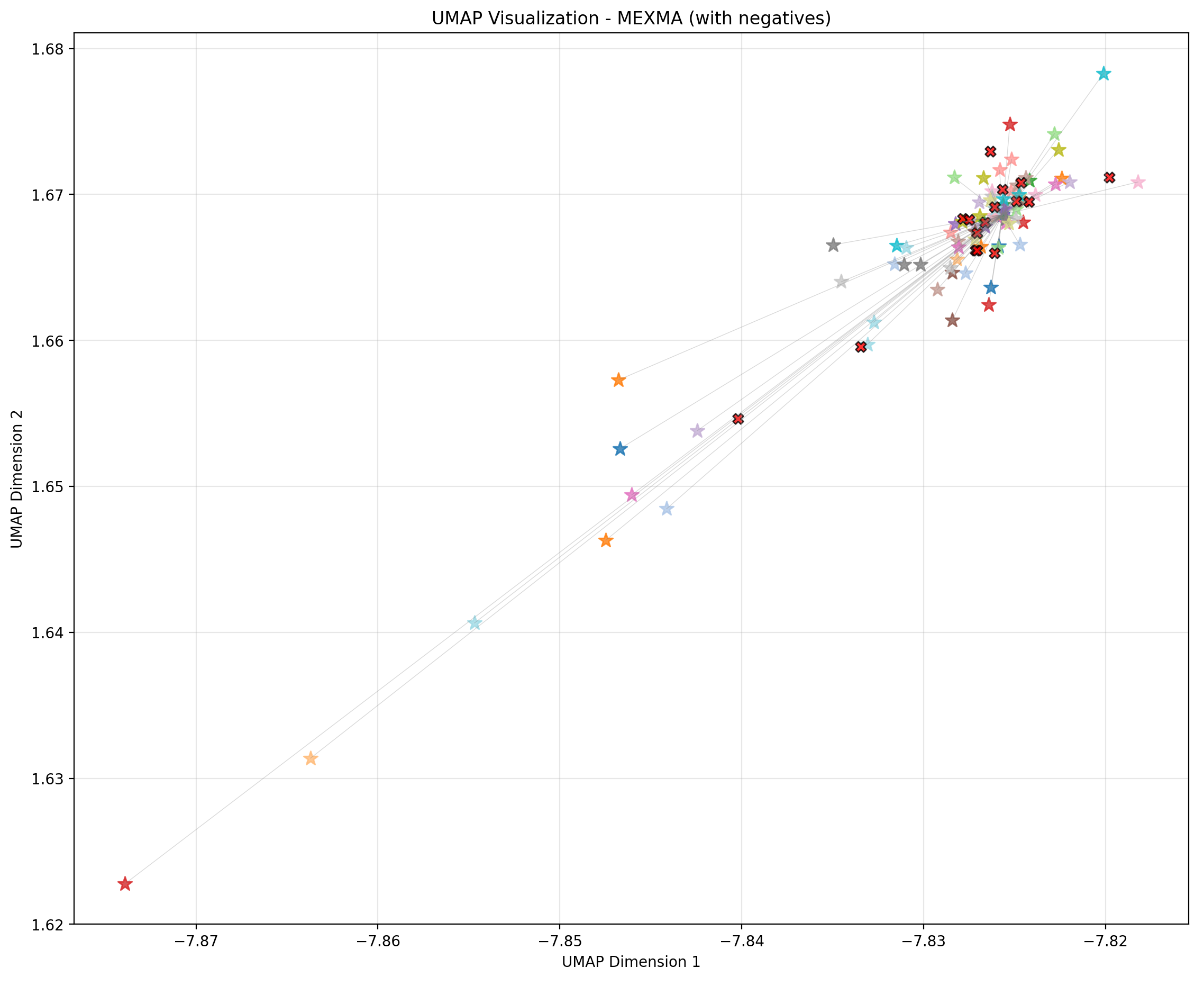}
        \caption{MEXMA}
    \end{subfigure}

    \vspace{1em}
    \par\medskip

    \begin{subfigure}{0.43\linewidth}
        \centering
        \includegraphics[width=\linewidth]{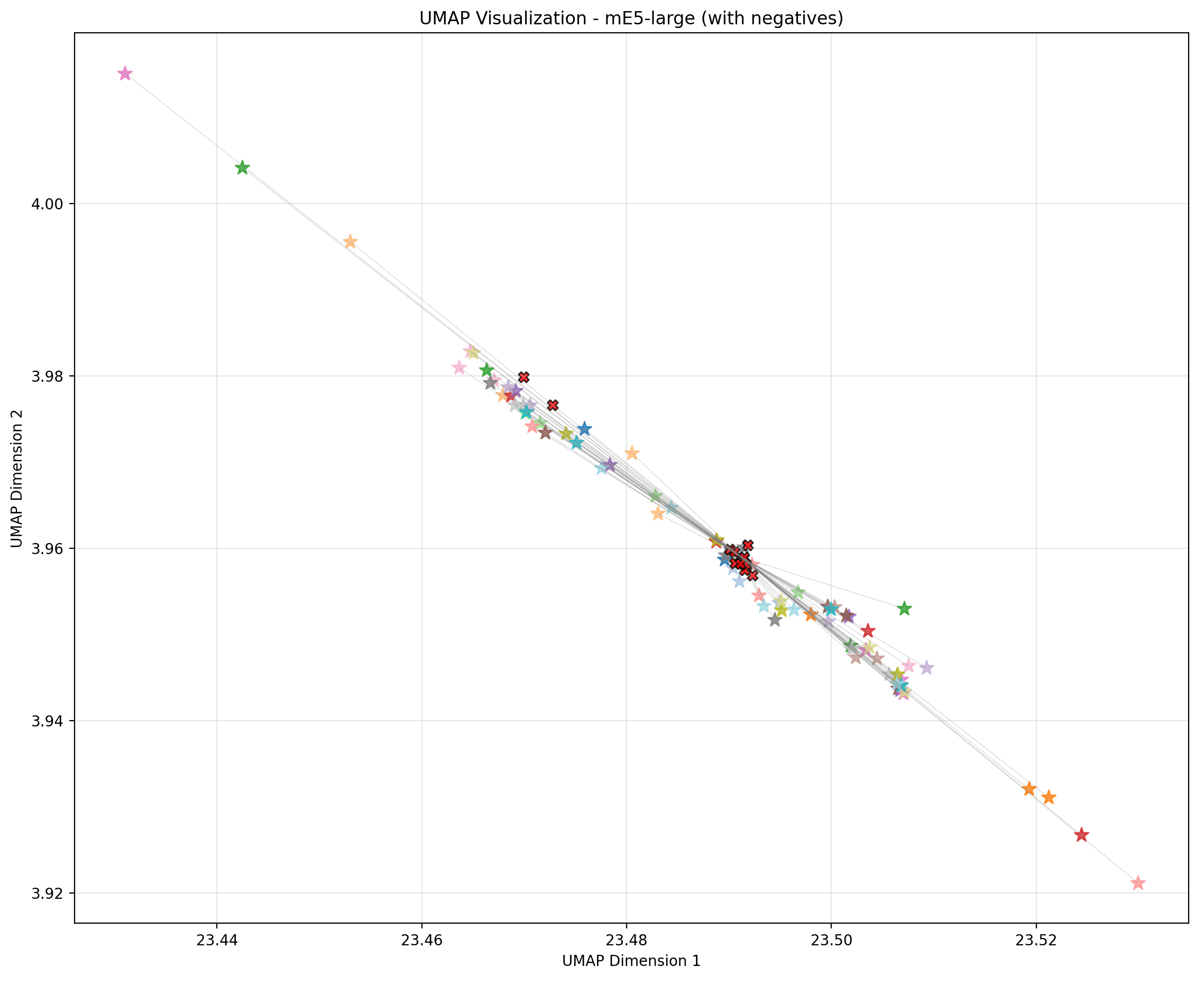}
        \caption{mE5\textsubscript{large}}
    \end{subfigure}
    \hfill
    \begin{subfigure}{0.56\linewidth}
        \centering
        \includegraphics[width=\linewidth]{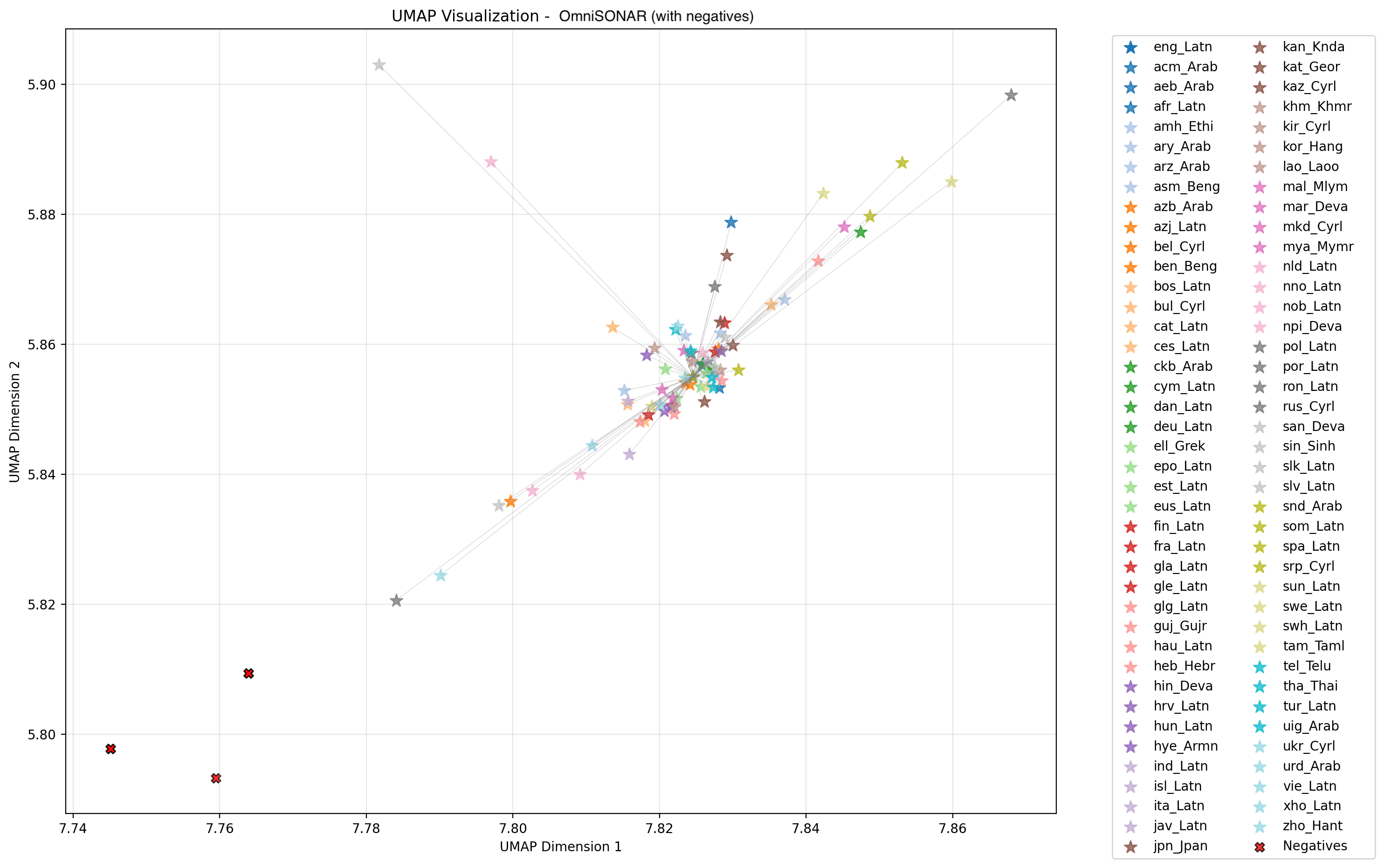}
        \caption{\sonar{}}
    \end{subfigure}

    \caption{UMAP visualization of the sentence ``\textit{During his time with the team, he scored 403 goals in 468 appearances.}'' from FLORES devset along closest hard negatives, shown as red crosses. Lines connect the translations to their English counterpart.}
    \label{fig:umap_models_grid}
\end{figure}

So far, quantitative results have showcased the efficacy of \sonar{} as an embedding space. Although visualization approaches such as UMAP \cite{mcinnes2020umapuniformmanifoldapproximation} may lead to misinterpretations of the embedding spaces, they can provide visual support to our cross-lingual alignment results. To illustrate this, we fit a UMAP projection on the FLORES devset and plot one randomly sampled English sentence alongside its translations, with the hard negatives from \citet{chen-etal-2023-xsim}. To ensure fairness, we only plot the languages common to our baselines. As visualized in \autoref{fig:umap_models_grid}, \sonar{} is the only model for which hard negatives are not within the cluster defined around the English sentence. %

\begin{figure}[htbp]
    \centering

    \begin{subfigure}{0.32\linewidth}
        \centering
        \includegraphics[width=\linewidth]{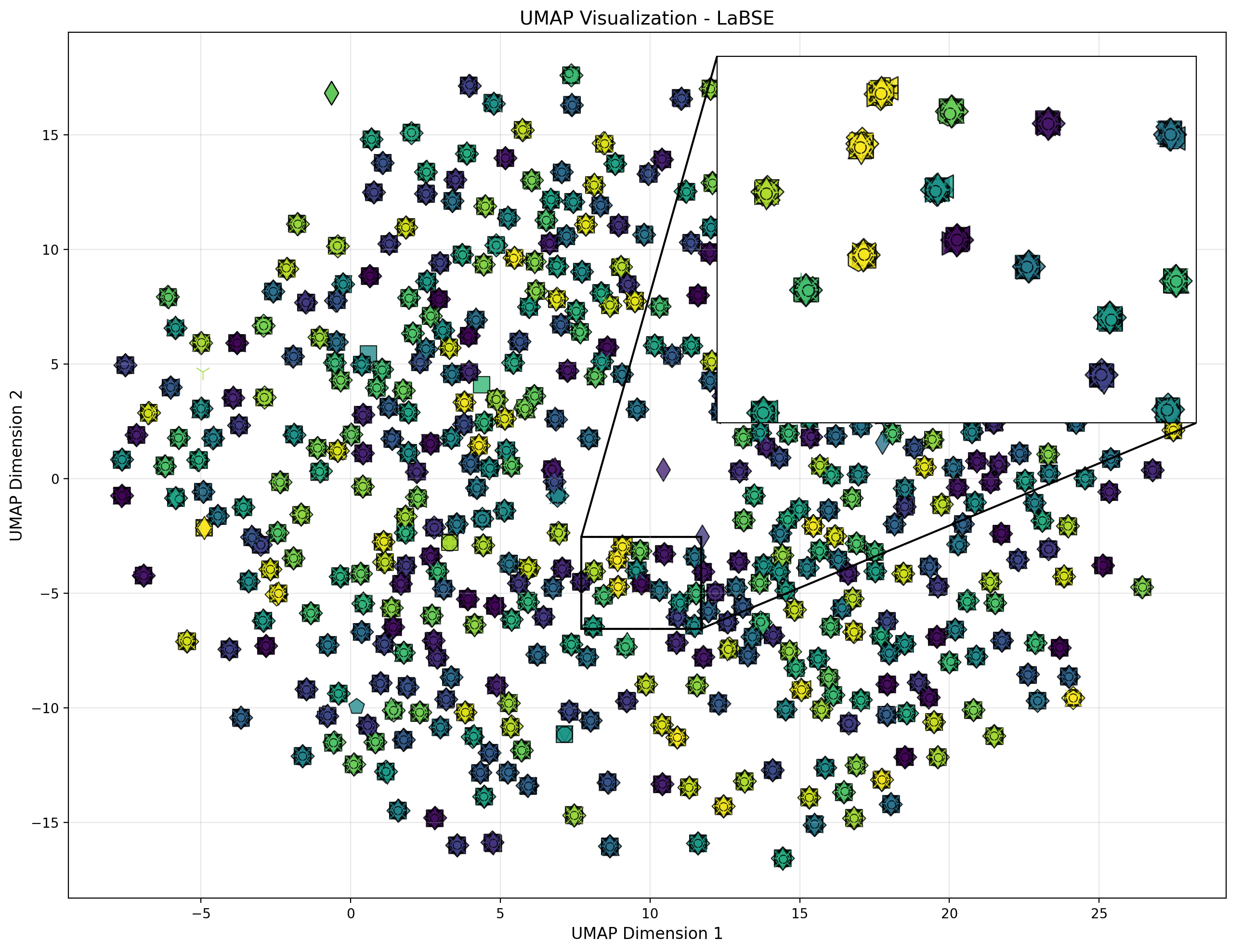}
        \caption{LaBSE}
    \end{subfigure}
    \hfill
    \begin{subfigure}{0.32\linewidth}
        \centering
        \includegraphics[width=\linewidth]{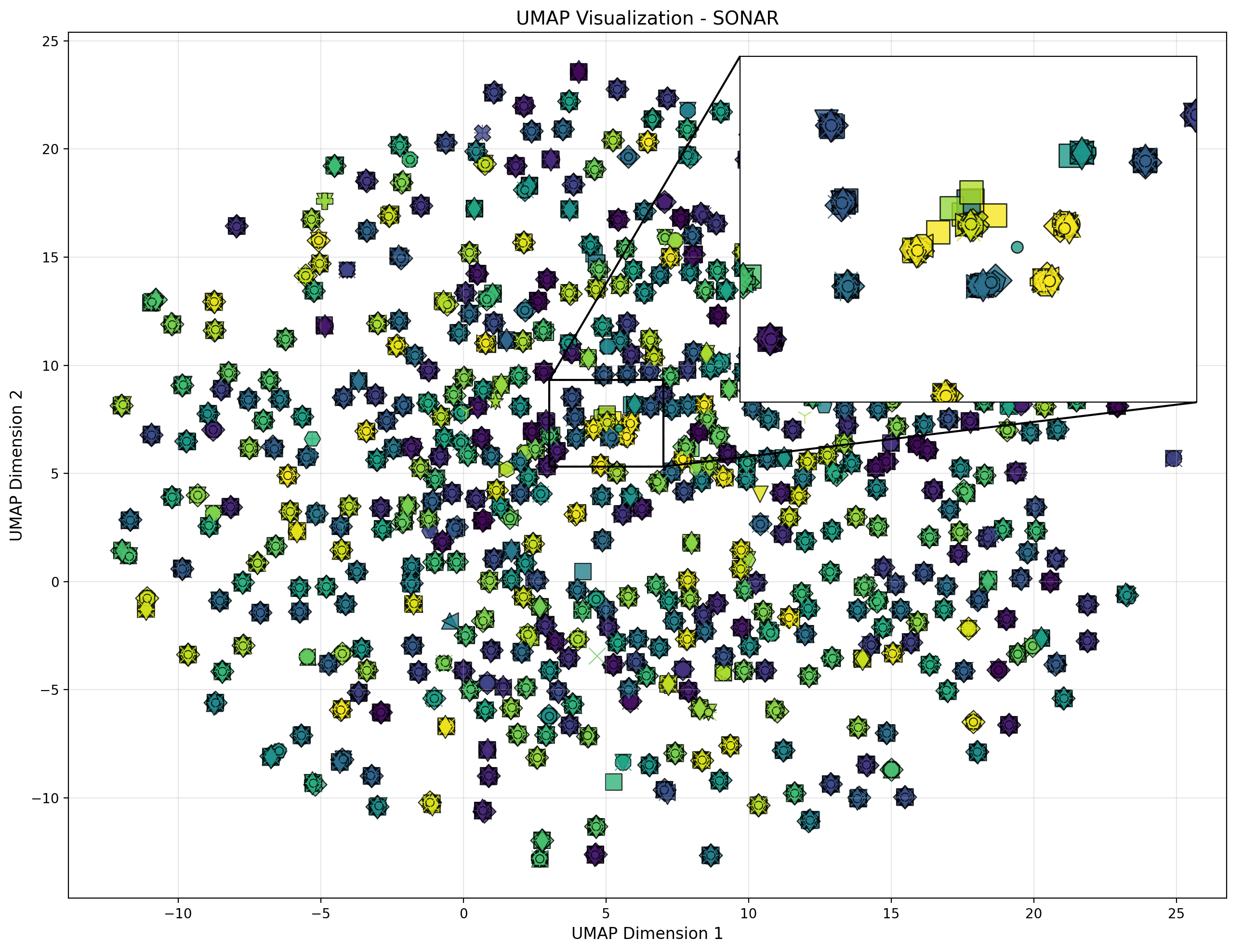}
        \caption{SONAR}
    \end{subfigure}
    \hfill
    \begin{subfigure}{0.32\linewidth}
        \centering
        \includegraphics[width=\linewidth]{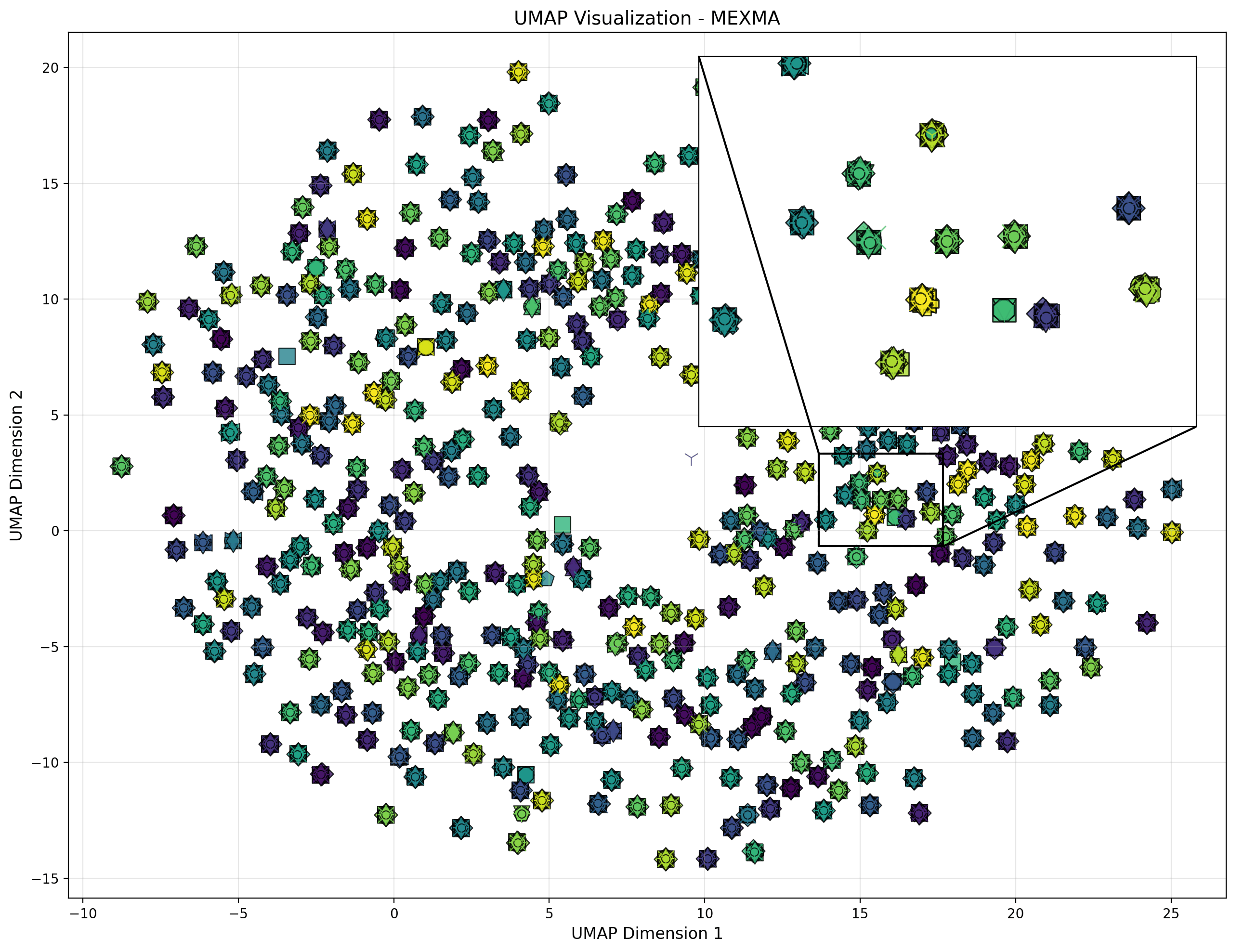}
        \caption{MEXMA}
    \end{subfigure}

    \vspace{1em}
    \par\medskip

    \begin{subfigure}{0.46\linewidth}
        \centering
        \includegraphics[width=\linewidth]{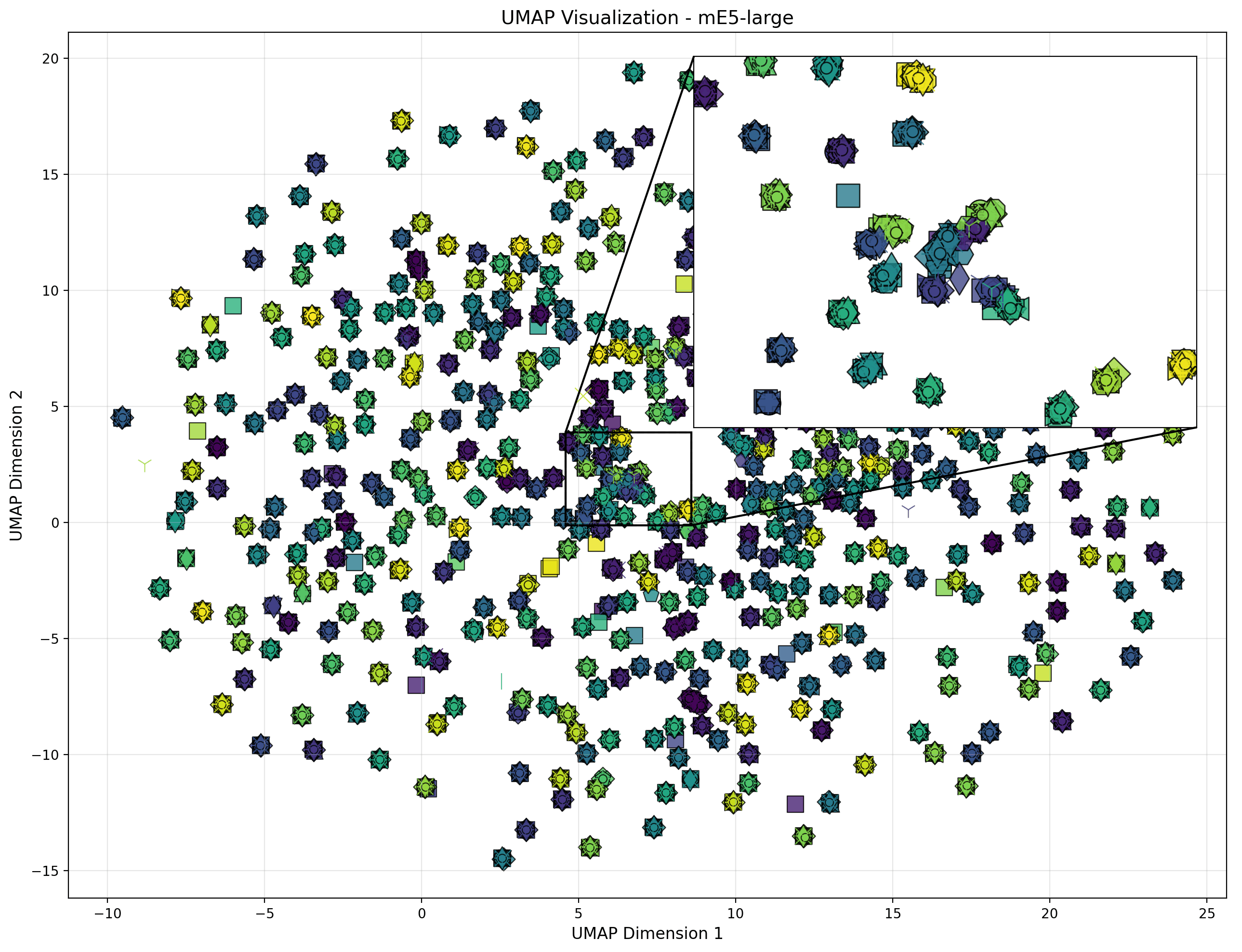}
        \caption{mE5\textsubscript{large}}
    \end{subfigure}
    \hfill
    \begin{subfigure}{0.53\linewidth}
        \centering
        \includegraphics[width=\linewidth]{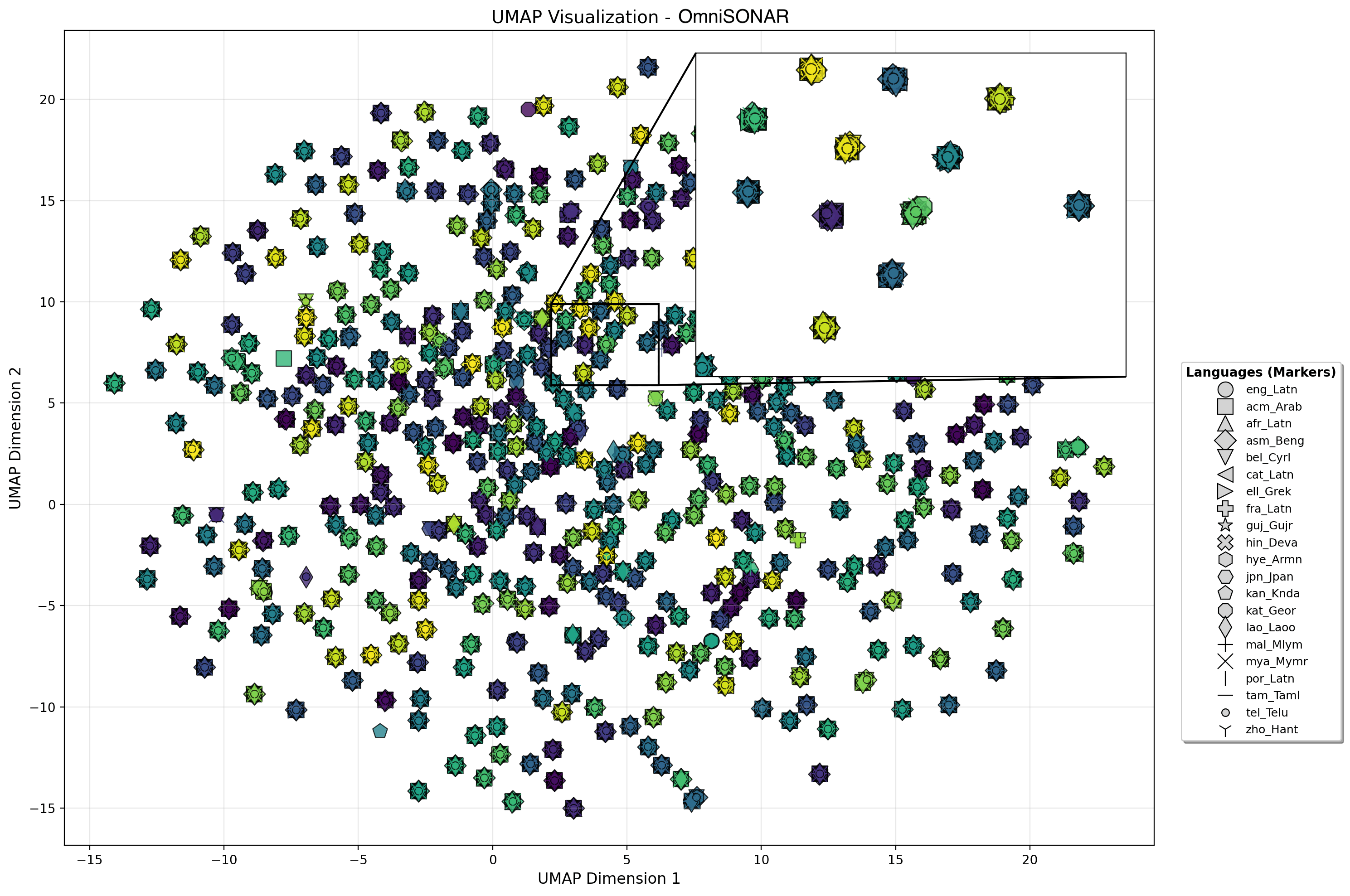}
        \caption{\sonar{}}
    \end{subfigure}

    \caption{UMAP visualization of the whole space defined by the FLORES devset for 20 languages with different scripts.}
    \label{fig:umap_models_whole_space}
\end{figure}

For a broader perspective, \autoref{fig:umap_models_whole_space} displays 500 sentences from the devset, excluding hard negatives. Across models, clusters consistently form around the same sentence in different languages, with MEXMA, LaBSE, and \sonar{} exhibiting fewer outliers.
However, when hard negatives are introduced (see \autoref{fig:umap_models_grid}), most models fail to separate them from the target cluster. This visualization highlights the trade-off between xsim and xsim++ performance discussed in \autoref{section:results}: \sonar{}'s contrastive training enables it to push hard negatives away (improving xsim++, as per \autoref{fig:umap_models_grid}), without compromising its cross-lingual alignment (xsim, as per \autoref{fig:umap_models_whole_space}).

\newpage
\section{Omninilingual Extension Training Algorithm} \label{sec:appendix/omni_algorithm}

\begin{algorithm*}[]
\caption{Omnilingual Student-Teacher Distillation Training}
\label{alg:omnilingual_distillation}
\begin{algorithmic}[1]
\Require Batch $\mathcal{B} = \{(x_i, y_i)\}_{i=1}^N$ of parallel sentences
\Require Student encoder $f_S$, frozen teacher encoder $f_T$
\Require Base language hyperparameters: $\lambda^{\text{MSE}}_{\text{base}}, \lambda^{student \rightarrow teacher}_{\text{base}}, \lambda^{teacher \rightarrow student}_{\text{base}}, \tau_{\text{base}}$
\Require New language hyperparameters: $\lambda^{\text{MSE}}_{\text{new}}, \lambda^{student \rightarrow teacher}_{\text{new}}, \lambda^{teacher \rightarrow student}_{\text{new}}, \tau_{\text{new}}$

\State \textbf{Compute student embeddings:}
\For{$i = 1$ to $N$}
    \State $x^s_i \gets f_S(x_i)$ \Comment{Student embedding of source sentence}
\EndFor

\State \textbf{Compute teacher embeddings:}
\For{$i = 1$ to $N$}
    \State $x^t_i \gets f_T(x_i)$ \Comment{Teacher embedding of source sentence}
    \State $y^t_i \gets f_T(y_i)$ \Comment{Teacher embedding of target sentence}
    \If{$x_i$ is a base language}
        \State $z^t_i \gets \frac{1}{2}(x^t_i + y^t_i)$ \Comment{Interpolated teacher embedding}
        \State $(\lambda^{\text{MSE}}_i, \lambda^{student \rightarrow teacher}_i, \lambda^{teacher \rightarrow student}_i, \tau_i) \gets
               (\lambda^{\text{MSE}}_{\text{base}}, \lambda^{student \rightarrow teacher}_{\text{base}},
                \lambda^{teacher \rightarrow student}_{\text{base}}, \tau_{\text{base}})$
    \Else
        \State $z^t_i \gets y^t_i$ \Comment{Target-only teacher embedding (new language)}
        \State $(\lambda^{\text{MSE}}_i, \lambda^{student \rightarrow teacher}_i, \lambda^{teacher \rightarrow student}_i, \tau_i) \gets
               (\lambda^{\text{MSE}}_{\text{new}}, \lambda^{student \rightarrow teacher}_{\text{new}},
                \lambda^{teacher \rightarrow student}_{\text{new}}, \tau_{\text{new}})$
    \EndIf
\EndFor

\State \textbf{Compute losses:}
\State Initialize total loss $\mathcal{L} \gets 0$
\For{$i = 1$ to $N$}
    \State \textit{// Mean Squared Error (MSE) loss}
    \State $\mathcal{L}^i_{\text{MSE}} \gets \lVert x^s_i - z^t_i \rVert^2$

    \State \textit{// Student $\rightarrow$ Teacher contrastive loss}
    \State $\text{denom}_{S \rightarrow T} \gets \sum_{k=1}^{N} \exp(\cos(x^s_i, z^t_k) \cdot \tau_i)$
    \State $\mathcal{L}^i_{student \rightarrow teacher} \gets
           -\log \frac{\exp(\cos(x^s_i, z^t_i) \cdot \tau_i)}{\text{denom}_{S \rightarrow T}}$

    \State \textit{// Teacher $\rightarrow$ Student contrastive loss}
    \State $\text{denom}_{T \rightarrow S} \gets \sum_{k=1}^{N} \exp(\cos(z^t_i, x^s_k) \cdot \tau_i)$
    \State $\mathcal{L}^i_{teacher \rightarrow student} \gets
           -\log \frac{\exp(\cos(z^t_i, x^s_i) \cdot \tau_i)}{\text{denom}_{T \rightarrow S}}$

    \State \textit{// Combine losses for example $i$}
    \State $\mathcal{L}_i \gets
           \lambda^{student \rightarrow teacher}_i \cdot \mathcal{L}^i_{student \rightarrow teacher}
           + \lambda^{teacher \rightarrow student}_i \cdot \mathcal{L}^i_{teacher \rightarrow student}
           + \lambda^{\text{MSE}}_i \cdot \mathcal{L}^i_{\text{MSE}}$
    \State $\mathcal{L} \gets \mathcal{L} + \mathcal{L}_i$
\EndFor

\State \textbf{Batch loss:} $\mathcal{L} \gets \frac{1}{N}\mathcal{L}$
\Return $\mathcal{L}$

\end{algorithmic}
\end{algorithm*}

\end{document}